\documentclass[10pt]{article} 
\usepackage[accepted]{tmlr}

\usepackage{amsmath,amsfonts,bm}

\def\eqref#1{equation~\ref{#1}}

\def\1{\bm{1}}

\DeclareMathAlphabet{\mathsfit}{\encodingdefault}{\sfdefault}{m}{sl}
\SetMathAlphabet{\mathsfit}{bold}{\encodingdefault}{\sfdefault}{bx}{n}

\usepackage{hyperref}
\usepackage{url}
\usepackage{algorithm}
\usepackage{algorithmic}
\usepackage{latexsym}
\usepackage{amsthm}
\usepackage{booktabs}
\usepackage{enumitem}
\usepackage{subcaption}
\usepackage{amsmath,amssymb}
\usepackage{xspace,stmaryrd}
\usepackage{nicefrac}
\usepackage{comment}
\usepackage{multirow}
\usepackage{xcolor}
\usepackage{tikz}
\usepackage{varwidth}
\usetikzlibrary{arrows.meta, shapes, positioning}
\tikzset{thickarrow/.style={->, thick, >=Stealth}}
\usepackage{colortbl}

\title{SPACR: Single-Pass Adaptive Training \\of Uncertainty-Aware Conformal Regressors}

\author{\name Soundouss Messoudi \email soundouss.messoudi@hds.utc.fr \\
      \addr Heudiasyc - UMR CNRS 7253, Université de Technologie de Compiègne
      \AND
      \name Sylvain Rousseau \email sylvain.rousseau@hds.utc.fr  \\
      \addr Heudiasyc - UMR CNRS 7253, Université de Technologie de Compiègne
      \AND
      \name Sébastien Destercke \email sebastien.destercke@hds.utc.fr\\
      \addr Heudiasyc - UMR CNRS 7253, Université de Technologie de Compiègne}

\def\month{10}  
\def\year{2026} 
\def\openreview{\url{https://openreview.net/forum?id=dScMsfkwoC}} 

\begin{document}

\maketitle

\begin{abstract}
Conformal Prediction (CP) provides robust uncertainty guarantees for predictive models, but is typically applied post hoc, which misaligns model training with the conformal goal of producing efficient (i.e., narrow) intervals. We propose SPACR (Single-Pass Adaptive Conformal Regressor), a novel method for directly training uncertainty-aware regressors within a differentiable loss. SPACR jointly optimizes accuracy, efficiency, and validity without batch-splitting or a predefined confidence level during training. As a result, a single SPACR model yields valid prediction intervals at multiple confidence levels during inference, avoiding the costly retraining required by methods like Directly Optimized Inductive Conformal Regression (DOICR). Experiments on diverse tabular and image datasets show that SPACR consistently gives tighter intervals and better coverage-efficiency trade-offs compared to standard CP and DOICR, while significantly reducing computational costs relative to retraining-dependent baselines. The code is available at \url{https://github.com/M-Soundouss/SPACR}.
\end{abstract}

\section{Introduction}
\label{sec:intro}

Quantifying uncertainty is a key challenge in Machine Learning (ML), particularly where incorrect predictions can lead to serious consequences. In healthcare, a wrong diagnosis can harm patients \citep{vazquez2022conformal,seoni2023application}, while in autonomous driving, mistakes can put pedestrians and drivers at risk \citep{michelmore2020uncertainty}. In environmental monitoring, uncertainty-aware methods are vital in natural disasters for faster responses to protect people and property \citep{han2017bayesian,allaire2021novel}. These high-stakes fields increasingly require models to not only produce accurate predictions but also quantify their reliability. This enables experts to step in when needed, thereby creating a collaborative human-AI decision-making framework instead of a blind reliance on overconfident but incorrect predictions.

Conformal Prediction (CP) is a powerful distribution-free framework for generating uncertainty estimates, popular for its theoretical guarantees and model-agnostic nature \citep{vovk2005algorithmic,angelopoulos2021gentle}. Instead of single point predictions, CP produces sets (for classification) or intervals (for regression) that come with statistical guarantees. By selecting a confidence level, users can control the probability that the true value falls within the predicted range, offering a robust and interpretable measure of uncertainty. Unlike baselines such as Deep Ensembles \citep{lakshminarayanan2017simple} or Bayesian deep learning that explicitly model aleatoric and epistemic uncertainty \citep{kendall2017uncertainties}, CP requires no assumptions beyond data exchangeability, making it easily and robustly applicable to diverse ML models.

While CP has demonstrated significant potential, conventional methods treat the model as a ``black box''. Thus, the optimization objective is often misaligned with conformal goals, which is maximizing predictive efficiency (producing the narrowest possible intervals) while maintaining empirical validity (ensuring the target coverage rate is strictly met). This means that conventional CP approaches can lead to unnecessarily wide intervals and less effective uncertainty estimates.

This challenge has motivated the development of methods that directly train models to maximize predictive efficiency. For classification, \cite{colombo2020training} proposed a direct approach to training CP, where the model learns a probabilistic measure of efficiency through optimization. However, this approach involves multiple step functions, which are non-differentiable. To overcome this,~\cite{bellotti2021optimized} suggested approximating the CP goal with a differentiable function that can be optimized using gradient descent.~\cite{Stutz2022ICLR} further improved this work by accounting for the calibration step during training, which plays a key role in reducing inefficiency. The broader shift toward differentiable conformal objectives is further evidenced by recent frameworks exploring the end-to-end optimization of conformal risk control \citep{yeh2026conformal}.

While these differentiable objectives have significantly advanced conformal classification, extending end-to-end conformal training to regression presents distinct structural challenges. Classification CP constructs predictive subsets from a finite, discrete label space, often leveraging standard softmax probabilities. In contrast, conformal regression must construct intervals within a continuous, infinite target space. The primary challenge lies in jointly optimizing a precise continuous point prediction alongside an instance-specific uncertainty scale that dynamically shapes the interval width. This requires a loss formulation that tightly bounds heteroscedastic residuals while maintaining stable gradients, all without relying on the non-differentiable sorting operations typically required to compute empirical quantiles.

It is also important to clarify the data requirements inherent to all end-to-end conformal training frameworks. Whether applied to classification or regression, these training procedures are fundamentally heuristic. They uniformly still require a standard Inductive Conformal Prediction (ICP) post-hoc calibration step to establish distribution-free mathematical guarantees. Thus, the goal of conformal training is not to replace post-hoc calibration, but to learn representations during training that yield highly efficient intervals once that post-hoc calibration is applied.

To our knowledge, \cite{lei2023reliable} is the only work that successfully extends this conformal training paradigm to regression. Their method, Directly Optimized Inductive Conformal Regression (DOICR), minimizes ICP interval width while enforcing validity via a Deep Neural Network (DNN) loss. However, DOICR intensifies the fundamental data requirements of ICP by mandating an \textit{additional} internal batch splitting during training to simulate the calibration process. Furthermore, it incorporates a predefined, fixed $\alpha$ directly into the loss function. As a result, the model must be completely retrained for each different confidence level, which limits its sample efficiency and increases computational costs. 

To address these limitations, we propose a new framework: \emph{Single-Pass Adaptive Conformal Regressor (SPACR)}. Its novelty lies in integrating known components (accuracy, efficiency, and validity) into a unified, differentiable training objective. Although SPACR still needs standard post-hoc calibration to achieve distribution-free marginal coverage, it eliminates the additional training-time batch splitting required by DOICR. Moreover, by decoupling the uncertainty estimate from a fixed confidence level during training, a single trained SPACR model can adaptively generate valid prediction intervals at any confidence level during inference without the costly retraining required by prior end-to-end approaches. To our knowledge, SPACR is the first method to enable such single-pass adaptive conformal regression with these capabilities. Our contributions are:

\begin{itemize}
    \item We introduce SPACR, a differentiable framework for training uncertainty-aware regressors by directly embedding conformal objectives into the loss function.
    \item We propose a unified optimization strategy that simultaneously minimizes point prediction error, penalizes interval width for efficiency, and enforces a validity penalty to ensure reliable coverage.
    \item SPACR mathematically decouples its uncertainty estimate from the target confidence level during training. This eliminates the sample-inefficient batch-splitting required by DOICR, enabling a single model to adaptively generate valid intervals across any confidence level during inference.
    \item We empirically demonstrate that SPACR achieves tighter intervals and superior coverage-efficiency trade-offs compared to standard CP methods and state-of-the-art baselines like DOICR, while significantly reducing computational costs compared to retraining-dependent methods.
\end{itemize}

The remainder of this paper is organized as follows: Section~\ref{sec:related} provides the background on CP and related work in training uncertainty-aware regressors. Section~\ref{sec:SPACR} details the SPACR architecture and its unified optimization objective. Sections~\ref{sec:exp} and \ref{sec:results} outline the experimental setup and evaluate our method against existing baselines. Finally, Section~\ref{sec:conclu} concludes and discusses future work.


\section{Related work}
\label{sec:related}

\subsection{Conformal Prediction for Regression}
Conformal Prediction (CP) \citep{vovk2005algorithmic,angelopoulos2021gentle} is a model-agnostic framework that provides distribution-free uncertainty guarantees. For regression, CP constructs prediction intervals $\hat{C}_{1-\alpha}(x)$ that satisfy marginal coverage : $\mathbb{P}\big(y \in \hat{C}_{1-\alpha}(x)\big) \geq 1 - \alpha,$
where $\alpha \in (0,1)$ is the target error rate. 
To avoid the impractical $\mathcal{O}(n)$ per-prediction cost of Transductive CP, \cite{papadopoulos2008inductive} introduced Inductive CP (ICP) as an efficient alternative that adopts a split-conformal approach to reduce prediction complexity to $\mathcal{O}(1)$, making ICP the practical standard.

ICP uses an exchangeable dataset partitioned into a proper training set $\mathcal{D}_{train} = \{(x_i, y_i)\}_{i=1}^n$ and a calibration set $\mathcal{D}_{cal}=\{(x_{j},y_{j})\}_{j=1}^{n_{cal}}$, where each input $x \in \mathcal{X}$ is associated with a real-valued target $y \in \mathbb{R}$. A regression model $f: \mathcal{X} \to \mathbb{R}$ is trained on $\mathcal{D}_{train}$. Next, standard non-conformity scores $s_j$ are computed on $\mathcal{D}_{cal}$ to measure how much an example deviates from the model's predictions, with:
\begin{equation}\label{eq:alpha_scp}
    s_j = |y_j - f(x_j)|.
\end{equation}
Using the $(1 - \alpha)$-quantile of these scores as the threshold $\hat{q}$, the prediction interval for a new point $x_{test}$ is:
\begin{equation}\label{eq:final_interval_scp}
    \hat{C}_{1-\alpha}(x_{test}) = \Big[f(x_{test}) - \hat{q},\ f(x_{test}) + \hat{q}\Big].
\end{equation}

For a normalized method,~\cite{papadopoulos2011reliable} calculate scores on $\mathcal{D}_{cal}$ as: 
\begin{equation}\label{eq:alpha_ncp}
    s_j = \frac{|y_j - f(x_j)|}{\sigma_j},
\end{equation}
where $\sigma_j$ estimates the instance-specific difficulty. For a new input $x_{test}$, the final prediction interval becomes:
\begin{equation}\label{eq:final_interval_ncp}
    \hat{C}_{1-\alpha}(x_{test}) = \Big[f(x_{test}) - \hat{q}\sigma_{test},\ f(x_{test}) + \hat{q}\sigma_{test}\Big].
\end{equation}

The normalization term $\sigma$ enables adaptive intervals: simple instances with small $\sigma$ get tighter bounds, while challenging cases receive wider intervals. This generalizes early ICP implementations that used fixed-width intervals ($\sigma \equiv 1$), demonstrating how modern approaches improve efficiency without sacrificing coverage.

While ICP guarantees marginal coverage under exchangeability and uses $\sigma$ for efficiency, its required data-splitting reduces effective training data. To address this, particularly for small datasets, end-to-end conformal learning integrates calibration into training to simultaneously optimize accuracy and interval width.

\subsection{Training Conformal Regressors}\label{subsec:doicr}

To overcome ICP limitations, recent methods directly train models to produce well-calibrated uncertainty estimates. This aligns standard training objectives (like point prediction accuracy) with the conformal goals of maximizing predictive efficiency (i.e., minimizing the size of prediction intervals) while ensuring validity. 

\cite{lei2023reliable} presented Directly Optimized Inductive Conformal Regression (DOICR), which explicitly adapts the classification conformal training framework ConfTr proposed by~\cite{Stutz2022ICLR} into the continuous regression domain. While classification methods optimize over discrete label spaces, DOICR translates this objective to regression by minimizing the prediction intervals' average width while maintaining the desired coverage via a specialized Deep Neural Network (DNN) loss function.

DOICR relies on an iterative training procedure where, in each epoch, $\mathcal{D}_{train}$ is randomly split into a primary training set ($\mathcal{D}_1$) and an embedded calibration set ($\mathcal{D}_2$). The DNN outputs both a predicted mean $m(x)$ and a log-uncertainty $u(x)$. For $(x_j, y_j) \in \mathcal{D}_2$, normalized non-conformity scores are computed as $s_j = \frac{|y_j - m(x_j)|}{\exp(u(x_j))}$. The $(1-\alpha)$-quantile $q$ of these defines the conformal threshold, which is integrated into the loss function as:
$$\mathcal{L}_{\text{DOICR}} = \frac{2q}{|\mathcal{D}_1|} \sum_{(x_i, y_i) \in \mathcal{D}_1} \exp(u(x_i)),$$
which combines the quantile $q$ with the sum of $\exp(u(x_i))$ over the primary training set $\mathcal{D}_1$. By back-propagating through this loss, DOICR jointly optimizes predictions and uncertainty estimates. However, the reliance on batch-wise data splitting, which seems less sample efficient, can lead to issues when there is limited access to training data. Also, it requires a fixed significance level $\alpha$ during training through the quantile $q$. Consequently, various $\alpha$ values demand retraining, which increases the computational complexity.


\section{SPACR : Single-Pass Adaptive Conformal Regressor}
\label{sec:SPACR}

Addressing the limitations of prior approaches, we introduce SPACR, a novel framework that integrates conformal goals into the training loop without data splitting and without fixing a significance level $\alpha$ during training. 

Given a proper training set $\mathcal{D}_{train} = \{(x_i, y_i)\}_{i=1}^n$, our regressor $f_\theta: \mathcal{X} \rightarrow \mathbb{R} \times \mathbb{R}$ outputs both a predicted mean $\hat{y}_i$ and a raw uncertainty score $u_i$ for each input $x_i$. The prediction interval during training is then:
$$[\hat{y}_i - \hat{\sigma}_i, \hat{y}_i + \hat{\sigma}_i],$$
where the uncertainty term $\hat{\sigma}_i$ dictates the interval size and is parameterized through an exponential mapping:
$$\hat{\sigma}_i = \exp(u_i) \quad \text{with} \quad u_i \in \mathbb{R}.$$
This mapping ensures positive interval widths, stable gradients, and adaptation to input difficulty (narrow intervals for predictable samples and wide intervals for uncertain ones) implemented end-to-end.

To train SPACR, we propose a custom loss function that jointly optimizes three objectives:
\begin{itemize}
    \item \textbf{Accuracy ($\mathcal{L}_{\text{Accuracy}}$)}: Measured by Mean Absolute Error (MAE) for precise point predictions.
    \item \textbf{Efficiency ($\mathcal{L}_{\text{Efficiency}}$)}: A term that penalizes large uncertainties to encourage narrow intervals.
    \item \textbf{Validity ($\mathcal{L}_{\text{Validity}}$)}: A loss that imposes a penalty when the true value falls outside the predicted interval, thus forcing the model to produce genuine interval predictions rather than point predictions.
\end{itemize}

The overall loss is defined as:
\begin{equation} \label{eq:our_loss}
\mathcal{L}_{\text{SPACR}} = \underbrace{\frac{1}{n} \sum_{i=1}^n |\hat{y}_i - y_i|}_{\mathcal{L}_{\text{Accuracy}}} + \underbrace{\frac{1}{n}\sum_{i=1}^n \hat{\sigma}_i}_{\mathcal{L}_{\text{Efficiency}}} + \lambda \underbrace{\left[\frac{1}{n}\sum_{i=1}^n \phi(y_i, \hat{y}_i, \hat{\sigma}_i)\right]}_{{\mathcal{L}_{\text{Validity}}}},
\end{equation}
where $\lambda > 0$ controls the importance of the validity term in the overall loss, and $\phi(\cdot)$ computes the distance the true target falls outside the predicted bounds, given by $\phi(y, \hat{y}, \hat{\sigma}) = \max(|y - \hat{y}| - \hat{\sigma}, 0).$

Since $\mathcal{L}_{\text{Accuracy}}$ and $\mathcal{L}_{\text{Efficiency}}$ share the same scale (units of $y$), they are weighted equally. This leaves only $\lambda$ to control coverage, restricting the hyperparameter search space and preserving computational efficiency.

\begin{figure}[ht]
\centerline{\includegraphics[trim={1cm 0.86cm 1cm 0.68cm}, clip, width=0.55\textwidth]{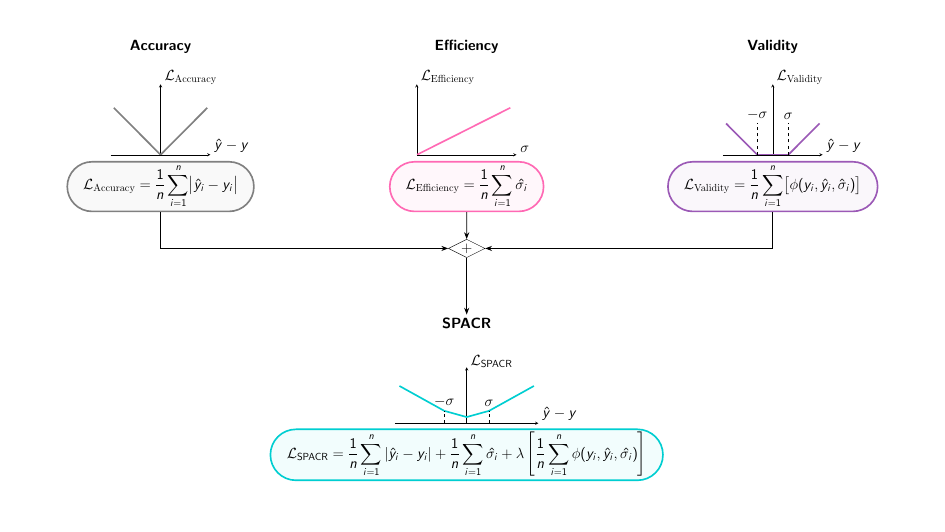}}
\caption{Decomposition of the SPACR loss. Top: the three functions ($\mathcal{L}_{\text{Accuracy}}$, $\mathcal{L}_{\text{Efficiency}}$, and $\mathcal{L}_{\text{Validity}}$). Center: their summation node. Bottom: the resulting overall SPACR loss.} \label{fig:loss_decomp}
\end{figure}

As illustrated in Figure~\ref{fig:loss_decomp}, this loss encourages the model to dynamically adjust interval widths based on empirical coverage violations, which aligns its training objective with the CP validity requirements. Note that $\mathcal{L}_{\text{Validity}}$ is equivalent to the $\alpha$-insensitivity loss used in robust SVMs \citep{vapnik1996support}. 

Similar to other recent conformal training methods such as DOICR and ConfTr, SPACR relies on a post-hoc ICP calibration step to enforce distribution-free validity. Its contribution lies in learning better uncertainty estimates during training that align with conformal objectives. Although SPACR requires a differentiable model and thus departs from the model-agnostic spirit of classical CP, this assumption enables efficient, end-to-end optimization and supports dynamic confidence levels at inference without retraining. This trade-off aligns with practical needs in deep learning, where differentiability is standard.

Algorithm \ref{alg:SPACR_training} outlines the SPACR training procedure. A key distinction of SPACR is that it does not require splitting the proper training set $\mathcal{D}_{train}$ into a primary training set and an embedded calibration set. This design allows us to be more sample-efficient when batch processing.

\let\AND\relax
\begin{algorithm}[t]
\caption{SPACR Training Procedure}
\label{alg:SPACR_training}
\textbf{Input:} Proper training dataset $\mathcal{D}_{train}$, DNN $f_\theta = (f_\theta^{(1)}, f_\theta^{(2)})$, number of epochs $E_{\text{max}}$, learning rate $\eta$\\
\textbf{Output:} Trained conformalized model $f_\theta$
\begin{algorithmic}[1]
\STATE Initialize model parameters $\theta$
\FOR{epoch $= 1$ \TO $E_{\text{max}}$}
    \STATE Shuffle and partition $\mathcal{D}_{train}$ into mini-batches $\mathcal{B}$
    \FOR{each batch $B \in \mathcal{B}$}
        \STATE \textbf{Forward pass:} For each $(x_i, y_i) \in B$, compute predicted mean $\hat{y}_i = f_\theta^{(1)}(x_i)$ and uncertainty $\hat{\sigma}_i = \exp(f_\theta^{(2)}(x_i))$
        \STATE \textbf{Compute loss:} Evaluate $\mathcal{L}_{\text{SPACR}}(B; \theta)$ over the current batch $B$ using Eq.~\ref{eq:our_loss}
        \STATE \textbf{Backward pass:} Compute gradients $\nabla_\theta \mathcal{L}_{\text{SPACR}}$
        \STATE \textbf{Parameter update:} Update $\theta$ using an optimizer (e.g., Adam) with learning rate $\eta$
    \ENDFOR
\ENDFOR
\end{algorithmic}
\end{algorithm}

During inference, SPACR follows the standard post-hoc ICP calibration procedure, identical to traditional conformal methods and training-based techniques such as DOICR. Given a chosen significance level $\alpha$ and a separate calibration set $\mathcal{D}_{cal}=\{(x_{j},y_{j})\}_{j=1}^{n_{cal}}$, the model outputs predictions $\hat{y}_{j}$ and uncertainty estimates $\hat{\sigma}_{j}$. To prevent numerical instability for very small $\hat{\sigma}_{j}$, we compute the non-conformity scores as:

\begin{equation}
s_{j} = \frac{|y_{j} - \hat{y}_{j}|}{\hat{\sigma}_{j} + \beta} \quad \text{for } j=1,\dots,n_{cal}.
\label{eq:nonconformity_scores}
\end{equation}

We set $\beta = 0.05 \mu_{\hat{\sigma}}$ as a smoothing constant, where $\mu_{\hat{\sigma}}$ is the mean calibration uncertainty. For a fair comparison, this applies to all baselines using normalized non-conformity measures.

The conformity threshold $\hat{q}$ is set as the $\lceil (n_{cal}+1)(1-\alpha) \rceil$-th smallest value of these non-conformity scores. For a new input $x_{test}$, the final $(1-\alpha)$-CP interval is given by:

\begin{equation}
\hat{C}_{1-\alpha}(x_{test}) = \Big[\hat{y}_{test} - \hat{q}(\hat{\sigma}_{test} + \beta),\ \hat{y}_{test} + \hat{q}(\hat{\sigma}_{test} + \beta)\Big].
\label{eq:final_interval_our_method}
\end{equation}

It is important to note that the validity term $\mathcal{L}_{\text{Validity}}$ operates on the raw training-time band $[\hat{y}_{i} - \hat{\sigma}_{i}, \hat{y}_{i} + \hat{\sigma}_{i}]$, which is related to, but distinct from, the final calibrated interval $\hat{C}_{1-\alpha}(x_{test})$ defined above. During training, minimizing $\mathcal{L}_{\text{Validity}}$ acts as a heuristic that forces $\hat{\sigma}_{i}$ to dynamically track the empirical residual magnitude $|y_{i} - \hat{y}_{i}|$. By shaping $\hat{\sigma}_{i}$ to encapsulate the raw prediction error, the non-conformity scores $s_{j}$ computed during the post-hoc ICP calibration phase remain stable and concentrated. This prevents the calibration quantile $\hat{q}$ from inflating, ensuring that the theoretical marginal coverage guarantee provided by ICP translates into narrow, highly adaptive intervals in practice.

Another key distinction of SPACR is that it does not require specifying a significance level $\alpha$ during training. The complexity of the calibration step remains identical to that of the standard ICP, but it crucially requires no model updates. This enables efficient exploration of multiple $\alpha$ values using a single trained model, which significantly reduces computational cost compared to methods that necessitate retraining for each desired $\alpha$.


\section{Experimental Setting}
\label{sec:exp}

\subsection{Methodology}

To evaluate SPACR against existing approaches, we adopt a unified experimental setting where all methods share the same underlying neural network architecture, with differences tailored to each one:

\begin{itemize}
    \item \textbf{Standard ICP (SICP)}: A conformal regression baseline trained via Mean Absolute Error (MAE) to predict $\hat{y}_i$, followed by post-hoc calibration to construct constant-width intervals (Eq.~\ref{eq:alpha_scp}, Eq.~\ref{eq:final_interval_scp}).
    
    \item \textbf{Normalized ICP (NICP)}: Extends SICP by scaling uncertainties through absolute error-based difficulty estimation, leading to adaptive intervals (Eq.~\ref{eq:alpha_ncp}, Eq.~\ref{eq:final_interval_ncp}).

    \item \textbf{Conformalized Quantile Regression (CQR)}: ~\cite{romano2019conformalized} trains a model to predict lower and upper quantiles using the pinball loss then ensures validity with conformal calibration.
    
    \item \textbf{DOICR}: The Directly Optimized Inductive Conformal Regression method by~\cite{lei2023reliable}, as described in Subsection~\ref{subsec:doicr}, which incorporates conformal calibration into training.
    
    \item \textbf{SPACR (Ours)}: The proposed Single-Pass Conformal Regressor (Section~\ref{sec:SPACR}) integrates conformal goals directly into training, avoiding batch splitting and a fixed $\alpha$.
\end{itemize}

To show SPACR’s applicability to tabular and image regression, we evaluate our framework using two distinct differentiable architectures. For tabular data, all methods use a residual Multi-Layer Perceptron (MLP) with a 64-dimensional input projection and three residual blocks, each containing two 64-unit fully connected layers. The output structure varies depending on the method: SICP and NICP use a single output neuron for point predictions, while DOICR and SPACR employ two output neurons, with one for the predicted mean and another for the uncertainty estimate. CQR uses two quantile outputs to estimate the lower and upper conditional quantiles at levels $\alpha/2$ and $1-\alpha/2$, respectively, which are subsequently conformalized using the calibration set. Additionally, for NICP, a separate layer models the difficulty estimator using detached absolute residuals as training targets. For the computer vision tasks, the architecture instead consists of a Convolutional Neural Network (CNN), specifically fine-tuning a ResNet18 model pretrained on ImageNet1K.

To ensure a fair and controlled evaluation of the conformal frameworks themselves, we maintain a consistent underlying network hyperparameter configuration across all baselines. All models are trained using the Adam optimizer with learning rates of $\eta = 10^{-4}$ for tabular experiments and $\eta = 10^{-3}$ for image experiments, a batch size of $256$, and a maximum of $200$ epochs. Similarly, for SPACR, the regularization parameter $\lambda$ is fixed to 5 across all datasets. To ensure reproducibility, experiments are repeated across five independent random seeds. The framework is implemented in PyTorch.

\subsection{Datasets}

The experiments were conducted on various regression datasets, primarily tabular data and, to a lesser extent, image data. Each dataset was split into proper training, calibration, and testing sets with a ratio of $60\%$, $20\%$, $20\%$. The tabular datasets were sourced from the benchmark collection introduced by \cite{grinsztajn2022tree} with different characteristics in terms of features and sample sizes. For all tabular datasets, targets are transformed using \(\log(1+y)\) and MAE and interval widths are reported on this transformed scale. For the image-based datasets, we used Drift by \cite{li2024get}, which contains satellite imagery for monitoring trees across five countries. Specifically, we selected 13,094 examples from Denmark to predict the number of trees per image. We also used the UTK Face dataset by \cite{zhang2017age}, consisting of over 20,000 face images, to predict age. A comprehensive summary of all dataset characteristics, including sample sizes and feature dimensions, is detailed in Table~\ref{tab:dataset_info} in the Appendix.

\subsection{Performance Metrics}

To assess each method’s effectiveness, we measure:
\begin{itemize}
    \item \textbf{Accuracy (MAE)}: The Mean Absolute Error between the point predictions and the true target values. This verifies that optimizing for conformal bounds does not degrade the underlying predictive performance of the model.
    \item \textbf{Coverage (\%)}: The percentage of test samples for which the true target value falls within the predicted conformal interval. A well-calibrated method should ideally achieve a coverage rate close to the chosen confidence level $1-\alpha$.
    \item \textbf{Efficiency}: The width of the prediction intervals. Smaller interval widths indicate higher efficiency, meaning the method provides more precise uncertainty estimates. To comprehensively characterize the distribution of interval widths, we report both the median and the mean. We use the median as our primary metric because it is robust to extreme outliers.
\end{itemize}


\section{Results}
\label{sec:results}

Due to space constraints, we present a subset of the experimental results in the main text and provide additional results in the Appendix \ref{sec:added_results}.

\begin{table*}[!htbp]
\caption{MAE and Marginal Coverage (Mean $\pm$ Std) across datasets. Invalid coverage is marked in \textcolor{red}{red}.}
\centering\scriptsize\resizebox{\textwidth}{!}{
\begin{tabular}{lccccccc}
\toprule\textbf{Dataset} & \textbf{Method} & \multicolumn{2}{c}{$\alpha = 0.10$ (90\%)} & \multicolumn{2}{c}{$\alpha = 0.05$ (95\%)} & \multicolumn{2}{c}{$\alpha = 0.01$ (99\%)} \\
\cmidrule(lr){3-4} \cmidrule(lr){5-6} \cmidrule(lr){7-8}
& & MAE $\downarrow$ & Cov. (\%) & MAE $\downarrow$ & Cov. (\%) & MAE $\downarrow$ & Cov. (\%) \\ \midrule
\multirow{5}{*}{\textbf{Bike Sharing}} & SICP & 0.31 $\pm$ 0.01 & 90.13 $\pm$ 0.50 & 0.31 $\pm$ 0.01 & 94.79 $\pm$ 0.38 & 0.31 $\pm$ 0.01 & 99.13 $\pm$ 0.24 \\
 & NICP & 0.31 $\pm$ 0.01 & 90.09 $\pm$ 0.14 & 0.31 $\pm$ 0.01 & 94.89 $\pm$ 0.49 & 0.31 $\pm$ 0.01 & 99.10 $\pm$ 0.27 \\
 & CQR & 0.32 $\pm$ 0.01 & 89.76 $\pm$ 0.22 & 0.32 $\pm$ 0.01 & 94.69 $\pm$ 0.30 & 0.33 $\pm$ 0.01 & 99.00 $\pm$ 0.35 \\
 & DOICR & 0.68 $\pm$ 0.05 & 90.07 $\pm$ 0.48 & 0.66 $\pm$ 0.05 & 95.20 $\pm$ 0.34 & 0.71 $\pm$ 0.05 & 99.03 $\pm$ 0.38 \\
 & \cellcolor{gray!20} SPACR & \cellcolor{gray!20} 0.32 $\pm$ 0.02 & \cellcolor{gray!20} 90.06 $\pm$ 0.52 & \cellcolor{gray!20} 0.32 $\pm$ 0.02 & \cellcolor{gray!20} 95.09 $\pm$ 0.41 & \cellcolor{gray!20} 0.32 $\pm$ 0.02 & \cellcolor{gray!20} 99.02 $\pm$ 0.22 \\
\midrule
\multirow{5}{*}{\textbf{Brazilian Houses}} & SICP & 0.11 $\pm$ 0.01 & 89.93 $\pm$ 0.52 & 0.11 $\pm$ 0.01 & 95.15 $\pm$ 0.28 & 0.11 $\pm$ 0.01 & 99.21 $\pm$ 0.16 \\
 & NICP & 0.11 $\pm$ 0.01 & 89.72 $\pm$ 0.29 & 0.11 $\pm$ 0.01 & 94.98 $\pm$ 0.32 & 0.11 $\pm$ 0.01 & 99.23 $\pm$ 0.19 \\
 & CQR & 0.13 $\pm$ 0.02 & 89.97 $\pm$ 0.46 & 0.14 $\pm$ 0.03 & 94.60 $\pm$ 0.73 & 0.13 $\pm$ 0.03 & 99.23 $\pm$ 0.26 \\
 & DOICR & 0.55 $\pm$ 0.16 & \textcolor{red}{\textbf{71.91 $\pm$ 40.21}} & 0.68 $\pm$ 0.10 & \textcolor{red}{\textbf{75.89 $\pm$ 42.42}} & 0.76 $\pm$ 0.30 & 99.14 $\pm$ 0.12 \\
 & \cellcolor{gray!20} SPACR & \cellcolor{gray!20} 0.34 $\pm$ 0.03 & \cellcolor{gray!20} 89.86 $\pm$ 0.69 & \cellcolor{gray!20} 0.34 $\pm$ 0.03 & \cellcolor{gray!20} 95.00 $\pm$ 0.88 & \cellcolor{gray!20} 0.34 $\pm$ 0.03 & \cellcolor{gray!20} 99.21 $\pm$ 0.22 \\
\midrule
\multirow{5}{*}{\textbf{California}} & SICP & 0.06 $\pm$ 0.00 & 90.41 $\pm$ 0.80 & 0.06 $\pm$ 0.00 & 95.16 $\pm$ 0.62 & 0.06 $\pm$ 0.00 & 99.01 $\pm$ 0.15 \\
 & NICP & 0.06 $\pm$ 0.00 & 90.34 $\pm$ 0.79 & 0.06 $\pm$ 0.00 & 95.08 $\pm$ 0.60 & 0.06 $\pm$ 0.00 & 99.01 $\pm$ 0.20 \\
 & CQR & 0.07 $\pm$ 0.00 & 90.14 $\pm$ 0.92 & 0.07 $\pm$ 0.00 & 95.10 $\pm$ 0.36 & 0.07 $\pm$ 0.00 & 99.08 $\pm$ 0.17 \\
 & DOICR & 0.07 $\pm$ 0.00 & 90.18 $\pm$ 1.16 & 0.07 $\pm$ 0.00 & 95.22 $\pm$ 0.56 & 0.07 $\pm$ 0.00 & 99.02 $\pm$ 0.06 \\
 & \cellcolor{gray!20} SPACR & \cellcolor{gray!20} 0.06 $\pm$ 0.00 & \cellcolor{gray!20} 90.06 $\pm$ 0.94 & \cellcolor{gray!20} 0.06 $\pm$ 0.00 & \cellcolor{gray!20} 95.04 $\pm$ 0.57 & \cellcolor{gray!20} 0.06 $\pm$ 0.00 & \cellcolor{gray!20} 99.02 $\pm$ 0.15 \\
\midrule
\multirow{5}{*}{\textbf{CPU Act}} & SICP & 0.04 $\pm$ 0.00 & 89.75 $\pm$ 1.37 & 0.04 $\pm$ 0.00 & 94.97 $\pm$ 1.22 & 0.04 $\pm$ 0.00 & 99.11 $\pm$ 0.31 \\
 & NICP & 0.04 $\pm$ 0.00 & 89.79 $\pm$ 1.36 & 0.04 $\pm$ 0.00 & 95.25 $\pm$ 0.68 & 0.04 $\pm$ 0.00 & 99.12 $\pm$ 0.23 \\
 & CQR & 0.04 $\pm$ 0.01 & 89.87 $\pm$ 0.94 & 0.04 $\pm$ 0.01 & 95.06 $\pm$ 1.18 & 0.04 $\pm$ 0.01 & 99.28 $\pm$ 0.26 \\
 & DOICR & 0.52 $\pm$ 0.05 & 90.30 $\pm$ 0.47 & 0.48 $\pm$ 0.05 & 95.25 $\pm$ 0.76 & 0.48 $\pm$ 0.09 & 99.02 $\pm$ 0.31 \\
 & \cellcolor{gray!20} SPACR & \cellcolor{gray!20} 0.08 $\pm$ 0.02 & \cellcolor{gray!20} 89.96 $\pm$ 0.52 & \cellcolor{gray!20} 0.08 $\pm$ 0.02 & \cellcolor{gray!20} 94.78 $\pm$ 0.71 & \cellcolor{gray!20} 0.08 $\pm$ 0.02 & \cellcolor{gray!20} 99.13 $\pm$ 0.15 \\
\midrule
\multirow{5}{*}{\textbf{Diamonds}} & SICP & 0.09 $\pm$ 0.00 & 89.99 $\pm$ 0.26 & 0.09 $\pm$ 0.00 & 95.06 $\pm$ 0.15 & 0.09 $\pm$ 0.00 & 99.05 $\pm$ 0.11 \\
 & NICP & 0.09 $\pm$ 0.00 & 89.89 $\pm$ 0.36 & 0.09 $\pm$ 0.00 & 94.91 $\pm$ 0.05 & 0.09 $\pm$ 0.00 & 99.03 $\pm$ 0.10 \\
 & CQR & 0.09 $\pm$ 0.00 & 89.94 $\pm$ 0.23 & 0.09 $\pm$ 0.00 & 94.91 $\pm$ 0.17 & 0.09 $\pm$ 0.00 & 98.98 $\pm$ 0.10 \\
 & DOICR & 0.10 $\pm$ 0.00 & 89.86 $\pm$ 0.27 & 0.10 $\pm$ 0.00 & 94.91 $\pm$ 0.16 & 0.11 $\pm$ 0.00 & 98.94 $\pm$ 0.10 \\
 & \cellcolor{gray!20} SPACR & \cellcolor{gray!20} 0.09 $\pm$ 0.00 & \cellcolor{gray!20} 90.05 $\pm$ 0.36 & \cellcolor{gray!20} 0.09 $\pm$ 0.00 & \cellcolor{gray!20} 95.02 $\pm$ 0.11 & \cellcolor{gray!20} 0.09 $\pm$ 0.00 & \cellcolor{gray!20} 99.00 $\pm$ 0.08 \\
\midrule
\multirow{5}{*}{\textbf{Fifa}} & SICP & 0.07 $\pm$ 0.00 & 90.22 $\pm$ 0.59 & 0.07 $\pm$ 0.00 & 95.15 $\pm$ 0.52 & 0.07 $\pm$ 0.00 & 99.08 $\pm$ 0.15 \\
 & NICP & 0.07 $\pm$ 0.00 & 90.11 $\pm$ 0.46 & 0.07 $\pm$ 0.00 & 95.23 $\pm$ 0.47 & 0.07 $\pm$ 0.00 & 99.04 $\pm$ 0.29 \\
 & CQR & 0.07 $\pm$ 0.00 & 90.07 $\pm$ 0.70 & 0.07 $\pm$ 0.00 & 94.86 $\pm$ 0.52 & 0.07 $\pm$ 0.00 & 99.21 $\pm$ 0.27 \\
 & DOICR & 0.08 $\pm$ 0.00 & 90.10 $\pm$ 0.47 & 0.09 $\pm$ 0.00 & 94.98 $\pm$ 0.52 & 0.09 $\pm$ 0.00 & 99.08 $\pm$ 0.28 \\
 & \cellcolor{gray!20} SPACR & \cellcolor{gray!20} 0.07 $\pm$ 0.00 & \cellcolor{gray!20} 90.27 $\pm$ 0.51 & \cellcolor{gray!20} 0.07 $\pm$ 0.00 & \cellcolor{gray!20} 94.90 $\pm$ 0.57 & \cellcolor{gray!20} 0.07 $\pm$ 0.00 & \cellcolor{gray!20} 98.93 $\pm$ 0.16 \\
\midrule
\multirow{5}{*}{\textbf{House Sales}} & SICP & 0.13 $\pm$ 0.00 & 89.87 $\pm$ 0.74 & 0.13 $\pm$ 0.00 & 94.76 $\pm$ 0.39 & 0.13 $\pm$ 0.00 & 98.85 $\pm$ 0.32 \\
 & NICP & 0.13 $\pm$ 0.00 & 89.65 $\pm$ 0.65 & 0.13 $\pm$ 0.00 & 94.53 $\pm$ 0.51 & 0.13 $\pm$ 0.00 & 98.84 $\pm$ 0.31 \\
 & CQR & 0.13 $\pm$ 0.00 & 89.63 $\pm$ 0.85 & 0.13 $\pm$ 0.00 & 94.66 $\pm$ 0.44 & 0.13 $\pm$ 0.00 & 98.80 $\pm$ 0.36 \\
 & DOICR & 0.90 $\pm$ 0.15 & 90.28 $\pm$ 0.45 & 0.99 $\pm$ 0.19 & 94.99 $\pm$ 0.32 & 0.90 $\pm$ 0.13 & 98.84 $\pm$ 0.08 \\
 & \cellcolor{gray!20} SPACR & \cellcolor{gray!20} 0.14 $\pm$ 0.01 & \cellcolor{gray!20} 89.83 $\pm$ 1.04 & \cellcolor{gray!20} 0.14 $\pm$ 0.01 & \cellcolor{gray!20} 94.74 $\pm$ 0.41 & \cellcolor{gray!20} 0.14 $\pm$ 0.01 & \cellcolor{gray!20} 98.87 $\pm$ 0.31 \\
\midrule
\multirow{5}{*}{\textbf{Isolet}} & SICP & 0.20 $\pm$ 0.01 & 90.29 $\pm$ 1.69 & 0.20 $\pm$ 0.01 & 94.73 $\pm$ 1.45 & 0.20 $\pm$ 0.01 & 99.40 $\pm$ 0.12 \\
 & NICP & 0.20 $\pm$ 0.01 & 89.71 $\pm$ 1.49 & 0.20 $\pm$ 0.01 & 95.06 $\pm$ 1.60 & 0.20 $\pm$ 0.01 & 99.33 $\pm$ 0.11 \\
 & CQR & 0.20 $\pm$ 0.01 & 89.68 $\pm$ 1.51 & 0.20 $\pm$ 0.01 & 94.69 $\pm$ 1.47 & 0.21 $\pm$ 0.01 & 99.00 $\pm$ 0.18 \\
 & DOICR & 0.31 $\pm$ 0.02 & 90.23 $\pm$ 0.89 & 0.31 $\pm$ 0.01 & 94.82 $\pm$ 0.41 & 0.35 $\pm$ 0.02 & 99.03 $\pm$ 0.28 \\
 & \cellcolor{gray!20} SPACR & \cellcolor{gray!20} 0.20 $\pm$ 0.01 & \cellcolor{gray!20} 89.05 $\pm$ 1.26 & \cellcolor{gray!20} 0.20 $\pm$ 0.01 & \cellcolor{gray!20} 94.54 $\pm$ 0.70 & \cellcolor{gray!20} 0.20 $\pm$ 0.01 & \cellcolor{gray!20} 99.13 $\pm$ 0.28 \\
\midrule
\multirow{5}{*}{\textbf{Medical Charges}} & SICP & 0.04 $\pm$ 0.00 & 89.85 $\pm$ 0.19 & 0.04 $\pm$ 0.00 & 94.91 $\pm$ 0.22 & 0.04 $\pm$ 0.00 & 98.99 $\pm$ 0.12 \\
 & NICP & 0.04 $\pm$ 0.00 & 89.95 $\pm$ 0.15 & 0.04 $\pm$ 0.00 & 94.93 $\pm$ 0.19 & 0.04 $\pm$ 0.00 & 99.04 $\pm$ 0.09 \\
 & CQR & 0.04 $\pm$ 0.00 & 89.96 $\pm$ 0.08 & 0.04 $\pm$ 0.00 & 94.96 $\pm$ 0.17 & 0.04 $\pm$ 0.00 & 98.98 $\pm$ 0.11 \\
 & DOICR & 0.05 $\pm$ 0.00 & 90.12 $\pm$ 0.17 & 0.05 $\pm$ 0.00 & 94.92 $\pm$ 0.25 & 0.07 $\pm$ 0.00 & 99.00 $\pm$ 0.05 \\
 & \cellcolor{gray!20} SPACR & \cellcolor{gray!20} 0.04 $\pm$ 0.00 & \cellcolor{gray!20} 89.90 $\pm$ 0.09 & \cellcolor{gray!20} 0.04 $\pm$ 0.00 & \cellcolor{gray!20} 95.00 $\pm$ 0.20 & \cellcolor{gray!20} 0.04 $\pm$ 0.00 & \cellcolor{gray!20} 99.00 $\pm$ 0.07 \\
\midrule
\multirow{5}{*}{\textbf{Pol}} & SICP & 0.10 $\pm$ 0.01 & 90.07 $\pm$ 0.80 & 0.10 $\pm$ 0.01 & 95.12 $\pm$ 0.77 & 0.10 $\pm$ 0.01 & 98.97 $\pm$ 0.15 \\
 & NICP & 0.10 $\pm$ 0.01 & 90.06 $\pm$ 0.83 & 0.10 $\pm$ 0.01 & 95.11 $\pm$ 0.71 & 0.10 $\pm$ 0.01 & 99.03 $\pm$ 0.13 \\
 & CQR & 0.10 $\pm$ 0.01 & 90.28 $\pm$ 0.68 & 0.10 $\pm$ 0.01 & 95.27 $\pm$ 0.23 & 0.10 $\pm$ 0.01 & 98.99 $\pm$ 0.16 \\
 & DOICR & 0.39 $\pm$ 0.04 & 89.39 $\pm$ 0.87 & 0.41 $\pm$ 0.03 & 94.92 $\pm$ 0.19 & 0.62 $\pm$ 0.04 & 98.95 $\pm$ 0.32 \\
 & \cellcolor{gray!20} SPACR & \cellcolor{gray!20} 0.09 $\pm$ 0.01 & \cellcolor{gray!20} 90.37 $\pm$ 0.43 & \cellcolor{gray!20} 0.09 $\pm$ 0.01 & \cellcolor{gray!20} 95.67 $\pm$ 0.37 & \cellcolor{gray!20} 0.09 $\pm$ 0.01 & \cellcolor{gray!20} 99.14 $\pm$ 0.17 \\
\midrule
\multirow{5}{*}{\textbf{Superconduct}} & SICP & 0.35 $\pm$ 0.01 & 89.83 $\pm$ 0.56 & 0.35 $\pm$ 0.01 & 94.88 $\pm$ 0.27 & 0.35 $\pm$ 0.01 & 98.96 $\pm$ 0.28 \\
 & NICP & 0.35 $\pm$ 0.01 & 89.92 $\pm$ 0.61 & 0.35 $\pm$ 0.01 & 94.86 $\pm$ 0.44 & 0.35 $\pm$ 0.01 & 98.94 $\pm$ 0.24 \\
 & CQR & 0.36 $\pm$ 0.01 & 89.56 $\pm$ 0.47 & 0.36 $\pm$ 0.01 & 94.59 $\pm$ 0.23 & 0.36 $\pm$ 0.01 & 98.93 $\pm$ 0.38 \\
 & DOICR & 0.47 $\pm$ 0.01 & 89.74 $\pm$ 0.61 & 0.52 $\pm$ 0.01 & 94.85 $\pm$ 0.34 & 0.63 $\pm$ 0.02 & 98.91 $\pm$ 0.30 \\
 & \cellcolor{gray!20} SPACR & \cellcolor{gray!20} 0.38 $\pm$ 0.01 & \cellcolor{gray!20} 89.79 $\pm$ 0.36 & \cellcolor{gray!20} 0.38 $\pm$ 0.01 & \cellcolor{gray!20} 95.02 $\pm$ 0.27 & \cellcolor{gray!20} 0.38 $\pm$ 0.01 & \cellcolor{gray!20} 98.83 $\pm$ 0.16 \\
\midrule
\multirow{5}{*}{\textbf{Wine Quality}} & SICP & 0.08 $\pm$ 0.00 & 90.46 $\pm$ 1.03 & 0.08 $\pm$ 0.00 & 94.95 $\pm$ 0.64 & 0.08 $\pm$ 0.00 & 99.00 $\pm$ 0.37 \\
 & NICP & 0.08 $\pm$ 0.00 & 91.31 $\pm$ 1.22 & 0.08 $\pm$ 0.00 & 95.32 $\pm$ 0.89 & 0.08 $\pm$ 0.00 & 98.92 $\pm$ 0.26 \\
 & CQR & 0.08 $\pm$ 0.00 & 90.26 $\pm$ 1.06 & 0.08 $\pm$ 0.00 & 94.80 $\pm$ 0.70 & 0.08 $\pm$ 0.00 & 99.11 $\pm$ 0.46 \\
 & DOICR & 0.11 $\pm$ 0.01 & 90.02 $\pm$ 0.75 & 0.10 $\pm$ 0.01 & 95.46 $\pm$ 0.77 & 0.10 $\pm$ 0.01 & 99.18 $\pm$ 0.20 \\
 & \cellcolor{gray!20} SPACR & \cellcolor{gray!20} 0.08 $\pm$ 0.00 & \cellcolor{gray!20} 90.48 $\pm$ 0.85 & \cellcolor{gray!20} 0.08 $\pm$ 0.00 & \cellcolor{gray!20} 95.31 $\pm$ 0.80 & \cellcolor{gray!20} 0.08 $\pm$ 0.00 & \cellcolor{gray!20} 99.02 $\pm$ 0.33 \\
\midrule
\multirow{5}{*}{\textbf{Drift}} & SICP & 6.03 $\pm$ 0.40 & 89.70 $\pm$ 0.71 & 6.03 $\pm$ 0.40 & 94.89 $\pm$ 0.54 & 6.03 $\pm$ 0.40 & 98.97 $\pm$ 0.32 \\
 & NICP & 6.09 $\pm$ 0.85 & 90.31 $\pm$ 0.47 & 6.09 $\pm$ 0.85 & 95.38 $\pm$ 0.67 & 6.09 $\pm$ 0.85 & 99.18 $\pm$ 0.11 \\
 & CQR & 6.32 $\pm$ 1.11 & 89.48 $\pm$ 0.69 & 6.94 $\pm$ 1.22 & 95.08 $\pm$ 0.49 & 7.02 $\pm$ 0.91 & 99.06 $\pm$ 0.33 \\
 & DOICR & 8.17 $\pm$ 2.08 & \textcolor{red}{\textbf{53.65 $\pm$ 48.98}} & 6.45 $\pm$ 0.42 & \textcolor{red}{\textbf{57.04 $\pm$ 52.08}} & 18.82 $\pm$ 16.91 & 99.04 $\pm$ 0.30 \\
 & \cellcolor{gray!20} SPACR & \cellcolor{gray!20} 6.00 $\pm$ 0.36 & \cellcolor{gray!20} 90.03 $\pm$ 1.29 & \cellcolor{gray!20} 6.00 $\pm$ 0.36 & \cellcolor{gray!20} 95.17 $\pm$ 0.71 & \cellcolor{gray!20} 6.00 $\pm$ 0.36 & \cellcolor{gray!20} 99.27 $\pm$ 0.28 \\
\midrule
\multirow{5}{*}{\textbf{UTK Face}} & SICP & 7.09 $\pm$ 0.68 & 89.97 $\pm$ 0.67 & 7.09 $\pm$ 0.68 & 94.98 $\pm$ 0.45 & 7.09 $\pm$ 0.68 & 98.99 $\pm$ 0.23 \\
 & NICP & 7.14 $\pm$ 1.59 & 90.16 $\pm$ 0.35 & 7.14 $\pm$ 1.59 & 94.92 $\pm$ 0.34 & 7.14 $\pm$ 1.59 & 99.08 $\pm$ 0.03 \\
 & CQR & 6.37 $\pm$ 0.88 & 90.01 $\pm$ 0.33 & 8.96 $\pm$ 2.85 & 95.01 $\pm$ 0.43 & 7.07 $\pm$ 1.52 & 99.00 $\pm$ 0.14 \\
 & DOICR & 8.10 $\pm$ 0.52 & 89.46 $\pm$ 0.29 & 8.04 $\pm$ 2.31 & 95.00 $\pm$ 0.22 & 7.05 $\pm$ 0.66 & 98.94 $\pm$ 0.25 \\
 & \cellcolor{gray!20} SPACR & \cellcolor{gray!20} 5.43 $\pm$ 0.18 & \cellcolor{gray!20} 90.13 $\pm$ 0.46 & \cellcolor{gray!20} 5.43 $\pm$ 0.18 & \cellcolor{gray!20} 95.02 $\pm$ 0.30 & \cellcolor{gray!20} 5.43 $\pm$ 0.18 & \cellcolor{gray!20} 98.95 $\pm$ 0.14 \\
\bottomrule\end{tabular}}
\label{tab:main_results_cov_mae}\end{table*}

\begin{table*}[!htbp]
\caption{Median and Mean Efficiency (Mean $\pm$ Std) across datasets. Best results are highlighted in bold.}
\centering\scriptsize\resizebox{\textwidth}{!}{
\begin{tabular}{lccccccc}
\toprule\textbf{Dataset} & \textbf{Method} & \multicolumn{2}{c}{$\alpha = 0.10$ (90\%)} & \multicolumn{2}{c}{$\alpha = 0.05$ (95\%)} & \multicolumn{2}{c}{$\alpha = 0.01$ (99\%)} \\
\cmidrule(lr){3-4} \cmidrule(lr){5-6} \cmidrule(lr){7-8}
& & Med. Eff. $\downarrow$ & Mean Eff. $\downarrow$ & Med. Eff. $\downarrow$ & Mean Eff. $\downarrow$ & Med. Eff. $\downarrow$ & Mean Eff. $\downarrow$ \\ \midrule
\multirow{5}{*}{\textbf{Bike Sharing}} & SICP & 1.36 $\pm$ 0.02 & 1.36 $\pm$ 0.02 & 1.68 $\pm$ 0.02 & 1.68 $\pm$ 0.02 & 2.57 $\pm$ 0.09 & 2.57 $\pm$ 0.09 \\
 & NICP & \textbf{1.29 $\pm$ 0.04} & \textbf{1.34 $\pm$ 0.04} & 1.55 $\pm$ 0.05 & 1.61 $\pm$ 0.06 & 2.20 $\pm$ 0.07 & 2.29 $\pm$ 0.07 \\
 & CQR & 1.39 $\pm$ 0.06 & 1.46 $\pm$ 0.06 & 1.74 $\pm$ 0.05 & 1.83 $\pm$ 0.05 & 3.09 $\pm$ 0.08 & 3.10 $\pm$ 0.07 \\
 & DOICR & 2.48 $\pm$ 0.22 & 2.66 $\pm$ 0.19 & 2.68 $\pm$ 0.22 & 2.88 $\pm$ 0.19 & 3.30 $\pm$ 0.23 & 3.48 $\pm$ 0.19 \\
 & \cellcolor{gray!20} SPACR & \cellcolor{gray!20} \textbf{1.29 $\pm$ 0.09} & \cellcolor{gray!20} 1.35 $\pm$ 0.08 & \cellcolor{gray!20} \textbf{1.54 $\pm$ 0.11} & \cellcolor{gray!20} \textbf{1.60 $\pm$ 0.10} & \cellcolor{gray!20} \textbf{2.07 $\pm$ 0.09} & \cellcolor{gray!20} \textbf{2.16 $\pm$ 0.08} \\
\midrule
\multirow{5}{*}{\textbf{Brazilian Houses}} & SICP & 0.46 $\pm$ 0.04 & \textbf{0.46 $\pm$ 0.04} & \textbf{0.59 $\pm$ 0.04} & \textbf{0.59 $\pm$ 0.04} & 0.94 $\pm$ 0.08 & \textbf{0.94 $\pm$ 0.08} \\
 & NICP & \textbf{0.45 $\pm$ 0.05} & 0.47 $\pm$ 0.05 & \textbf{0.59 $\pm$ 0.06} & 0.61 $\pm$ 0.06 & \textbf{0.93 $\pm$ 0.08} & 0.96 $\pm$ 0.07 \\
 & CQR & 0.51 $\pm$ 0.07 & 0.53 $\pm$ 0.06 & 0.66 $\pm$ 0.08 & 0.69 $\pm$ 0.07 & 1.23 $\pm$ 0.15 & 1.30 $\pm$ 0.14 \\
 & DOICR & 1.53 $\pm$ 0.29 & 1.54 $\pm$ 0.29 & 1.87 $\pm$ 0.12 & 1.90 $\pm$ 0.13 & 2.47 $\pm$ 0.70 & 2.53 $\pm$ 0.70 \\
 & \cellcolor{gray!20} SPACR & \cellcolor{gray!20} 1.07 $\pm$ 0.12 & \cellcolor{gray!20} 818.78 $\pm$ 1828.32 & \cellcolor{gray!20} 1.22 $\pm$ 0.15 & \cellcolor{gray!20} 918.38 $\pm$ 2050.66 & \cellcolor{gray!20} 1.59 $\pm$ 0.18 & \cellcolor{gray!20} 1203.90 $\pm$ 2688.22 \\
\midrule
\multirow{5}{*}{\textbf{California}} & SICP & 0.28 $\pm$ 0.01 & 0.28 $\pm$ 0.01 & 0.36 $\pm$ 0.01 & 0.36 $\pm$ 0.01 & 0.56 $\pm$ 0.01 & 0.56 $\pm$ 0.01 \\
 & NICP & 0.28 $\pm$ 0.01 & 31.31 $\pm$ 69.38 & 0.35 $\pm$ 0.01 & 39.75 $\pm$ 88.08 & 0.56 $\pm$ 0.02 & 62.74 $\pm$ 139.03 \\
 & CQR & 0.30 $\pm$ 0.01 & 0.30 $\pm$ 0.01 & 0.38 $\pm$ 0.01 & 0.38 $\pm$ 0.01 & 0.61 $\pm$ 0.01 & 0.60 $\pm$ 0.01 \\
 & DOICR & 0.28 $\pm$ 0.02 & 0.28 $\pm$ 0.01 & 0.34 $\pm$ 0.01 & 0.34 $\pm$ 0.01 & \textbf{0.47 $\pm$ 0.01} & \textbf{0.48 $\pm$ 0.01} \\
 & \cellcolor{gray!20} SPACR & \cellcolor{gray!20} \textbf{0.26 $\pm$ 0.01} & \cellcolor{gray!20} \textbf{0.26 $\pm$ 0.01} & \cellcolor{gray!20} \textbf{0.33 $\pm$ 0.01} & \cellcolor{gray!20} \textbf{0.32 $\pm$ 0.01} & \cellcolor{gray!20} 0.53 $\pm$ 0.03 & \cellcolor{gray!20} 0.52 $\pm$ 0.03 \\
\midrule
\multirow{5}{*}{\textbf{CPU Act}} & SICP & \textbf{0.15 $\pm$ 0.01} & \textbf{0.15 $\pm$ 0.01} & \textbf{0.21 $\pm$ 0.02} & \textbf{0.21 $\pm$ 0.02} & 0.58 $\pm$ 0.06 & 0.58 $\pm$ 0.06 \\
 & NICP & \textbf{0.15 $\pm$ 0.01} & 0.16 $\pm$ 0.01 & 0.22 $\pm$ 0.02 & 0.23 $\pm$ 0.02 & 0.66 $\pm$ 0.17 & 0.71 $\pm$ 0.22 \\
 & CQR & 0.16 $\pm$ 0.01 & 0.23 $\pm$ 0.02 & 0.28 $\pm$ 0.04 & 0.37 $\pm$ 0.04 & 0.78 $\pm$ 0.11 & 0.86 $\pm$ 0.11 \\
 & DOICR & 0.97 $\pm$ 0.09 & 1.04 $\pm$ 0.12 & 1.12 $\pm$ 0.08 & 1.29 $\pm$ 0.10 & 1.28 $\pm$ 0.15 & 1.52 $\pm$ 0.17 \\
 & \cellcolor{gray!20} SPACR & \cellcolor{gray!20} 0.24 $\pm$ 0.05 & \cellcolor{gray!20} 0.28 $\pm$ 0.06 & \cellcolor{gray!20} 0.27 $\pm$ 0.05 & \cellcolor{gray!20} 0.32 $\pm$ 0.06 & \cellcolor{gray!20} \textbf{0.37 $\pm$ 0.05} & \cellcolor{gray!20} \textbf{0.44 $\pm$ 0.06} \\
\midrule
\multirow{5}{*}{\textbf{Diamonds}} & SICP & 0.38 $\pm$ 0.01 & 0.38 $\pm$ 0.01 & 0.47 $\pm$ 0.00 & 0.47 $\pm$ 0.00 & 0.69 $\pm$ 0.01 & 0.69 $\pm$ 0.01 \\
 & NICP & 0.37 $\pm$ 0.01 & 0.38 $\pm$ 0.01 & 0.45 $\pm$ 0.00 & 0.46 $\pm$ 0.00 & 0.67 $\pm$ 0.01 & 0.68 $\pm$ 0.01 \\
 & CQR & 0.38 $\pm$ 0.01 & 0.39 $\pm$ 0.01 & 0.47 $\pm$ 0.01 & 0.49 $\pm$ 0.01 & 0.83 $\pm$ 0.02 & 0.84 $\pm$ 0.02 \\
 & DOICR & 0.40 $\pm$ 0.01 & 0.40 $\pm$ 0.01 & 0.46 $\pm$ 0.01 & 0.46 $\pm$ 0.01 & 0.61 $\pm$ 0.01 & 0.62 $\pm$ 0.01 \\
 & \cellcolor{gray!20} SPACR & \cellcolor{gray!20} \textbf{0.34 $\pm$ 0.00} & \cellcolor{gray!20} \textbf{0.35 $\pm$ 0.00} & \cellcolor{gray!20} \textbf{0.42 $\pm$ 0.00} & \cellcolor{gray!20} \textbf{0.42 $\pm$ 0.00} & \cellcolor{gray!20} \textbf{0.60 $\pm$ 0.01} & \cellcolor{gray!20} \textbf{0.61 $\pm$ 0.01} \\
\midrule
\multirow{5}{*}{\textbf{Fifa}} & SICP & 0.30 $\pm$ 0.01 & 0.30 $\pm$ 0.01 & 0.43 $\pm$ 0.01 & 0.43 $\pm$ 0.01 & 0.65 $\pm$ 0.02 & 0.65 $\pm$ 0.02 \\
 & NICP & 0.30 $\pm$ 0.01 & 0.30 $\pm$ 0.01 & 0.44 $\pm$ 0.01 & 0.44 $\pm$ 0.02 & 0.63 $\pm$ 0.02 & 0.64 $\pm$ 0.02 \\
 & CQR & 0.33 $\pm$ 0.00 & 0.32 $\pm$ 0.00 & \textbf{0.36 $\pm$ 0.00} & 0.37 $\pm$ 0.00 & \textbf{0.43 $\pm$ 0.01} & 0.46 $\pm$ 0.01 \\
 & DOICR & 0.32 $\pm$ 0.01 & 0.33 $\pm$ 0.01 & 0.38 $\pm$ 0.01 & 0.39 $\pm$ 0.01 & 0.44 $\pm$ 0.02 & \textbf{0.45 $\pm$ 0.01} \\
 & \cellcolor{gray!20} SPACR & \cellcolor{gray!20} \textbf{0.28 $\pm$ 0.00} & \cellcolor{gray!20} \textbf{0.29 $\pm$ 0.00} & \cellcolor{gray!20} \textbf{0.36 $\pm$ 0.01} & \cellcolor{gray!20} \textbf{0.36 $\pm$ 0.00} & \cellcolor{gray!20} 0.61 $\pm$ 0.04 & \cellcolor{gray!20} 0.62 $\pm$ 0.04 \\
\midrule
\multirow{5}{*}{\textbf{House Sales}} & SICP & 0.56 $\pm$ 0.01 & 0.56 $\pm$ 0.01 & 0.72 $\pm$ 0.02 & 0.72 $\pm$ 0.02 & 1.13 $\pm$ 0.06 & 1.13 $\pm$ 0.06 \\
 & NICP & \textbf{0.54 $\pm$ 0.02} & \textbf{0.55 $\pm$ 0.02} & \textbf{0.68 $\pm$ 0.02} & \textbf{0.69 $\pm$ 0.02} & 1.04 $\pm$ 0.03 & 1.06 $\pm$ 0.03 \\
 & CQR & \textbf{0.54 $\pm$ 0.03} & 0.56 $\pm$ 0.03 & 0.69 $\pm$ 0.03 & 0.70 $\pm$ 0.03 & 1.13 $\pm$ 0.05 & 1.14 $\pm$ 0.05 \\
 & DOICR & 2.30 $\pm$ 0.28 & 2.31 $\pm$ 0.29 & 2.56 $\pm$ 0.37 & 2.60 $\pm$ 0.40 & 2.64 $\pm$ 0.28 & 2.70 $\pm$ 0.28 \\
 & \cellcolor{gray!20} SPACR & \cellcolor{gray!20} 0.55 $\pm$ 0.02 & \cellcolor{gray!20} 0.57 $\pm$ 0.02 & \cellcolor{gray!20} \textbf{0.68 $\pm$ 0.03} & \cellcolor{gray!20} 0.70 $\pm$ 0.02 & \cellcolor{gray!20} \textbf{1.00 $\pm$ 0.03} & \cellcolor{gray!20} \textbf{1.03 $\pm$ 0.04} \\
\midrule
\multirow{5}{*}{\textbf{Isolet}} & SICP & 0.88 $\pm$ 0.03 & 0.88 $\pm$ 0.03 & 1.19 $\pm$ 0.08 & 1.19 $\pm$ 0.08 & 2.33 $\pm$ 0.22 & 2.33 $\pm$ 0.22 \\
 & NICP & 0.87 $\pm$ 0.03 & 0.87 $\pm$ 0.03 & 1.20 $\pm$ 0.11 & 1.20 $\pm$ 0.10 & 2.23 $\pm$ 0.17 & 2.24 $\pm$ 0.17 \\
 & CQR & 0.87 $\pm$ 0.03 & 0.90 $\pm$ 0.03 & 1.15 $\pm$ 0.03 & 1.19 $\pm$ 0.03 & 2.01 $\pm$ 0.04 & 2.05 $\pm$ 0.05 \\
 & DOICR & 1.13 $\pm$ 0.04 & 1.18 $\pm$ 0.03 & 1.29 $\pm$ 0.04 & 1.36 $\pm$ 0.02 & 1.70 $\pm$ 0.09 & 1.79 $\pm$ 0.10 \\
 & \cellcolor{gray!20} SPACR & \cellcolor{gray!20} \textbf{0.78 $\pm$ 0.06} & \cellcolor{gray!20} \textbf{0.85 $\pm$ 0.07} & \cellcolor{gray!20} \textbf{1.01 $\pm$ 0.07} & \cellcolor{gray!20} \textbf{1.09 $\pm$ 0.07} & \cellcolor{gray!20} \textbf{1.61 $\pm$ 0.11} & \cellcolor{gray!20} \textbf{1.75 $\pm$ 0.11} \\
\midrule
\multirow{5}{*}{\textbf{Medical Charges}} & SICP & 0.18 $\pm$ 0.00 & 0.18 $\pm$ 0.00 & 0.27 $\pm$ 0.01 & 0.27 $\pm$ 0.01 & 0.59 $\pm$ 0.02 & 0.59 $\pm$ 0.02 \\
 & NICP & 0.18 $\pm$ 0.00 & 0.18 $\pm$ 0.00 & 0.26 $\pm$ 0.01 & 0.26 $\pm$ 0.01 & 0.55 $\pm$ 0.02 & 0.56 $\pm$ 0.02 \\
 & CQR & 0.19 $\pm$ 0.01 & 0.20 $\pm$ 0.01 & 0.29 $\pm$ 0.01 & 0.31 $\pm$ 0.01 & 0.80 $\pm$ 0.04 & 0.83 $\pm$ 0.04 \\
 & DOICR & 0.21 $\pm$ 0.01 & 0.22 $\pm$ 0.01 & 0.26 $\pm$ 0.01 & 0.27 $\pm$ 0.01 & \textbf{0.37 $\pm$ 0.01} & \textbf{0.39 $\pm$ 0.01} \\
 & \cellcolor{gray!20} SPACR & \cellcolor{gray!20} \textbf{0.17 $\pm$ 0.00} & \cellcolor{gray!20} \textbf{0.17 $\pm$ 0.00} & \cellcolor{gray!20} \textbf{0.24 $\pm$ 0.00} & \cellcolor{gray!20} \textbf{0.24 $\pm$ 0.00} & \cellcolor{gray!20} 0.50 $\pm$ 0.01 & \cellcolor{gray!20} 0.50 $\pm$ 0.02 \\
\midrule
\multirow{5}{*}{\textbf{Pol}} & SICP & 0.35 $\pm$ 0.02 & 0.35 $\pm$ 0.02 & 0.94 $\pm$ 0.08 & 0.94 $\pm$ 0.08 & 3.44 $\pm$ 0.34 & 3.44 $\pm$ 0.34 \\
 & NICP & 0.34 $\pm$ 0.02 & \textbf{0.34 $\pm$ 0.02} & 0.91 $\pm$ 0.07 & 0.91 $\pm$ 0.07 & 3.33 $\pm$ 0.27 & 3.32 $\pm$ 0.27 \\
 & CQR & 0.49 $\pm$ 0.08 & 0.57 $\pm$ 0.04 & 1.32 $\pm$ 0.19 & 1.33 $\pm$ 0.23 & 2.48 $\pm$ 0.14 & 2.95 $\pm$ 0.09 \\
 & DOICR & 0.79 $\pm$ 0.15 & 1.00 $\pm$ 0.10 & 1.20 $\pm$ 0.12 & 1.35 $\pm$ 0.08 & 1.72 $\pm$ 0.14 & 1.96 $\pm$ 0.13 \\
 & \cellcolor{gray!20} SPACR & \cellcolor{gray!20} \textbf{0.09 $\pm$ 0.03} & \cellcolor{gray!20} \textbf{0.34 $\pm$ 0.05} & \cellcolor{gray!20} \textbf{0.14 $\pm$ 0.04} & \cellcolor{gray!20} \textbf{0.54 $\pm$ 0.10} & \cellcolor{gray!20} \textbf{0.33 $\pm$ 0.08} & \cellcolor{gray!20} \textbf{1.23 $\pm$ 0.15} \\
\midrule
\multirow{5}{*}{\textbf{Superconduct}} & SICP & 1.66 $\pm$ 0.04 & 1.66 $\pm$ 0.04 & 2.24 $\pm$ 0.06 & 2.24 $\pm$ 0.06 & 3.65 $\pm$ 0.18 & 3.65 $\pm$ 0.18 \\
 & NICP & 1.52 $\pm$ 0.07 & 1.60 $\pm$ 0.06 & 2.06 $\pm$ 0.09 & 2.16 $\pm$ 0.08 & 3.50 $\pm$ 0.15 & 3.69 $\pm$ 0.14 \\
 & CQR & 1.59 $\pm$ 0.02 & 1.64 $\pm$ 0.02 & 2.02 $\pm$ 0.02 & 2.08 $\pm$ 0.01 & 3.26 $\pm$ 0.11 & 3.19 $\pm$ 0.08 \\
 & DOICR & 1.55 $\pm$ 0.05 & 1.61 $\pm$ 0.05 & 1.85 $\pm$ 0.04 & 1.93 $\pm$ 0.04 & \textbf{2.60 $\pm$ 0.09} & \textbf{2.63 $\pm$ 0.09} \\
 & \cellcolor{gray!20} SPACR & \cellcolor{gray!20} \textbf{1.29 $\pm$ 0.04} & \cellcolor{gray!20} \textbf{1.36 $\pm$ 0.03} & \cellcolor{gray!20} \textbf{1.69 $\pm$ 0.05} & \cellcolor{gray!20} \textbf{1.78 $\pm$ 0.05} & \cellcolor{gray!20} 3.03 $\pm$ 0.11 & \cellcolor{gray!20} 3.19 $\pm$ 0.14 \\
\midrule
\multirow{5}{*}{\textbf{Wine Quality}} & SICP & \textbf{0.34 $\pm$ 0.01} & \textbf{0.34 $\pm$ 0.01} & \textbf{0.43 $\pm$ 0.02} & \textbf{0.43 $\pm$ 0.02} & \textbf{0.65 $\pm$ 0.06} & \textbf{0.65 $\pm$ 0.06} \\
 & NICP & 0.36 $\pm$ 0.02 & 0.37 $\pm$ 0.03 & 0.44 $\pm$ 0.03 & 0.45 $\pm$ 0.04 & 0.67 $\pm$ 0.05 & 0.69 $\pm$ 0.07 \\
 & CQR & 0.35 $\pm$ 0.01 & 0.36 $\pm$ 0.01 & 0.46 $\pm$ 0.02 & 0.47 $\pm$ 0.02 & 0.77 $\pm$ 0.05 & 0.79 $\pm$ 0.05 \\
 & DOICR & 0.43 $\pm$ 0.04 & 0.43 $\pm$ 0.04 & 0.50 $\pm$ 0.04 & 0.51 $\pm$ 0.04 & 0.68 $\pm$ 0.04 & 0.69 $\pm$ 0.04 \\
 & \cellcolor{gray!20} SPACR & \cellcolor{gray!20} 0.35 $\pm$ 0.01 & \cellcolor{gray!20} 0.35 $\pm$ 0.01 & \cellcolor{gray!20} \textbf{0.43 $\pm$ 0.01} & \cellcolor{gray!20} \textbf{0.43 $\pm$ 0.01} & \cellcolor{gray!20} \textbf{0.65 $\pm$ 0.05} & \cellcolor{gray!20} \textbf{0.65 $\pm$ 0.05} \\
\midrule
\multirow{5}{*}{\textbf{Drift}} & SICP & 26.41 $\pm$ 2.73 & 26.41 $\pm$ 2.73 & 35.33 $\pm$ 3.64 & 35.33 $\pm$ 3.64 & 63.70 $\pm$ 9.28 & 63.70 $\pm$ 9.28 \\
 & NICP & 23.55 $\pm$ 3.32 & 25.85 $\pm$ 3.12 & 29.35 $\pm$ 4.74 & 32.20 $\pm$ 4.43 & 44.13 $\pm$ 10.32 & 48.24 $\pm$ 9.23 \\
 & CQR & 23.94 $\pm$ 3.59 & 24.88 $\pm$ 3.75 & 33.06 $\pm$ 6.29 & 34.50 $\pm$ 6.12 & 66.99 $\pm$ 19.81 & 69.61 $\pm$ 19.82 \\
 & DOICR & 29.70 $\pm$ 6.46 & 31.52 $\pm$ 7.03 & 29.56 $\pm$ 1.34 & 31.29 $\pm$ 1.96 & 98.84 $\pm$ 89.02 & 100.07 $\pm$ 84.18 \\
 & \cellcolor{gray!20} SPACR & \cellcolor{gray!20} \textbf{22.51 $\pm$ 2.28} & \cellcolor{gray!20} \textbf{24.72 $\pm$ 2.46} & \cellcolor{gray!20} \textbf{27.59 $\pm$ 2.98} & \cellcolor{gray!20} \textbf{30.31 $\pm$ 3.34} & \cellcolor{gray!20} \textbf{41.19 $\pm$ 5.90} & \cellcolor{gray!20} \textbf{45.28 $\pm$ 6.74} \\
\midrule
\multirow{5}{*}{\textbf{UTK Face}} & SICP & 32.15 $\pm$ 4.12 & 32.15 $\pm$ 4.12 & 40.67 $\pm$ 5.42 & 40.67 $\pm$ 5.42 & 60.17 $\pm$ 6.94 & 60.17 $\pm$ 6.94 \\
 & NICP & 30.36 $\pm$ 9.72 & 34.98 $\pm$ 12.32 & 38.97 $\pm$ 13.57 & 45.17 $\pm$ 17.84 & 58.48 $\pm$ 16.74 & 69.21 $\pm$ 30.68 \\
 & CQR & 26.29 $\pm$ 3.96 & 26.96 $\pm$ 4.18 & 41.82 $\pm$ 8.57 & 42.82 $\pm$ 8.43 & 55.22 $\pm$ 6.05 & 56.69 $\pm$ 6.33 \\
 & DOICR & 31.67 $\pm$ 2.57 & 31.64 $\pm$ 3.55 & 38.38 $\pm$ 6.59 & 37.11 $\pm$ 6.54 & 47.42 $\pm$ 5.11 & 48.37 $\pm$ 3.85 \\
 & \cellcolor{gray!20} SPACR & \cellcolor{gray!20} \textbf{21.84 $\pm$ 0.84} & \cellcolor{gray!20} \textbf{22.54 $\pm$ 0.43} & \cellcolor{gray!20} \textbf{27.49 $\pm$ 0.98} & \cellcolor{gray!20} \textbf{28.39 $\pm$ 0.91} & \cellcolor{gray!20} \textbf{42.51 $\pm$ 2.06} & \cellcolor{gray!20} \textbf{43.93 $\pm$ 2.76} \\
\bottomrule\end{tabular}}
\label{tab:main_results_eff}\end{table*}

\subsection{Performance (Accuracy, Coverage, and Efficiency)}

\begin{figure*}[ht]
  \centering
  \begin{subfigure}[t]{0.32\textwidth}
    \includegraphics[width=\textwidth]{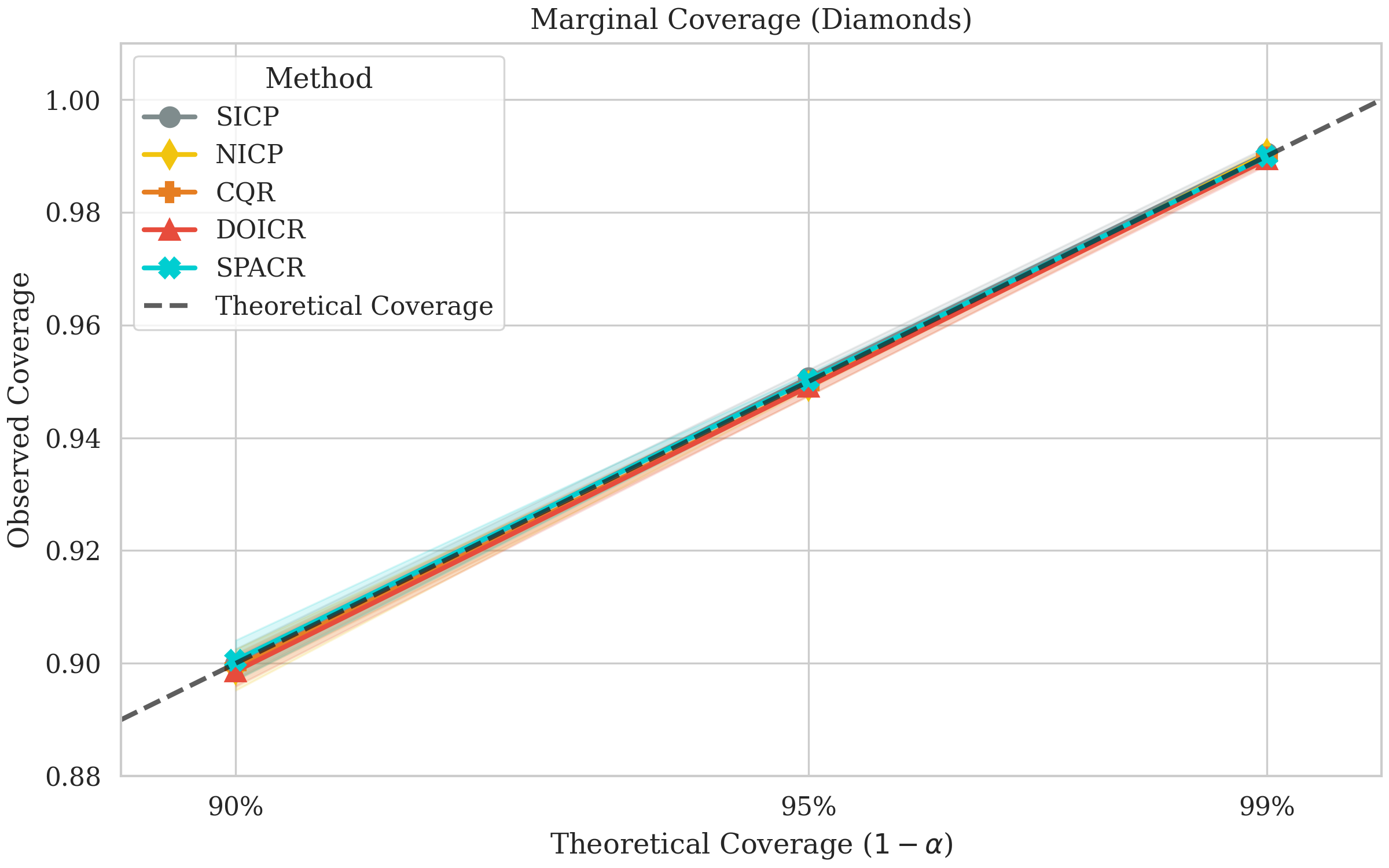}
    \caption{Marginal coverage vs. target confidence.}
    \label{fig:coverage_methods_Diamonds}
  \end{subfigure}
  \hfill
  \begin{subfigure}[t]{0.32\textwidth}
    \includegraphics[width=\textwidth]{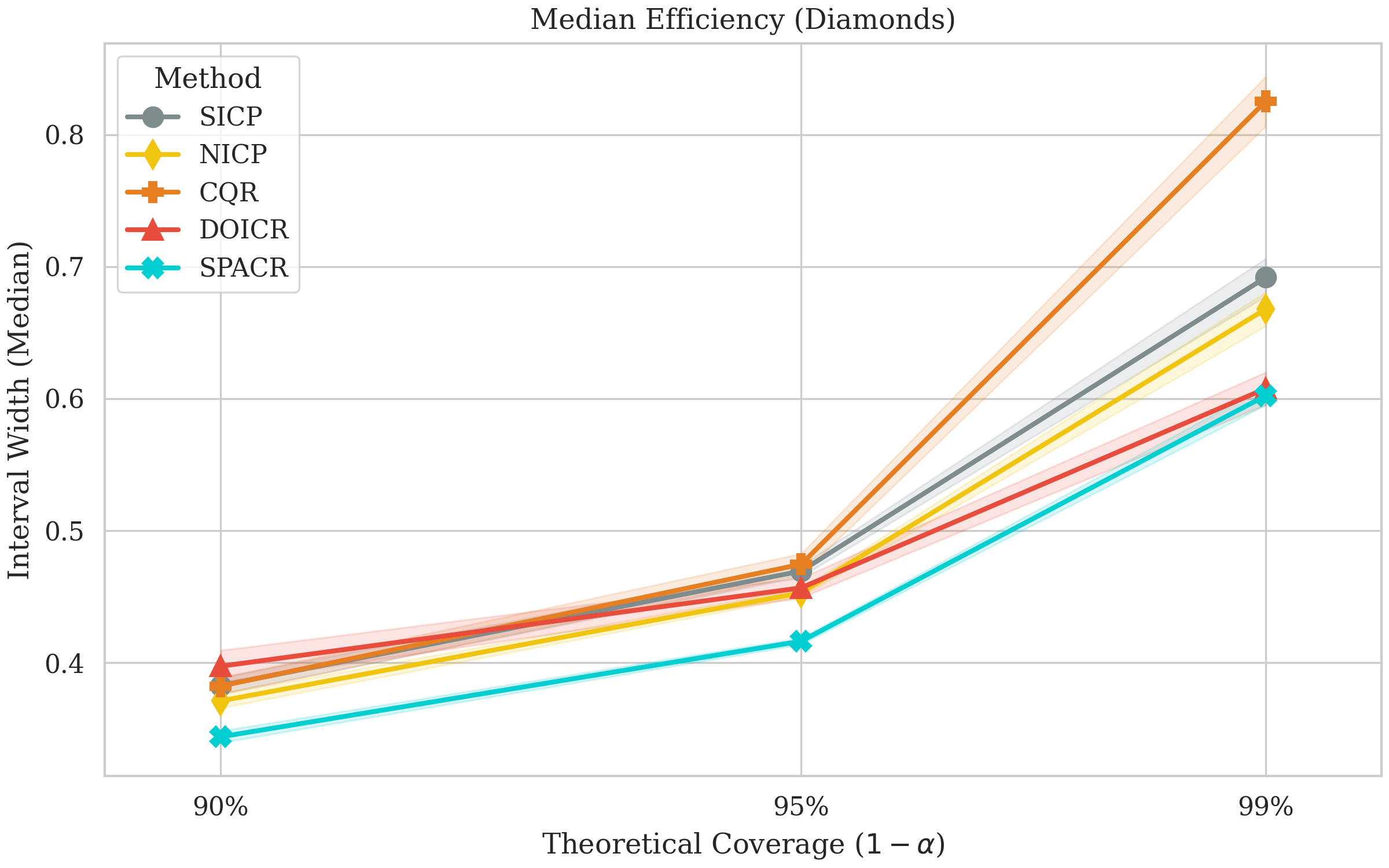}
    \caption{Efficiency (median interval width).}
    \label{fig:efficiency_methods_Diamonds}
  \end{subfigure}
  \hfill
  \begin{subfigure}[t]{0.32\textwidth}
    \includegraphics[width=\textwidth]{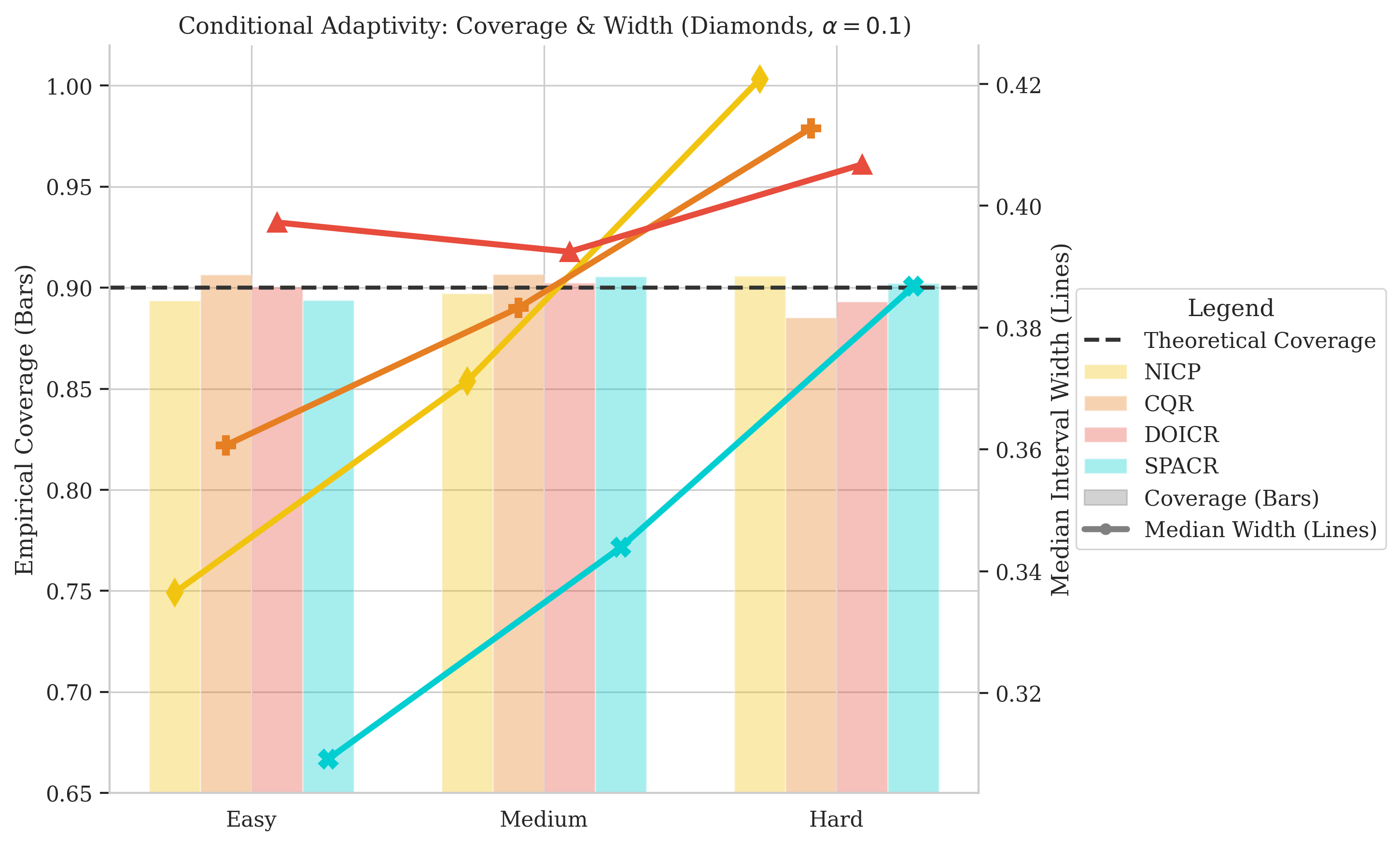}
    \caption{Conditional coverage (bars) and efficiency (lines) by internal difficulty.}
    \label{fig:iqr_methods_Diamonds}
  \end{subfigure}
  \caption{Performance of conformal regression methods on the Diamonds dataset.}
  \label{fig:methods_figures_Diamonds}
\end{figure*}

\begin{figure*}[ht]
  \centering
  \begin{subfigure}[t]{0.32\textwidth}
    \includegraphics[width=\textwidth]{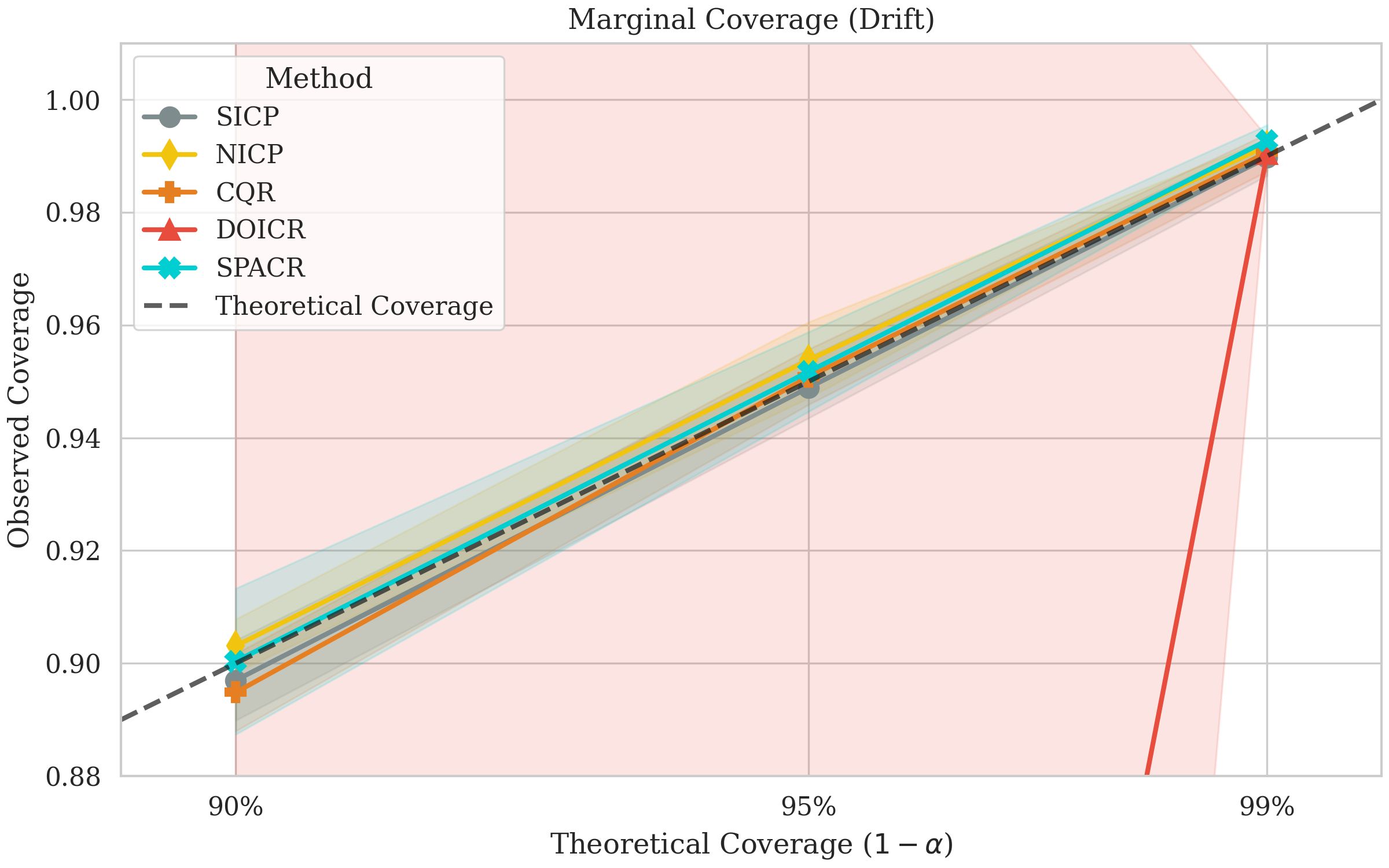}
    \caption{Marginal coverage vs. target confidence.}
    \label{fig:coverage_methods_Drift}
  \end{subfigure}
  \hfill
  \begin{subfigure}[t]{0.32\textwidth}
    \includegraphics[width=\textwidth]{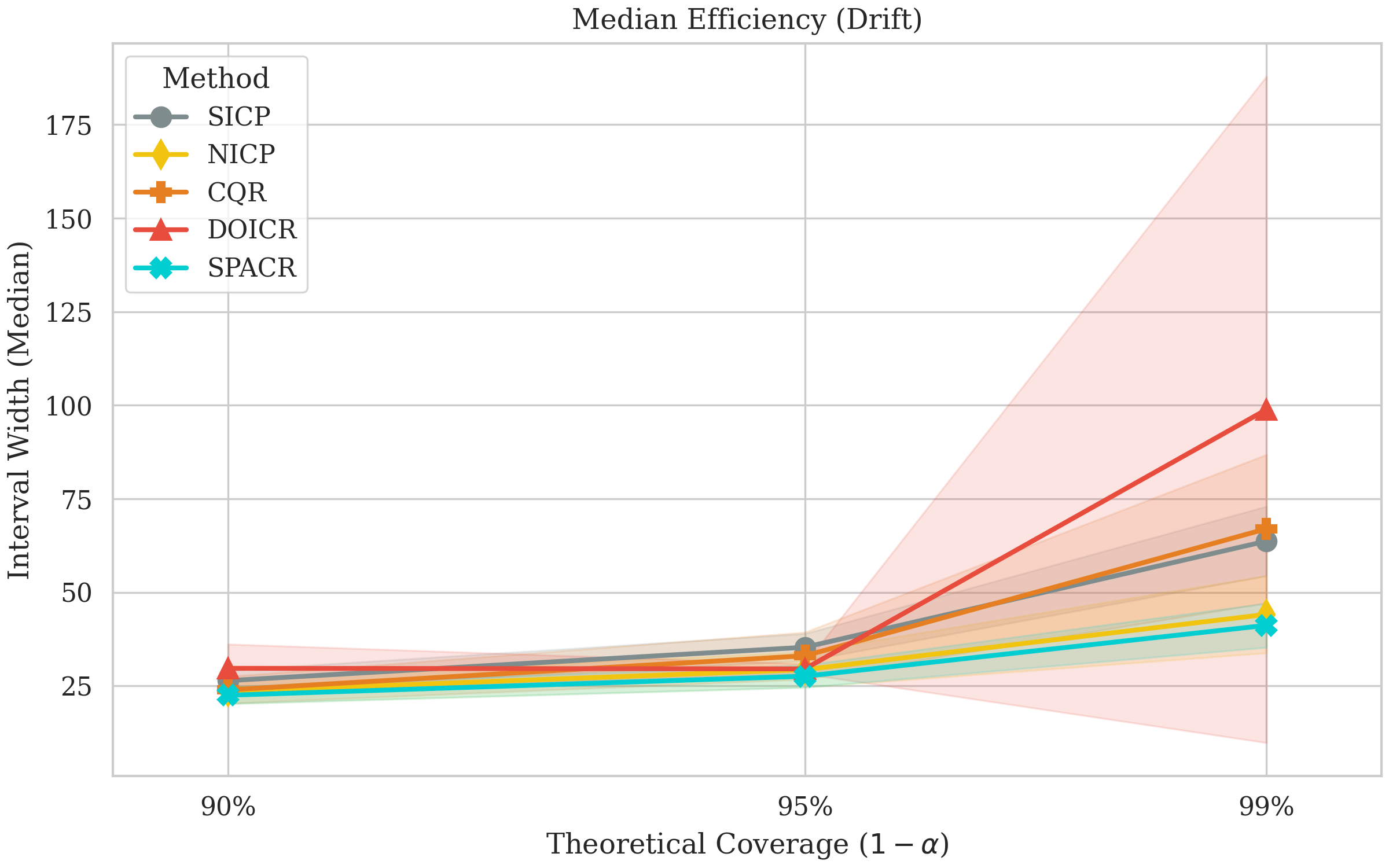}
    \caption{Efficiency (median interval width).}
    \label{fig:efficiency_methods_Drift}
  \end{subfigure}
  \hfill
  \begin{subfigure}[t]{0.32\textwidth}
    \includegraphics[width=\textwidth]{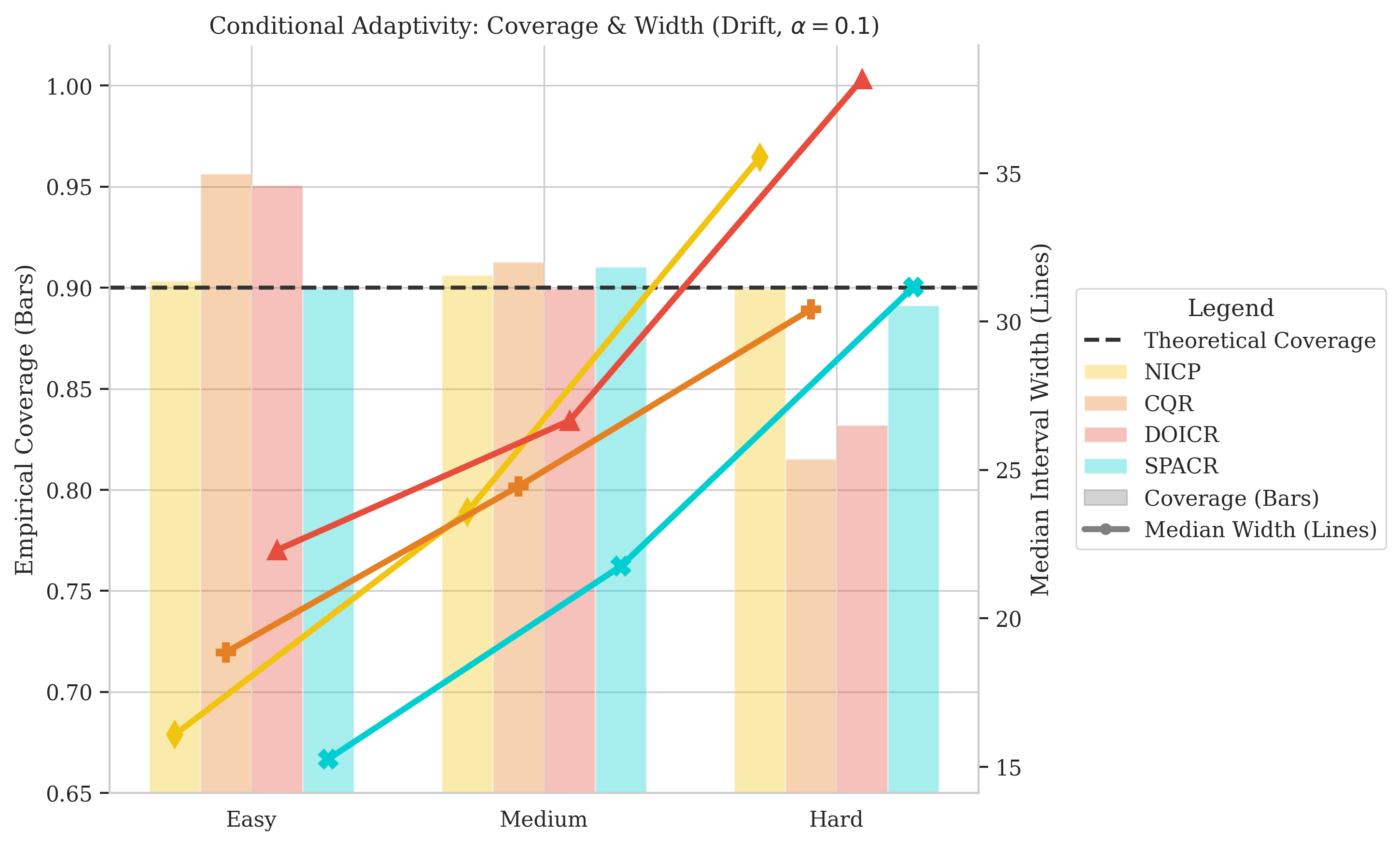}
    \caption{Conditional coverage (bars) and efficiency (lines) by internal difficulty.}
    \label{fig:iqr_methods_Drift}
  \end{subfigure}
  \caption{Performance of conformal regression methods on the Drift dataset (image regression).}
  \label{fig:methods_figures_Drift}
\end{figure*}

The experimental results comparing point prediction accuracy (MAE), marginal coverage, and efficiency (median and mean interval widths) of SICP, NICP, CQR, DOICR, and our proposed SPACR method across different datasets at various significance levels ($\alpha \in \{0.10, 0.05, 0.01\}$) are presented in Tables~\ref{tab:main_results_cov_mae} and~\ref{tab:main_results_eff}. Illustrative performance curves for the Diamonds and Drift datasets are shown in Figures~\ref{fig:methods_figures_Diamonds} and~\ref{fig:methods_figures_Drift}, respectively.

In terms of point prediction accuracy (MAE), SPACR demonstrates robust performance, though integrating conformal objectives directly into the loss function introduces a slight trade-off on certain tabular datasets. Specifically, as shown in Table~\ref{tab:main_results_cov_mae}, SPACR exhibits a higher MAE relative to the SICP baseline on Brazilian Houses, CPU Act, and Superconduct. However, across the majority of tabular datasets, SPACR largely preserves the underlying model's predictive accuracy and successfully avoids the severe accuracy degradation often incurred by other end-to-end conformal training approaches like DOICR. Crucially, on complex image datasets (Drift and UTK Face), SPACR outright outperforms all baselines, dropping the MAE on UTK Face from $7.09$ (SICP) to $5.43$. This improvement stems from the joint training objective acting as a multi-task structural regularizer. By forcing the high-capacity ResNet18 backbone to simultaneously optimize for point accuracy and valid uncertainty bounds, the network resists overfitting label noise and learns more robust feature representations.

For validity, the evaluated methods generally maintain marginal coverage within $\pm 1.5\%$ of the target confidence levels. As observed in Figures~\ref{fig:coverage_methods_Diamonds} and~\ref{fig:coverage_methods_Drift}, SPACR successfully satisfies the validity constraint, yielding coverage rates that track the nominal theoretical calibration line perfectly. A notable exception to this general robustness is DOICR, which exhibits severe under-coverage on certain datasets. For example, Table~\ref{tab:main_results_cov_mae} shows DOICR achieving only $71.91 \pm 40.21\%$ for a target validity of $90\%$ on Brazilian Houses, and severely failing on Drift with $57.04 \pm 52.08\%$ at $95\%$. This failure on Drift is further illustrated in Figure~\ref{fig:coverage_methods_Drift}, where the DOICR coverage curve deviates drastically from the theoretical baseline.

While providing valid marginal coverage, SPACR consistently delivers superior efficiency, giving the tightest prediction intervals in the majority of test cases. Across the 14 evaluated datasets in Table~\ref{tab:main_results_eff}, SPACR generally achieves the lowest median and mean interval widths. This efficiency advantage is particularly pronounced on image datasets. For Drift, Figure~\ref{fig:efficiency_methods_Drift} shows that the SPACR curve consistently lies below the competing baselines. However, evaluating mean efficiency exposes SPACR's sensitivity to extreme, one-sided outliers, as observed in the Brazilian Houses dataset. This highlights a structural limitation of SPACR's symmetric interval parameterization ($[\hat{y}_i - \hat{\sigma}_i, \hat{y}_i + \hat{\sigma}_i]$) compared to CQR. Because CQR independently predicts lower and upper quantiles, it natively adapts to skewed residual distributions through asymmetric intervals. Consequently, while CQR can efficiently extend only the affected boundary to cover a one-sided outlier, SPACR is forced to symmetrically inflate its single $\hat{\sigma}_i$ parameter, which disproportionately increases the mean interval width. Despite this limitation, SPACR's median efficiency remains highly robust across skewed distributions, underscoring its utility as the most stable efficiency metric. The other exceptions to SPACR's median efficiency dominance sometimes occur at the $99\%$ confidence level, where competing baselines achieve slightly tighter intervals on a small subset of the datasets (e.g., DOICR on California, Medical Charges, and Superconduct), though SPACR remains highly competitive.

\subsection{Adaptability of SPACR}\label{subsec:adaptability_spacr}

The empirical results highlight SPACR's robust conditional adaptivity across varying data characteristics and user requirements. By decoupling training from the choice of $\alpha$, SPACR adapts seamlessly to different confidence requirements where a single trained model produces valid intervals at $90\%$, $95\%$, and $99\%$ coverage (Table~\ref{tab:main_results_cov_mae}) without the extra training computation costs required by other methods such as CQR and DOICR.

Figures~\ref{fig:iqr_methods_Diamonds} and~\ref{fig:iqr_methods_Drift} highlight SPACR's conditional adaptivity across difficulty bins (``Easy'', ``Medium'', ``Hard'') at $\alpha = 0.10$. To evaluate this, the test samples are divided into three equal-sized tertiles constructed dynamically for each method based on its own internal difficulty estimator $\sigma$. By sorting the test instances according to a method's specific $\hat{\sigma}$ predictions and splitting them into three bins, we assess how well each method's perceived difficulty translates into valid conditional coverage and efficient intervals. Although exact distribution-free conditional coverage is theoretically impossible without strong assumptions~\citep{foygel2021limits}, SPACR's optimization serves as a highly effective heuristic, maintaining empirical coverage near the target across subsets. In contrast, baselines struggle with consistent conditional calibration. For instance, on the Drift dataset (Figure~\ref{fig:iqr_methods_Drift}), CQR and DOICR suffer from severe under-coverage in the ``Hard'' bin, which is compensated by an over-coverage in the ``Easy'' bin.

Furthermore, SPACR achieves this stable coverage while maintaining highly adaptive and efficient prediction intervals. As shown by the median interval width lines in Figures~\ref{fig:iqr_methods_Diamonds} and~\ref{fig:iqr_methods_Drift}, SPACR's interval width scales smoothly and monotonically in response to instance difficulty, widening appropriately for ``Hard'' samples while remaining significantly tighter than the other methods across all bins. For instance, in Figure~\ref{fig:iqr_methods_Diamonds}, DOICR exhibits a nearly flat median interval width line across the difficulty bins. This lack of an upward trajectory indicates that DOICR fails to scale its uncertainty estimates according to instance difficulty, essentially outputting rigid, unadaptive intervals. These observations confirm that SPACR's learned $\hat{\sigma}$ successfully adapts to the residual distributions without requiring a predefined $\alpha$ parameter during training.

In conclusion, SPACR flexibly adjusts to diverse data settings (from low-dimensional tabular benchmarks to high-dimensional image tasks) and dynamically optimizes the trade-off between efficiency and coverage, making it a versatile tool for uncertainty quantification in supervised regression tasks.

\subsection{Sensitivity Analysis of the Regularization Parameter $\lambda$}
\label{subsec:lambda_sensitivity}

\begin{figure*}[ht]
  \centering
  \begin{subfigure}[t]{0.32\textwidth}
    \includegraphics[width=\textwidth]{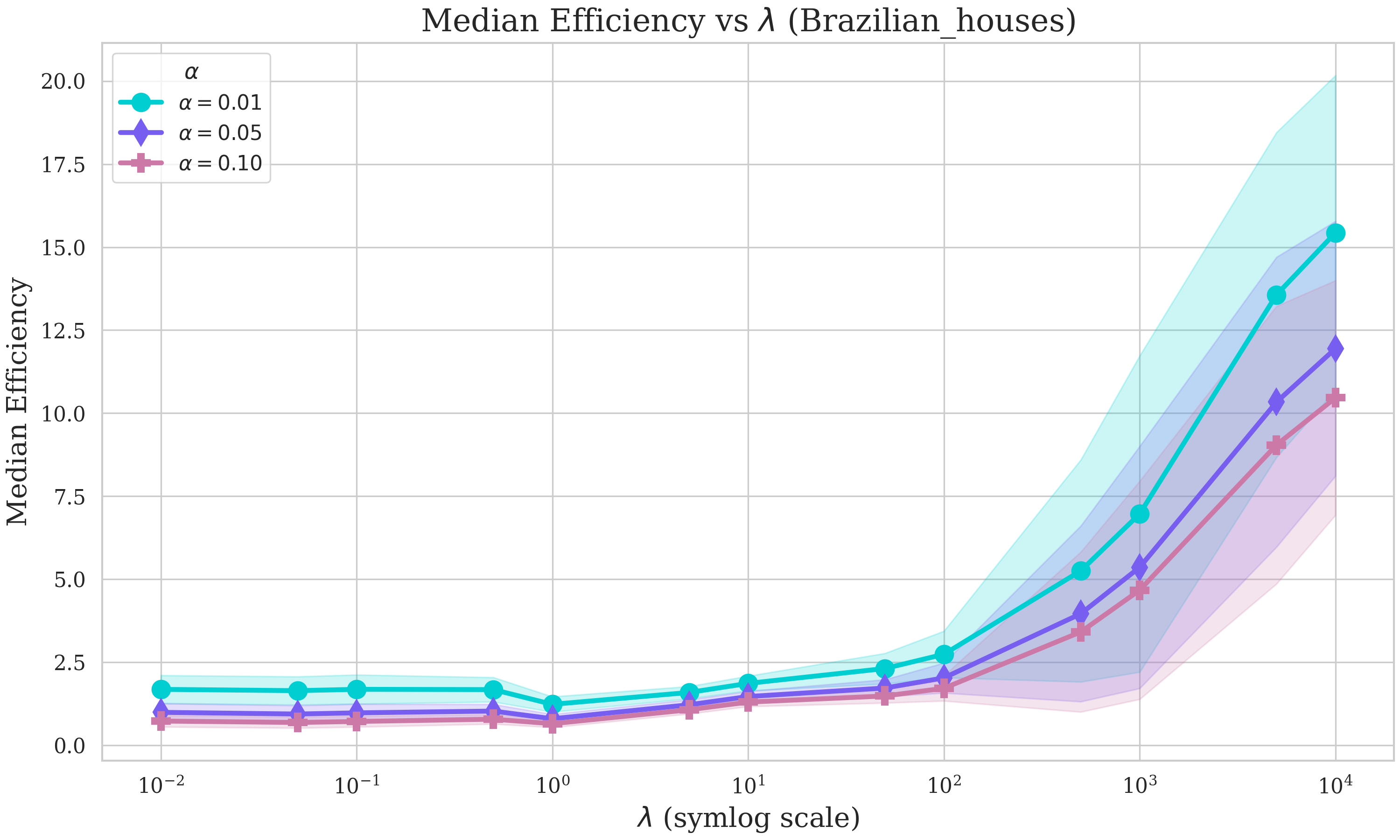}
    \caption{Median efficiency.}
    \label{fig:efficiency_lambda_Brazilian_houses}
  \end{subfigure}
  \hfill
  \begin{subfigure}[t]{0.32\textwidth}
    \includegraphics[width=\textwidth]{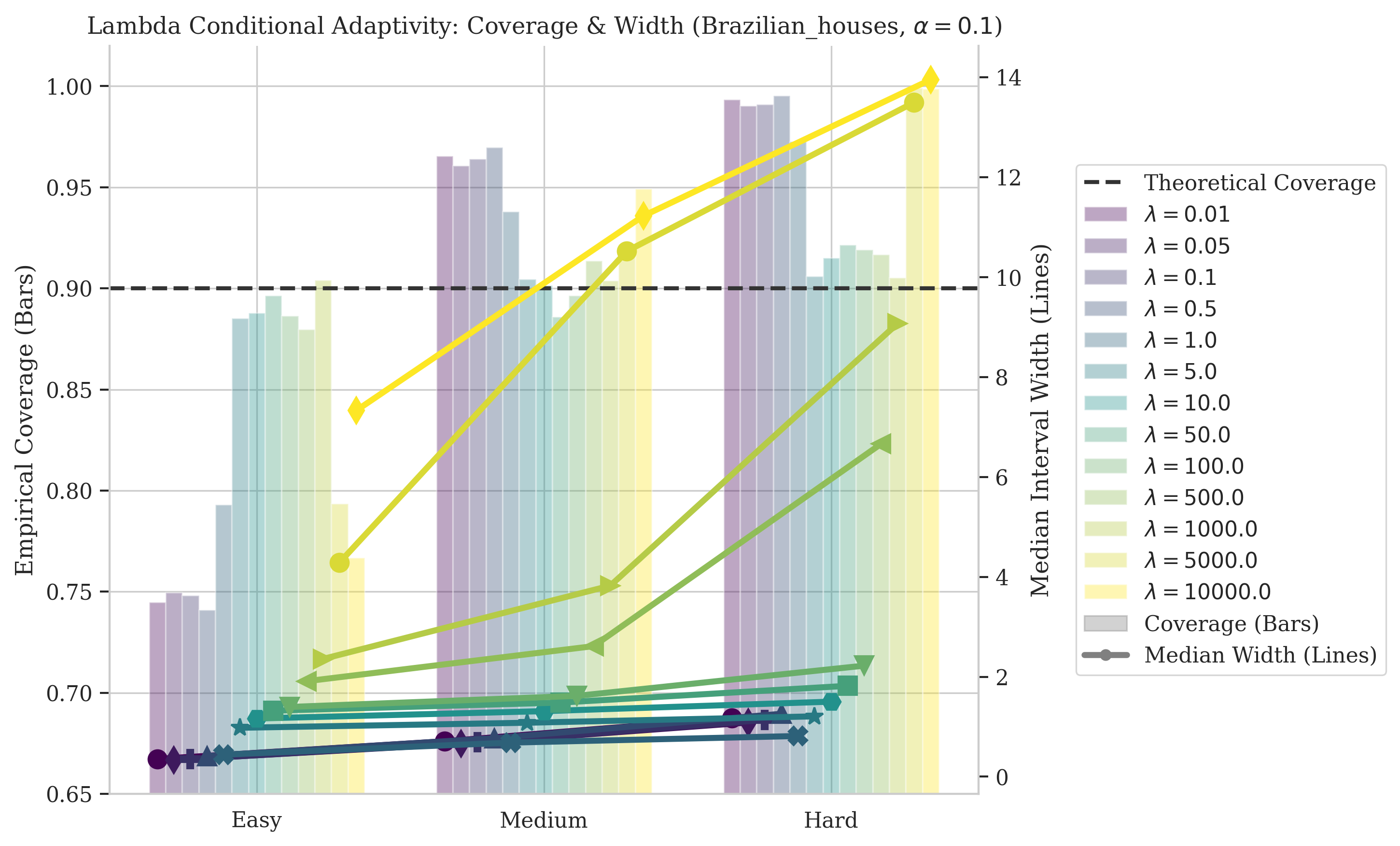}
    \caption{Conditional coverage (bars) and efficiency (lines) by internal difficulty.}
    \label{fig:box_plot_lambda_Brazilian_houses}
  \end{subfigure}
    \hfill
  \begin{subfigure}[t]{0.32\textwidth}
    \includegraphics[width=\textwidth]{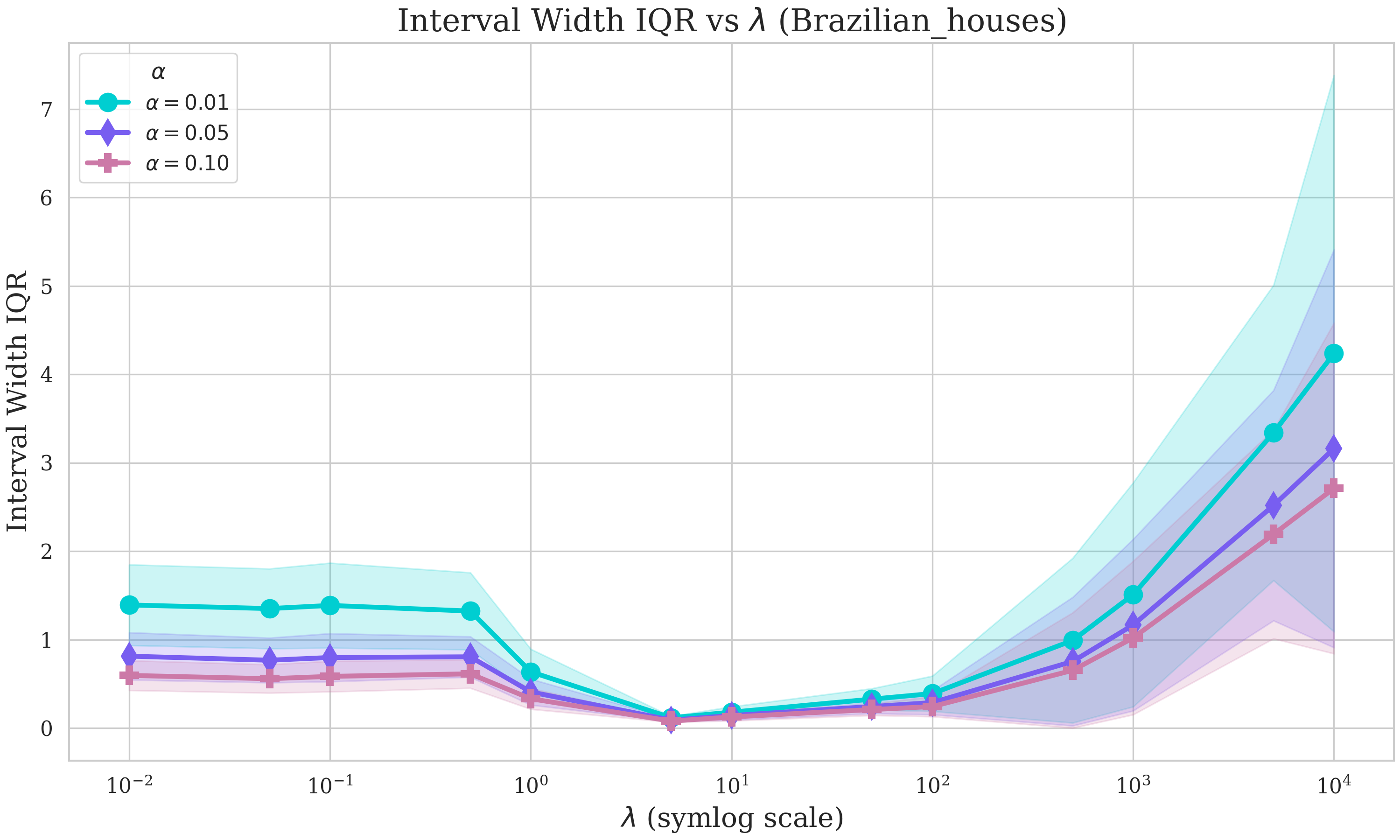}
    \caption{Interval width IQR.}
    \label{fig:iqr_lambda_Brazilian_houses}
  \end{subfigure}
  \caption{Sensitivity analysis of the regularization parameter $\lambda$ on the Brazilian Houses dataset.}
  \label{fig:lambda_figures_Brazilian_houses}
\end{figure*}

\begin{figure*}[ht]
  \centering
  \begin{subfigure}[t]{0.32\textwidth}
    \includegraphics[width=\textwidth]{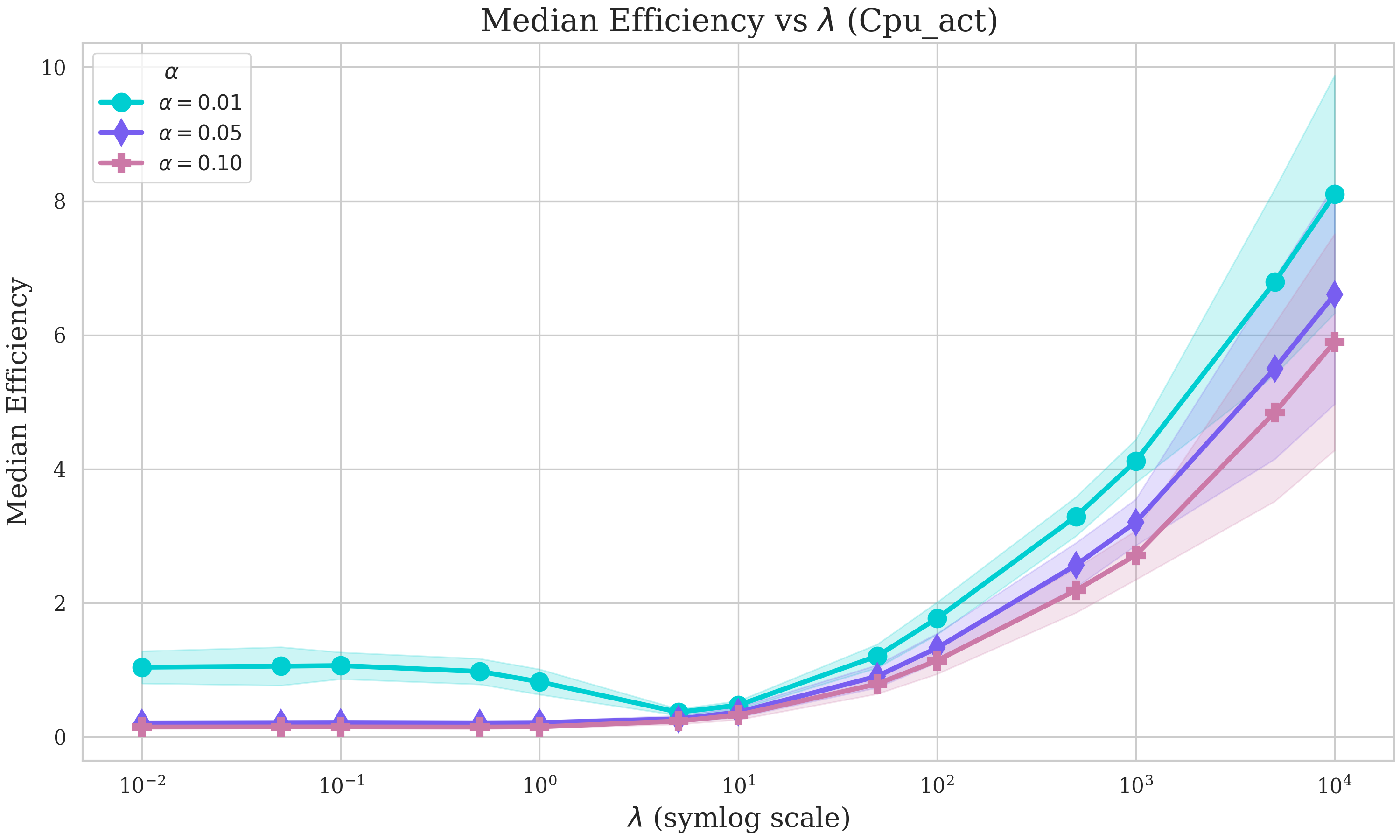}
    \caption{Median efficiency.}
    \label{fig:efficiency_lambda_Cpu_act}
  \end{subfigure}
  \hfill
  \begin{subfigure}[t]{0.32\textwidth}
    \includegraphics[width=\textwidth]{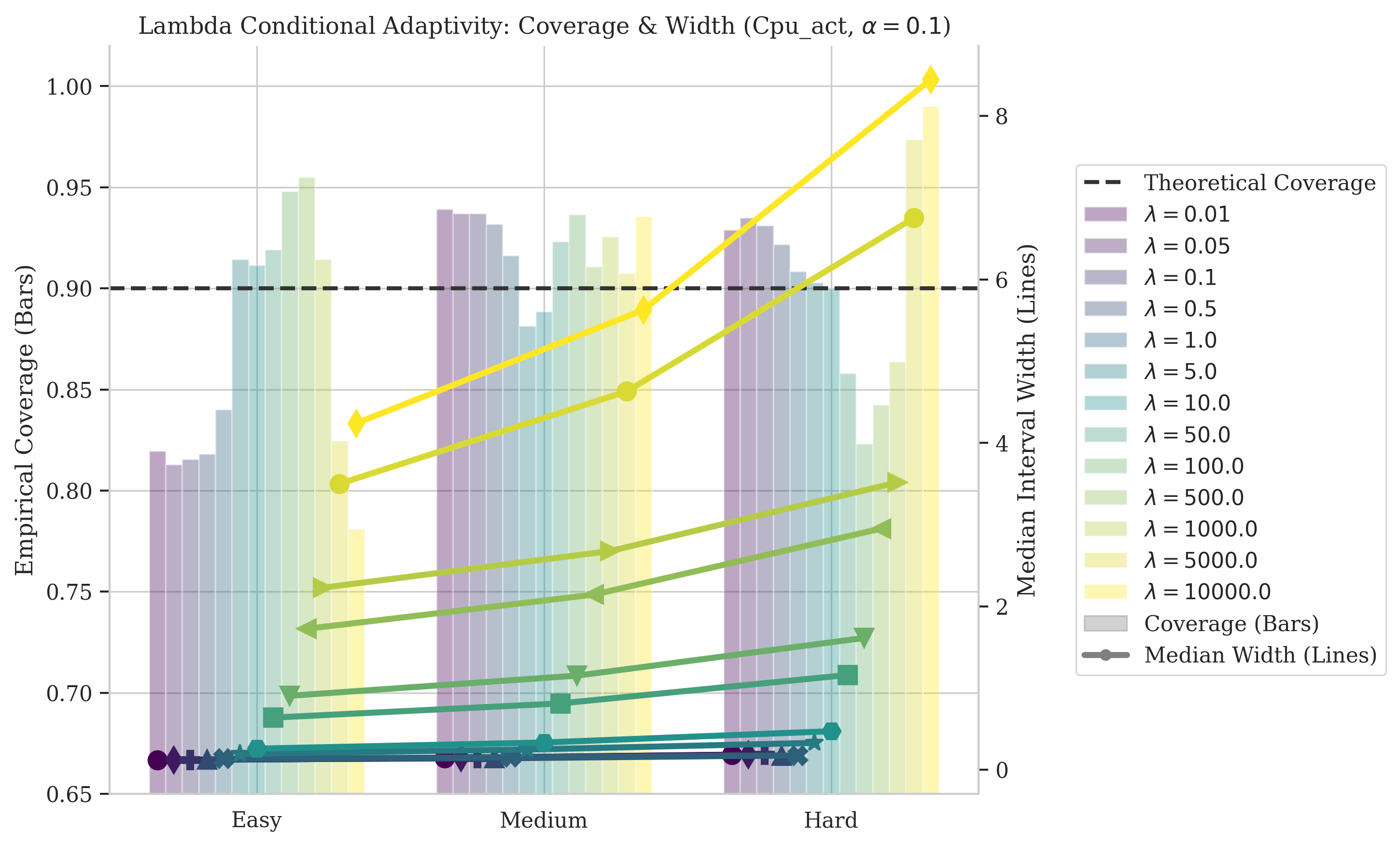}
    \caption{Conditional coverage (bars) and efficiency (lines) by internal difficulty.}
    \label{fig:box_plot_lambda_Cpu_act}
  \end{subfigure}  
    \hfill
  \begin{subfigure}[t]{0.32\textwidth}
    \includegraphics[width=\textwidth]{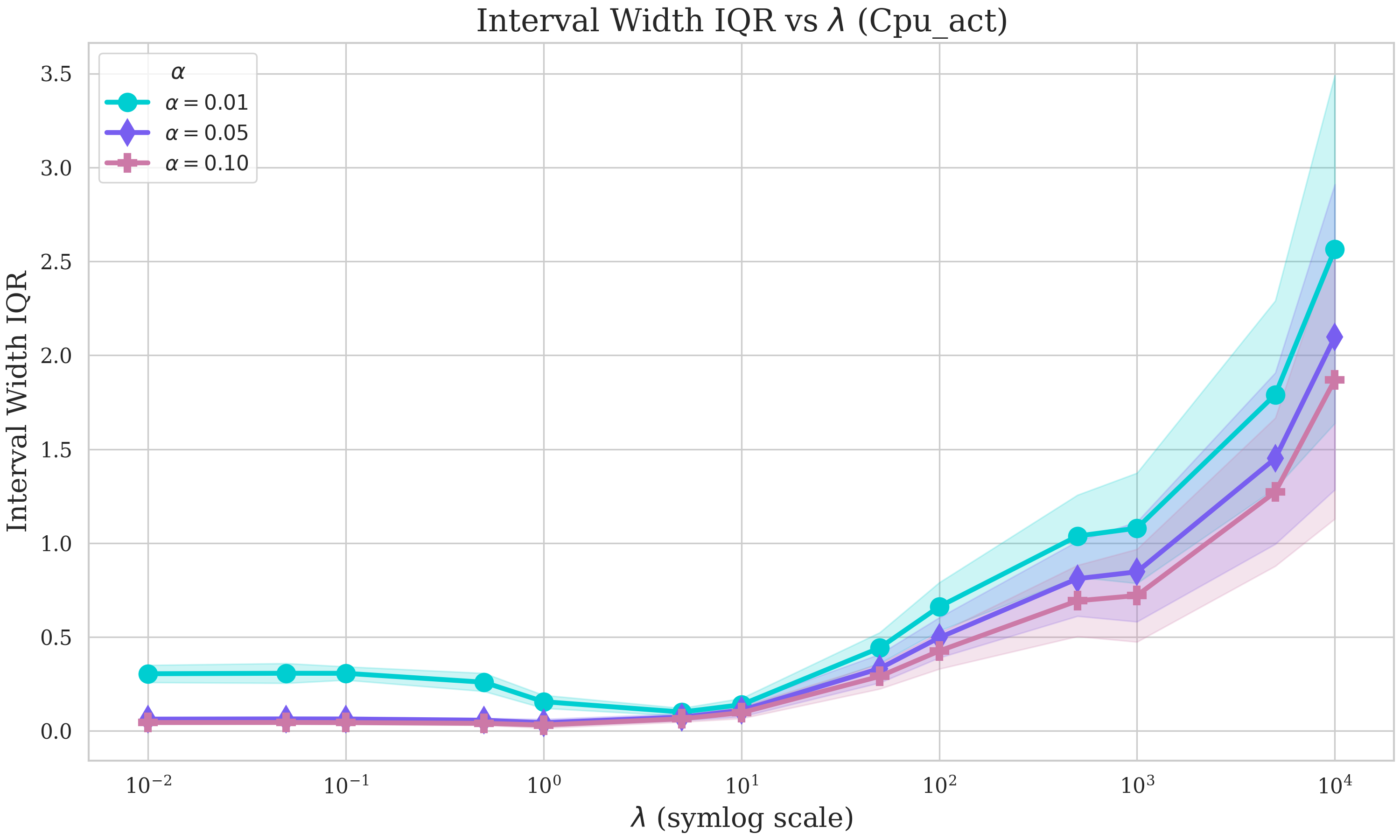}
    \caption{Interval width IQR.}
    \label{fig:iqr_lambda_Cpu_act}
  \end{subfigure}
    \caption{Sensitivity analysis of the regularization parameter $\lambda$ on the CPU Act dataset.}
  \label{fig:lambda_figures_Cpu_act}
\end{figure*}

\begin{figure*}[ht]
  \centering
  \begin{subfigure}[t]{0.32\textwidth}
    \includegraphics[width=\textwidth]{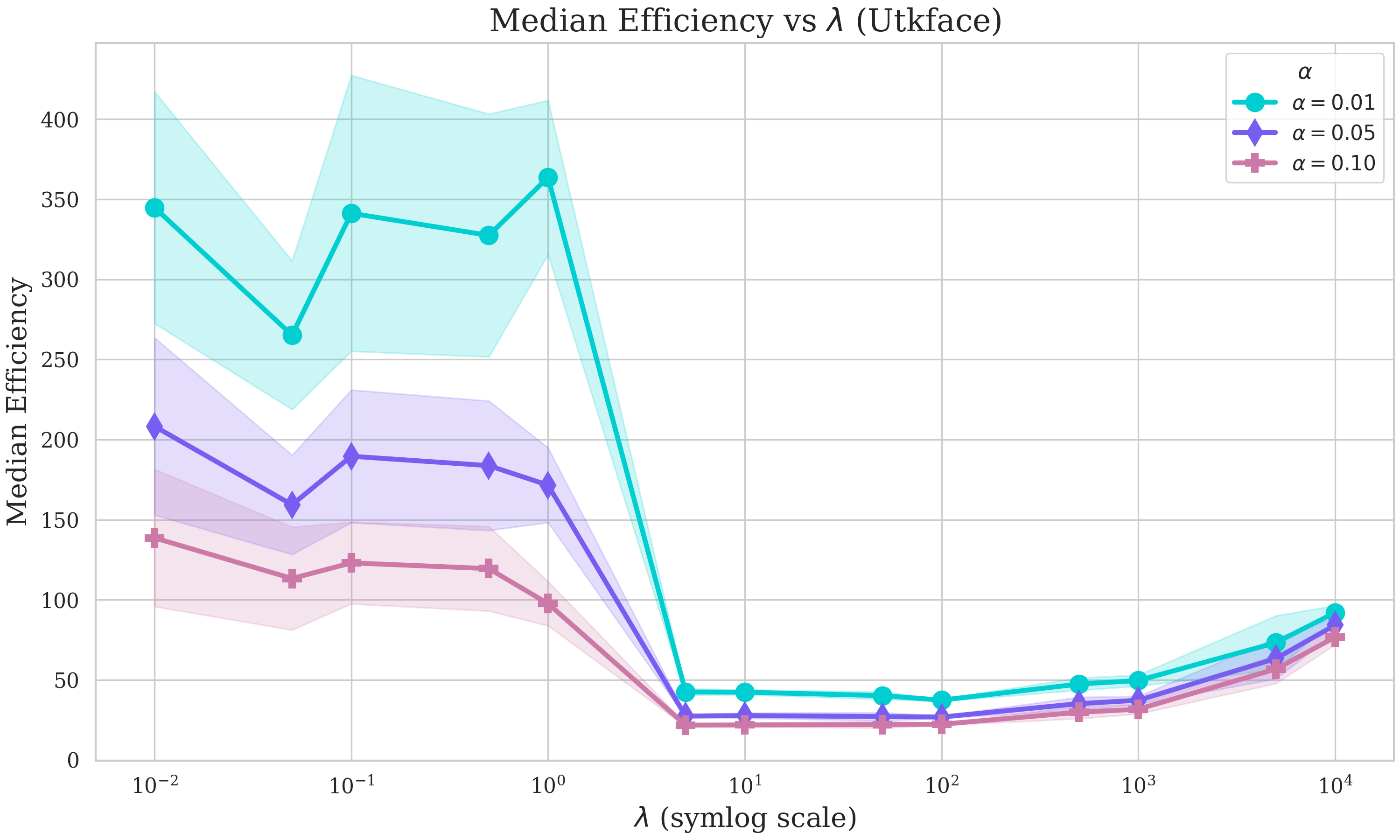}
    \caption{Median efficiency.}
    \label{fig:efficiency_lambda_Utkface}
  \end{subfigure}
  \hfill
  \begin{subfigure}[t]{0.32\textwidth}
    \includegraphics[width=\textwidth]{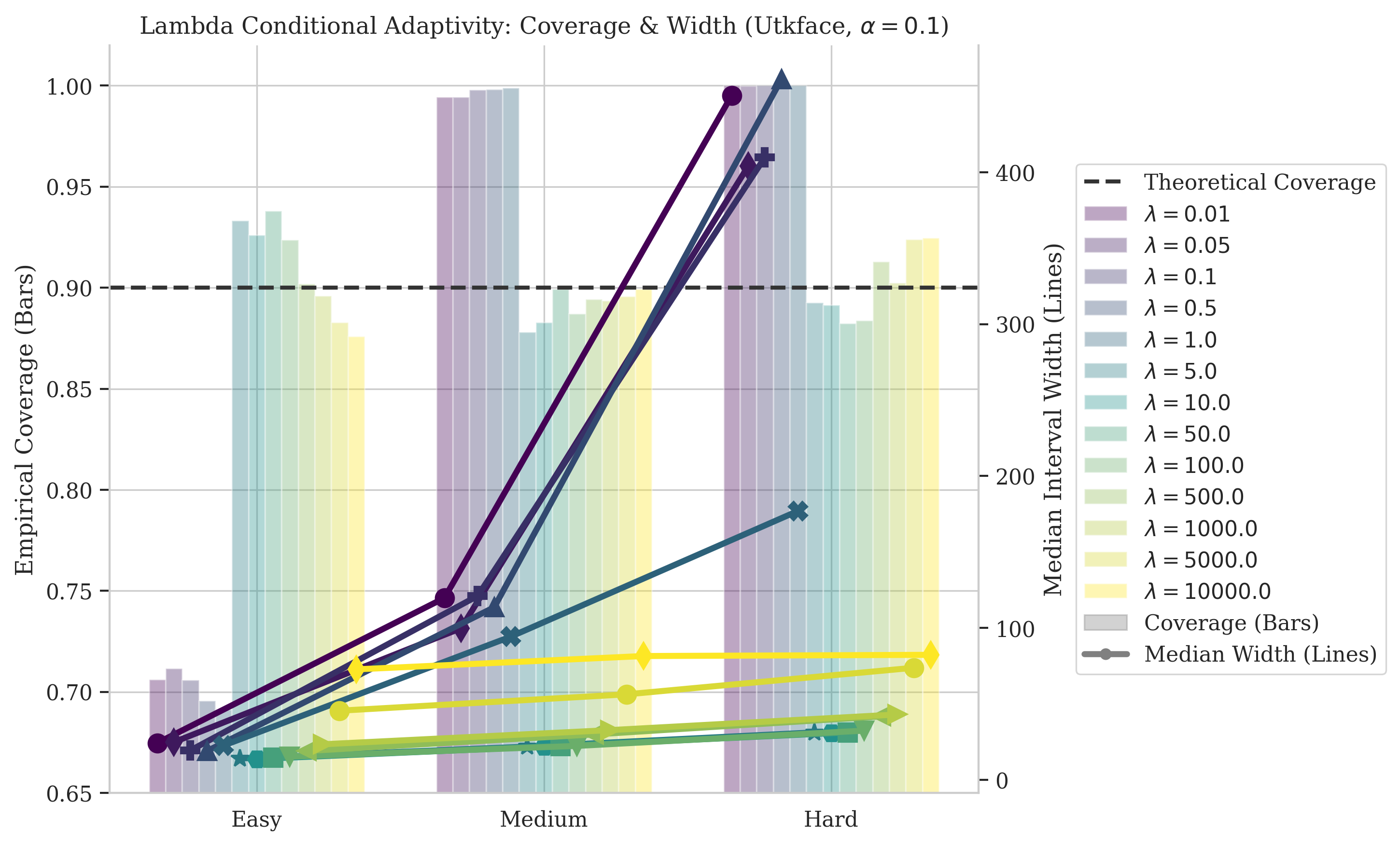}
    \caption{Conditional coverage (bars) and efficiency (lines) by internal difficulty.}
    \label{fig:box_plot_lambda_Utkface}
  \end{subfigure}
    \hfill
  \begin{subfigure}[t]{0.32\textwidth}
    \includegraphics[width=\textwidth]{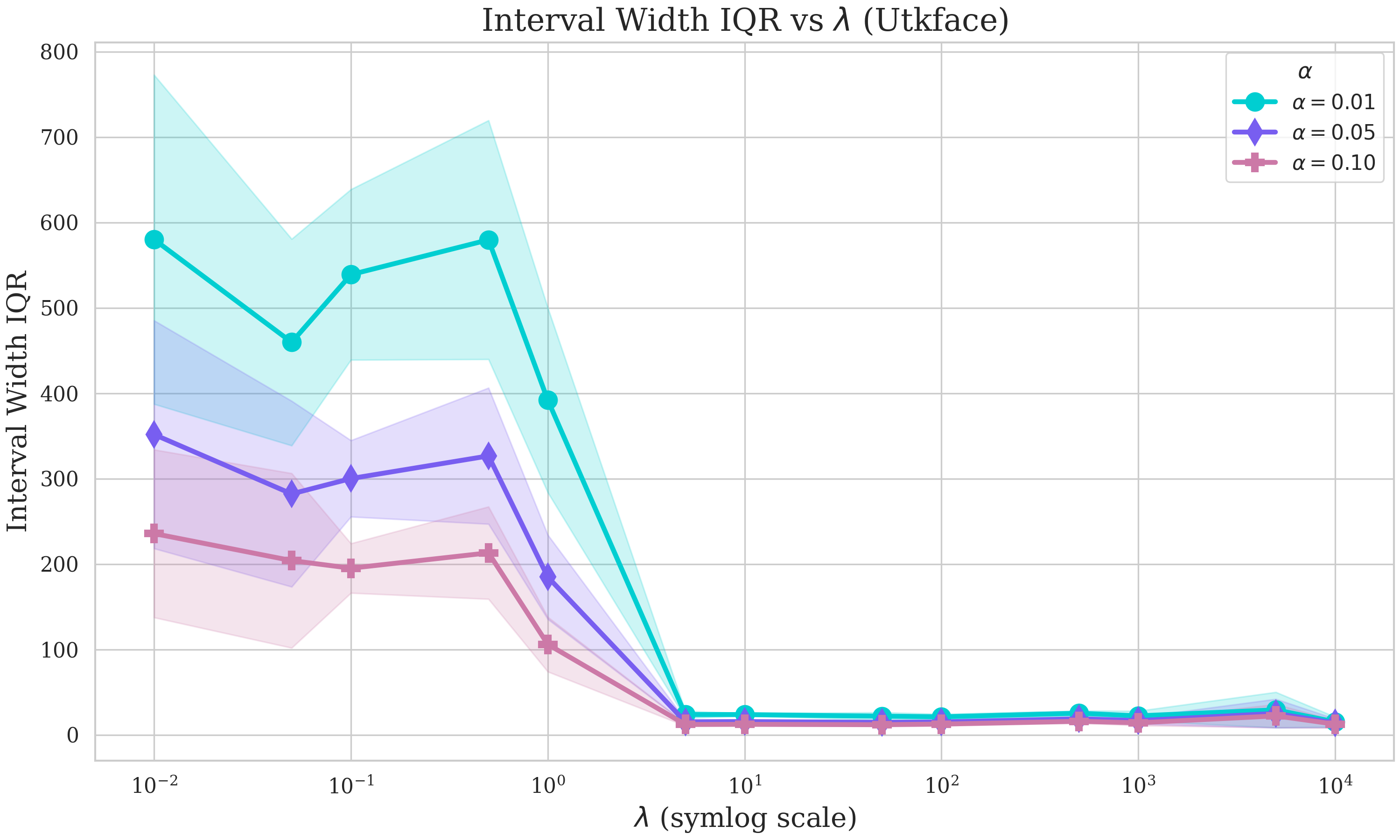}
    \caption{Interval width IQR.}
    \label{fig:iqr_lambda_Utkface}
  \end{subfigure}
    \caption{Sensitivity analysis of the regularization parameter $\lambda$ on the UTK Face dataset (image regression).}
  \label{fig:lambda_figures_Utkface}
\end{figure*}

The regularization parameter $\lambda$ in the $\mathcal{L}_{\text{SPACR}}$ objective governs the trade-off between point prediction accuracy ($\mathcal{L}_{\text{Accuracy}}$), interval efficiency ($\mathcal{L}_{\text{Efficiency}}$), and empirical coverage validity ($\mathcal{L}_{\text{Validity}}$). By modeling this trade-off with a single hyperparameter, we enable a highly efficient 1D logarithmic search, completely avoiding the exponential scaling and computational burden associated with multi-parameter grid searches. To evaluate hyperparameter sensitivity, performance is analyzed across a logarithmic grid from $\lambda = 0.01$ to $10000$. For conciseness, Table~\ref{tab:lambda_results_10} reports these metrics at $\alpha = 0.10$, supplemented by visual analysis for the Brazilian Houses (Figure~\ref{fig:lambda_figures_Brazilian_houses}), CPU Act (Figure~\ref{fig:lambda_figures_Cpu_act}), and UTK Face (Figure~\ref{fig:lambda_figures_Utkface}) datasets.

A primary observation is that the marginal coverage remains stable near the nominal $1-\alpha$ target across all $\lambda$ values, confirming that post-hoc ICP guarantees marginal validity independent of the model's training dynamics. However, varying $\lambda$ significantly dictates the balance between MAE, efficiency, and adaptivity:

\begin{itemize}
    \item \textbf{Under-regularization ($\lambda \to 0$):} The validity term $\mathcal{L}_{\text{Validity}}$ becomes negligible, driving uncertainty estimates toward zero. The impact on point prediction accuracy diverges based on the data modality and network capacity. For lower-dimensional tabular tasks optimized via MLPs, MAE remains near its minimum as the network strictly prioritizes $\mathcal{L}_{\text{Accuracy}}$. On datasets like CPU Act, the network uniformly shrinks $\hat{\sigma}_i$ to a tiny constant, which yields non-adaptive intervals resembling Standard ICP. Specifically, an IQR close to zero (e.g., in Figure~\ref{fig:iqr_lambda_Cpu_act}, for $\alpha = 0.10$ and $\lambda = 0.01$, $\text{IQR} = 0.04 \approx 0$) correctly indicates nearly constant-width intervals, which is also confirmed by the conditional adaptivity analysis showing a completely flat median interval width line across the difficulty bins (Figure~\ref{fig:box_plot_lambda_Cpu_act}, purple hues). In high-dimensional image regression tasks using deep convolutional architectures (e.g., ResNet18 on UTK Face and Drift), removing the validity penalty surprisingly degrades MAE (e.g., UTK Face MAE inflates from $5.43$ at $\lambda=5$ to $7.80$ at $\lambda=0.01$ as in Table~\ref{tab:lambda_results_10}), which leaves the network susceptible to overfitting noisy targets, yielding suboptimal local minima. Simultaneously, driving $\hat{\sigma}_i \to 0$ compromises efficiency. On UTK Face, vanishing denominators in the non-conformity scores force massive calibration quantiles $\hat{q}$, resulting in impractically wide, high-variance intervals. For instance, at $\alpha=0.10$, setting $\lambda=0.01$ yields a highly unstable median efficiency of $138.69$. The conditional adaptivity plot (Figure~\ref{fig:box_plot_lambda_Utkface}, purple hues) reveals erratic, massive widths across all strata, confirming that this large IQR is a symptom of extreme instability rather than meaningful adaptivity. Moreover, because the collapsed $\hat{\sigma}_i$ values are effectively meaningless, the model loses its capacity to correctly rank instance difficulty. As a result, sorting instances into subsets based on these broken estimates yields severe conditional miscalibration: the intervals are too narrow for the actual unranked errors in the ``Easy'' bin (causing under-coverage) and unnecessarily massive in the ``Hard'' bin (causing over-coverage).
    
    \item \textbf{Over-regularization ($\lambda \to \infty$):} The validity term $\mathcal{L}_{\text{Validity}}$ dominates, forcing the model to prioritize the coverage constraint $\phi(y_i, \hat{y}_i, \hat{\sigma}_i)$ at the direct expense of both point prediction and efficiency. As a result, the underlying regressor accuracy degrades universally across both evaluated modalities. For instance, on Brazilian Houses, MAE inflates from $0.12$ at $\lambda=0.05$ to $3.37$ at $\lambda=10000$ (Table~\ref{tab:lambda_results_10}). Crucially, on datasets with heavy-tailed target distributions such as Brazilian Houses, over-regularization forces the exponential parameterization ($\hat{\sigma}_i = \exp(u_i)$) to inflate massively to cover extreme outliers. This triggers an exponential explosion in the \textit{mean} interval width, while the \textit{median} width remains comparatively stable. The conditional adaptivity plots (Figure~\ref{fig:box_plot_lambda_Brazilian_houses}, yellow hues) further illustrate this dynamic, showing severe width inflation particularly in the 'Hard' difficulty bin in order to maintain coverage. This dynamic empirically validates our structural choice to report median efficiency as the primary robustness metric, as the arithmetic mean becomes mathematically unstable and unrepresentative under strong validity penalties.
    
    \item \textbf{Optimal Balance ($\lambda \in [1, 50]$):} Moderate values safely preserve point prediction accuracy in tabular datasets and yield a global minimum for MAE in complex image datasets (forming a U-shaped accuracy curve). Across datasets, setting $\lambda = 5$ consistently yields a near-optimal balance between MAE and median efficiency without interval inflation, while maintaining true conditional adaptivity. The conditional analyses in Figures~\ref{fig:box_plot_lambda_Brazilian_houses}, \ref{fig:box_plot_lambda_Cpu_act}, and \ref{fig:box_plot_lambda_Utkface} (green hues) demonstrate that moderate $\lambda$ values allow interval widths to scale monotonically with instance difficulty. The intervals remain appropriately narrow for ``Easy'' samples and widen dynamically for ``Hard'' samples, while empirical coverage remains highly stable across all strata. Furthermore, fine-tuning $\lambda$ allows for strictly superior efficiency. For instance, observing the Brazilian Houses dataset at $\alpha = 0.10$ (Table~\ref{tab:lambda_results_10}), choosing $\lambda = 1$ yields an optimal median width of $0.65$ and an MAE of $0.14$, safely avoiding both the accuracy degradation of higher $\lambda$ and the instability of lower $\lambda$, and strictly improving upon the baseline comparisons in Tables~\ref{tab:main_results_cov_mae} and~\ref{tab:main_results_eff}. Consequently, for practical applications, one can simply tune $\lambda$ using a small holdout validation set to measure coverage, accuracy, and interval width at the desired target $\alpha$, sweeping across $\lambda \in [1, 50]$ to find the optimal configuration for a specific dataset.
\end{itemize}

\begin{table*}[htb]
\caption{Impact of $\lambda$ on MAE, Marginal Coverage, Median/Mean Efficiency, and Interval Width IQR. Results are shown for $\alpha = 0.10$. Best results per dataset are highlighted in bold. The row corresponding to $\lambda = 5$ is shaded in gray.}
\centering\scriptsize\resizebox{\textwidth}{!}{
\begin{tabular}{lcccccc}
\toprule
\textbf{Dataset} & \textbf{Lambda ($\lambda$)} & MAE $\downarrow$ & Cov. (\%) & Med. Eff. $\downarrow$ & Mean Eff. $\downarrow$ & IQR $\downarrow$ \\ \midrule
\multirow{13}{*}{\textbf{Brazilian Houses}}
& 0.01 & 0.13 $\pm$ 0.02 & 90.12 $\pm$ 1.25 & 0.73 $\pm$ 0.17 & 0.78 $\pm$ 0.19 & 0.59 $\pm$ 0.17 \\
& 0.05 & \textbf{0.12 $\pm$ 0.02} & 90.01 $\pm$ 0.84 & 0.69 $\pm$ 0.17 & 0.74 $\pm$ 0.18 & 0.56 $\pm$ 0.16 \\
& 0.1 & 0.13 $\pm$ 0.02 & 90.11 $\pm$ 0.83 & 0.72 $\pm$ 0.16 & 0.77 $\pm$ 0.18 & 0.58 $\pm$ 0.17 \\
& 0.5 & 0.14 $\pm$ 0.02 & 90.20 $\pm$ 0.83 & 0.79 $\pm$ 0.14 & 0.84 $\pm$ 0.16 & 0.61 $\pm$ 0.16 \\
& 1 & 0.14 $\pm$ 0.02 & 90.12 $\pm$ 0.16 & \textbf{0.65 $\pm$ 0.11} & \textbf{0.67 $\pm$ 0.11} & 0.33 $\pm$ 0.12 \\
& \cellcolor{gray!20} 5 & \cellcolor{gray!20} 0.34 $\pm$ 0.03 & \cellcolor{gray!20} 89.86 $\pm$ 0.69 & \cellcolor{gray!20} 1.07 $\pm$ 0.12 & \cellcolor{gray!20} 818.78 $\pm$ 1828.32 & \cellcolor{gray!20} \textbf{0.08 $\pm$ 0.01} \\
& 10 & 0.41 $\pm$ 0.03 & 90.14 $\pm$ 0.52 & 1.30 $\pm$ 0.13 & $4.90 \times 10^7 \pm 1.10 \times 10^8$ & 0.12 $\pm$ 0.04 \\
& 50 & 0.43 $\pm$ 0.04 & 90.13 $\pm$ 0.89 & 1.49 $\pm$ 0.21 & $2.15 \times 10^8 \pm 4.80 \times 10^8$ & 0.21 $\pm$ 0.07 \\
& 100 & 0.48 $\pm$ 0.10 & 90.07 $\pm$ 0.56 & 1.72 $\pm$ 0.38 & $1.10 \times 10^9 \pm 2.46 \times 10^9$ & 0.24 $\pm$ 0.11 \\
& 500 & 1.02 $\pm$ 0.81 & 90.33 $\pm$ 0.99 & 3.41 $\pm$ 2.41 & $1.06 \times 10^9 \pm 2.38 \times 10^9$ & 0.65 $\pm$ 0.65 \\
& 1000 & 1.41 $\pm$ 1.15 & 90.43 $\pm$ 0.80 & 4.67 $\pm$ 3.28 & $5.67 \times 10^9 \pm 1.27 \times 10^{10}$ & 1.02 $\pm$ 0.87 \\
& 5000 & 2.92 $\pm$ 1.74 & 90.48 $\pm$ 0.81 & 9.04 $\pm$ 4.18 & $5.20 \times 10^{18} \pm 1.16 \times 10^{19}$ & 2.19 $\pm$ 1.18 \\
& 10000 & 3.37 $\pm$ 1.54 & 90.49 $\pm$ 0.97 & 10.46 $\pm$ 3.53 & $1.56 \times 10^{21} \pm 3.49 \times 10^{21}$ & 2.71 $\pm$ 1.87 \\
\midrule
\multirow{13}{*}{\textbf{CPU Act}}
& 0.01 & \textbf{0.04 $\pm$ 0.01} & 89.59 $\pm$ 1.36 & \textbf{0.15 $\pm$ 0.01} & 0.19 $\pm$ 0.02 & 0.04 $\pm$ 0.01 \\
& 0.05 & \textbf{0.04 $\pm$ 0.01} & 89.49 $\pm$ 1.07 & \textbf{0.15 $\pm$ 0.01} & 0.19 $\pm$ 0.03 & 0.04 $\pm$ 0.01 \\
& 0.1 & \textbf{0.04 $\pm$ 0.01} & 89.46 $\pm$ 1.23 & \textbf{0.15 $\pm$ 0.01} & 0.19 $\pm$ 0.03 & 0.04 $\pm$ 0.01 \\
& 0.5 & \textbf{0.04 $\pm$ 0.01} & 89.05 $\pm$ 0.78 & \textbf{0.15 $\pm$ 0.01} & 0.18 $\pm$ 0.03 & 0.04 $\pm$ 0.01 \\
& 1 & \textbf{0.04 $\pm$ 0.01} & 88.82 $\pm$ 1.03 & \textbf{0.15 $\pm$ 0.02} & \textbf{0.17 $\pm$ 0.03} & \textbf{0.03 $\pm$ 0.02} \\
& \cellcolor{gray!20} 5 & \cellcolor{gray!20} 0.08 $\pm$ 0.02 & \cellcolor{gray!20} 89.96 $\pm$ 0.52 & \cellcolor{gray!20} 0.24 $\pm$ 0.05 & \cellcolor{gray!20} 0.28 $\pm$ 0.06 & \cellcolor{gray!20} 0.06 $\pm$ 0.02 \\
& 10 & 0.11 $\pm$ 0.03 & 90.01 $\pm$ 0.59 & 0.33 $\pm$ 0.07 & 0.40 $\pm$ 0.09 & 0.10 $\pm$ 0.03 \\
& 50 & 0.29 $\pm$ 0.06 & 90.01 $\pm$ 0.62 & 0.79 $\pm$ 0.15 & 0.95 $\pm$ 0.18 & 0.29 $\pm$ 0.07 \\
& 100 & 0.39 $\pm$ 0.08 & 90.26 $\pm$ 1.20 & 1.14 $\pm$ 0.20 & 1.33 $\pm$ 0.23 & 0.43 $\pm$ 0.10 \\
& 500 & 0.66 $\pm$ 0.11 & 90.27 $\pm$ 0.85 & 2.19 $\pm$ 0.33 & 2.57 $\pm$ 0.48 & 0.69 $\pm$ 0.19 \\
& 1000 & 0.80 $\pm$ 0.13 & 90.13 $\pm$ 0.63 & 2.71 $\pm$ 0.37 & 3.21 $\pm$ 0.55 & 0.72 $\pm$ 0.25 \\
& 5000 & 1.54 $\pm$ 0.53 & 90.19 $\pm$ 0.83 & 4.84 $\pm$ 1.33 & 5.81 $\pm$ 1.60 & 1.27 $\pm$ 0.39 \\
& 10000 & 1.86 $\pm$ 0.54 & 90.23 $\pm$ 0.71 & 5.89 $\pm$ 1.61 & 7.46 $\pm$ 2.45 & 1.87 $\pm$ 0.74 \\
\midrule
\multirow{13}{*}{\textbf{UTK Face}}
& 0.01 & 7.80 $\pm$ 2.48 & 90.02 $\pm$ 0.65 & 138.69 $\pm$ 42.87 & 378.64 $\pm$ 138.88 & 235.76 $\pm$ 98.26 \\
& 0.05 & 8.11 $\pm$ 2.55 & 90.19 $\pm$ 0.72 & 113.36 $\pm$ 32.09 & 440.49 $\pm$ 284.55 & 204.10 $\pm$ 102.21 \\
& 0.1 & 5.98 $\pm$ 0.38 & 90.13 $\pm$ 0.48 & 123.09 $\pm$ 25.53 & 295.86 $\pm$ 21.04 & 195.18 $\pm$ 28.97 \\
& 0.5 & 6.01 $\pm$ 0.20 & 89.79 $\pm$ 0.46 & 119.50 $\pm$ 26.40 & 297.73 $\pm$ 50.03 & 213.17 $\pm$ 54.07 \\
& 1 & 5.89 $\pm$ 0.40 & 89.68 $\pm$ 0.76 & 97.68 $\pm$ 13.83 & 111.02 $\pm$ 23.27 & 106.05 $\pm$ 31.99 \\
& \cellcolor{gray!20} 5 & \cellcolor{gray!20} \textbf{5.43 $\pm$ 0.18} & \cellcolor{gray!20} 90.13 $\pm$ 0.46 & \cellcolor{gray!20} \textbf{21.84 $\pm$ 0.84} & \cellcolor{gray!20} \textbf{22.54 $\pm$ 0.43} & \cellcolor{gray!20} 12.18 $\pm$ 1.61 \\
& 10 & 5.53 $\pm$ 0.23 & 90.02 $\pm$ 0.20 & 21.96 $\pm$ 1.16 & 22.22 $\pm$ 0.94 & \textbf{12.34 $\pm$ 1.18} \\
& 50 & 5.65 $\pm$ 0.30 & 90.66 $\pm$ 0.85 & 22.20 $\pm$ 2.17 & 22.71 $\pm$ 0.77 & 11.97 $\pm$ 2.00 \\
& 100 & 5.96 $\pm$ 0.14 & 89.81 $\pm$ 0.38 & 22.38 $\pm$ 0.66 & 23.40 $\pm$ 0.55 & 12.65 $\pm$ 1.90 \\
& 500 & 7.72 $\pm$ 1.36 & 90.30 $\pm$ 0.34 & 29.87 $\pm$ 3.94 & 31.26 $\pm$ 5.54 & 15.93 $\pm$ 2.13 \\
& 1000 & 8.27 $\pm$ 1.04 & 89.73 $\pm$ 0.64 & 31.78 $\pm$ 2.96 & 32.90 $\pm$ 2.64 & 14.13 $\pm$ 3.13 \\
& 5000 & 14.01 $\pm$ 2.92 & 90.08 $\pm$ 0.35 & 56.83 $\pm$ 9.02 & 60.51 $\pm$ 10.58 & 22.25 $\pm$ 13.65 \\
& 10000 & 19.20 $\pm$ 1.61 & 90.00 $\pm$ 0.34 & 76.74 $\pm$ 4.66 & 79.09 $\pm$ 4.97 & 12.56 $\pm$ 4.47 \\
\bottomrule\end{tabular}}
\label{tab:lambda_results_10}\end{table*}

\subsection{Ablation Study: Impact of Validity and Efficiency Loss Terms}

To evaluate the individual contributions of the components in our proposed loss function, we compare the full SPACR model against two variants: \textbf{SPACR\_VAL}, which ablates the efficiency penalty ($\mathcal{L}_{\text{Efficiency}}$), and \textbf{SPACR\_EFF}, which ablates the validity penalty ($\mathcal{L}_{\text{Validity}}$). Table~\ref{tab:ablation} reports the point prediction error (MAE), marginal coverage, median efficiency, mean efficiency, and interquartile range (IQR) across the evaluated datasets.

\textbf{Marginal Coverage:} As expected, all variants successfully achieve the target marginal coverage across the different confidence levels due to the post-hoc ICP calibration step, which enforces distribution-free validity regardless of the training objective.

\textbf{No Drop in Accuracy (MAE):} The results show that adding efficiency and validity penalties does not hurt point prediction accuracy. Instead, training these goals together actually helps the model learn better. For example, on the UTK Face dataset, the full SPACR model gets the best MAE of $5.43$. If we remove the efficiency term, the MAE gets worse ($6.51$), and if we remove the validity term, it drops further to $6.87$. By simultaneously optimizing for accuracy, narrow intervals, and valid bounds, the network learns more robust feature representations.

\textbf{Matching the $\lambda$ Analysis:} Ablating specific loss components directly mirrors the boundary conditions of the $\lambda$ sensitivity analysis, with failure severity scaling according to data complexity and model capacity. For tabular datasets using MLPs, the underlying accuracy loss ($\mathcal{L}_{\text{Accuracy}}$) maintains relatively stable base predictions, allowing the calibration step to default to safe intervals without severe numerical instability. However, for high-dimensional image regression tasks using deep convolutional networks (ResNet18), removing the validity penalty (SPACR\_EFF) triggers a structural breakdown identical to extreme under-regularization ($\lambda \to 0$). Without the validity constraint, the model fails to produce bounded intervals prior to calibration, resulting in massive, impractical median widths and IQRs on datasets such as Drift and UTK Face. In contrast, ablating the efficiency penalty (SPACR\_VAL) simulates extreme over-regularization ($\lambda \to \infty$). Without any pressure to keep intervals narrow, the exponential function ($\hat{\sigma}_i = \exp(u_i)$) inflates to cover extreme outliers, which causes exploding mean widths (e.g., Brazilian Houses). Furthermore, for standard samples, SPACR\_VAL frequently collapses to an IQR of exactly zero (e.g., California and Medical Charges). While a large IQR does not inherently guarantee \textit{correct} adaptivity, an IQR of exactly zero definitively proves a complete failure of conditional adaptivity, resulting in rigid, constant-width intervals.

Ultimately, retaining all three loss terms is strictly necessary to prevent both constant-width failure modes and mean interval inflation, thereby ensuring that the resulting intervals are valid, stable, and appropriately adaptive to instance-specific difficulty.

\begin{table*}[ht]
\caption{Ablation Study: Impact of Validity and Efficiency loss terms on SPACR's performance.}
\centering\scriptsize
\setlength{\tabcolsep}{3pt}
\resizebox{\textwidth}{!}{
\begin{tabular}{lcccccccccccccc}
\toprule\textbf{Dataset} & \textbf{Variant} & \textbf{MAE}$\downarrow$ & \multicolumn{4}{c}{$\alpha = 0.10$ (90\%)} & \multicolumn{4}{c}{$\alpha = 0.05$ (95\%)} & \multicolumn{4}{c}{$\alpha = 0.01$ (99\%)} \\
\cmidrule(lr){4-7} \cmidrule(lr){8-11} \cmidrule(lr){12-15}
& & & Cov.(\%) & Med. Eff.$\downarrow$ & Mean Eff.$\downarrow$ & IQR$\downarrow$ & Cov.(\%) & Med. Eff.$\downarrow$ & Mean Eff.$\downarrow$ & IQR$\downarrow$ & Cov.(\%) & Med. Eff.$\downarrow$ & Mean Eff.$\downarrow$ & IQR$\downarrow$ \\ \midrule
\multirow{3}{*}{\begin{tabular}{@{}l@{}}\textbf{Bike} \\ \textbf{Sharing}\end{tabular}} & SPACR\_VAL & 0.34 $\pm$ 0.01 & 89.96 $\pm$ 0.69 & 1.59 $\pm$ 0.12 & 1.59 $\pm$ 0.11 & 0.44 $\pm$ 0.18 & 94.95 $\pm$ 0.42 & 2.02 $\pm$ 0.17 & 2.02 $\pm$ 0.16 & 0.56 $\pm$ 0.24 & 99.00 $\pm$ 0.21 & 3.19 $\pm$ 0.31 & 3.19 $\pm$ 0.30 & 0.88 $\pm$ 0.39 \\
 & SPACR\_EFF & 0.31 $\pm$ 0.01 & 90.23 $\pm$ 0.20 & 1.58 $\pm$ 0.15 & 2.48 $\pm$ 0.10 & 1.98 $\pm$ 0.14 & 95.51 $\pm$ 0.53 & 2.23 $\pm$ 0.30 & 3.48 $\pm$ 0.15 & 2.79 $\pm$ 0.23 & 99.09 $\pm$ 0.14 & 3.96 $\pm$ 1.25 & 6.08 $\pm$ 1.23 & 4.87 $\pm$ 1.01 \\
 & \cellcolor{gray!20} SPACR & \cellcolor{gray!20} 0.32 $\pm$ 0.02 & \cellcolor{gray!20} 90.06 $\pm$ 0.52 & \cellcolor{gray!20} 1.29 $\pm$ 0.09 & \cellcolor{gray!20} 1.35 $\pm$ 0.08 & \cellcolor{gray!20} 0.49 $\pm$ 0.06 & \cellcolor{gray!20} 95.09 $\pm$ 0.41 & \cellcolor{gray!20} 1.54 $\pm$ 0.11 & \cellcolor{gray!20} 1.60 $\pm$ 0.10 & \cellcolor{gray!20} 0.58 $\pm$ 0.07 & \cellcolor{gray!20} 99.02 $\pm$ 0.22 & \cellcolor{gray!20} 2.07 $\pm$ 0.09 & \cellcolor{gray!20} 2.16 $\pm$ 0.08 & \cellcolor{gray!20} 0.79 $\pm$ 0.12 \\
\midrule
\multirow{3}{*}{\begin{tabular}{@{}l@{}}\textbf{Brazilian} \\ \textbf{Houses}\end{tabular}} & SPACR\_VAL & 0.19 $\pm$ 0.03 & 90.40 $\pm$ 0.46 & 0.71 $\pm$ 0.10 & $(2.78 \pm 6.21) \times 10^{11}$ & 0.27 $\pm$ 0.16 & 95.32 $\pm$ 0.16 & 0.86 $\pm$ 0.12 & $(3.38 \pm 7.56) \times 10^{11}$ & 0.33 $\pm$ 0.20 & 99.13 $\pm$ 0.26 & 1.25 $\pm$ 0.15 & $(4.78 \pm 10.70) \times 10^{11}$ & 0.48 $\pm$ 0.28 \\
 & SPACR\_EFF & 0.13 $\pm$ 0.02 & 89.92 $\pm$ 0.86 & 0.73 $\pm$ 0.19 & 0.78 $\pm$ 0.21 & 0.59 $\pm$ 0.18 & 95.16 $\pm$ 0.64 & 0.99 $\pm$ 0.29 & 1.06 $\pm$ 0.32 & 0.81 $\pm$ 0.28 & 99.08 $\pm$ 0.24 & 1.68 $\pm$ 0.47 & 1.80 $\pm$ 0.51 & 1.38 $\pm$ 0.48 \\
 & \cellcolor{gray!20} SPACR & \cellcolor{gray!20} 0.34 $\pm$ 0.03 & \cellcolor{gray!20} 89.86 $\pm$ 0.69 & \cellcolor{gray!20} 1.07 $\pm$ 0.12 & \cellcolor{gray!20} $(8.19 \pm 18.30) \times 10^2$ & \cellcolor{gray!20} 0.08 $\pm$ 0.01 & \cellcolor{gray!20} 95.00 $\pm$ 0.88 & \cellcolor{gray!20} 1.22 $\pm$ 0.15 & \cellcolor{gray!20} $(9.18 \pm 20.50) \times 10^2$ & \cellcolor{gray!20} 0.09 $\pm$ 0.01 & \cellcolor{gray!20} 99.21 $\pm$ 0.22 & \cellcolor{gray!20} 1.59 $\pm$ 0.18 & \cellcolor{gray!20} $(1.20 \pm 2.69) \times 10^3$ & \cellcolor{gray!20} 0.12 $\pm$ 0.02 \\
\midrule
\multirow{3}{*}{\textbf{California}} & SPACR\_VAL & 0.06 $\pm$ 0.00 & 90.40 $\pm$ 0.80 & 0.28 $\pm$ 0.01 & 0.28 $\pm$ 0.01 & 0.00 $\pm$ 0.00 & 95.24 $\pm$ 0.62 & 0.36 $\pm$ 0.01 & 0.36 $\pm$ 0.01 & 0.00 $\pm$ 0.00 & 99.03 $\pm$ 0.16 & 0.56 $\pm$ 0.01 & 0.56 $\pm$ 0.01 & 0.00 $\pm$ 0.00 \\
 & SPACR\_EFF & 0.07 $\pm$ 0.00 & 90.47 $\pm$ 1.04 & 0.30 $\pm$ 0.01 & 0.30 $\pm$ 0.01 & 0.06 $\pm$ 0.01 & 95.42 $\pm$ 0.75 & 0.39 $\pm$ 0.02 & 0.39 $\pm$ 0.02 & 0.07 $\pm$ 0.02 & 99.11 $\pm$ 0.21 & 0.66 $\pm$ 0.03 & 0.67 $\pm$ 0.03 & 0.12 $\pm$ 0.02 \\
 & \cellcolor{gray!20} SPACR & \cellcolor{gray!20} 0.06 $\pm$ 0.00 & \cellcolor{gray!20} 90.06 $\pm$ 0.94 & \cellcolor{gray!20} 0.26 $\pm$ 0.01 & \cellcolor{gray!20} 0.26 $\pm$ 0.01 & \cellcolor{gray!20} 0.06 $\pm$ 0.00 & \cellcolor{gray!20} 95.04 $\pm$ 0.57 & \cellcolor{gray!20} 0.33 $\pm$ 0.01 & \cellcolor{gray!20} 0.32 $\pm$ 0.01 & \cellcolor{gray!20} 0.07 $\pm$ 0.00 & \cellcolor{gray!20} 99.02 $\pm$ 0.15 & \cellcolor{gray!20} 0.53 $\pm$ 0.03 & \cellcolor{gray!20} 0.52 $\pm$ 0.03 & \cellcolor{gray!20} 0.12 $\pm$ 0.01 \\
\midrule
\multirow{3}{*}{\begin{tabular}{@{}l@{}}\textbf{CPU} \\ \textbf{Act}\end{tabular}} & SPACR\_VAL & 0.06 $\pm$ 0.01 & 89.98 $\pm$ 1.18 & 0.23 $\pm$ 0.05 & 0.24 $\pm$ 0.05 & 0.04 $\pm$ 0.01 & 94.95 $\pm$ 0.15 & 0.32 $\pm$ 0.07 & 0.34 $\pm$ 0.06 & 0.06 $\pm$ 0.01 & 99.07 $\pm$ 0.38 & 1.33 $\pm$ 0.82 & 1.38 $\pm$ 0.82 & 0.22 $\pm$ 0.10 \\
 & SPACR\_EFF & 0.04 $\pm$ 0.01 & 89.47 $\pm$ 1.39 & 0.15 $\pm$ 0.01 & 0.19 $\pm$ 0.02 & 0.04 $\pm$ 0.01 & 94.61 $\pm$ 1.32 & 0.21 $\pm$ 0.01 & 0.26 $\pm$ 0.03 & 0.06 $\pm$ 0.01 & 99.04 $\pm$ 0.29 & 1.06 $\pm$ 0.28 & 1.31 $\pm$ 0.30 & 0.31 $\pm$ 0.05 \\
 & \cellcolor{gray!20} SPACR & \cellcolor{gray!20} 0.08 $\pm$ 0.02 & \cellcolor{gray!20} 89.96 $\pm$ 0.52 & \cellcolor{gray!20} 0.24 $\pm$ 0.05 & \cellcolor{gray!20} 0.28 $\pm$ 0.06 & \cellcolor{gray!20} 0.06 $\pm$ 0.02 & \cellcolor{gray!20} 94.78 $\pm$ 0.71 & \cellcolor{gray!20} 0.27 $\pm$ 0.05 & \cellcolor{gray!20} 0.32 $\pm$ 0.06 & \cellcolor{gray!20} 0.07 $\pm$ 0.02 & \cellcolor{gray!20} 99.13 $\pm$ 0.15 & \cellcolor{gray!20} 0.37 $\pm$ 0.05 & \cellcolor{gray!20} 0.44 $\pm$ 0.06 & \cellcolor{gray!20} 0.10 $\pm$ 0.02 \\
\midrule
\multirow{3}{*}{\textbf{Diamonds}} & SPACR\_VAL & 0.09 $\pm$ 0.00 & 89.93 $\pm$ 0.28 & 0.38 $\pm$ 0.00 & 0.39 $\pm$ 0.01 & 0.05 $\pm$ 0.01 & 95.02 $\pm$ 0.13 & 0.47 $\pm$ 0.00 & 0.48 $\pm$ 0.00 & 0.06 $\pm$ 0.02 & 99.00 $\pm$ 0.11 & 0.66 $\pm$ 0.01 & 0.67 $\pm$ 0.01 & 0.08 $\pm$ 0.02 \\
 & SPACR\_EFF & 0.09 $\pm$ 0.00 & 89.89 $\pm$ 0.46 & 0.54 $\pm$ 0.02 & 0.63 $\pm$ 0.04 & 0.62 $\pm$ 0.08 & 94.98 $\pm$ 0.31 & 0.84 $\pm$ 0.05 & 0.97 $\pm$ 0.08 & 0.97 $\pm$ 0.13 & 99.09 $\pm$ 0.09 & 1.76 $\pm$ 0.07 & 2.05 $\pm$ 0.11 & 2.03 $\pm$ 0.21 \\
 & \cellcolor{gray!20} SPACR & \cellcolor{gray!20} 0.09 $\pm$ 0.00 & \cellcolor{gray!20} 90.05 $\pm$ 0.36 & \cellcolor{gray!20} 0.34 $\pm$ 0.00 & \cellcolor{gray!20} 0.35 $\pm$ 0.00 & \cellcolor{gray!20} 0.06 $\pm$ 0.00 & \cellcolor{gray!20} 95.02 $\pm$ 0.11 & \cellcolor{gray!20} 0.42 $\pm$ 0.00 & \cellcolor{gray!20} 0.42 $\pm$ 0.00 & \cellcolor{gray!20} 0.07 $\pm$ 0.01 & \cellcolor{gray!20} 99.00 $\pm$ 0.08 & \cellcolor{gray!20} 0.60 $\pm$ 0.01 & \cellcolor{gray!20} 0.61 $\pm$ 0.01 & \cellcolor{gray!20} 0.10 $\pm$ 0.01 \\
\midrule
\multirow{3}{*}{\textbf{Fifa}} & SPACR\_VAL & 0.07 $\pm$ 0.00 & 90.15 $\pm$ 0.68 & 0.30 $\pm$ 0.01 & 0.30 $\pm$ 0.01 & 0.02 $\pm$ 0.00 & 95.13 $\pm$ 0.63 & 0.43 $\pm$ 0.02 & 0.43 $\pm$ 0.02 & 0.02 $\pm$ 0.01 & 99.06 $\pm$ 0.23 & 0.64 $\pm$ 0.02 & 0.64 $\pm$ 0.02 & 0.04 $\pm$ 0.01 \\
 & SPACR\_EFF & 0.07 $\pm$ 0.00 & 90.17 $\pm$ 0.59 & 0.34 $\pm$ 0.02 & 0.32 $\pm$ 0.02 & 0.06 $\pm$ 0.02 & 95.23 $\pm$ 0.20 & 0.49 $\pm$ 0.02 & 0.47 $\pm$ 0.02 & 0.08 $\pm$ 0.03 & 99.18 $\pm$ 0.08 & 0.97 $\pm$ 0.22 & 0.93 $\pm$ 0.20 & 0.16 $\pm$ 0.08 \\
 & \cellcolor{gray!20} SPACR & \cellcolor{gray!20} 0.07 $\pm$ 0.00 & \cellcolor{gray!20} 90.27 $\pm$ 0.51 & \cellcolor{gray!20} 0.28 $\pm$ 0.00 & \cellcolor{gray!20} 0.29 $\pm$ 0.00 & \cellcolor{gray!20} 0.05 $\pm$ 0.00 & \cellcolor{gray!20} 94.90 $\pm$ 0.57 & \cellcolor{gray!20} 0.36 $\pm$ 0.01 & \cellcolor{gray!20} 0.36 $\pm$ 0.00 & \cellcolor{gray!20} 0.07 $\pm$ 0.01 & \cellcolor{gray!20} 98.93 $\pm$ 0.16 & \cellcolor{gray!20} 0.61 $\pm$ 0.04 & \cellcolor{gray!20} 0.62 $\pm$ 0.04 & \cellcolor{gray!20} 0.12 $\pm$ 0.02 \\
\midrule
\multirow{3}{*}{\begin{tabular}{@{}l@{}}\textbf{House} \\ \textbf{Sales}\end{tabular}} & SPACR\_VAL & 0.14 $\pm$ 0.00 & 89.83 $\pm$ 0.91 & 0.57 $\pm$ 0.01 & 0.58 $\pm$ 0.01 & 0.06 $\pm$ 0.01 & 94.79 $\pm$ 0.35 & 0.73 $\pm$ 0.01 & 0.74 $\pm$ 0.01 & 0.08 $\pm$ 0.01 & 98.76 $\pm$ 0.24 & 1.12 $\pm$ 0.04 & 1.13 $\pm$ 0.04 & 0.12 $\pm$ 0.02 \\
 & SPACR\_EFF & 0.13 $\pm$ 0.00 & 89.72 $\pm$ 0.68 & 0.60 $\pm$ 0.02 & 0.61 $\pm$ 0.02 & 0.20 $\pm$ 0.01 & 94.80 $\pm$ 0.37 & 0.79 $\pm$ 0.03 & 0.80 $\pm$ 0.03 & 0.26 $\pm$ 0.01 & 98.92 $\pm$ 0.26 & 1.27 $\pm$ 0.06 & 1.30 $\pm$ 0.06 & 0.42 $\pm$ 0.02 \\
 & \cellcolor{gray!20} SPACR & \cellcolor{gray!20} 0.14 $\pm$ 0.01 & \cellcolor{gray!20} 89.83 $\pm$ 1.04 & \cellcolor{gray!20} 0.55 $\pm$ 0.02 & \cellcolor{gray!20} 0.57 $\pm$ 0.02 & \cellcolor{gray!20} 0.10 $\pm$ 0.01 & \cellcolor{gray!20} 94.74 $\pm$ 0.41 & \cellcolor{gray!20} 0.68 $\pm$ 0.03 & \cellcolor{gray!20} 0.70 $\pm$ 0.02 & \cellcolor{gray!20} 0.12 $\pm$ 0.01 & \cellcolor{gray!20} 98.87 $\pm$ 0.31 & \cellcolor{gray!20} 1.00 $\pm$ 0.03 & \cellcolor{gray!20} 1.03 $\pm$ 0.04 & \cellcolor{gray!20} 0.18 $\pm$ 0.01 \\
\midrule
\multirow{3}{*}{\textbf{Isolet}} & SPACR\_VAL & 0.21 $\pm$ 0.01 & 89.82 $\pm$ 1.36 & 0.92 $\pm$ 0.02 & 0.95 $\pm$ 0.02 & 0.27 $\pm$ 0.05 & 94.76 $\pm$ 1.26 & 1.26 $\pm$ 0.08 & 1.31 $\pm$ 0.08 & 0.37 $\pm$ 0.07 & 99.23 $\pm$ 0.26 & 2.39 $\pm$ 0.22 & 2.47 $\pm$ 0.25 & 0.70 $\pm$ 0.10 \\
 & SPACR\_EFF & 0.20 $\pm$ 0.01 & 90.32 $\pm$ 0.93 & 1.38 $\pm$ 0.25 & 2.05 $\pm$ 0.44 & 1.88 $\pm$ 0.52 & 95.27 $\pm$ 0.71 & 2.08 $\pm$ 0.32 & 3.09 $\pm$ 0.61 & 2.82 $\pm$ 0.70 & 99.06 $\pm$ 0.35 & 4.04 $\pm$ 0.66 & 6.02 $\pm$ 1.27 & 5.50 $\pm$ 1.43 \\
 & \cellcolor{gray!20} SPACR & \cellcolor{gray!20} 0.20 $\pm$ 0.01 & \cellcolor{gray!20} 89.05 $\pm$ 1.26 & \cellcolor{gray!20} 0.78 $\pm$ 0.06 & \cellcolor{gray!20} 0.85 $\pm$ 0.07 & \cellcolor{gray!20} 0.43 $\pm$ 0.03 & \cellcolor{gray!20} 94.54 $\pm$ 0.70 & \cellcolor{gray!20} 1.01 $\pm$ 0.07 & \cellcolor{gray!20} 1.09 $\pm$ 0.07 & \cellcolor{gray!20} 0.56 $\pm$ 0.05 & \cellcolor{gray!20} 99.13 $\pm$ 0.28 & \cellcolor{gray!20} 1.61 $\pm$ 0.11 & \cellcolor{gray!20} 1.75 $\pm$ 0.11 & \cellcolor{gray!20} 0.90 $\pm$ 0.09 \\
\midrule
\multirow{3}{*}{\begin{tabular}{@{}l@{}}\textbf{Medical} \\ \textbf{Charges}\end{tabular}} & SPACR\_VAL & 0.04 $\pm$ 0.00 & 89.95 $\pm$ 0.18 & 0.18 $\pm$ 0.00 & 0.18 $\pm$ 0.00 & 0.00 $\pm$ 0.00 & 94.93 $\pm$ 0.16 & 0.27 $\pm$ 0.01 & 0.27 $\pm$ 0.01 & 0.01 $\pm$ 0.00 & 99.01 $\pm$ 0.10 & 0.58 $\pm$ 0.03 & 0.59 $\pm$ 0.03 & 0.01 $\pm$ 0.01 \\
 & SPACR\_EFF & 0.04 $\pm$ 0.00 & 90.03 $\pm$ 0.22 & 0.22 $\pm$ 0.02 & 0.21 $\pm$ 0.02 & 0.06 $\pm$ 0.04 & 94.98 $\pm$ 0.22 & 0.35 $\pm$ 0.05 & 0.34 $\pm$ 0.05 & 0.10 $\pm$ 0.06 & 98.96 $\pm$ 0.10 & 0.93 $\pm$ 0.29 & 0.89 $\pm$ 0.28 & 0.28 $\pm$ 0.24 \\
 & \cellcolor{gray!20} SPACR & \cellcolor{gray!20} 0.04 $\pm$ 0.00 & \cellcolor{gray!20} 89.90 $\pm$ 0.09 & \cellcolor{gray!20} 0.17 $\pm$ 0.00 & \cellcolor{gray!20} 0.17 $\pm$ 0.00 & \cellcolor{gray!20} 0.01 $\pm$ 0.00 & \cellcolor{gray!20} 95.00 $\pm$ 0.20 & \cellcolor{gray!20} 0.24 $\pm$ 0.00 & \cellcolor{gray!20} 0.24 $\pm$ 0.00 & \cellcolor{gray!20} 0.02 $\pm$ 0.00 & \cellcolor{gray!20} 99.00 $\pm$ 0.07 & \cellcolor{gray!20} 0.50 $\pm$ 0.01 & \cellcolor{gray!20} 0.50 $\pm$ 0.02 & \cellcolor{gray!20} 0.04 $\pm$ 0.01 \\
\midrule
\multirow{3}{*}{\textbf{Pol}} & SPACR\_VAL & 0.11 $\pm$ 0.01 & 89.99 $\pm$ 0.83 & 0.53 $\pm$ 0.05 & 3.25 $\pm$ 1.25 & 0.87 $\pm$ 0.26 & 95.05 $\pm$ 0.68 & 1.33 $\pm$ 0.08 & 8.16 $\pm$ 3.06 & 2.19 $\pm$ 0.60 & 99.05 $\pm$ 0.27 & 4.89 $\pm$ 0.34 & 30.28 $\pm$ 11.65 & 8.10 $\pm$ 2.32 \\
 & SPACR\_EFF & 0.11 $\pm$ 0.01 & 90.07 $\pm$ 0.65 & 0.48 $\pm$ 0.04 & 0.50 $\pm$ 0.06 & 0.63 $\pm$ 0.05 & 95.47 $\pm$ 0.64 & 0.84 $\pm$ 0.10 & 0.87 $\pm$ 0.06 & 1.11 $\pm$ 0.07 & 99.07 $\pm$ 0.29 & 2.17 $\pm$ 0.25 & 2.26 $\pm$ 0.29 & 2.88 $\pm$ 0.32 \\
 & \cellcolor{gray!20} SPACR & \cellcolor{gray!20} 0.09 $\pm$ 0.01 & \cellcolor{gray!20} 90.37 $\pm$ 0.43 & \cellcolor{gray!20} 0.09 $\pm$ 0.03 & \cellcolor{gray!20} 0.34 $\pm$ 0.05 & \cellcolor{gray!20} 0.33 $\pm$ 0.03 & \cellcolor{gray!20} 95.67 $\pm$ 0.37 & \cellcolor{gray!20} 0.14 $\pm$ 0.04 & \cellcolor{gray!20} 0.54 $\pm$ 0.10 & \cellcolor{gray!20} 0.51 $\pm$ 0.06 & \cellcolor{gray!20} 99.14 $\pm$ 0.17 & \cellcolor{gray!20} 0.33 $\pm$ 0.08 & \cellcolor{gray!20} 1.23 $\pm$ 0.15 & \cellcolor{gray!20} 1.17 $\pm$ 0.08 \\
\midrule
\multirow{3}{*}{\textbf{Superconduct}} & SPACR\_VAL & 0.36 $\pm$ 0.01 & 89.89 $\pm$ 0.39 & 2.00 $\pm$ 0.05 & 2.02 $\pm$ 0.05 & 0.97 $\pm$ 0.12 & 94.88 $\pm$ 0.35 & 2.65 $\pm$ 0.10 & 2.68 $\pm$ 0.10 & 1.29 $\pm$ 0.18 & 99.08 $\pm$ 0.20 & 4.16 $\pm$ 0.18 & 4.21 $\pm$ 0.18 & 2.02 $\pm$ 0.30 \\
 & SPACR\_EFF & 0.36 $\pm$ 0.01 & 90.40 $\pm$ 0.60 & 1.70 $\pm$ 0.06 & 2.81 $\pm$ 0.15 & 2.45 $\pm$ 0.23 & 95.22 $\pm$ 0.61 & 2.69 $\pm$ 0.13 & 4.45 $\pm$ 0.33 & 3.88 $\pm$ 0.47 & 99.09 $\pm$ 0.28 & 6.06 $\pm$ 0.43 & 9.99 $\pm$ 0.67 & 8.68 $\pm$ 0.62 \\
 & \cellcolor{gray!20} SPACR & \cellcolor{gray!20} 0.38 $\pm$ 0.01 & \cellcolor{gray!20} 89.79 $\pm$ 0.36 & \cellcolor{gray!20} 1.29 $\pm$ 0.04 & \cellcolor{gray!20} 1.36 $\pm$ 0.03 & \cellcolor{gray!20} 0.47 $\pm$ 0.02 & \cellcolor{gray!20} 95.02 $\pm$ 0.27 & \cellcolor{gray!20} 1.69 $\pm$ 0.05 & \cellcolor{gray!20} 1.78 $\pm$ 0.05 & \cellcolor{gray!20} 0.62 $\pm$ 0.03 & \cellcolor{gray!20} 98.83 $\pm$ 0.16 & \cellcolor{gray!20} 3.03 $\pm$ 0.11 & \cellcolor{gray!20} 3.19 $\pm$ 0.14 & \cellcolor{gray!20} 1.11 $\pm$ 0.06 \\
\midrule
\multirow{3}{*}{\begin{tabular}{@{}l@{}}\textbf{Wine} \\ \textbf{Quality}\end{tabular}} & SPACR\_VAL & 0.08 $\pm$ 0.00 & 90.45 $\pm$ 1.29 & 0.35 $\pm$ 0.01 & 0.35 $\pm$ 0.01 & 0.05 $\pm$ 0.02 & 94.95 $\pm$ 0.79 & 0.44 $\pm$ 0.02 & 0.44 $\pm$ 0.02 & 0.06 $\pm$ 0.02 & 98.77 $\pm$ 0.34 & 0.63 $\pm$ 0.05 & 0.64 $\pm$ 0.05 & 0.08 $\pm$ 0.04 \\
 & SPACR\_EFF & 0.08 $\pm$ 0.00 & 90.38 $\pm$ 1.47 & 0.40 $\pm$ 0.02 & 0.41 $\pm$ 0.03 & 0.19 $\pm$ 0.03 & 95.42 $\pm$ 0.32 & 0.56 $\pm$ 0.03 & 0.57 $\pm$ 0.04 & 0.26 $\pm$ 0.05 & 98.95 $\pm$ 0.46 & 1.25 $\pm$ 0.17 & 1.28 $\pm$ 0.17 & 0.58 $\pm$ 0.08 \\
 & \cellcolor{gray!20} SPACR & \cellcolor{gray!20} 0.08 $\pm$ 0.00 & \cellcolor{gray!20} 90.48 $\pm$ 0.85 & \cellcolor{gray!20} 0.35 $\pm$ 0.01 & \cellcolor{gray!20} 0.35 $\pm$ 0.01 & \cellcolor{gray!20} 0.04 $\pm$ 0.01 & \cellcolor{gray!20} 95.31 $\pm$ 0.80 & \cellcolor{gray!20} 0.43 $\pm$ 0.01 & \cellcolor{gray!20} 0.43 $\pm$ 0.01 & \cellcolor{gray!20} 0.05 $\pm$ 0.01 & \cellcolor{gray!20} 99.02 $\pm$ 0.33 & \cellcolor{gray!20} 0.65 $\pm$ 0.05 & \cellcolor{gray!20} 0.65 $\pm$ 0.05 & \cellcolor{gray!20} 0.07 $\pm$ 0.01 \\
\midrule
\multirow{3}{*}{\textbf{Drift}} & SPACR\_VAL & 7.84 $\pm$ 1.68 & 89.81 $\pm$ 0.82 & 30.70 $\pm$ 8.50 & $(3.57 \pm 3.09) \times 10^3$ & 1.85 $\pm$ 2.09 & 94.77 $\pm$ 0.53 & 40.32 $\pm$ 11.34 & $(4.74 \pm 4.17) \times 10^3$ & 2.44 $\pm$ 2.76 & 99.11 $\pm$ 0.20 & 71.69 $\pm$ 18.68 & $(8.40 \pm 7.44) \times 10^3$ & 4.43 $\pm$ 5.08 \\
 & SPACR\_EFF & 6.93 $\pm$ 0.94 & 90.14 $\pm$ 0.72 & 96.38 $\pm$ 22.39 & 420.76 $\pm$ 62.71 & 376.23 $\pm$ 74.19 & 94.96 $\pm$ 0.80 & 144.93 $\pm$ 28.60 & 634.72 $\pm$ 79.02 & 566.86 $\pm$ 96.00 & 99.16 $\pm$ 0.25 & 274.57 $\pm$ 33.71 & $(1.21 \pm 0.14) \times 10^3$ & $(1.08 \pm 0.15) \times 10^3$ \\
 & \cellcolor{gray!20} SPACR & \cellcolor{gray!20} 6.00 $\pm$ 0.36 & \cellcolor{gray!20} 90.03 $\pm$ 1.29 & \cellcolor{gray!20} 22.51 $\pm$ 2.28 & \cellcolor{gray!20} 24.72 $\pm$ 2.46 & \cellcolor{gray!20} 13.04 $\pm$ 3.21 & \cellcolor{gray!20} 95.17 $\pm$ 0.71 & \cellcolor{gray!20} 27.59 $\pm$ 2.98 & \cellcolor{gray!20} 30.31 $\pm$ 3.34 & \cellcolor{gray!20} 16.02 $\pm$ 4.19 & \cellcolor{gray!20} 99.27 $\pm$ 0.28 & \cellcolor{gray!20} 41.19 $\pm$ 5.90 & \cellcolor{gray!20} 45.28 $\pm$ 6.74 & \cellcolor{gray!20} 24.02 $\pm$ 7.20 \\
\midrule
\multirow{3}{*}{\begin{tabular}{@{}l@{}}\textbf{UTK} \\ \textbf{Face}\end{tabular}} & SPACR\_VAL & 6.51 $\pm$ 0.81 & 90.20 $\pm$ 0.62 & 21.63 $\pm$ 3.95 & 347.25 $\pm$ 136.49 & 24.15 $\pm$ 15.76 & 94.97 $\pm$ 0.29 & 27.87 $\pm$ 5.06 & 446.24 $\pm$ 169.93 & 31.32 $\pm$ 20.44 & 98.98 $\pm$ 0.11 & 44.15 $\pm$ 6.77 & 701.95 $\pm$ 244.54 & 49.69 $\pm$ 31.53 \\
 & SPACR\_EFF & 6.87 $\pm$ 1.45 & 89.81 $\pm$ 0.73 & 127.94 $\pm$ 54.67 & 272.55 $\pm$ 29.87 & 180.93 $\pm$ 79.69 & 94.63 $\pm$ 0.44 & 210.41 $\pm$ 101.23 & 439.67 $\pm$ 36.56 & 296.36 $\pm$ 144.85 & 98.84 $\pm$ 0.27 & 391.71 $\pm$ 196.21 & 812.50 $\pm$ 70.12 & 551.08 $\pm$ 279.93 \\
 & \cellcolor{gray!20} SPACR & \cellcolor{gray!20} 5.43 $\pm$ 0.18 & \cellcolor{gray!20} 90.13 $\pm$ 0.46 & \cellcolor{gray!20} 21.84 $\pm$ 0.84 & \cellcolor{gray!20} 22.54 $\pm$ 0.43 & \cellcolor{gray!20} 12.18 $\pm$ 1.61 & \cellcolor{gray!20} 95.02 $\pm$ 0.30 & \cellcolor{gray!20} 27.49 $\pm$ 0.98 & \cellcolor{gray!20} 28.39 $\pm$ 0.91 & \cellcolor{gray!20} 15.36 $\pm$ 2.26 & \cellcolor{gray!20} 98.95 $\pm$ 0.14 & \cellcolor{gray!20} 42.51 $\pm$ 2.06 & \cellcolor{gray!20} 43.93 $\pm$ 2.76 & \cellcolor{gray!20} 23.82 $\pm$ 4.13 \\
\bottomrule\end{tabular}}
\label{tab:ablation}\end{table*}

\subsection{Computation Time Analysis}

Experiments were conducted on Ubuntu 24.04.3 LTS with an Intel Core i9-14900KF CPU, an NVIDIA GeForce RTX 5090 GPU, and 32GB of RAM. The timing results are shown in Table~\ref{tab:timing}.

\begin{table*}[ht]
\caption{Training time (in seconds) across datasets. "--" indicates no retraining needed.}
\centering
\scriptsize
\resizebox{\textwidth}{!}{
\begin{tabular}{llcccc|llcccc}
\toprule
\textbf{Dataset} & \textbf{Method} & $\alpha = 0.01$ & $\alpha = 0.05$ & $\alpha = 0.10$ & \textbf{Total (s)} & \textbf{Dataset} & \textbf{Method} & $\alpha = 0.01$ & $\alpha = 0.05$ & $\alpha = 0.10$ & \textbf{Total (s)}\\
\midrule
\multirow{5}{*}{\begin{tabular}{@{}l@{}}\textbf{Bike} \\ \textbf{Sharing}\end{tabular}}
& SICP & 28.80 $\pm$ 0.34 & -- & -- & 28.80 $\pm$ 0.34 &
\multirow{5}{*}{\begin{tabular}{@{}l@{}}\textbf{Brazilian} \\ \textbf{Houses}\end{tabular}}
& SICP & 19.93 $\pm$ 0.24 & -- & -- & 19.93 $\pm$ 0.24 \\
& NICP & 31.29 $\pm$ 0.62 & -- & -- & 31.29 $\pm$ 0.62 &
& NICP & 21.33 $\pm$ 0.40 & -- & -- & 21.33 $\pm$ 0.40 \\
& CQR & 34.31 $\pm$ 0.39 & 34.27 $\pm$ 0.45 & 33.94 $\pm$ 0.70 & 102.51 $\pm$ 1.05 &
& CQR & 22.86 $\pm$ 0.20 & 22.84 $\pm$ 0.68 & 22.58 $\pm$ 0.47 & 68.29 $\pm$ 0.95 \\
& DOICR & 35.42 $\pm$ 0.56 & 35.10 $\pm$ 0.66 & 35.19 $\pm$ 0.86 & 105.71 $\pm$ 1.35 &
& DOICR & 23.42 $\pm$ 0.60 & 23.76 $\pm$ 0.22 & 23.78 $\pm$ 0.48 & 70.96 $\pm$ 1.06 \\
& \cellcolor{gray!20}SPACR & \cellcolor{gray!20}32.80 $\pm$ 0.83 & \cellcolor{gray!20}-- & \cellcolor{gray!20}-- & \cellcolor{gray!20}32.80 $\pm$ 0.83 &
& \cellcolor{gray!20}SPACR & \cellcolor{gray!20}22.12 $\pm$ 0.47 & \cellcolor{gray!20}-- & \cellcolor{gray!20}-- & \cellcolor{gray!20}22.12 $\pm$ 0.47 \\
\midrule
\multirow{5}{*}{\textbf{California}}
& SICP & 33.96 $\pm$ 0.75 & -- & -- & 33.96 $\pm$ 0.75 &
\multirow{5}{*}{\begin{tabular}{@{}l@{}}\textbf{CPU} \\ \textbf{Act}\end{tabular}}
& SICP & 16.38 $\pm$ 0.44 & -- & -- & 16.38 $\pm$ 0.44 \\
& NICP & 36.76 $\pm$ 0.39 & -- & -- & 36.76 $\pm$ 0.39 &
& NICP & 17.17 $\pm$ 0.31 & -- & -- & 17.17 $\pm$ 0.31 \\
& CQR & 39.89 $\pm$ 0.95 & 39.94 $\pm$ 0.34 & 39.62 $\pm$ 0.65 & 119.45 $\pm$ 0.85 &
& CQR & 18.39 $\pm$ 0.44 & 18.65 $\pm$ 0.16 & 18.50 $\pm$ 0.35 & 55.54 $\pm$ 0.59 \\
& DOICR & 40.81 $\pm$ 0.39 & 41.52 $\pm$ 0.88 & 40.56 $\pm$ 0.23 & 122.89 $\pm$ 1.15 &
& DOICR & 19.24 $\pm$ 0.48 & 19.07 $\pm$ 0.29 & 19.20 $\pm$ 0.31 & 57.52 $\pm$ 0.33 \\
& \cellcolor{gray!20}SPACR & \cellcolor{gray!20}37.99 $\pm$ 0.62 & \cellcolor{gray!20}-- & \cellcolor{gray!20}-- & \cellcolor{gray!20}37.99 $\pm$ 0.62 &
& \cellcolor{gray!20}SPACR & \cellcolor{gray!20}17.38 $\pm$ 0.31 & \cellcolor{gray!20}-- & \cellcolor{gray!20}-- & \cellcolor{gray!20}17.38 $\pm$ 0.31 \\
\midrule
\multirow{5}{*}{\textbf{Diamonds}}
& SICP & 83.49 $\pm$ 2.05 & -- & -- & 83.49 $\pm$ 2.05 &
\multirow{5}{*}{\textbf{Fifa}}
& SICP & 30.08 $\pm$ 0.44 & -- & -- & 30.08 $\pm$ 0.44 \\
& NICP & 91.84 $\pm$ 2.58 & -- & -- & 91.84 $\pm$ 2.58 &
& NICP & 33.12 $\pm$ 0.65 & -- & -- & 33.12 $\pm$ 0.65 \\
& CQR & 97.91 $\pm$ 1.37 & 99.97 $\pm$ 3.44 & 99.12 $\pm$ 3.17 & 297.00 $\pm$ 7.00 &
& CQR & 35.28 $\pm$ 0.62 & 36.05 $\pm$ 1.59 & 35.24 $\pm$ 0.45 & 106.57 $\pm$ 1.41 \\
& DOICR & 104.67 $\pm$ 3.23 & 103.97 $\pm$ 4.24 & 105.06 $\pm$ 4.45 & 313.70 $\pm$ 11.53 &
& DOICR & 38.11 $\pm$ 3.86 & 37.19 $\pm$ 0.78 & 37.15 $\pm$ 0.43 & 112.46 $\pm$ 4.19 \\
& \cellcolor{gray!20}SPACR & \cellcolor{gray!20}95.11 $\pm$ 2.85 & \cellcolor{gray!20}-- & \cellcolor{gray!20}-- & \cellcolor{gray!20}95.11 $\pm$ 2.85 &
& \cellcolor{gray!20}SPACR & \cellcolor{gray!20}33.55 $\pm$ 0.83 & \cellcolor{gray!20}-- & \cellcolor{gray!20}-- & \cellcolor{gray!20}33.55 $\pm$ 0.83 \\
\midrule
\multirow{5}{*}{\begin{tabular}{@{}l@{}}\textbf{House} \\ \textbf{Sales}\end{tabular}}
& SICP & 36.04 $\pm$ 0.51 & -- & -- & 36.04 $\pm$ 0.51 &
\multirow{5}{*}{\textbf{Isolet}}
& SICP & 22.42 $\pm$ 1.09 & -- & -- & 22.42 $\pm$ 1.09 \\
& NICP & 39.00 $\pm$ 1.18 & -- & -- & 39.00 $\pm$ 1.18 &
& NICP & 23.91 $\pm$ 1.09 & -- & -- & 23.91 $\pm$ 1.09 \\
& CQR & 41.81 $\pm$ 1.04 & 42.34 $\pm$ 1.05 & 41.94 $\pm$ 0.83 & 126.09 $\pm$ 1.70 &
& CQR & 26.62 $\pm$ 0.59 & 26.49 $\pm$ 0.79 & 26.60 $\pm$ 0.85 & 79.70 $\pm$ 1.54 \\
& DOICR & 43.69 $\pm$ 0.97 & 43.72 $\pm$ 0.69 & 43.67 $\pm$ 0.25 & 131.08 $\pm$ 1.46 &
& DOICR & 27.49 $\pm$ 1.02 & 28.38 $\pm$ 0.94 & 28.05 $\pm$ 1.46 & 83.92 $\pm$ 0.35 \\
& \cellcolor{gray!20}SPACR & \cellcolor{gray!20}40.38 $\pm$ 0.95 & \cellcolor{gray!20}-- & \cellcolor{gray!20}-- & \cellcolor{gray!20}40.38 $\pm$ 0.95 &
& \cellcolor{gray!20}SPACR & \cellcolor{gray!20}25.86 $\pm$ 1.38 & \cellcolor{gray!20}-- & \cellcolor{gray!20}-- & \cellcolor{gray!20}25.86 $\pm$ 1.38 \\
\midrule
\multirow{5}{*}{\begin{tabular}{@{}l@{}}\textbf{Medical} \\ \textbf{Charges}\end{tabular}}
& SICP & 494.57 $\pm$ 24.55 & -- & -- & 494.57 $\pm$ 24.55 &
\multirow{5}{*}{\textbf{Pol}}
& SICP & 30.07 $\pm$ 4.50 & -- & -- & 30.07 $\pm$ 4.50 \\
& NICP & 567.78 $\pm$ 46.81 & -- & -- & 567.78 $\pm$ 46.81 &
& NICP & 31.81 $\pm$ 2.79 & -- & -- & 31.81 $\pm$ 2.79 \\
& CQR & 573.16 $\pm$ 34.55 & 581.53 $\pm$ 33.07 & 573.08 $\pm$ 38.54 & 1727.76 $\pm$ 89.51 &
& CQR & 34.14 $\pm$ 2.44 & 32.26 $\pm$ 0.77 & 32.84 $\pm$ 0.80 & 99.24 $\pm$ 3.12 \\
& DOICR & 564.86 $\pm$ 41.04 & 554.62 $\pm$ 16.41 & 555.47 $\pm$ 29.66 & 1674.96 $\pm$ 45.57 &
& DOICR & 34.18 $\pm$ 0.29 & 33.44 $\pm$ 0.98 & 33.93 $\pm$ 1.47 & 101.55 $\pm$ 2.16 \\
& \cellcolor{gray!20}SPACR & \cellcolor{gray!20}582.27 $\pm$ 56.20 & \cellcolor{gray!20}-- & \cellcolor{gray!20}-- & \cellcolor{gray!20}582.27 $\pm$ 56.20 &
& \cellcolor{gray!20}SPACR & \cellcolor{gray!20}31.11 $\pm$ 0.96 & \cellcolor{gray!20}-- & \cellcolor{gray!20}-- & \cellcolor{gray!20}31.11 $\pm$ 0.96 \\
\midrule
\multirow{5}{*}{\textbf{Superconduct}}
& SICP & 35.00 $\pm$ 0.55 & -- & -- & 35.00 $\pm$ 0.55 &
\multirow{5}{*}{\begin{tabular}{@{}l@{}}\textbf{Wine} \\ \textbf{Quality}\end{tabular}}
& SICP & 13.55 $\pm$ 0.28 & -- & -- & 13.55 $\pm$ 0.28 \\
& NICP & 38.61 $\pm$ 0.79 & -- & -- & 38.61 $\pm$ 0.79 &
& NICP & 14.44 $\pm$ 0.28 & -- & -- & 14.44 $\pm$ 0.28 \\
& CQR & 41.72 $\pm$ 0.28 & 41.55 $\pm$ 0.63 & 40.79 $\pm$ 0.66 & 124.06 $\pm$ 0.80 &
& CQR & 15.47 $\pm$ 0.56 & 15.49 $\pm$ 0.32 & 15.74 $\pm$ 1.12 & 46.70 $\pm$ 1.57 \\
& DOICR & 43.23 $\pm$ 0.69 & 42.89 $\pm$ 0.73 & 42.18 $\pm$ 0.51 & 128.30 $\pm$ 1.36 &
& DOICR & 16.41 $\pm$ 0.20 & 16.14 $\pm$ 0.33 & 16.06 $\pm$ 0.81 & 48.62 $\pm$ 0.53 \\
& \cellcolor{gray!20}SPACR & \cellcolor{gray!20}38.99 $\pm$ 0.82 & \cellcolor{gray!20}-- & \cellcolor{gray!20}-- & \cellcolor{gray!20}38.99 $\pm$ 0.82 &
& \cellcolor{gray!20}SPACR & \cellcolor{gray!20}14.97 $\pm$ 0.23 & \cellcolor{gray!20}-- & \cellcolor{gray!20}-- & \cellcolor{gray!20}14.97 $\pm$ 0.23 \\
\midrule
\multirow{5}{*}{\textbf{Drift}}
& SICP & 1922.86 $\pm$ 59.17 & -- & -- & 1922.86 $\pm$ 59.17 &
\multirow{5}{*}{\begin{tabular}{@{}l@{}}\textbf{UTK} \\ \textbf{Face}\end{tabular}}
& SICP & 2940.27 $\pm$ 109.72 & -- & -- & 2940.27 $\pm$ 109.72 \\
& NICP & 1918.58 $\pm$ 65.54 & -- & -- & 1918.58 $\pm$ 65.54 &
& NICP & 2934.61 $\pm$ 34.01 & -- & -- & 2934.61 $\pm$ 34.01 \\
& CQR & 1913.16 $\pm$ 52.41 & 1910.51 $\pm$ 77.94 & 1919.30 $\pm$ 115.10 & 5742.97 $\pm$ 178.93 &
& CQR & 2924.74 $\pm$ 170.04 & 2995.77 $\pm$ 40.08 & 2934.18 $\pm$ 123.56 & 8854.69 $\pm$ 223.88 \\
& DOICR & 1996.19 $\pm$ 181.77 & 2108.30 $\pm$ 334.54 & 2016.14 $\pm$ 210.32 & 6120.63 $\pm$ 643.41 &
& DOICR & 3022.59 $\pm$ 104.95 & 3015.58 $\pm$ 86.37 & 2964.23 $\pm$ 132.39 & 9002.39 $\pm$ 263.98 \\
& \cellcolor{gray!20}SPACR & \cellcolor{gray!20}1928.90 $\pm$ 69.87 & \cellcolor{gray!20}-- & \cellcolor{gray!20}-- & \cellcolor{gray!20}1928.90 $\pm$ 69.87 &
& \cellcolor{gray!20}SPACR & \cellcolor{gray!20}3024.50 $\pm$ 93.79 & \cellcolor{gray!20}-- & \cellcolor{gray!20}-- & \cellcolor{gray!20}3024.50 $\pm$ 93.79 \\
\bottomrule
\end{tabular}
}
\label{tab:timing}
\end{table*}

Timing results highlight SPACR's single-pass efficiency. While post-hoc methods (SICP, NICP) lack conformal training, approaches like CQR and DOICR scale linearly ($\mathcal{O}(k)$) by requiring retraining for each of the $k$ significance levels $\alpha$. In contrast, SPACR inherently supports multiple $\alpha$ values in constant time ($\mathcal{O}(1)$). For example, on the UTK Face dataset, SPACR requires roughly 50 minutes to evaluate three $\alpha$ values, whereas CQR and DOICR take nearly 2.5 hours, which is a $\sim 3\times$ computational cost. This efficiency gap widens in high-dimensional tasks like bounding box regression in $\mathbb{R}^4$~\citep{mukama2024copula}, making SPACR highly suited for production environments needing dynamic calibration without prohibitive retraining costs.

\section{Conclusion}
\label{sec:conclu}

In this paper, we introduced SPACR (Single-Pass Adaptive Conformal Regressor), a novel framework for directly training uncertainty-aware regressors via Conformal Prediction. By integrating accuracy, efficiency, and validity into a single differentiable loss, SPACR eliminates batch-splitting and fixed training confidence levels, which gives it a computational advantage over retraining-dependent methods. Experiments show that a single SPACR model efficiently computes intervals across arbitrary confidence levels, outperforming standard ICP, normalized ICP, CQR, and DOICR, a state-of-the-art method for training conformal regressors, by yielding tighter bounds without sacrificing coverage and while generally maintaining competitive point-prediction accuracy (MAE).

While we focused here on a specific symmetric form of $\mathcal{L}_{\text{Validity}}$, the fact that it is related to robust losses \citep{vapnik1996support} indicates that we could probably take inspiration from other robust forms of loss functions in the literature \citep{ghosh2017robust,barron2019general}. In particular, as we observed that skewed residual distributions can inflate the mean interval width compared to asymmetric methods like CQR, we could think of including some asymmetry in those losses. This has been done for the pinball loss function \citep{huang2013support}, for which some proposals already exist \citep{hu2012online}. However, this would probably require a better understanding of the interplay between the different losses used in our proposal, an aspect we intend to tackle in future work.


\bibliography{main}

@article{hu2012online,
  title={Online learning for quantile regression and support vector regression},
  author={Hu, Ting and Xiang, Dao-Hong and Zhou, Ding-Xuan},
  journal={Journal of Statistical Planning and Inference},
  volume={142},
  number={12},
  pages={3107--3122},
  year={2012},
  publisher={Elsevier}
}

@article{romano2019conformalized,
  title={Conformalized quantile regression},
  author={Romano, Yaniv and Patterson, Evan and Candes, Emmanuel},
  journal={Advances in neural information processing systems},
  volume={32},
  year={2019}
}

@article{huang2013support,
  title={Support vector machine classifier with pinball loss},
  author={Huang, Xiaolin and Shi, Lei and Suykens, Johan AK},
  journal={IEEE transactions on pattern analysis and machine intelligence},
  volume={36},
  number={5},
  pages={984--997},
  year={2013},
  publisher={IEEE}
}

@article{grinsztajn2022tree,
  title={Why do tree-based models still outperform deep learning on typical tabular data?},
  author={Grinsztajn, L{\'e}o and Oyallon, Edouard and Varoquaux, Ga{\"e}l},
  journal={Advances in neural information processing systems},
  volume={35},
  pages={507--520},
  year={2022}
}

@inproceedings{ghosh2017robust,
  title={Robust loss functions under label noise for deep neural networks},
  author={Ghosh, Aritra and Kumar, Himanshu and Sastry, P Shanti},
  booktitle={Proceedings of the AAAI conference on artificial intelligence},
  volume={31},
  year={2017}
}

@inproceedings{barron2019general,
  title={A general and adaptive robust loss function},
  author={Barron, Jonathan T},
  booktitle={Proceedings of the IEEE/CVF conference on computer vision and pattern recognition},
  pages={4331--4339},
  year={2019}
}

@article{vapnik1996support,
  title={Support vector method for function approximation, regression estimation and signal processing},
  author={Vapnik, Vladimir and Golowich, Steven and Smola, Alex},
  journal={Advances in neural information processing systems},
  volume={9},
  year={1996}
}

@article{vazquez2022conformal,
  title={Conformal prediction in clinical medical sciences},
  author={Vazquez, Janette and Facelli, Julio C},
  journal={Journal of Healthcare Informatics Research},
  volume={6},
  number={3},
  pages={241--252},
  year={2022},
  publisher={Springer}
}

@article{lakshminarayanan2017simple,
  title={Simple and scalable predictive uncertainty estimation using deep ensembles},
  author={Lakshminarayanan, Balaji and Pritzel, Alexander and Blundell, Charles},
  journal={Advances in neural information processing systems},
  volume={30},
  year={2017}
}

@article{foygel2021limits,
  title={The limits of distribution-free conditional predictive inference},
  author={Foygel Barber, Rina and Candes, Emmanuel J and Ramdas, Aaditya and Tibshirani, Ryan J},
  journal={Information and Inference: A Journal of the IMA},
  volume={10},
  number={2},
  pages={455--482},
  year={2021},
  publisher={Oxford University Press}
}

@article{kendall2017uncertainties,
  title={What uncertainties do we need in bayesian deep learning for computer vision?},
  author={Kendall, Alex and Gal, Yarin},
  journal={Advances in neural information processing systems},
  volume={30},
  year={2017}
}

@article{yeh2026conformal,
  title={Conformal risk training: End-to-end optimization of conformal risk control},
  author={Yeh, Christopher and Christianson, Nicolas and Wierman, Adam and Yue, Yisong},
  journal={Advances in Neural Information Processing Systems},
  volume={38},
  pages={70056--70092},
  year={2026}
}

@inproceedings{li2024get,
  title={Get Your Embedding Space in Order: Domain-Adaptive Regression for Forest Monitoring},
  author={Li, Sizhuo and Gominski, Dimitri and Brandt, Martin and Tong, Xiaoye and Ciais, Philippe},
  booktitle={European Conference on Computer Vision},
  pages={94--111},
  year={2024},
  organization={Springer}
}

@inproceedings{zhang2017age,
  title={Age progression/regression by conditional adversarial autoencoder},
  author={Zhang, Zhifei and Song, Yang and Qi, Hairong},
  booktitle={Proceedings of the IEEE conference on computer vision and pattern recognition},
  pages={5810--5818},
  year={2017}
}

@article{seoni2023application,
  title={Application of uncertainty quantification to artificial intelligence in healthcare: A review of last decade (2013--2023)},
  author={Seoni, Silvia and Jahmunah, Vicnesh and Salvi, Massimo and Barua, Prabal Datta and Molinari, Filippo and Acharya, U Rajendra},
  journal={Computers in Biology and Medicine},
  volume={165},
  pages={107441},
  year={2023},
  publisher={Elsevier}
}

@inproceedings{mukama2024copula,
  title={Copula-based conformal prediction for object detection: a more efficient approach},
  author={Mukama, Bruce Cyusa and Messoudi, Soundouss and Rousseau, Sylvain and Destercke, S{\'e}bastien},
  booktitle={13th Symposium on Conformal and Probabilistic Prediction with Applications (COPA 2024)},
  volume={230},
  pages={140--157},
  year={2024}
}

@inproceedings{michelmore2020uncertainty,
  title={Uncertainty quantification with statistical guarantees in end-to-end autonomous driving control},
  author={Michelmore, Rhiannon and Wicker, Matthew and Laurenti, Luca and Cardelli, Luca and Gal, Yarin and Kwiatkowska, Marta},
  booktitle={2020 IEEE international conference on robotics and automation (ICRA)},
  pages={7344--7350},
  year={2020},
  organization={IEEE}
}

@article{allaire2021novel,
  title={Novel method for a posteriori uncertainty quantification in wildland fire spread simulation},
  author={Allaire, Fr{\'e}d{\'e}ric and Mallet, Vivien and Filippi, Jean-Baptiste},
  journal={Applied Mathematical Modelling},
  volume={90},
  pages={527--546},
  year={2021},
  publisher={Elsevier}
}

@article{han2017bayesian,
  title={Bayesian flood forecasting methods: A review},
  author={Han, Shasha and Coulibaly, Paulin},
  journal={Journal of Hydrology},
  volume={551},
  pages={340--351},
  year={2017},
  publisher={Elsevier}
}

@book{vovk2005algorithmic,
  title={Algorithmic learning in a random world},
  author={Vovk, Vladimir and Gammerman, Alex and Shafer, Glenn},
  year={2005},
  publisher={Springer Science \& Business Media}
}

@book{papadopoulos2008inductive,
  title={Inductive conformal prediction: Theory and application to neural networks},
  author={Papadopoulos, Harris},
  year={2008},
  publisher={INTECH Open Access Publisher Rijeka}
}

@article{papadopoulos2011reliable,
  title={Reliable prediction intervals with regression neural networks},
  author={Papadopoulos, Harris and Haralambous, Haris},
  journal={Neural Networks},
  volume={24},
  number={8},
  pages={842--851},
  year={2011},
  publisher={Elsevier}
}

@article{angelopoulos2021gentle,
  title={Conformal prediction: A gentle introduction},
  author={Angelopoulos, Anastasios N and Bates, Stephen},
  journal={Foundations and Trends in Machine Learning},
  volume={16},
  number={4},
  pages={494--591},
  year={2023},
  publisher={Emerald Publishing Limited}
}

@inproceedings{colombo2020training,
  title={Training conformal predictors},
  author={Colombo, Nicolo and Vovk, Vladimir},
  booktitle={Conformal and Probabilistic Prediction and Applications},
  pages={55--64},
  year={2020},
  organization={PMLR}
}

@article{bellotti2021optimized,
  title={Optimized conformal classification using gradient descent approximation},
  author={Bellotti, Anthony},
  journal={arXiv preprint arXiv:2105.11255},
  year={2021}
}

@inproceedings{Stutz2022ICLR,
	author    = {Stutz, David and Dvijotham, Krishnamurthy and Cemgil, Ali Taylan and Doucet, Arnaud},
	title     = {Learning Optimal Conformal Classifiers},
	booktitle = {International Conference on Learning Representations},
	year      = {2022},
}

@article{lei2023reliable,
  title={Reliable prediction intervals with directly optimized inductive conformal regression for deep learning},
  author={Lei, Haocheng and Bellotti, Anthony},
  journal={Neural Networks},
  volume={168},
  pages={194--205},
  year={2023},
  publisher={Elsevier}
}
\bibliographystyle{tmlr}

\clearpage
\appendix

\section{Dataset Characteristics}
To ensure full reproducibility, Table~\ref{tab:dataset_info} details the sample sizes and feature dimensions of all datasets evaluated in our experiments. The tabular datasets are sourced from the benchmark collection by \cite{grinsztajn2022tree}. The image datasets are provided by \cite{li2024get} (Drift) and \cite{zhang2017age} (UTK Face).

\begin{table}[h]
\caption{Characteristics' summary of the datasets used in the experiments.}
\centering
\begin{tabular}{llcc}
\toprule
\textbf{Dataset} & \textbf{Type} & \textbf{Nbr. Features} & \textbf{Nbr. Examples} \\
\midrule
\textbf{Bike Sharing} & Tabular & 13 & 17379 \\
\textbf{Brazilian Houses} & Tabular & 13 & 10692 \\
\textbf{California} & Tabular & 9 & 20640 \\
\textbf{CPU Act} & Tabular & 22 & 8192 \\
\textbf{Diamonds} & Tabular & 10 & 53940 \\
\textbf{Fifa} & Tabular & 6 & 18063 \\
\textbf{House Sales} & Tabular & 17 & 21613 \\
\textbf{Isolet} & Tabular & 614 & 7797 \\
\textbf{Medical Charges} & Tabular & 12 & 163065 \\
\textbf{Pol} & Tabular & 49 & 15000 \\
\textbf{Superconduct} & Tabular & 80 & 21263 \\
\textbf{Wine Quality} & Tabular & 12 & 6497 \\
\midrule
\textbf{Drift} & Image & $3 \times 200 \times 200$ & 13094 \\
\textbf{UTK Face} & Image & $3 \times 200 \times 200$ & 23708 \\
\bottomrule
\end{tabular}
\label{tab:dataset_info}
\end{table}

\section{Additional Results}
\label{sec:added_results}

\subsection{Performance (coverage and efficiency)}

\begin{figure*}[!ht]
  \centering
  \begin{subfigure}[t]{0.32\textwidth}
    \includegraphics[width=\textwidth]{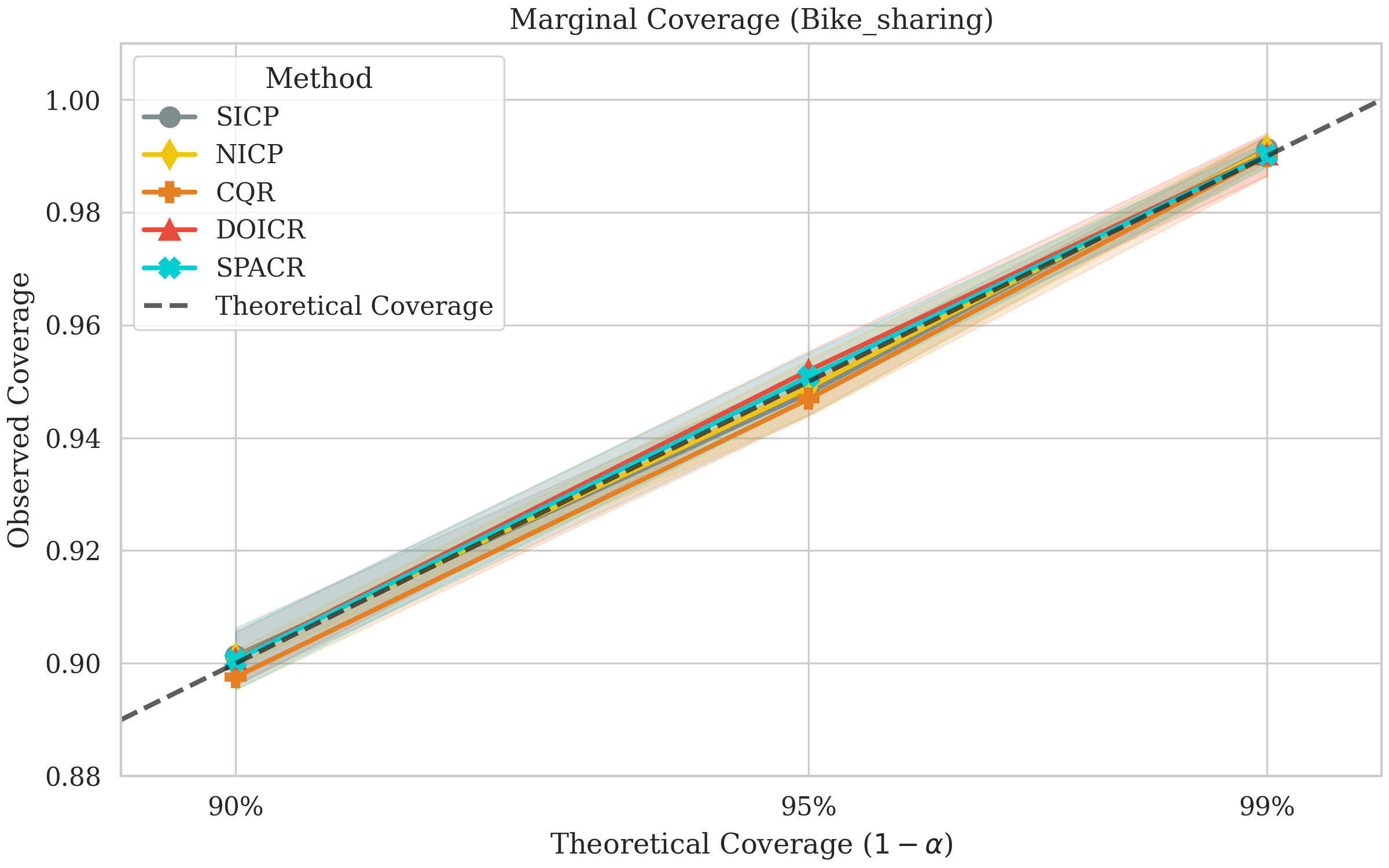}
    \caption{Marginal coverage vs. target confidence.}
    \label{fig:coverage_methods_bike_sharing}
  \end{subfigure}
  \hfill
  \begin{subfigure}[t]{0.32\textwidth}
    \includegraphics[width=\textwidth]{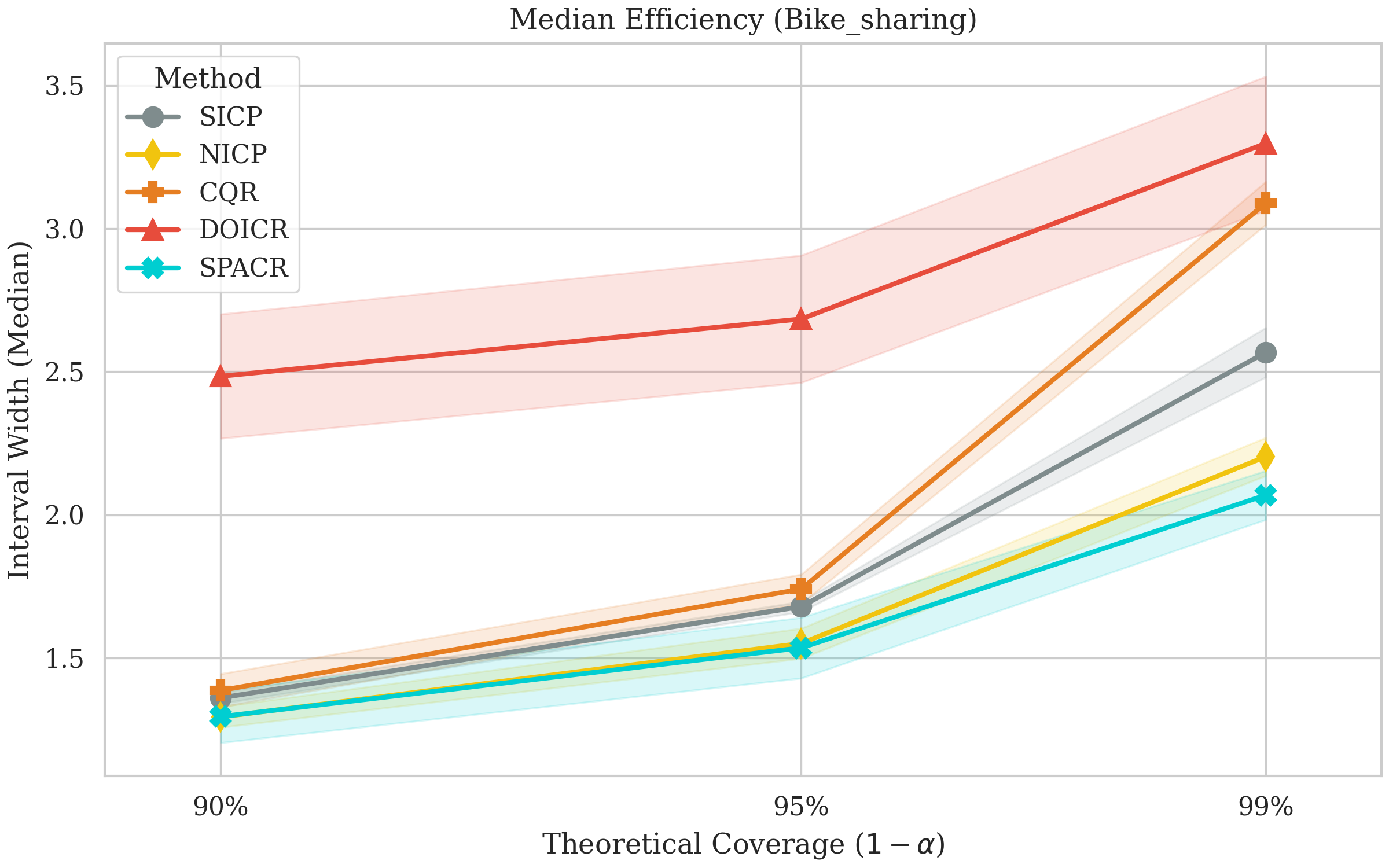}
    \caption{Efficiency (median interval width).}
    \label{fig:efficiency_methods_bike_sharing}
  \end{subfigure}
  \hfill
  \begin{subfigure}[t]{0.32\textwidth}
    \includegraphics[width=\textwidth]{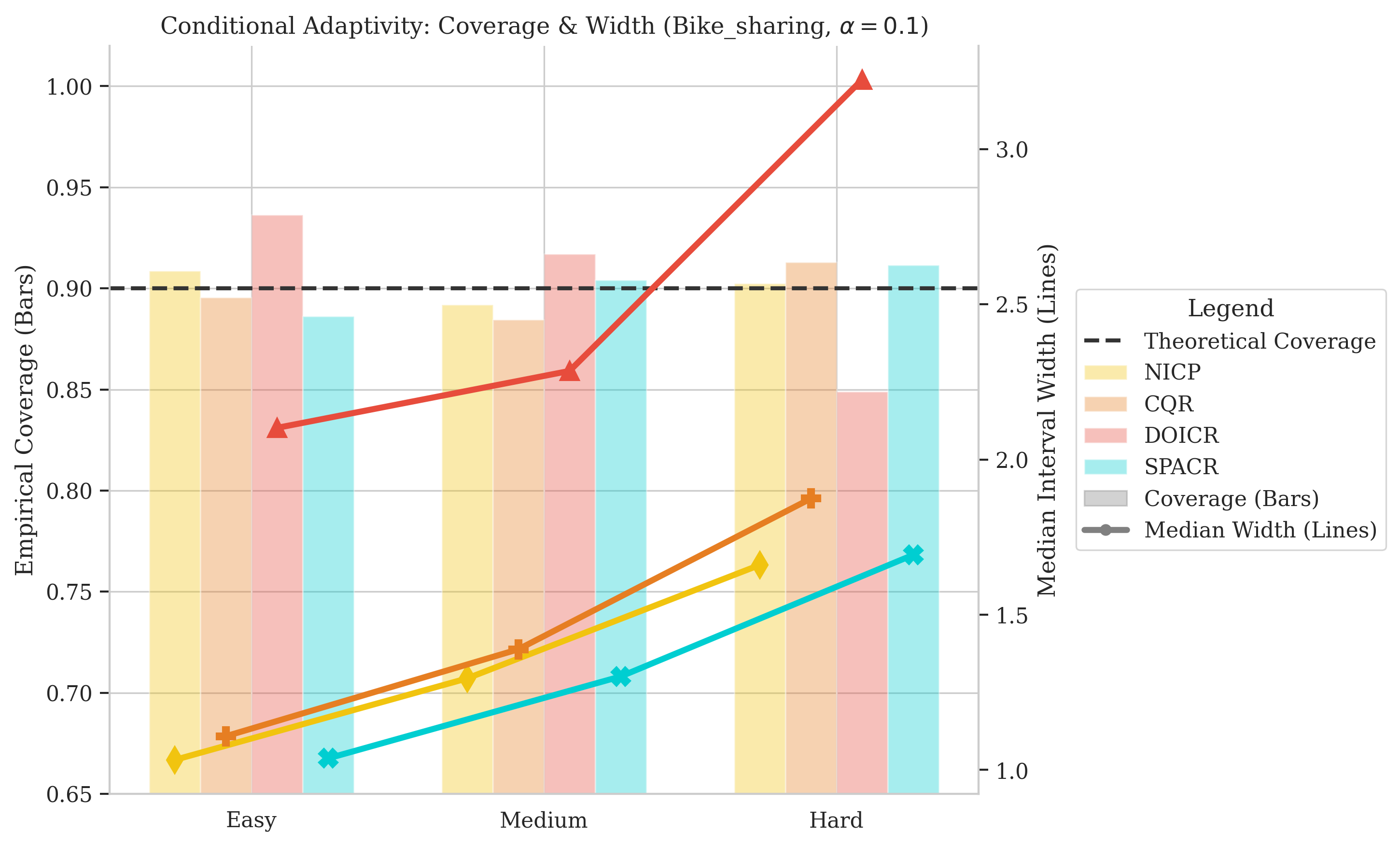}
    \caption{Conditional coverage (bars) and efficiency (lines) by internal difficulty.}
    \label{fig:iqr_methods_bike_sharing}
  \end{subfigure}
  \caption{Performance of conformal regression methods on the Bike Sharing dataset.}
  \label{fig:methods_figures_bike_sharing}
\end{figure*}

\begin{figure*}[!ht]
  \centering
  \begin{subfigure}[t]{0.32\textwidth}
    \includegraphics[width=\textwidth]{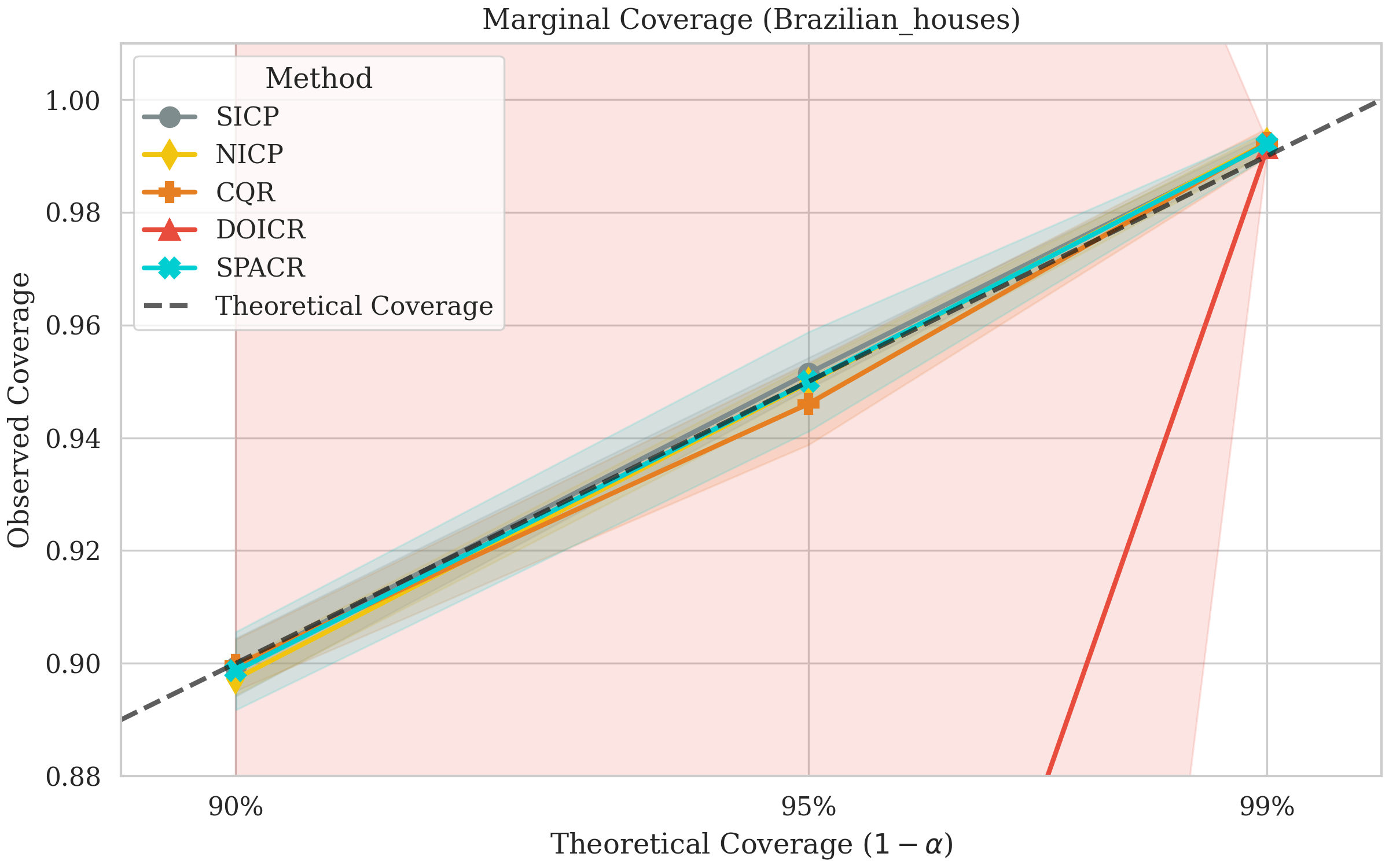}
    \caption{Marginal coverage vs. target confidence.}
    \label{fig:coverage_methods_br}
  \end{subfigure}
  \hfill
  \begin{subfigure}[t]{0.32\textwidth}
    \includegraphics[width=\textwidth]{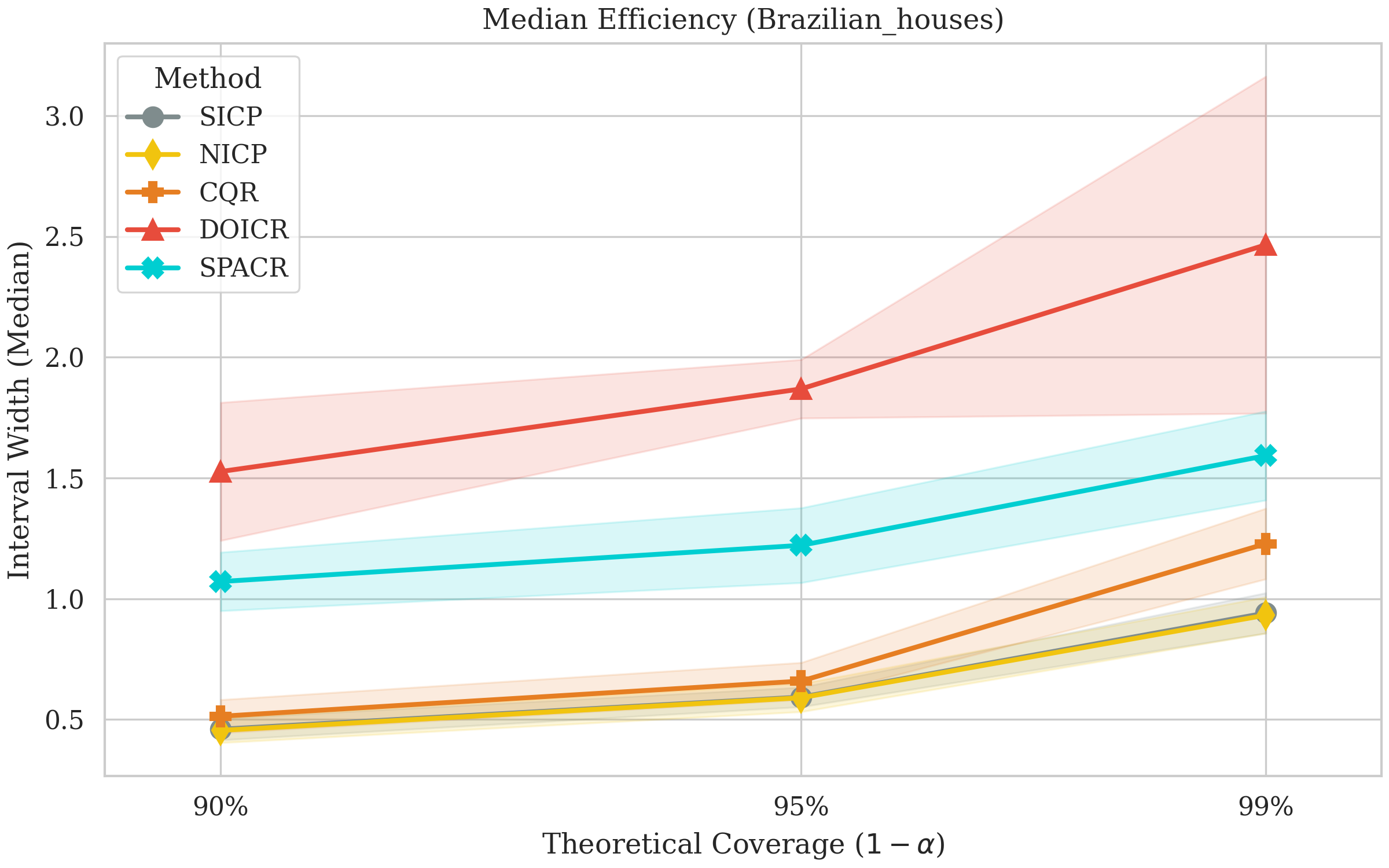}
    \caption{Efficiency (median interval width).}
    \label{fig:efficiency_methods_br}
  \end{subfigure}
  \hfill
  \begin{subfigure}[t]{0.32\textwidth}
    \includegraphics[width=\textwidth]{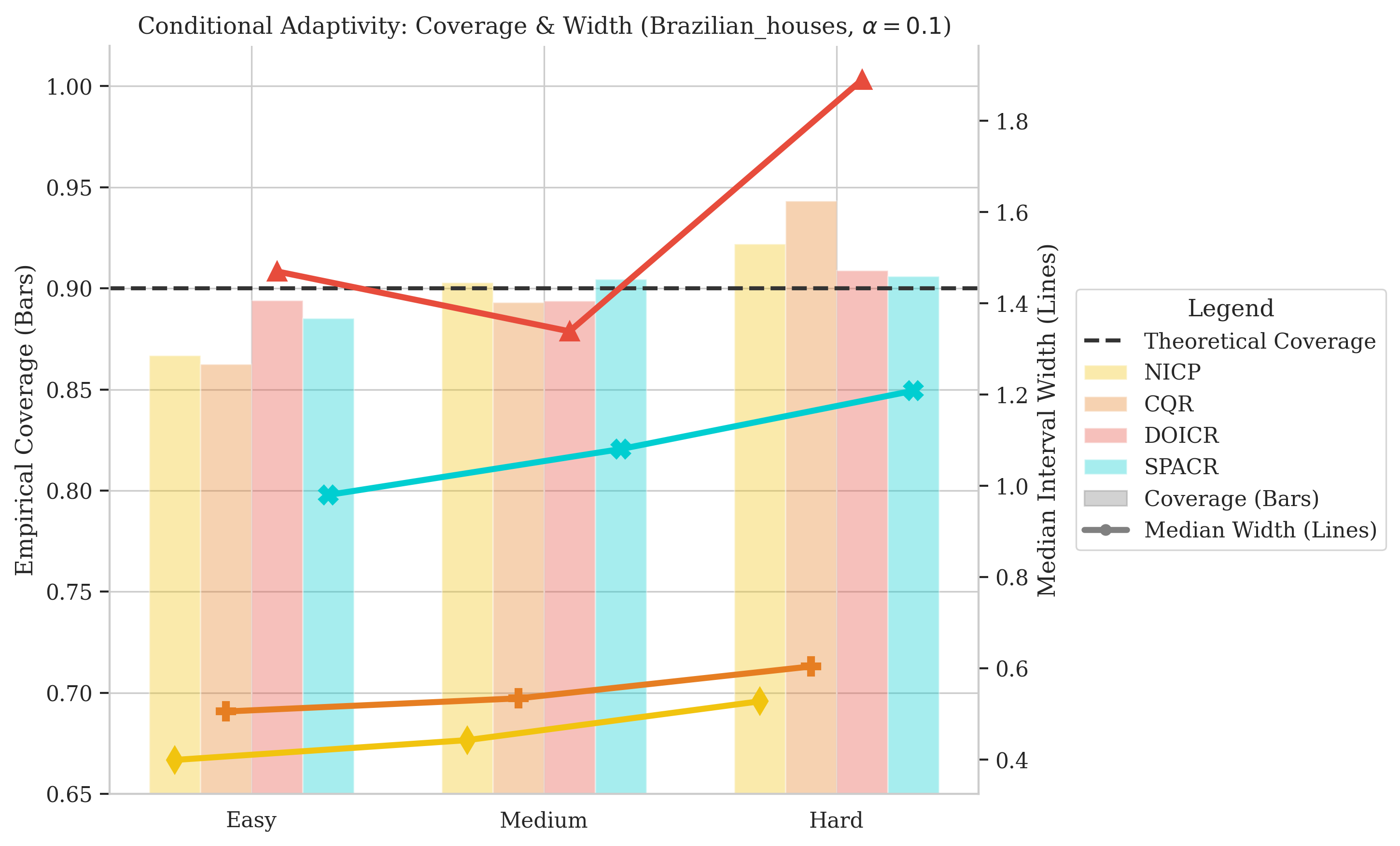}
    \caption{Conditional coverage (bars) and efficiency (lines) by internal difficulty.}
    \label{fig:iqr_methods_br}
  \end{subfigure}
    \caption{Performance of conformal regression methods on the Brazilian Houses dataset.}
      \label{fig:methods_figures_br}
\end{figure*}

\begin{figure*}[!ht]
  \centering
  \begin{subfigure}[t]{0.32\textwidth}
    \includegraphics[width=\textwidth]{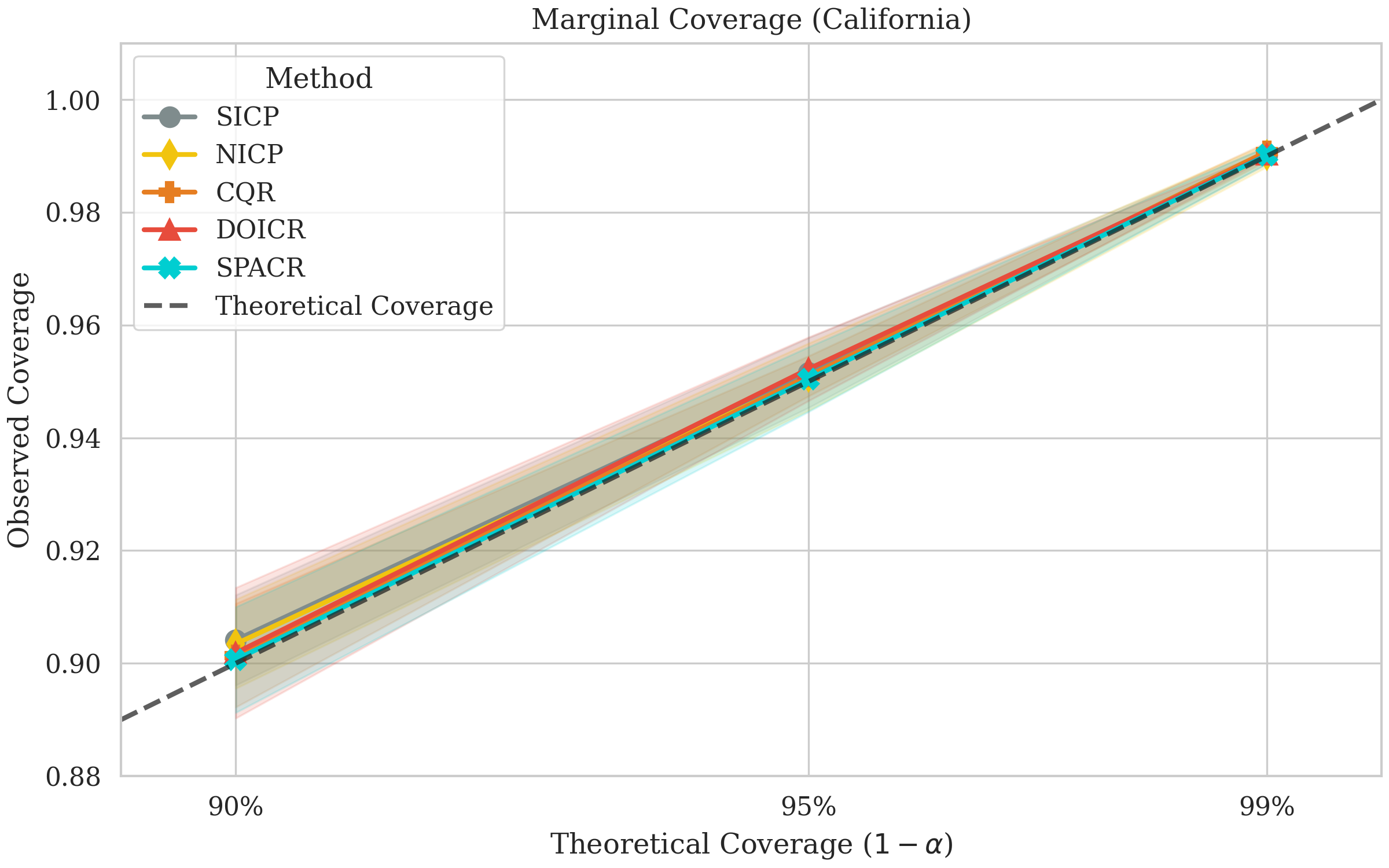}
    \caption{Marginal coverage vs. target confidence.}
    \label{fig:coverage_methods_California}
  \end{subfigure}
  \hfill
  \begin{subfigure}[t]{0.32\textwidth}
    \includegraphics[width=\textwidth]{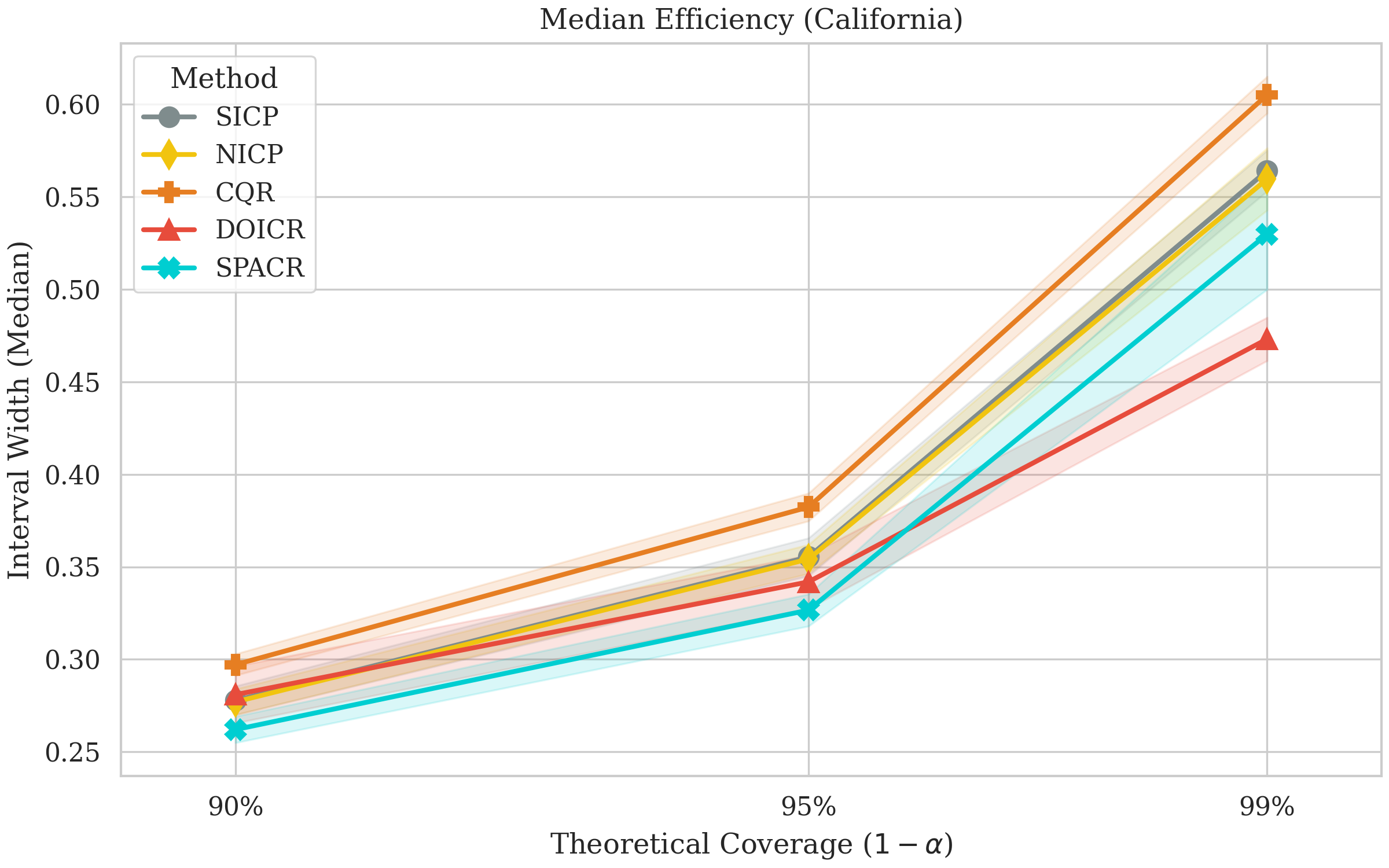}
    \caption{Efficiency (median interval width).}
    \label{fig:efficiency_methods_California}
  \end{subfigure}
  \hfill
  \begin{subfigure}[t]{0.32\textwidth}
    \includegraphics[width=\textwidth]{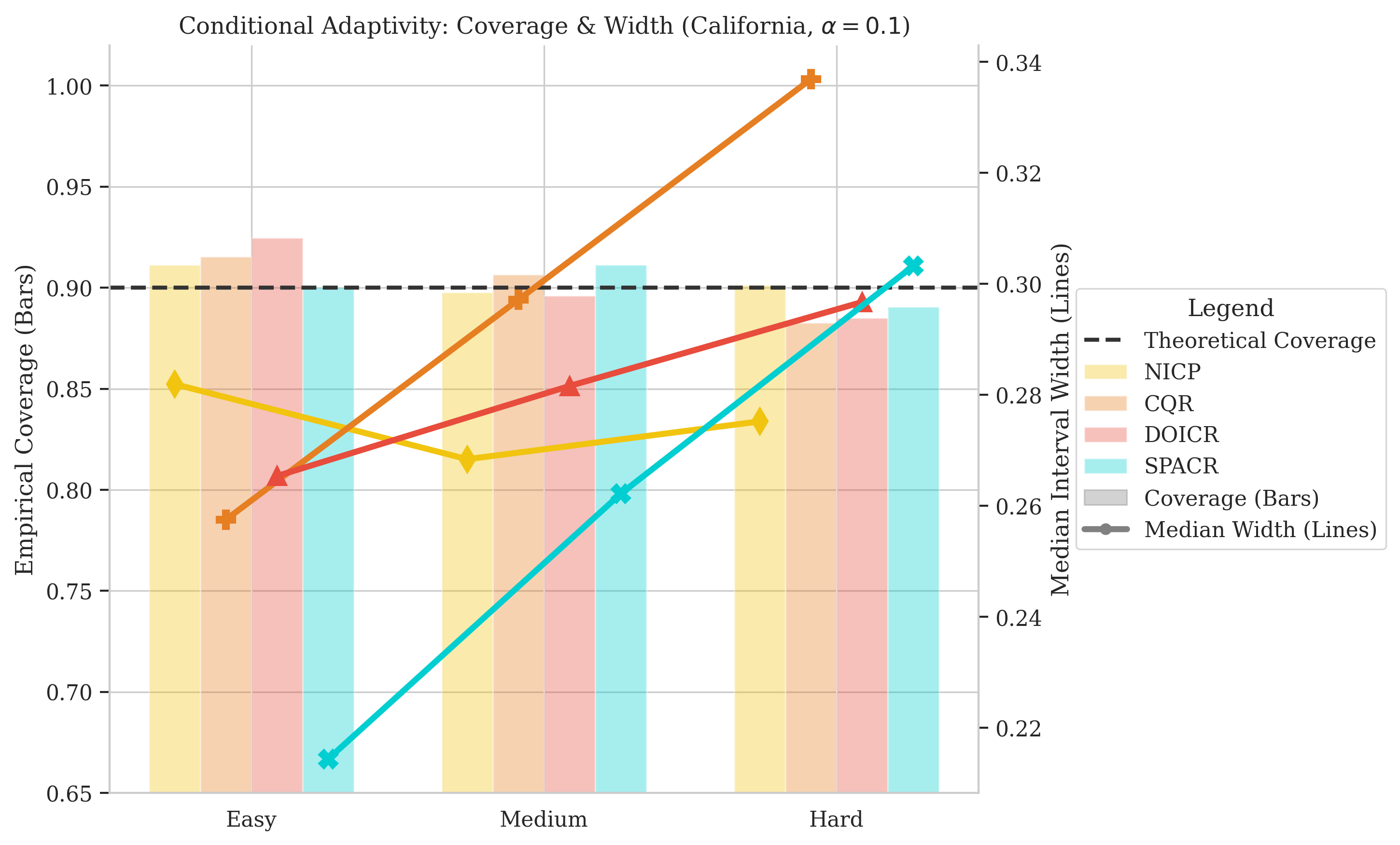}
    \caption{Conditional coverage (bars) and efficiency (lines) by internal difficulty.}
    \label{fig:iqr_methods_California}
  \end{subfigure}
    \caption{Performance of conformal regression methods on the California dataset.}
      \label{fig:methods_figures_California}
\end{figure*}

\begin{figure*}[!ht]
  \centering
  \begin{subfigure}[t]{0.32\textwidth}
    \includegraphics[width=\textwidth]{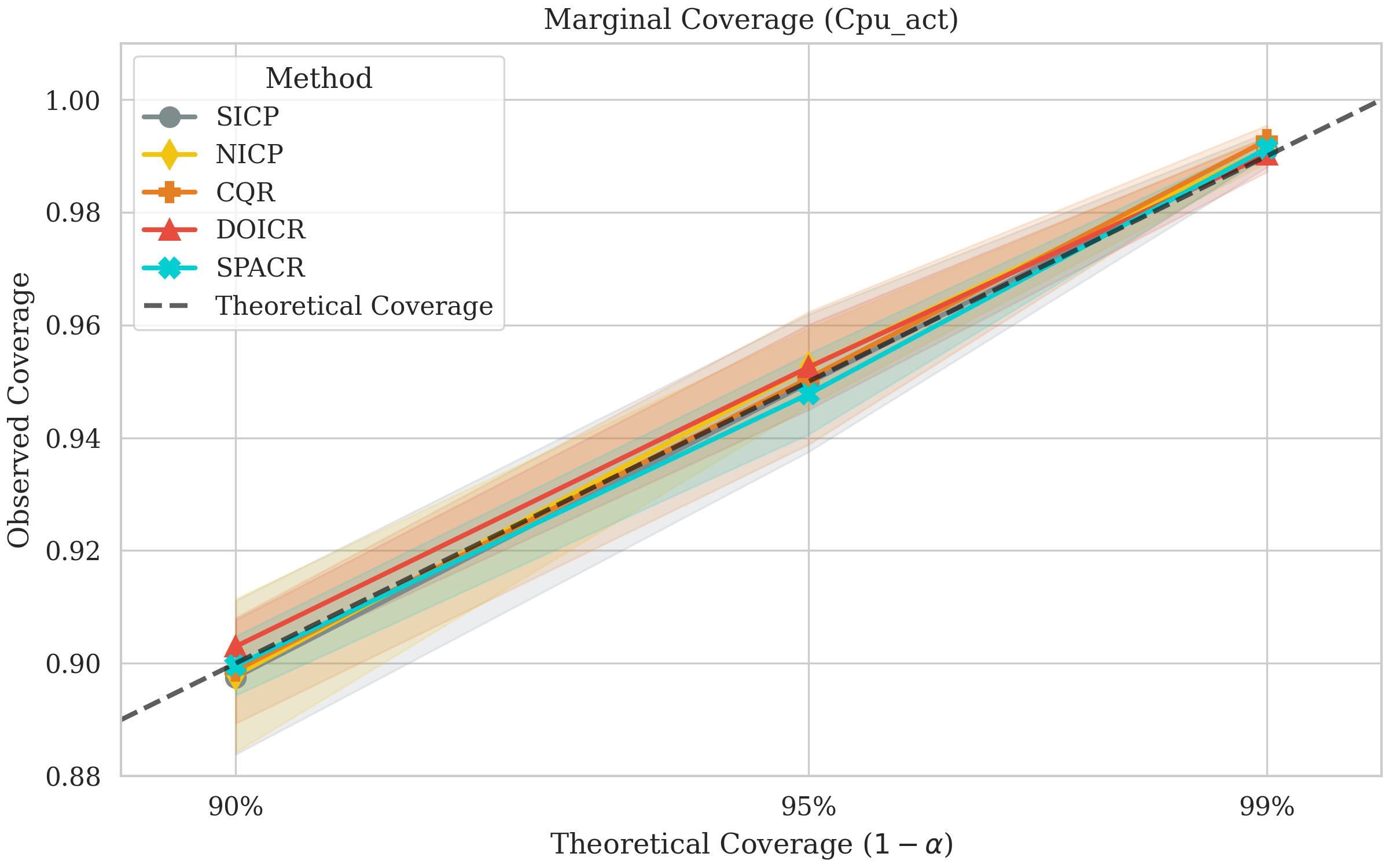}
    \caption{Marginal coverage vs. target confidence.}
    \label{fig:coverage_methods_Cpu_act}
  \end{subfigure}
  \hfill
  \begin{subfigure}[t]{0.32\textwidth}
    \includegraphics[width=\textwidth]{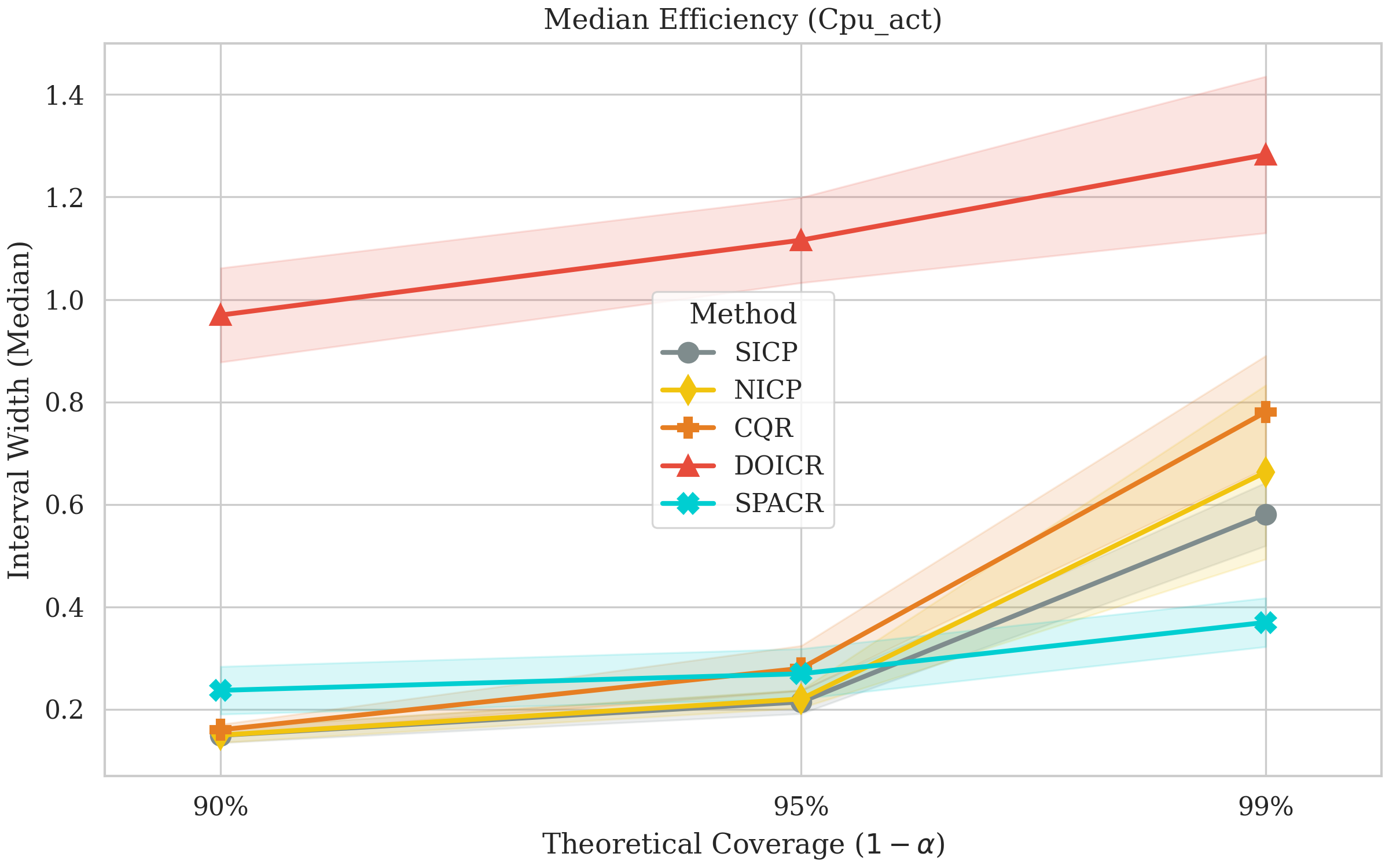}
    \caption{Efficiency (median interval width).}
    \label{fig:efficiency_methods_Cpu_act}
  \end{subfigure}
  \hfill
  \begin{subfigure}[t]{0.32\textwidth}
    \includegraphics[width=\textwidth]{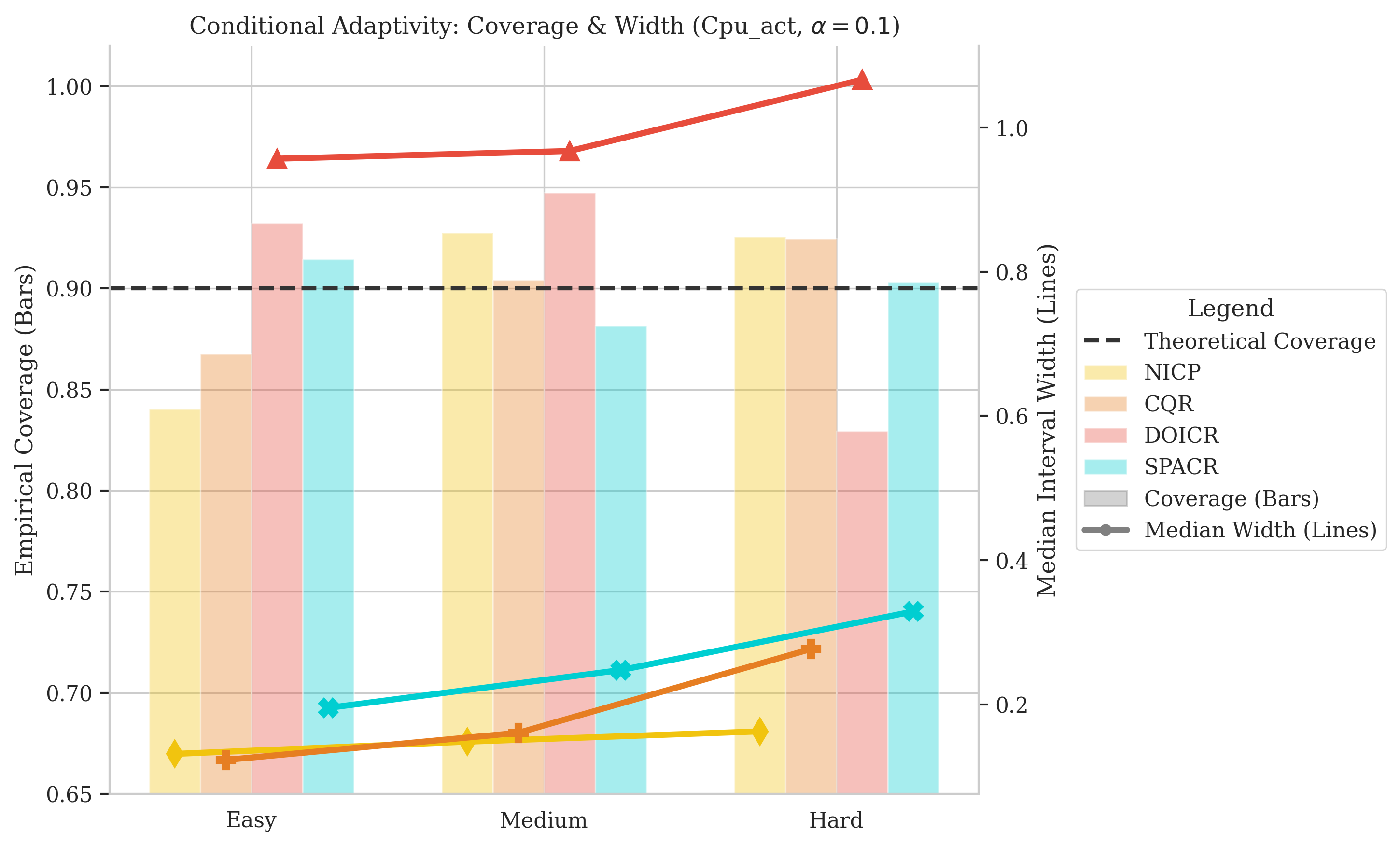}
    \caption{Conditional coverage (bars) and efficiency (lines) by internal difficulty.}
    \label{fig:iqr_methods_Cpu_act}
  \end{subfigure}
    \caption{Performance of conformal regression methods on the CPU Act dataset.}
      \label{fig:methods_figures_Cpu_act}
\end{figure*}

\begin{figure*}[!ht]
  \centering
  \begin{subfigure}[t]{0.32\textwidth}
    \includegraphics[width=\textwidth]{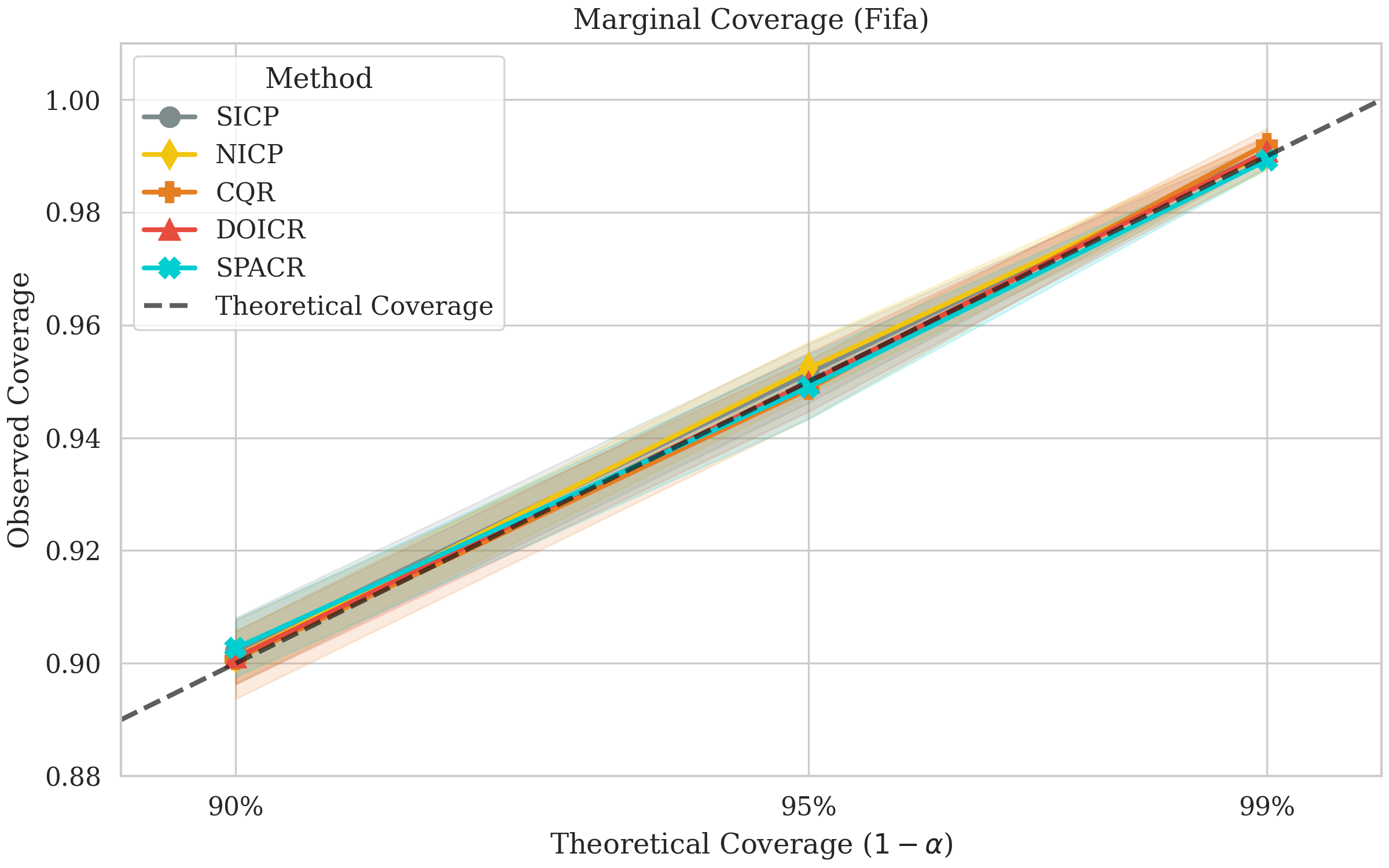}
    \caption{Marginal coverage vs. target confidence.}
    \label{fig:coverage_methods_Fifa}
  \end{subfigure}
  \hfill
  \begin{subfigure}[t]{0.32\textwidth}
    \includegraphics[width=\textwidth]{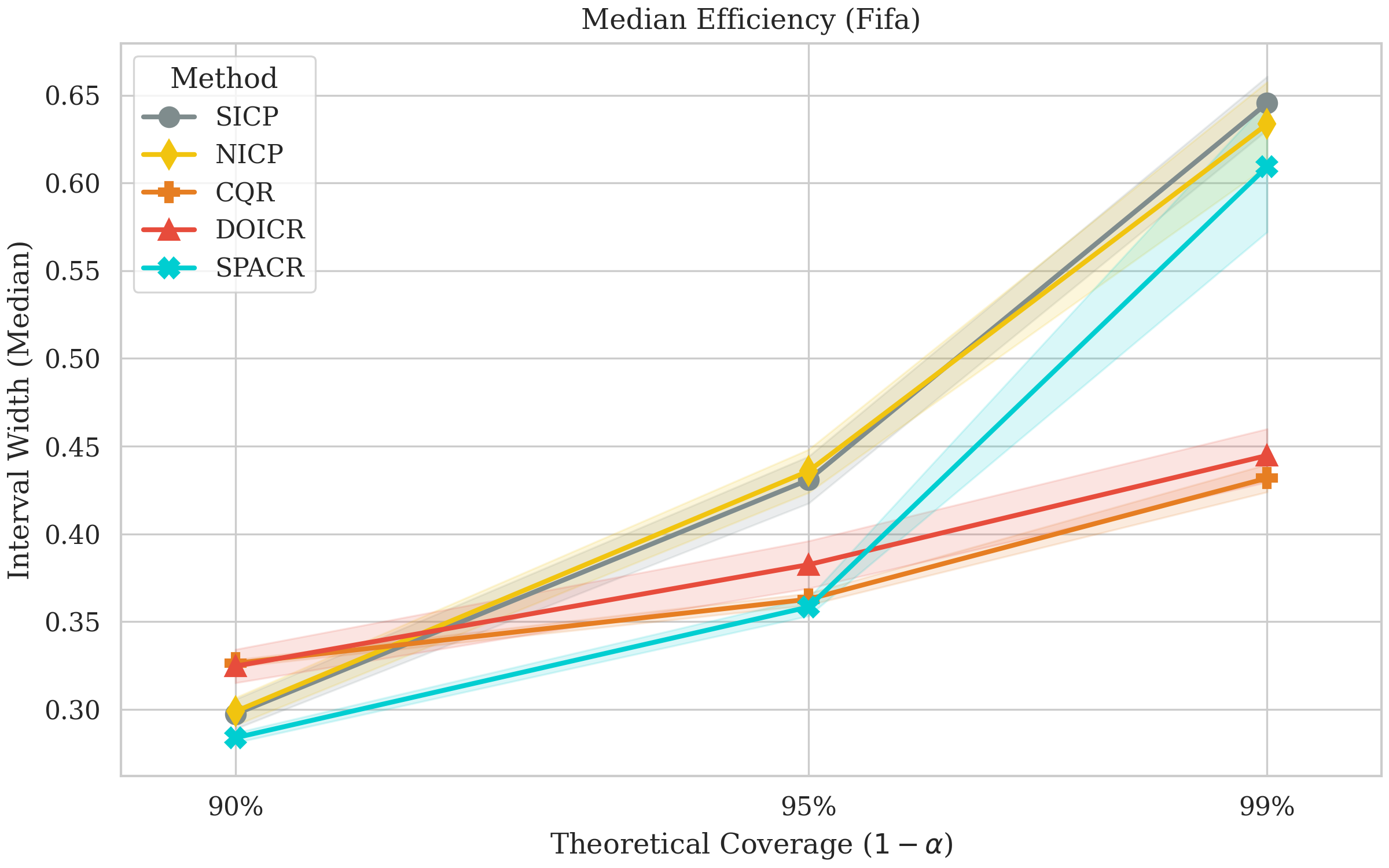}
    \caption{Efficiency (median interval width).}
    \label{fig:efficiency_methods_Fifa}
  \end{subfigure}
  \hfill
  \begin{subfigure}[t]{0.32\textwidth}
    \includegraphics[width=\textwidth]{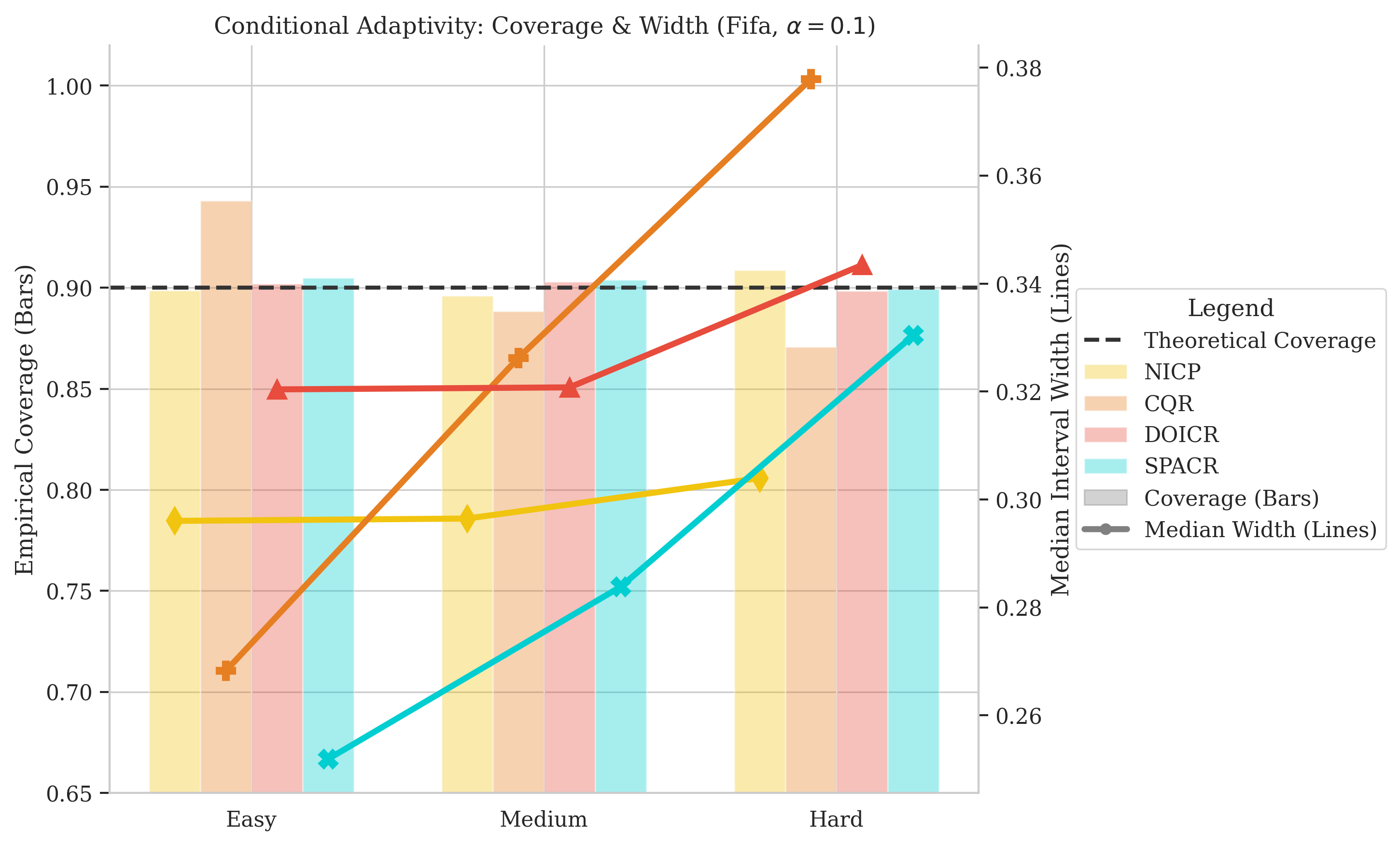}
    \caption{Conditional coverage (bars) and efficiency (lines) by internal difficulty.}
    \label{fig:iqr_methods_Fifa}
  \end{subfigure}
    \caption{Performance of conformal regression methods on the Fifa dataset.}
      \label{fig:methods_figures_Fifa}
\end{figure*}

\begin{figure*}[!ht]
  \centering
  \begin{subfigure}[t]{0.32\textwidth}
    \includegraphics[width=\textwidth]{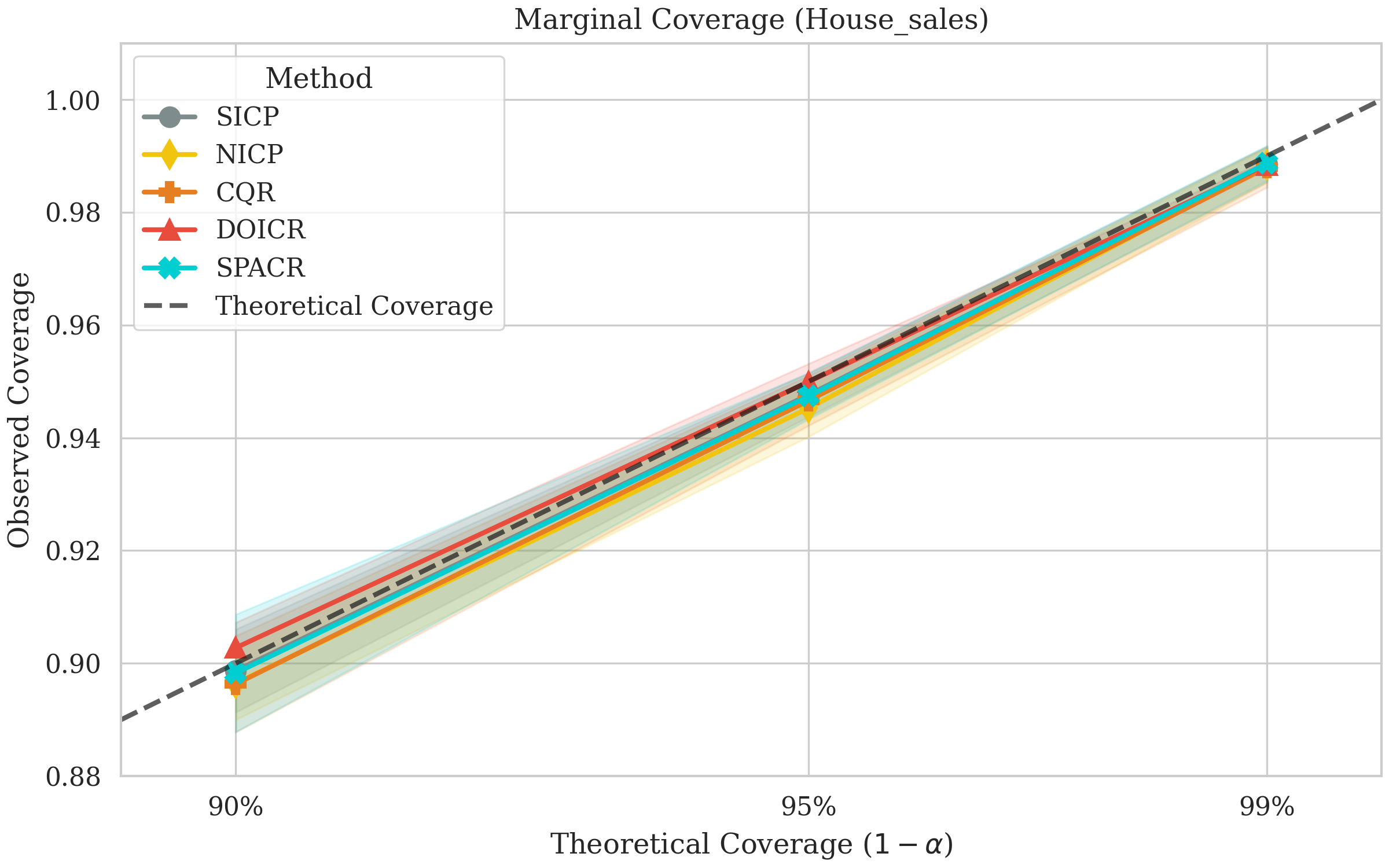}
    \caption{Marginal coverage vs. target confidence.}
    \label{fig:coverage_methods_House_sales}
  \end{subfigure}
  \hfill
  \begin{subfigure}[t]{0.32\textwidth}
    \includegraphics[width=\textwidth]{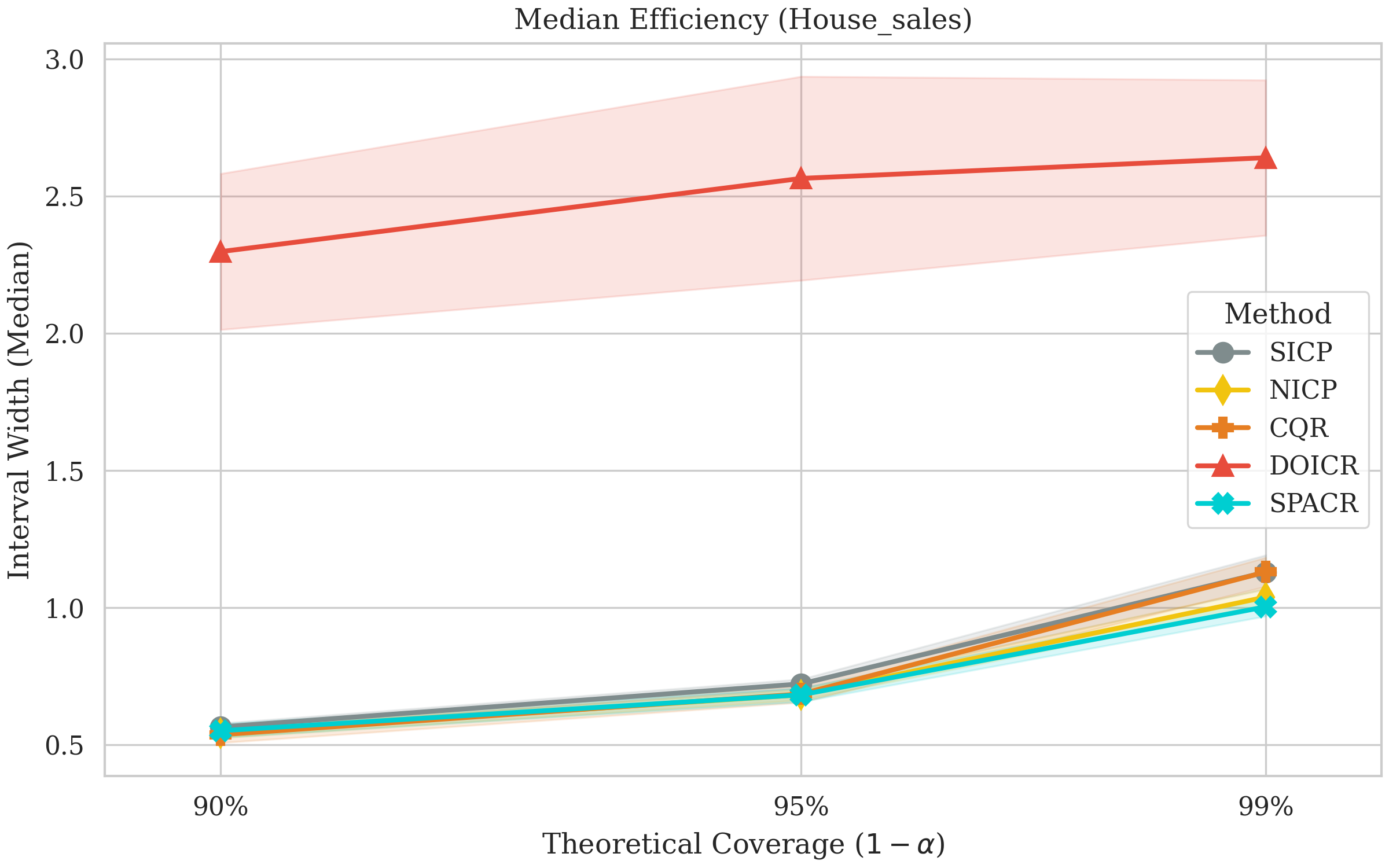}
    \caption{Efficiency (median interval width).}
    \label{fig:efficiency_methods_House_sales}
  \end{subfigure}
  \hfill
  \begin{subfigure}[t]{0.32\textwidth}
    \includegraphics[width=\textwidth]{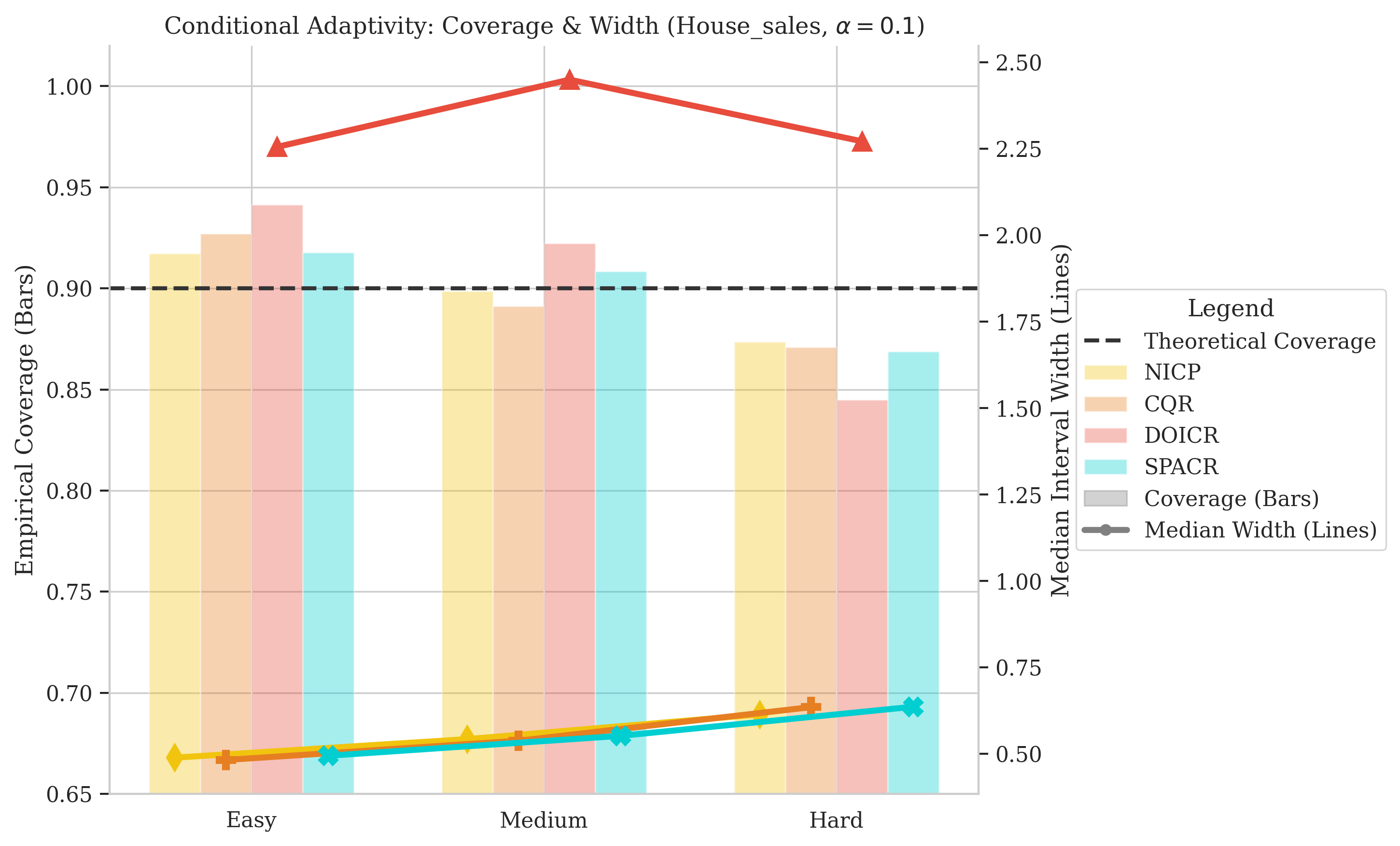}
    \caption{Conditional coverage (bars) and efficiency (lines) by internal difficulty.}
    \label{fig:iqr_methods_House_sales}
  \end{subfigure}
    \caption{Performance of conformal regression methods on the House Sales dataset.}
      \label{fig:methods_figures_House_sales}
\end{figure*}

\begin{figure*}[!ht]
  \centering
  \begin{subfigure}[t]{0.32\textwidth}
    \includegraphics[width=\textwidth]{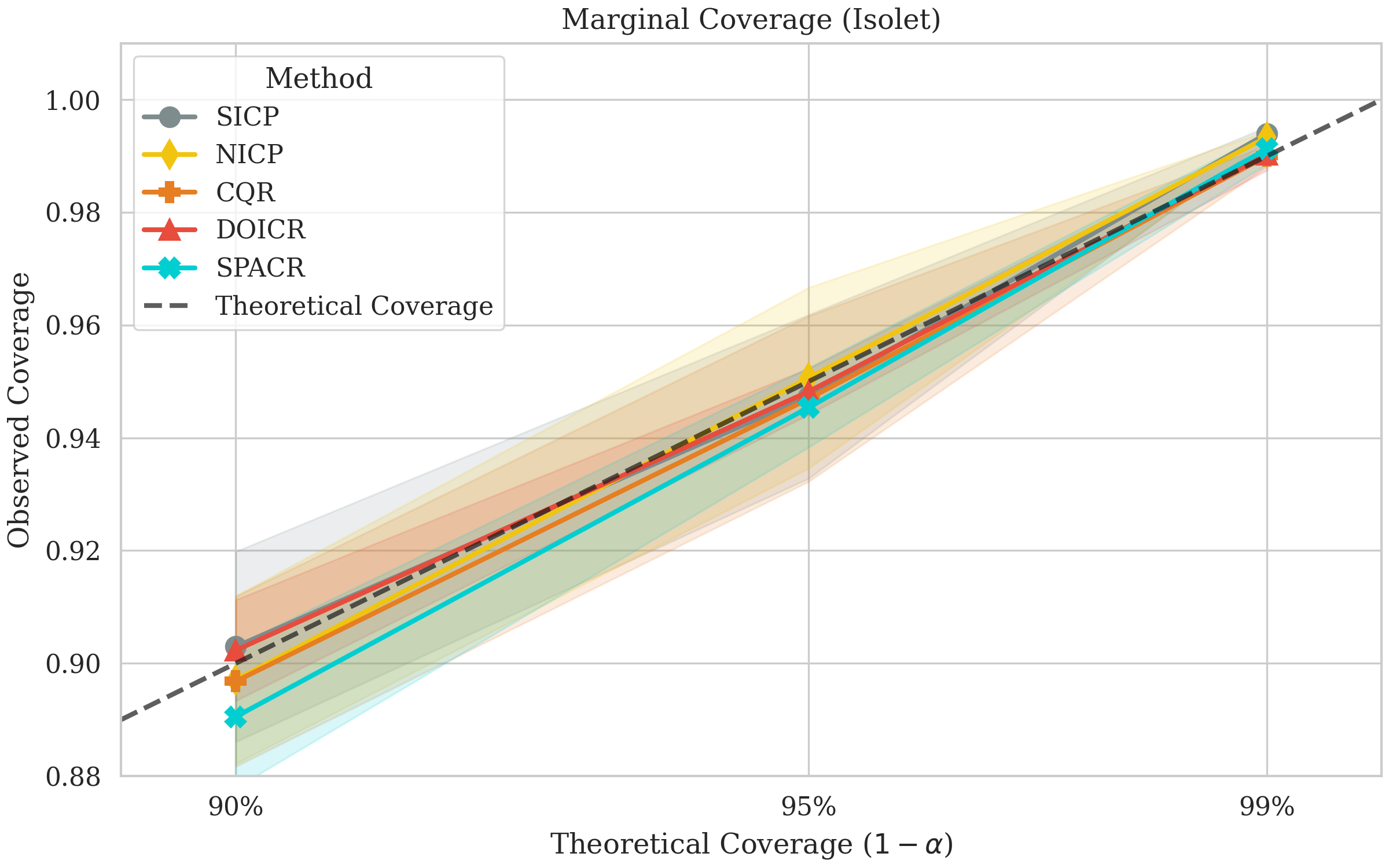}
    \caption{Marginal coverage vs. target confidence.}
    \label{fig:coverage_methods_Isolet}
  \end{subfigure}
  \hfill
  \begin{subfigure}[t]{0.32\textwidth}
    \includegraphics[width=\textwidth]{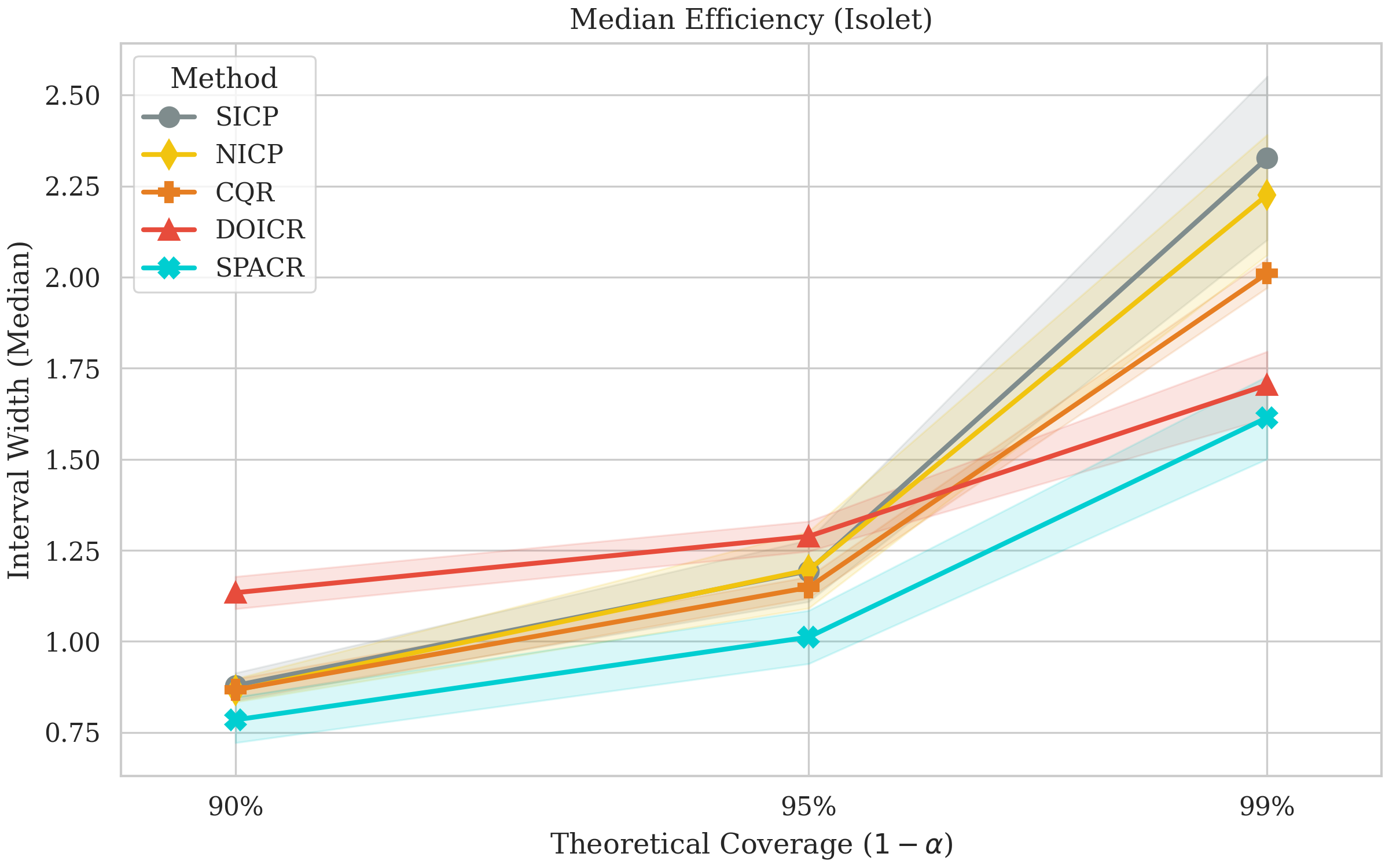}
    \caption{Efficiency (median interval width).}
    \label{fig:efficiency_methods_Isolet}
  \end{subfigure}
  \hfill
  \begin{subfigure}[t]{0.32\textwidth}
    \includegraphics[width=\textwidth]{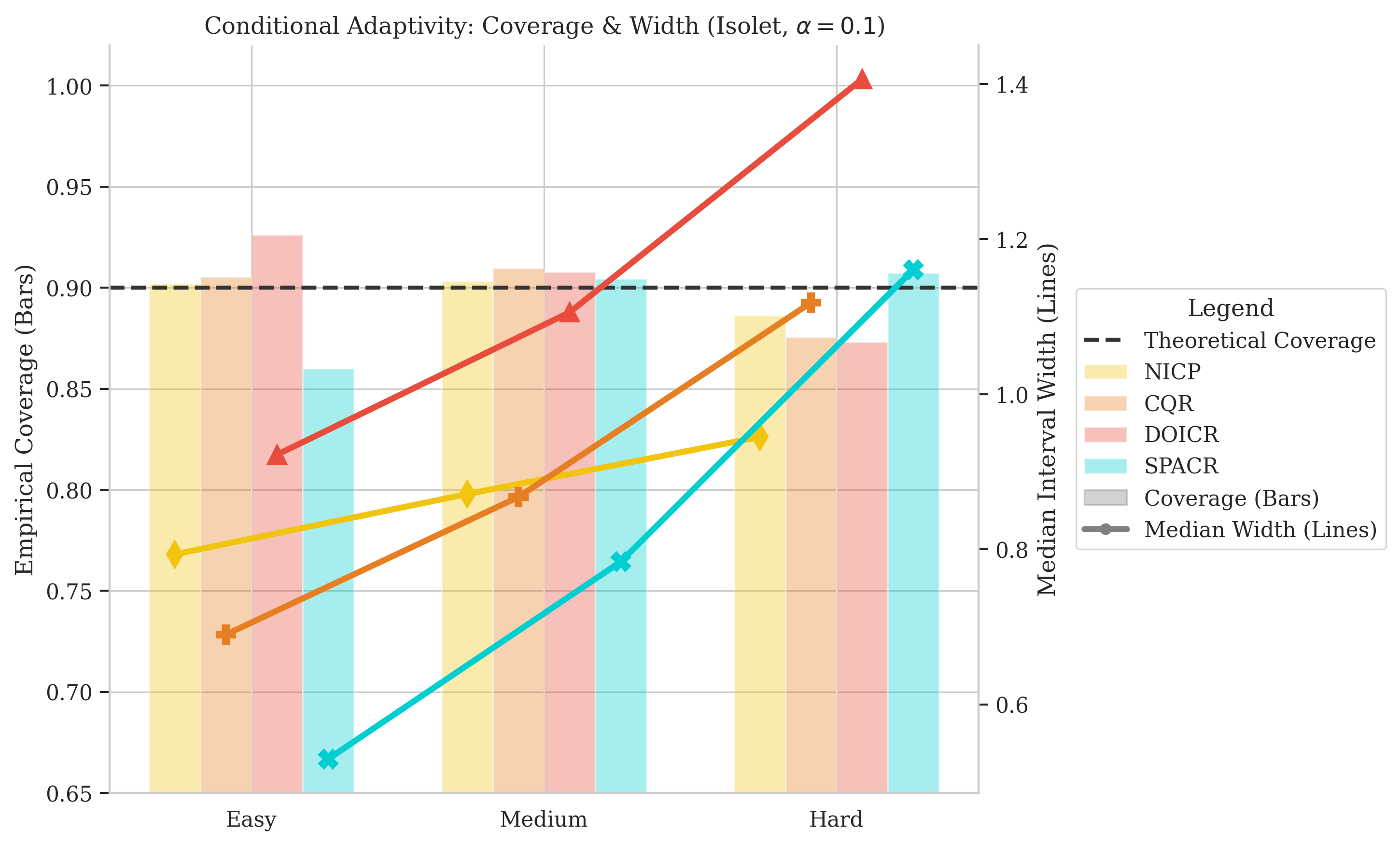}
    \caption{Conditional coverage (bars) and efficiency (lines) by internal difficulty.}
    \label{fig:iqr_methods_Isolet}
  \end{subfigure}
    \caption{Performance of conformal regression methods on the Isolet dataset.}
      \label{fig:methods_figures_Isolet}
\end{figure*}

\begin{figure*}[!ht]
  \centering
  \begin{subfigure}[t]{0.32\textwidth}
    \includegraphics[width=\textwidth]{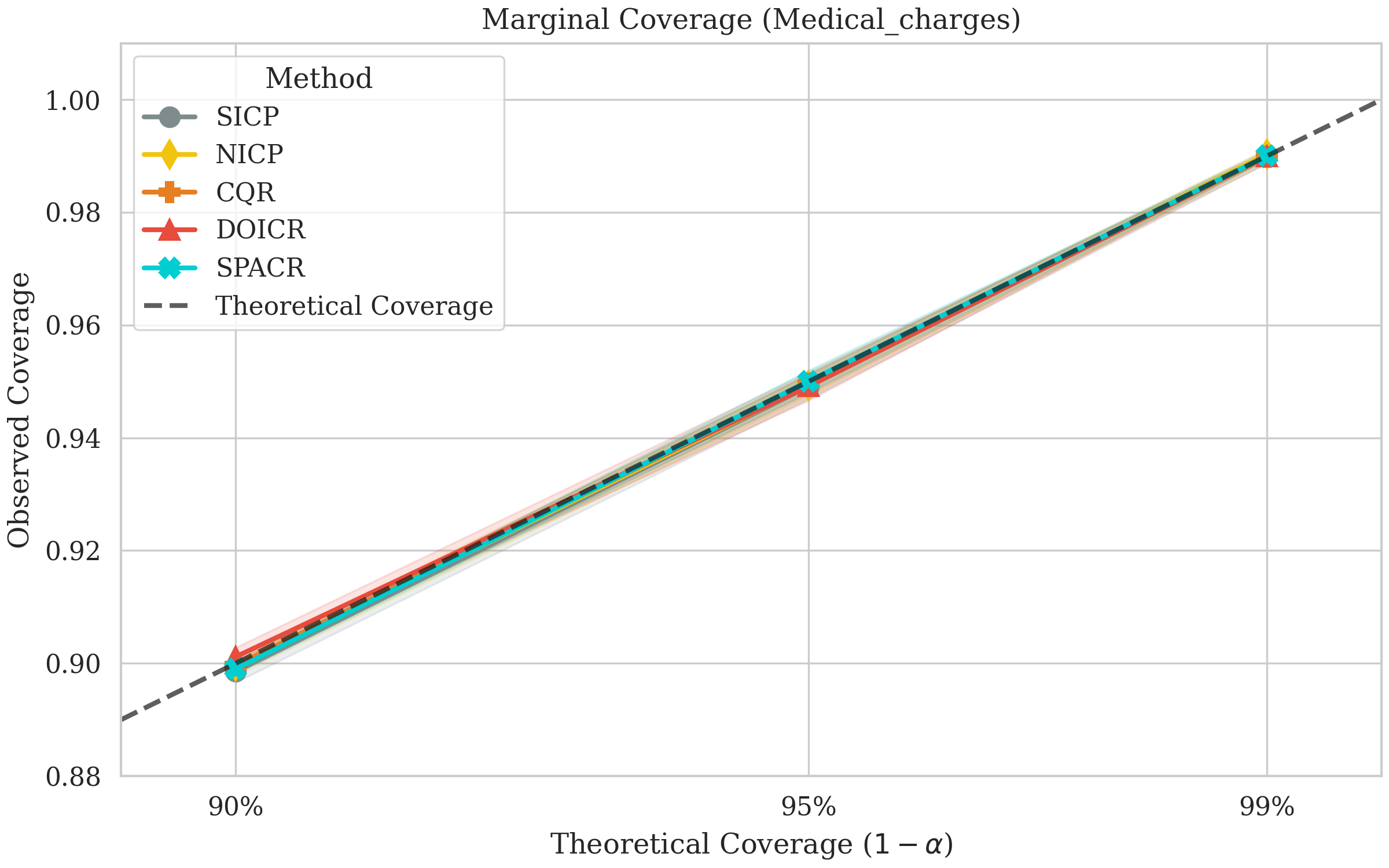}
    \caption{Marginal coverage vs. target confidence.}
    \label{fig:coverage_methods_Medical_charges}
  \end{subfigure}
  \hfill
  \begin{subfigure}[t]{0.32\textwidth}
    \includegraphics[width=\textwidth]{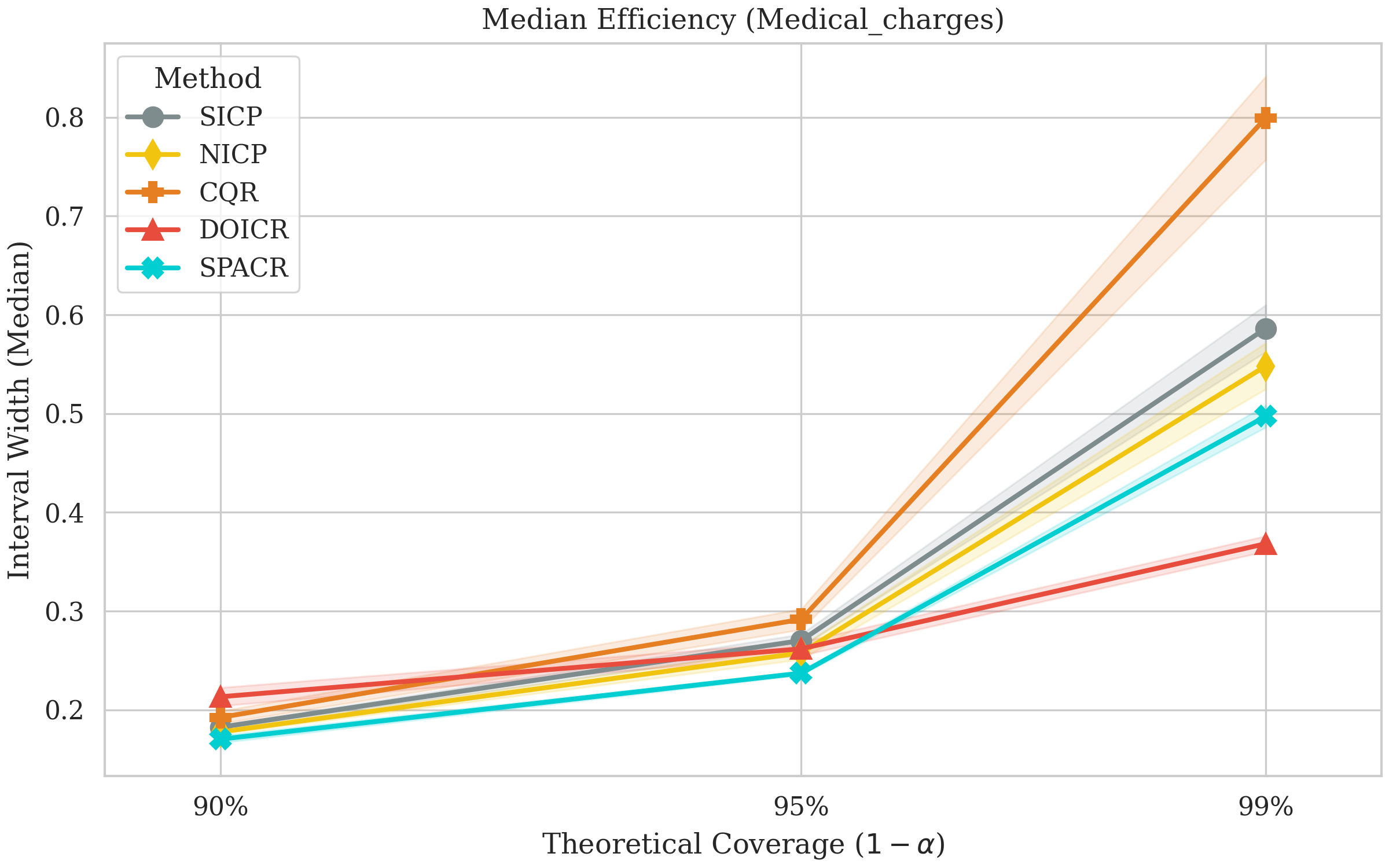}
    \caption{Efficiency (median interval width).}
    \label{fig:efficiency_methods_Medical_charges}
  \end{subfigure}
  \hfill
  \begin{subfigure}[t]{0.32\textwidth}
    \includegraphics[width=\textwidth]{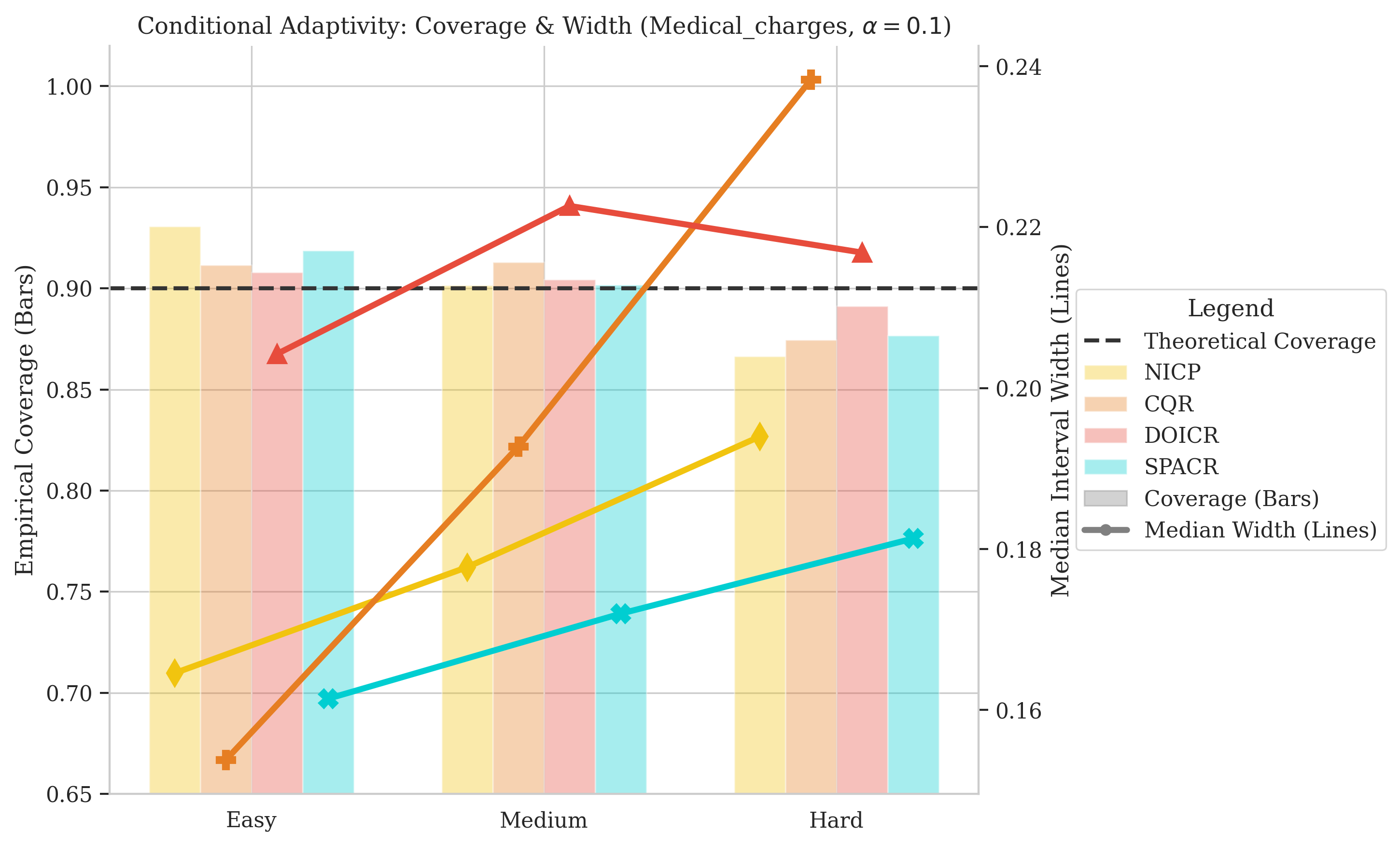}
    \caption{Conditional coverage (bars) and efficiency (lines) by internal difficulty.}
    \label{fig:iqr_methods_Medical_charges}
  \end{subfigure}
    \caption{Performance of conformal regression methods on the Medical Charges dataset.}
      \label{fig:methods_figures_Medical_charges}
\end{figure*}

\begin{figure*}[!ht]
  \centering
  \begin{subfigure}[t]{0.32\textwidth}
    \includegraphics[width=\textwidth]{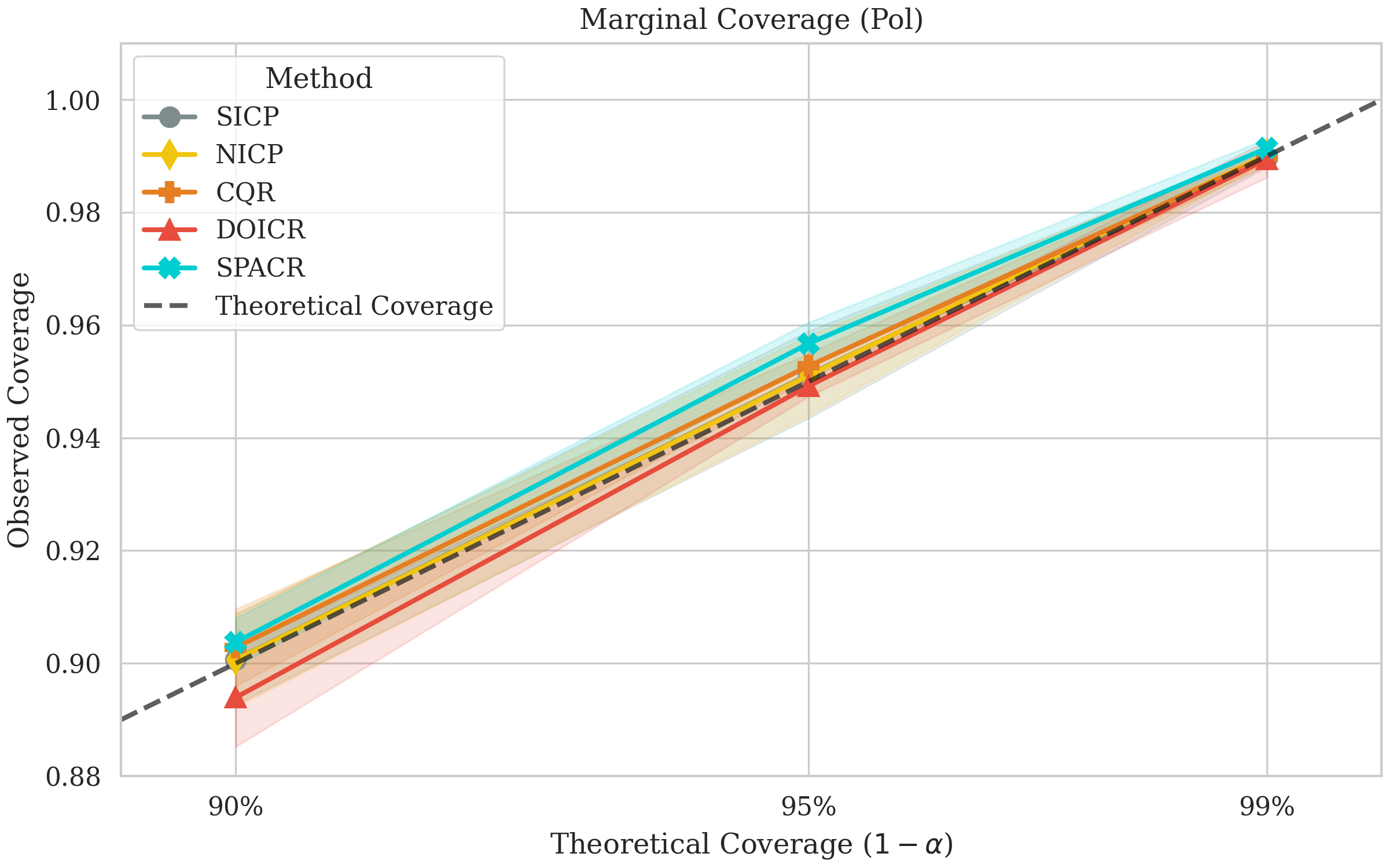}
    \caption{Marginal coverage vs. target confidence.}
    \label{fig:coverage_methods_Pol}
  \end{subfigure}
  \hfill
  \begin{subfigure}[t]{0.32\textwidth}
    \includegraphics[width=\textwidth]{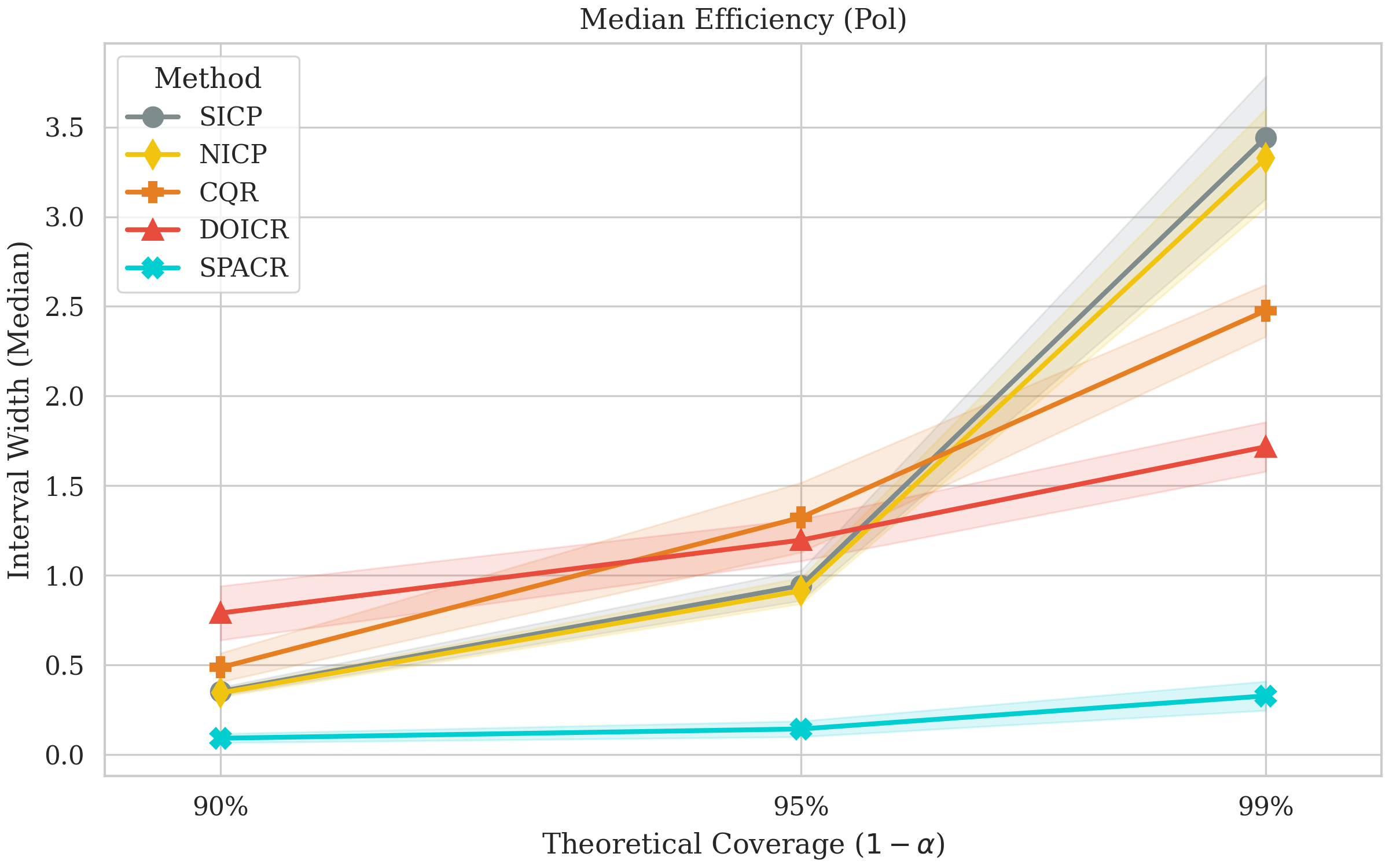}
    \caption{Efficiency (median interval width).}
    \label{fig:efficiency_methods_Pol}
  \end{subfigure}
  \hfill
  \begin{subfigure}[t]{0.32\textwidth}
    \includegraphics[width=\textwidth]{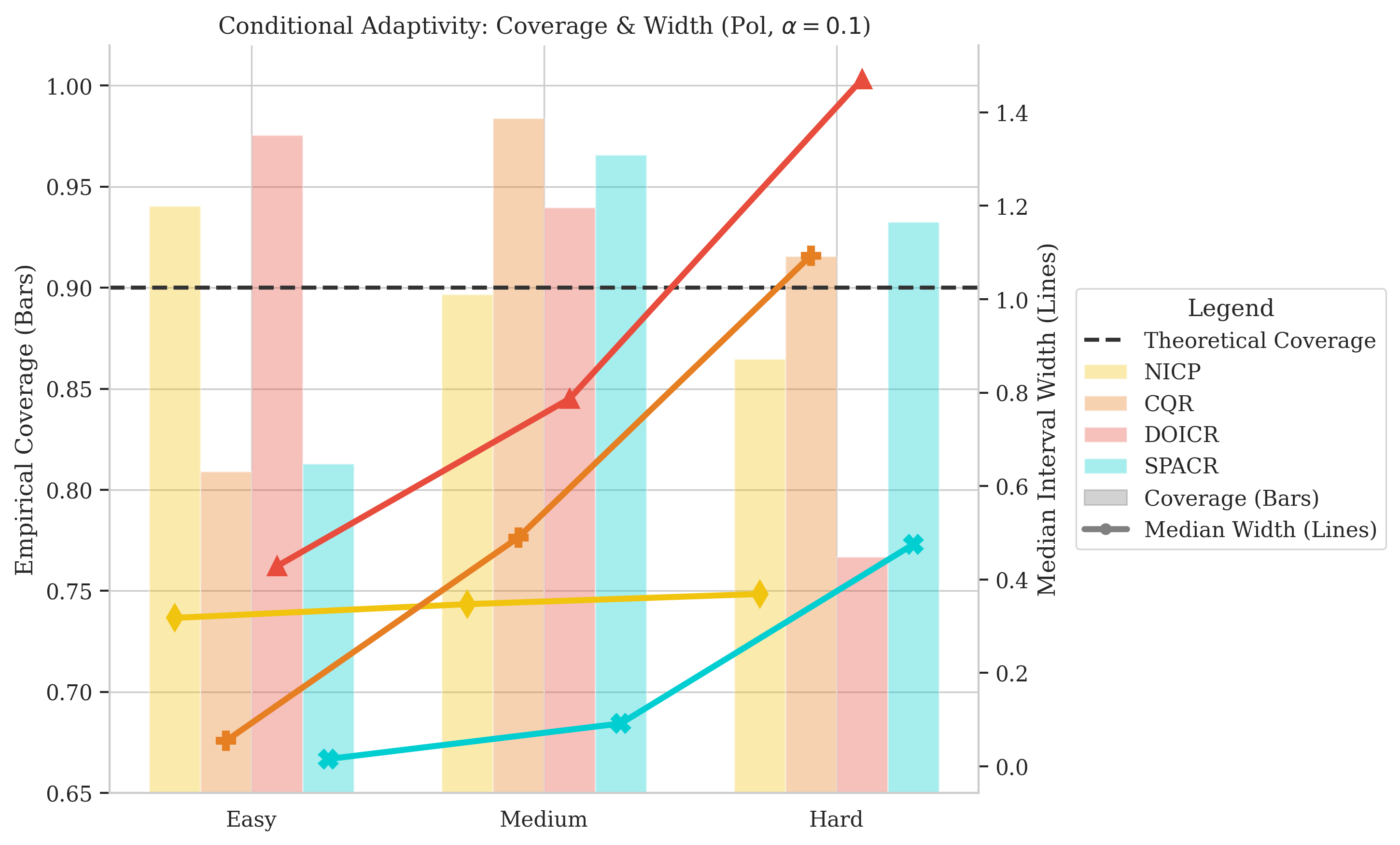}
    \caption{Conditional coverage (bars) and efficiency (lines) by internal difficulty.}
    \label{fig:iqr_methods_Pol}
  \end{subfigure}
    \caption{Performance of conformal regression methods on the Pol dataset.}
      \label{fig:methods_figures_Pol}
\end{figure*}

\begin{figure*}[!ht]
  \centering
  \begin{subfigure}[t]{0.32\textwidth}
    \includegraphics[width=\textwidth]{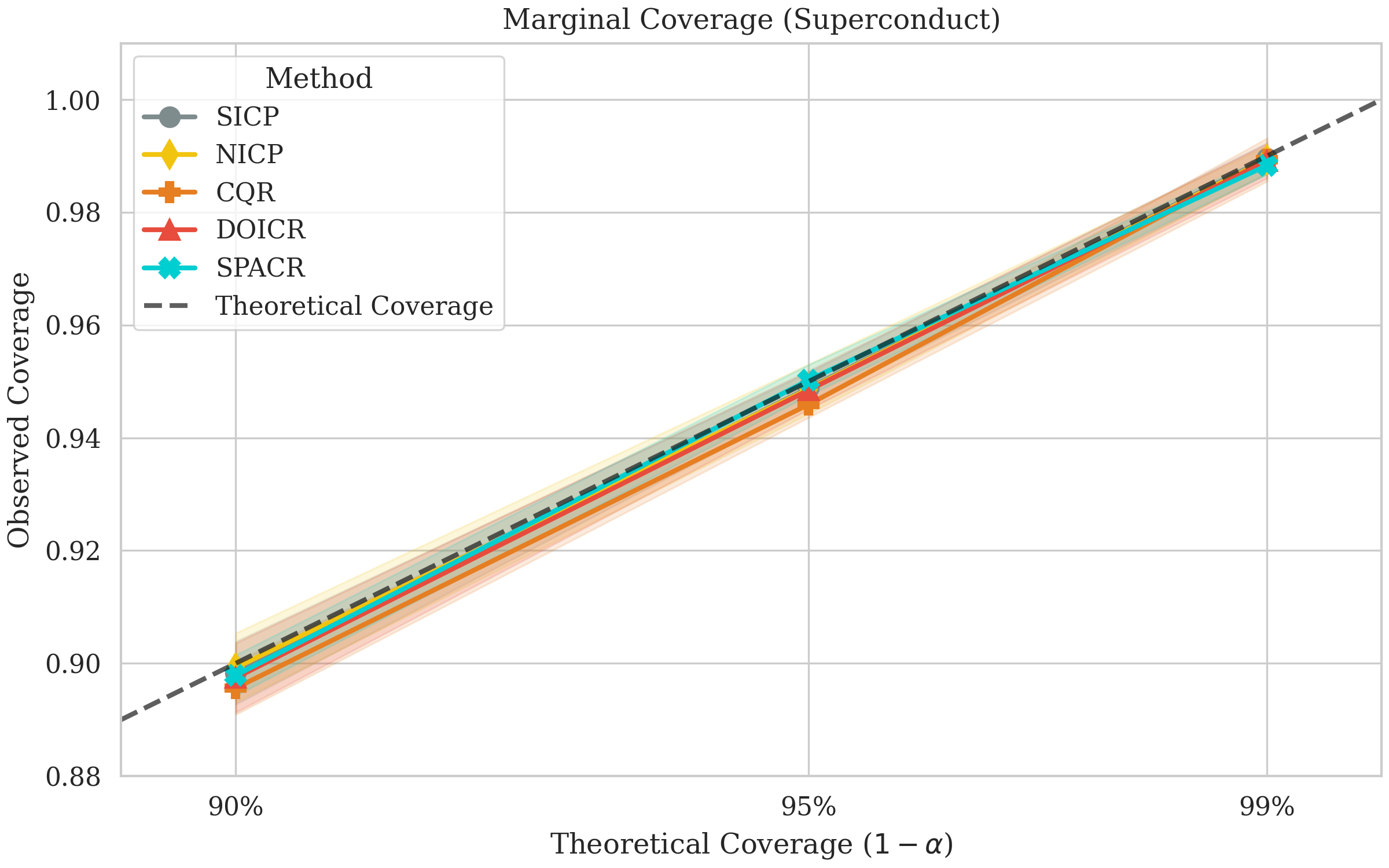}
    \caption{Marginal coverage vs. target confidence.}
    \label{fig:coverage_methods_Superconduct}
  \end{subfigure}
  \hfill
  \begin{subfigure}[t]{0.32\textwidth}
    \includegraphics[width=\textwidth]{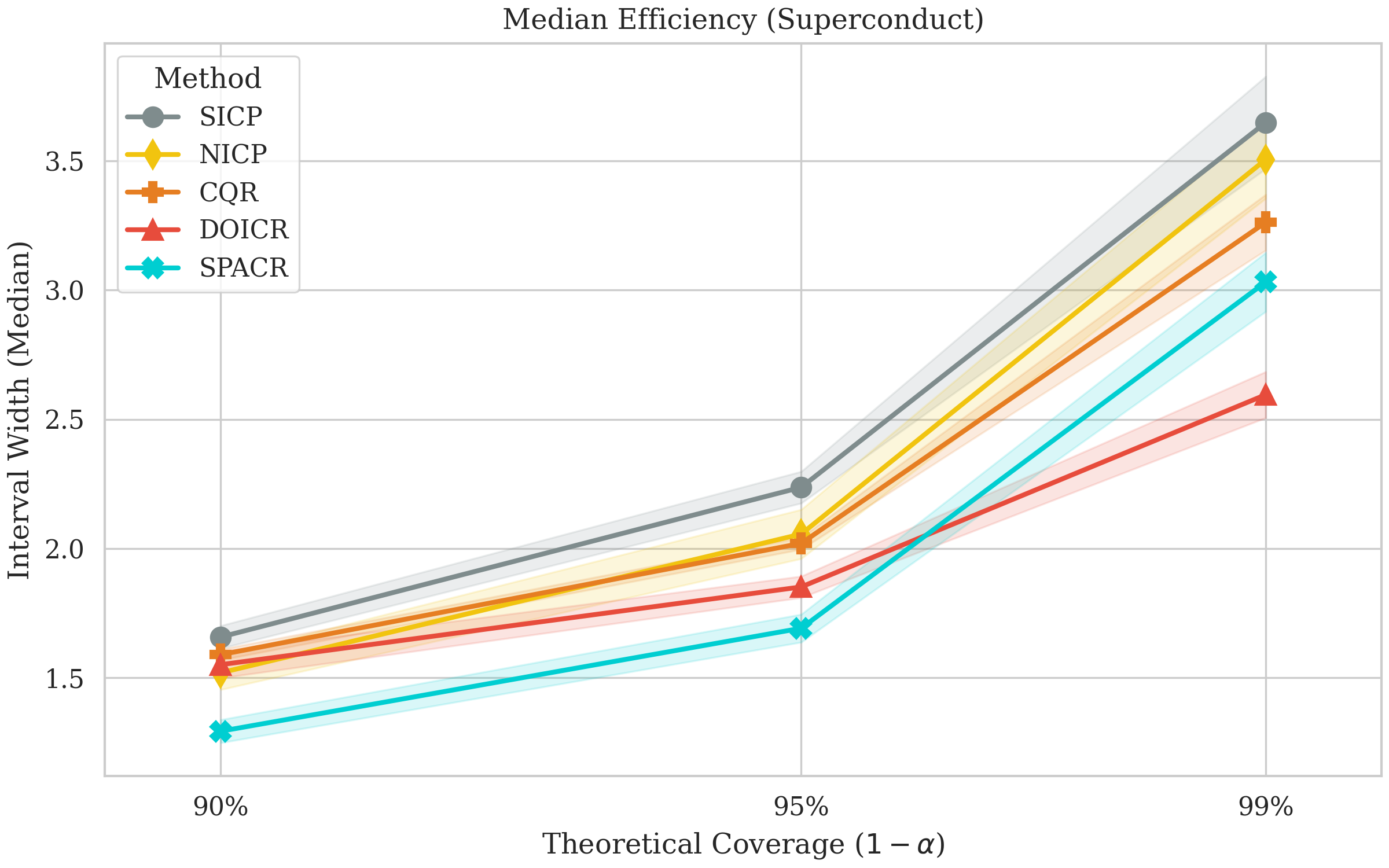}
    \caption{Efficiency (median interval width).}
    \label{fig:efficiency_methods_Superconduct}
  \end{subfigure}
  \hfill
  \begin{subfigure}[t]{0.32\textwidth}
    \includegraphics[width=\textwidth]{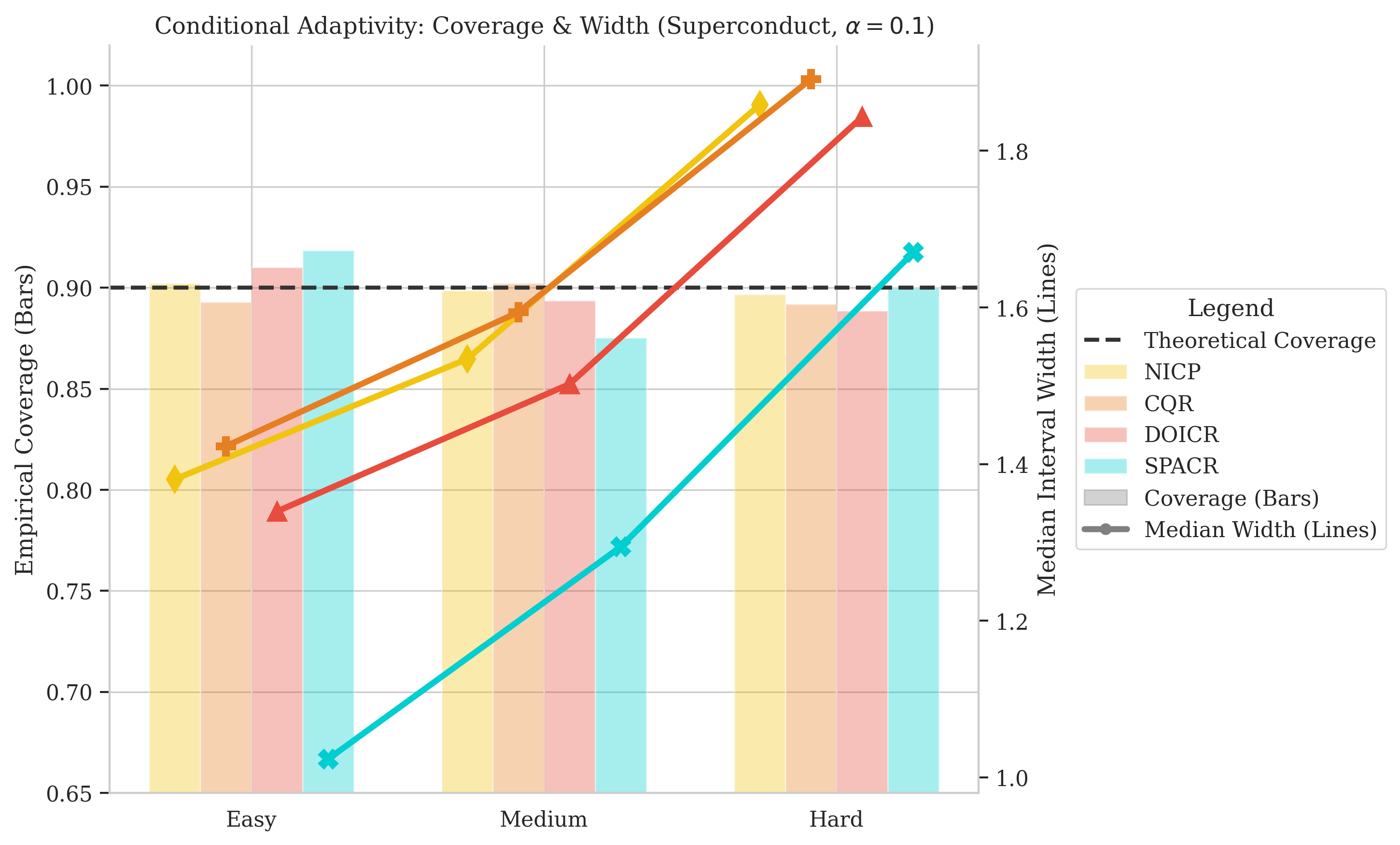}
    \caption{Conditional coverage (bars) and efficiency (lines) by internal difficulty.}
    \label{fig:iqr_methods_Superconduct}
  \end{subfigure}
    \caption{Performance of conformal regression methods on the Superconduct dataset.}
      \label{fig:methods_figures_Superconduct}
\end{figure*}

\begin{figure*}[!ht]
  \centering
  \begin{subfigure}[t]{0.32\textwidth}
    \includegraphics[width=\textwidth]{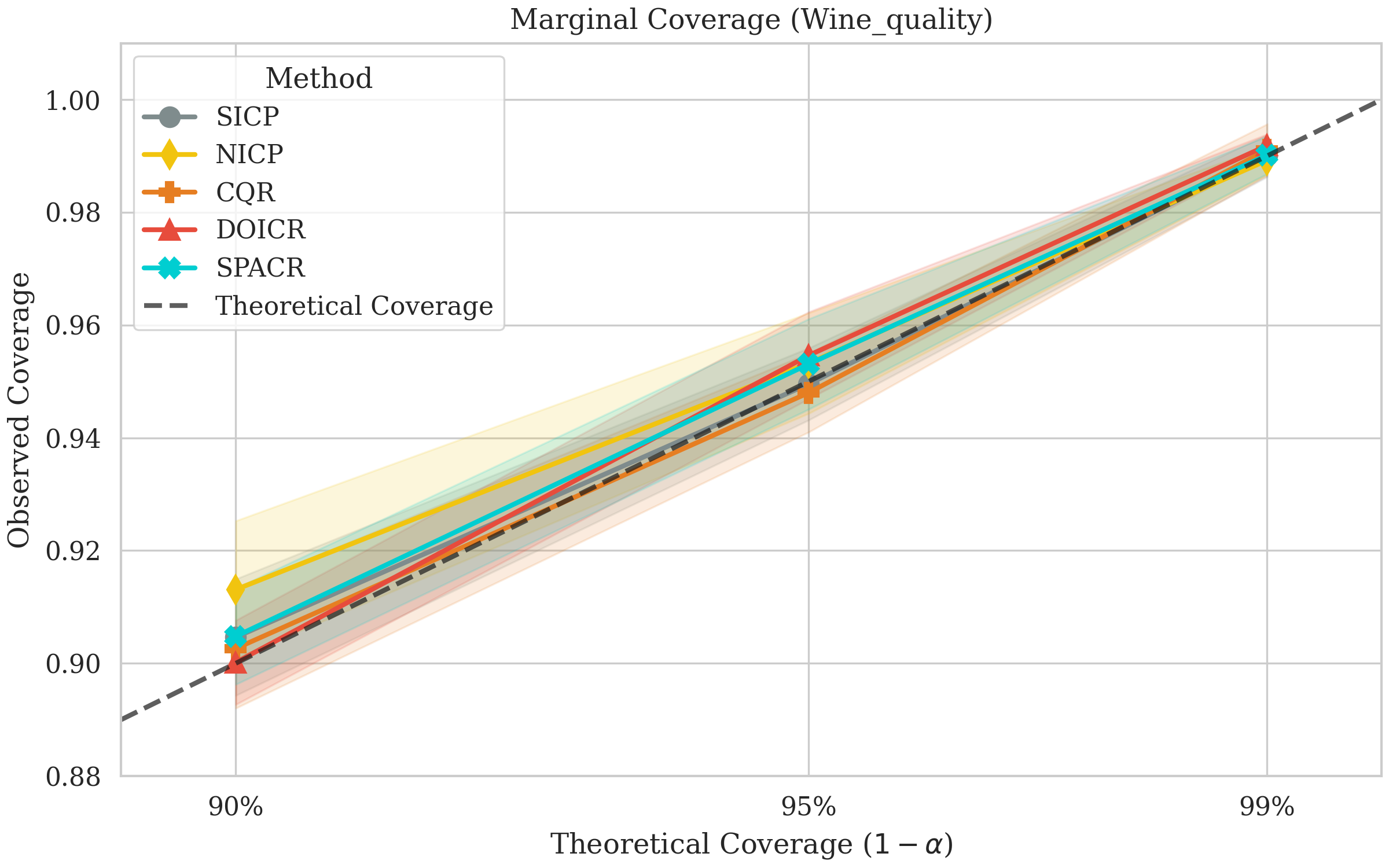}
    \caption{Marginal coverage vs. target confidence.}
    \label{fig:coverage_methods_Wine_quality}
  \end{subfigure}
  \hfill
  \begin{subfigure}[t]{0.32\textwidth}
    \includegraphics[width=\textwidth]{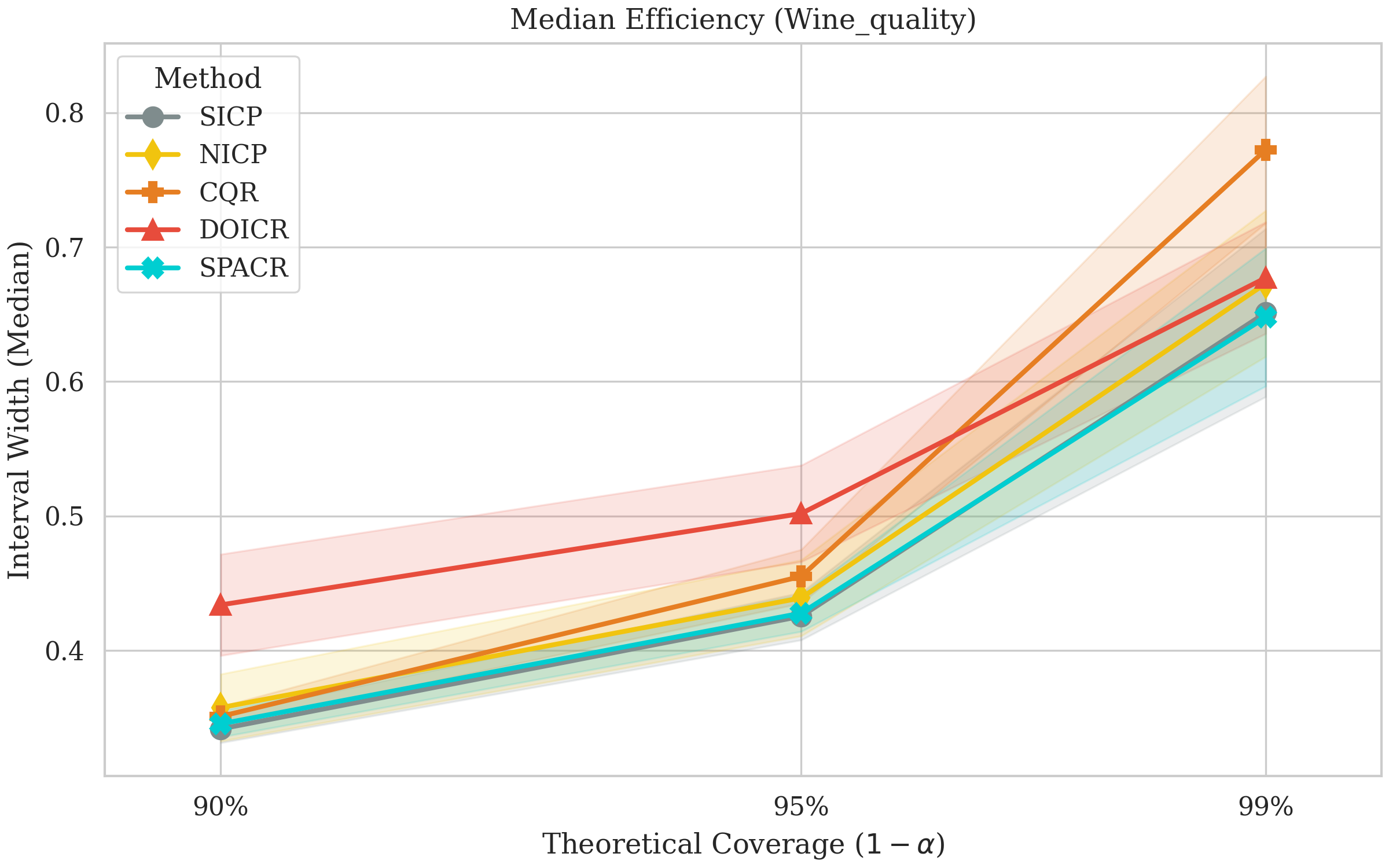}
    \caption{Efficiency (median interval width).}
    \label{fig:efficiency_methods_Wine_quality}
  \end{subfigure}
  \hfill
  \begin{subfigure}[t]{0.32\textwidth}
    \includegraphics[width=\textwidth]{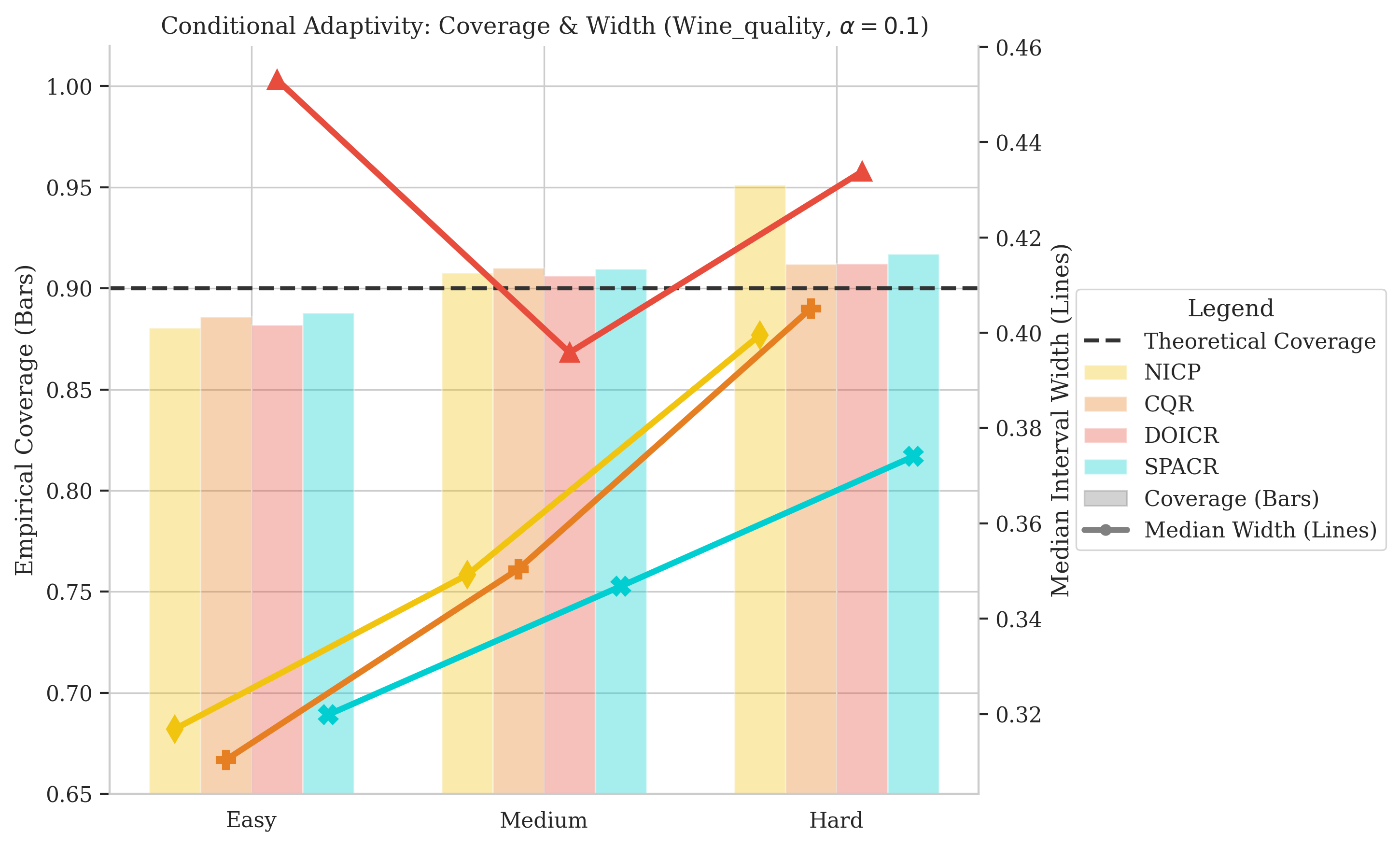}
    \caption{Conditional coverage (bars) and efficiency (lines) by internal difficulty.}
    \label{fig:iqr_methods_Wine_quality}
  \end{subfigure}
    \caption{Performance of conformal regression methods on the Wine Quality dataset.}
      \label{fig:methods_figures_Wine_quality}
\end{figure*}

\begin{figure*}[!ht]
  \centering
  \begin{subfigure}[t]{0.32\textwidth}
    \includegraphics[width=\textwidth]{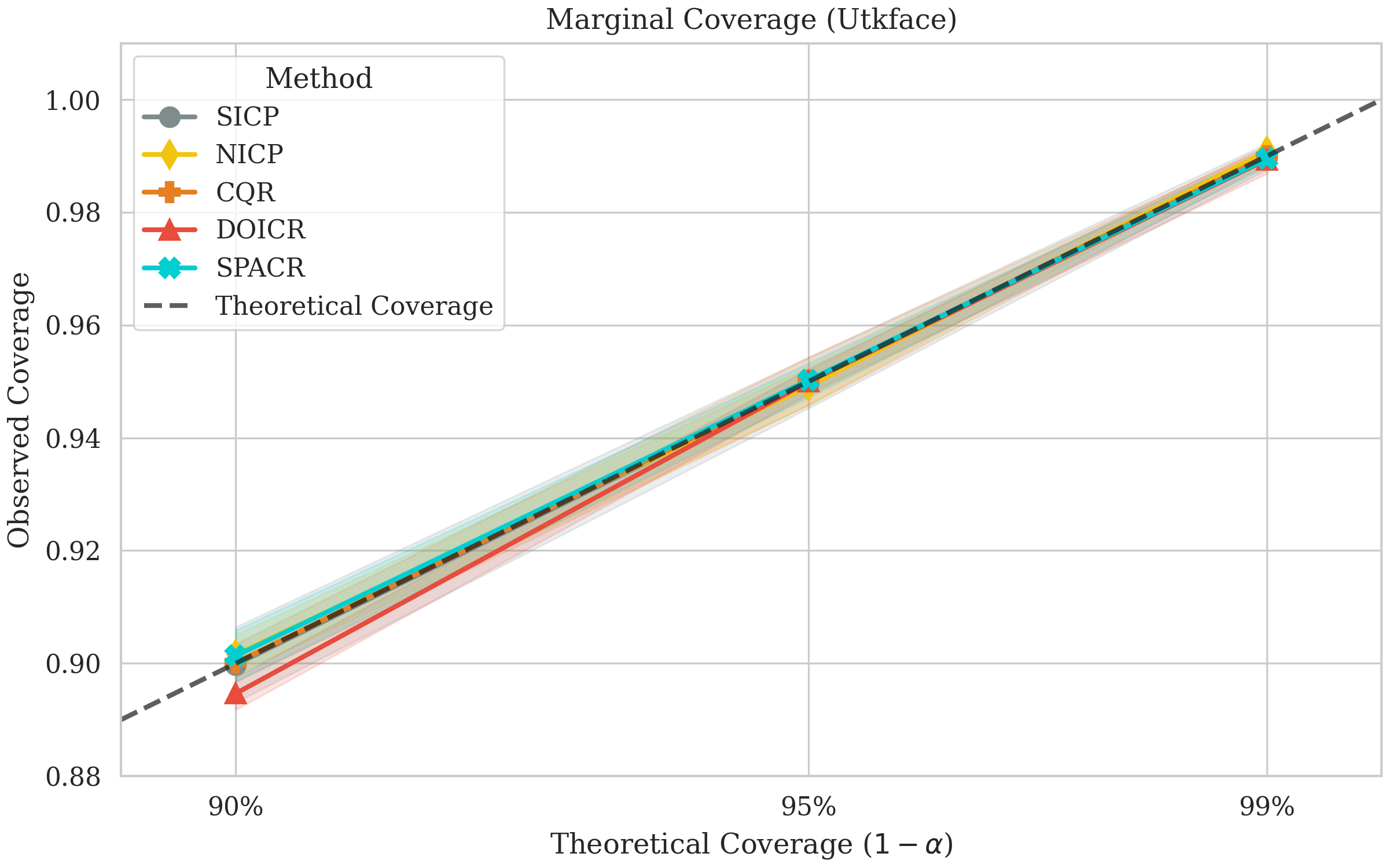}
    \caption{Marginal coverage vs. target confidence.}
    \label{fig:coverage_methods_Utkface}
  \end{subfigure}
  \hfill
  \begin{subfigure}[t]{0.32\textwidth}
    \includegraphics[width=\textwidth]{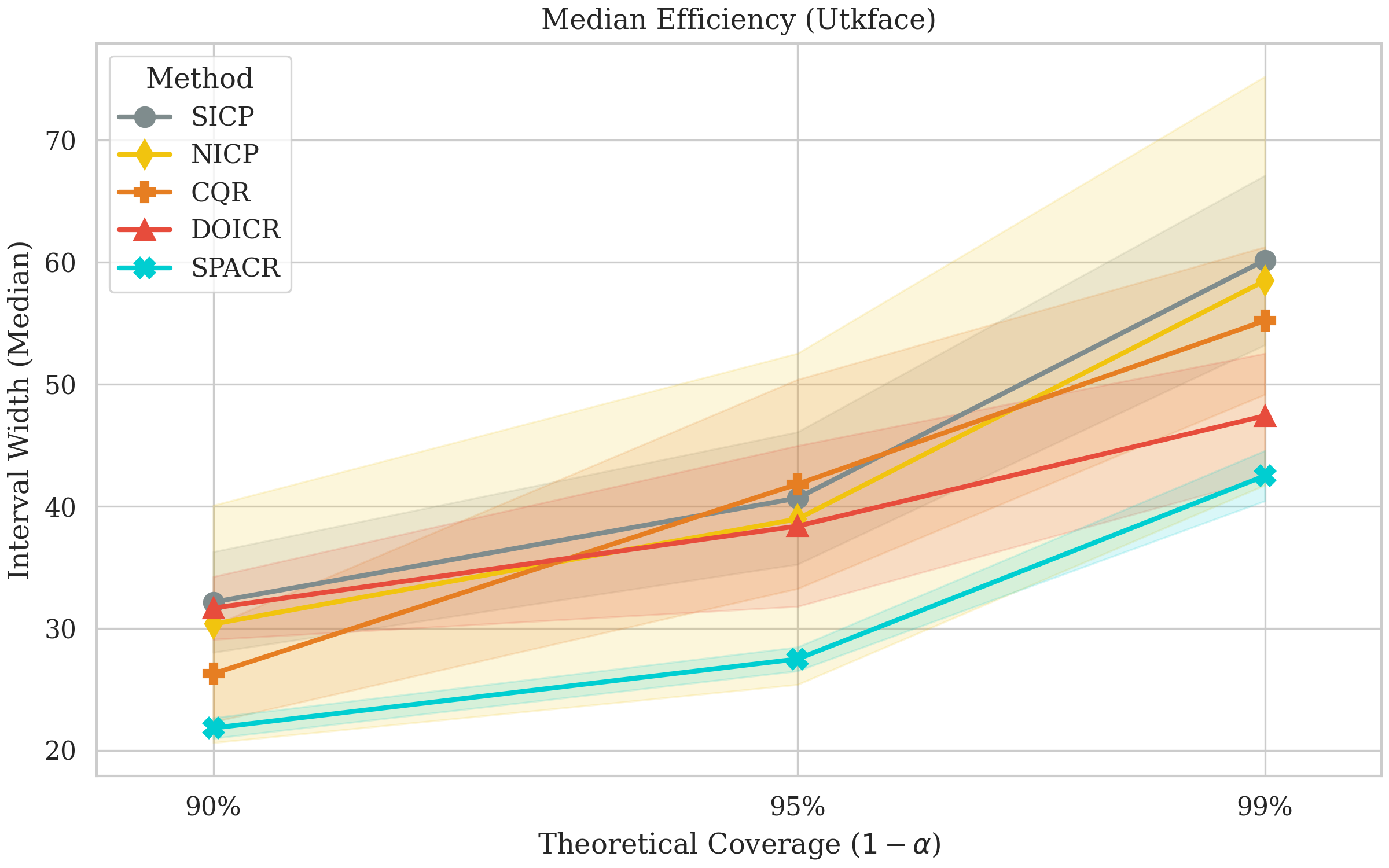}
    \caption{Efficiency (median interval width).}
    \label{fig:efficiency_methods_Utkface}
  \end{subfigure}
  \hfill
  \begin{subfigure}[t]{0.32\textwidth}
    \includegraphics[width=\textwidth]{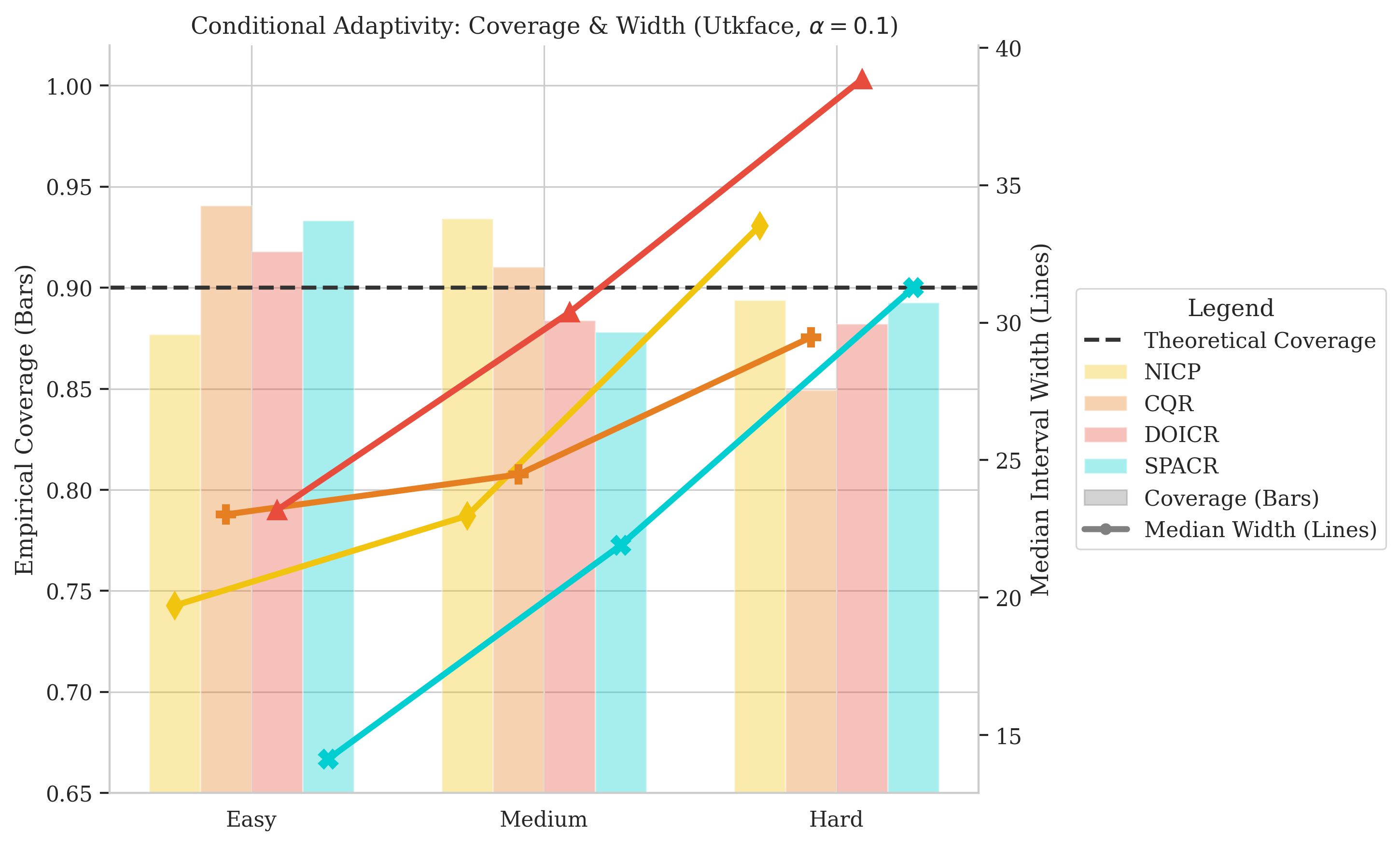}
    \caption{Conditional coverage (bars) and efficiency (lines) by internal difficulty.}
    \label{fig:iqr_methods_Utkface}
  \end{subfigure}
    \caption{Performance of conformal regression methods on the UTK Face dataset (image regression).}
      \label{fig:methods_figures_Utkface}
\end{figure*}

Figures \ref{fig:methods_figures_bike_sharing}, \ref{fig:methods_figures_br}, \ref{fig:methods_figures_California}, \ref{fig:methods_figures_Cpu_act}, \ref{fig:methods_figures_Fifa}, \ref{fig:methods_figures_House_sales}, \ref{fig:methods_figures_Isolet}, \ref{fig:methods_figures_Medical_charges}, \ref{fig:methods_figures_Pol}, \ref{fig:methods_figures_Superconduct}, \ref{fig:methods_figures_Wine_quality}, and \ref{fig:methods_figures_Utkface} detail the coverage and efficiency metrics for the remaining datasets. These results reinforce the main findings of the study: while all methods generally achieve the desired coverage, SPACR consistently produces the narrowest and most stable prediction intervals with robust conditional adaptivity.

\clearpage
\subsection{Sensitivity Analysis of the Regularization Parameter $\lambda$}

\begin{figure*}[!ht]
  \centering
  \begin{subfigure}[t]{0.32\textwidth}
    \includegraphics[width=\textwidth]{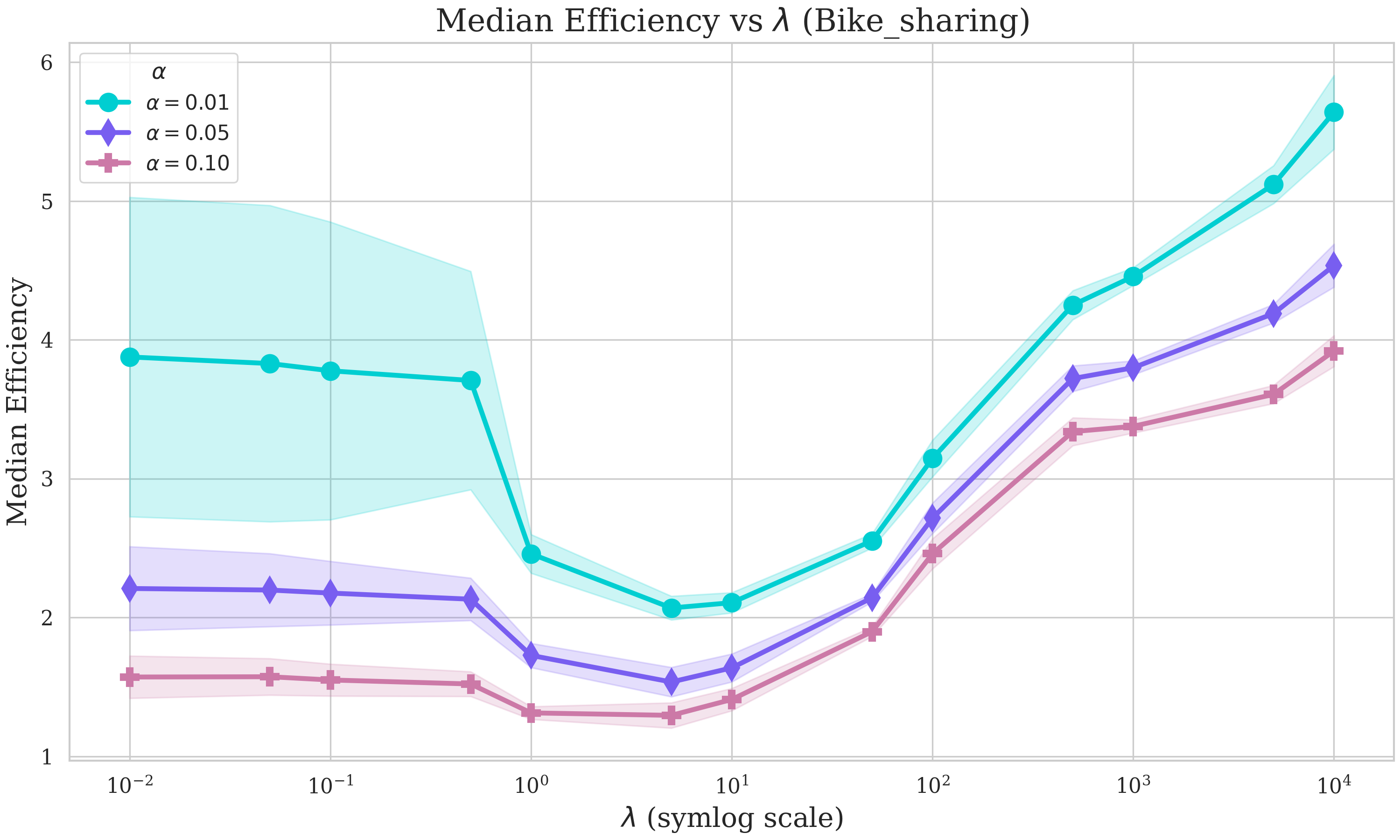}
    \caption{Median efficiency.}
    \label{fig:efficiency_lambda_Bike_sharing}
  \end{subfigure}
  \hfill
  \begin{subfigure}[t]{0.32\textwidth}
    \includegraphics[width=\textwidth]{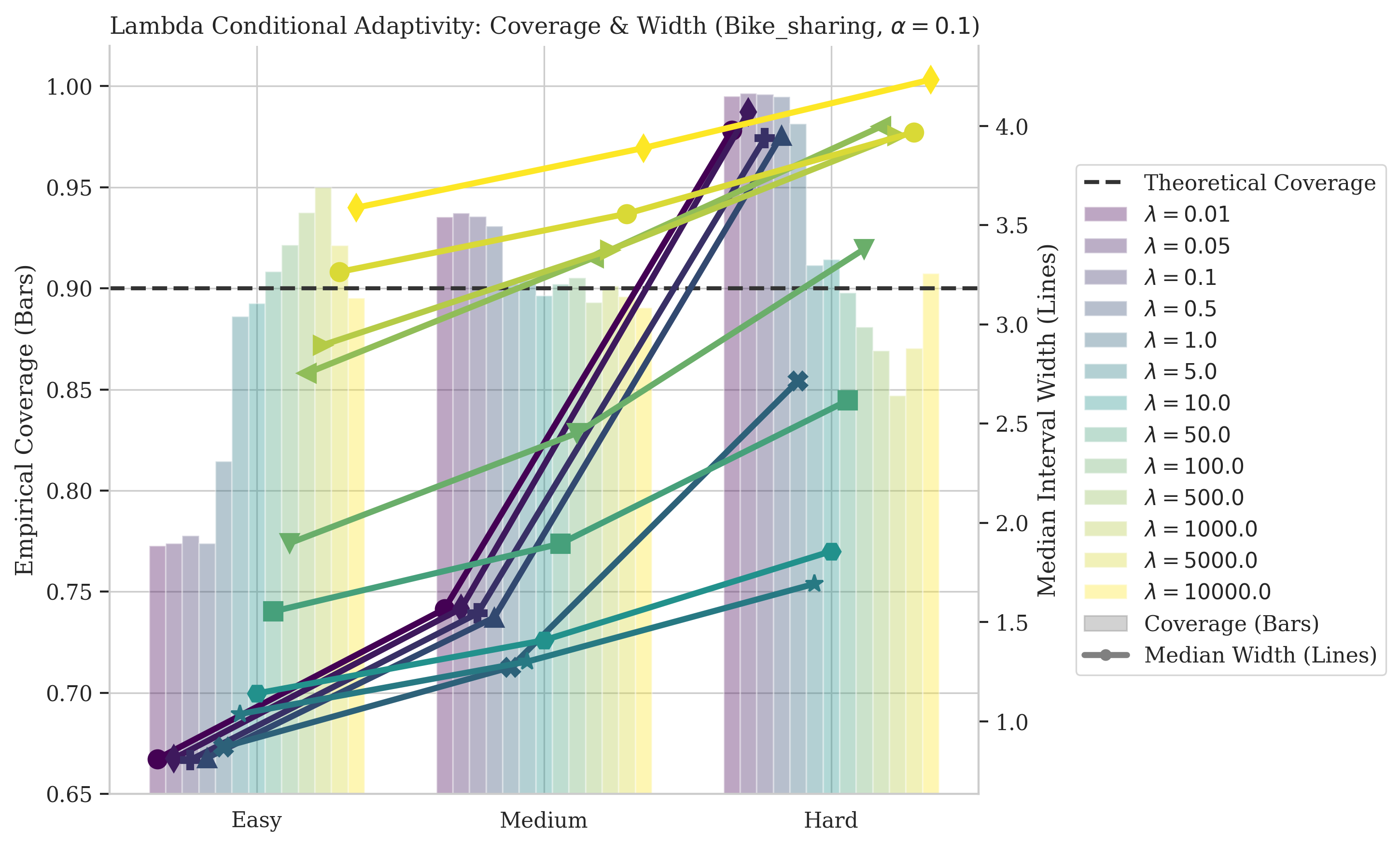}
    \caption{Conditional coverage (bars) and efficiency (lines) by internal difficulty.}
    \label{fig:box_plot_lambda_Bike_sharing}
  \end{subfigure}
    \hfill
  \begin{subfigure}[t]{0.32\textwidth}
    \includegraphics[width=\textwidth]{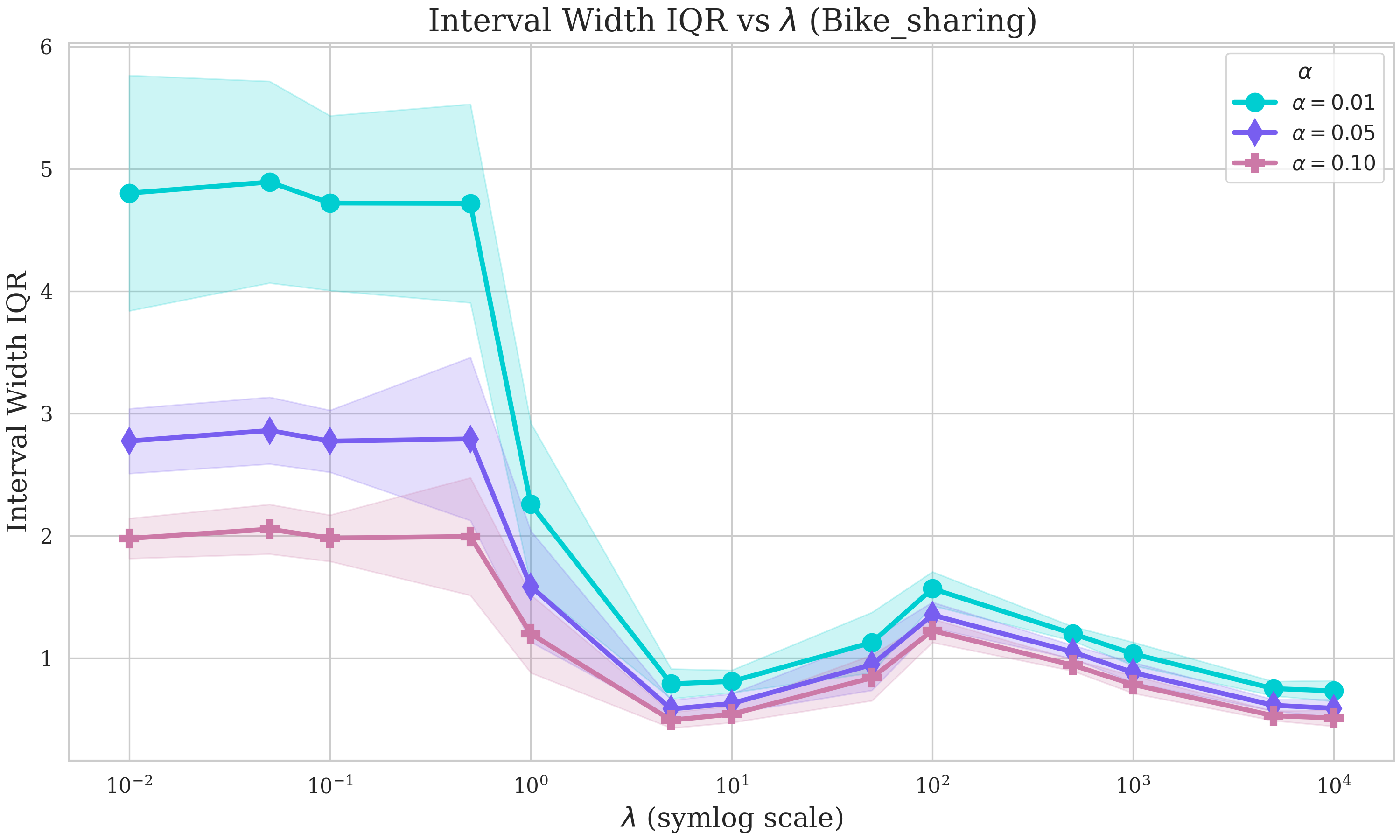}
    \caption{Interval width IQR.}
    \label{fig:iqr_lambda_Bike_sharing}
  \end{subfigure}
    \caption{Sensitivity analysis of the regularization parameter $\lambda$ on the Bike Sharing dataset.}
  \label{fig:lambda_figures_Bike_sharing}
\end{figure*}

\begin{figure*}[!ht]
  \centering
  \begin{subfigure}[t]{0.32\textwidth}
    \includegraphics[width=\textwidth]{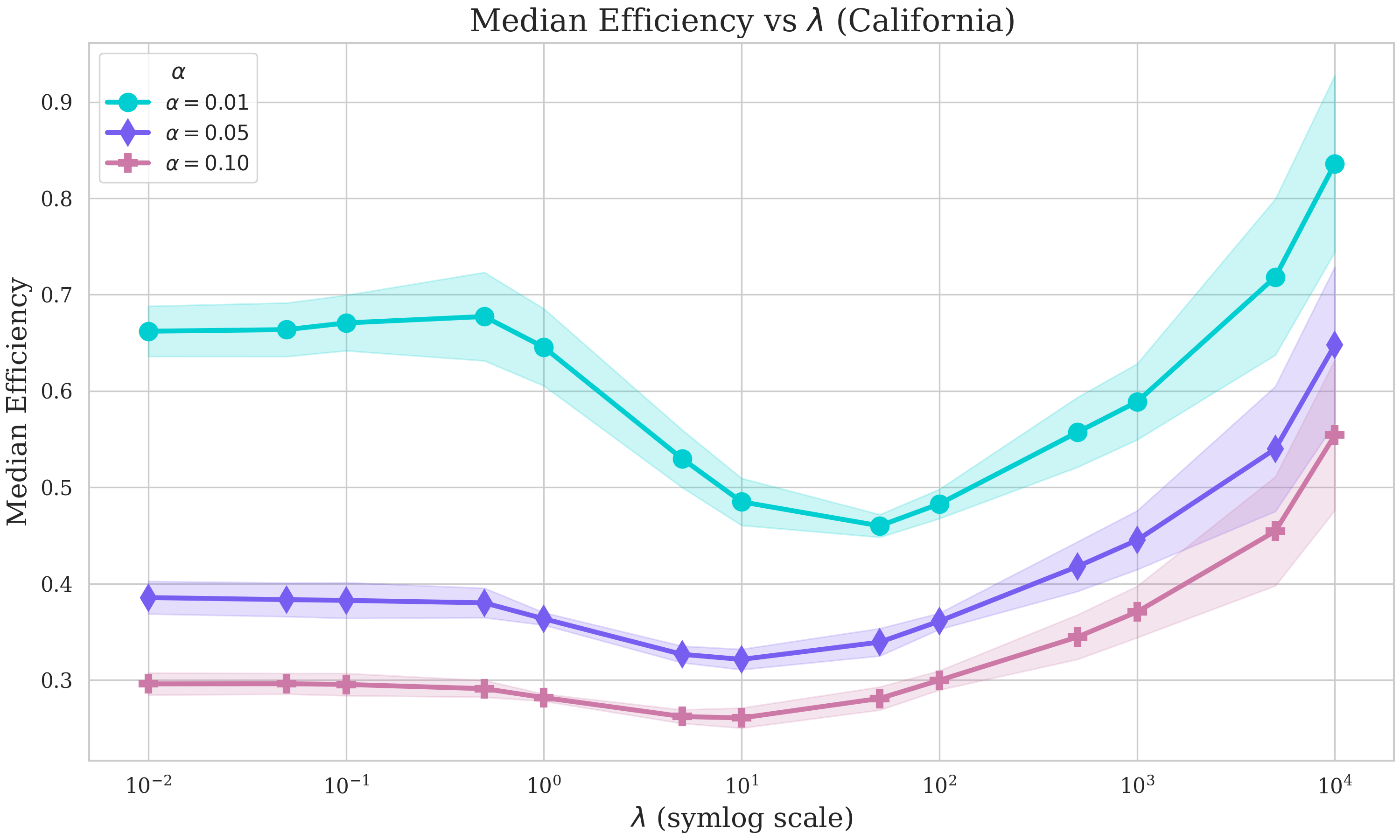}
    \caption{Median efficiency.}
    \label{fig:efficiency_lambda_California}
  \end{subfigure}
  \hfill
  \begin{subfigure}[t]{0.32\textwidth}
    \includegraphics[width=\textwidth]{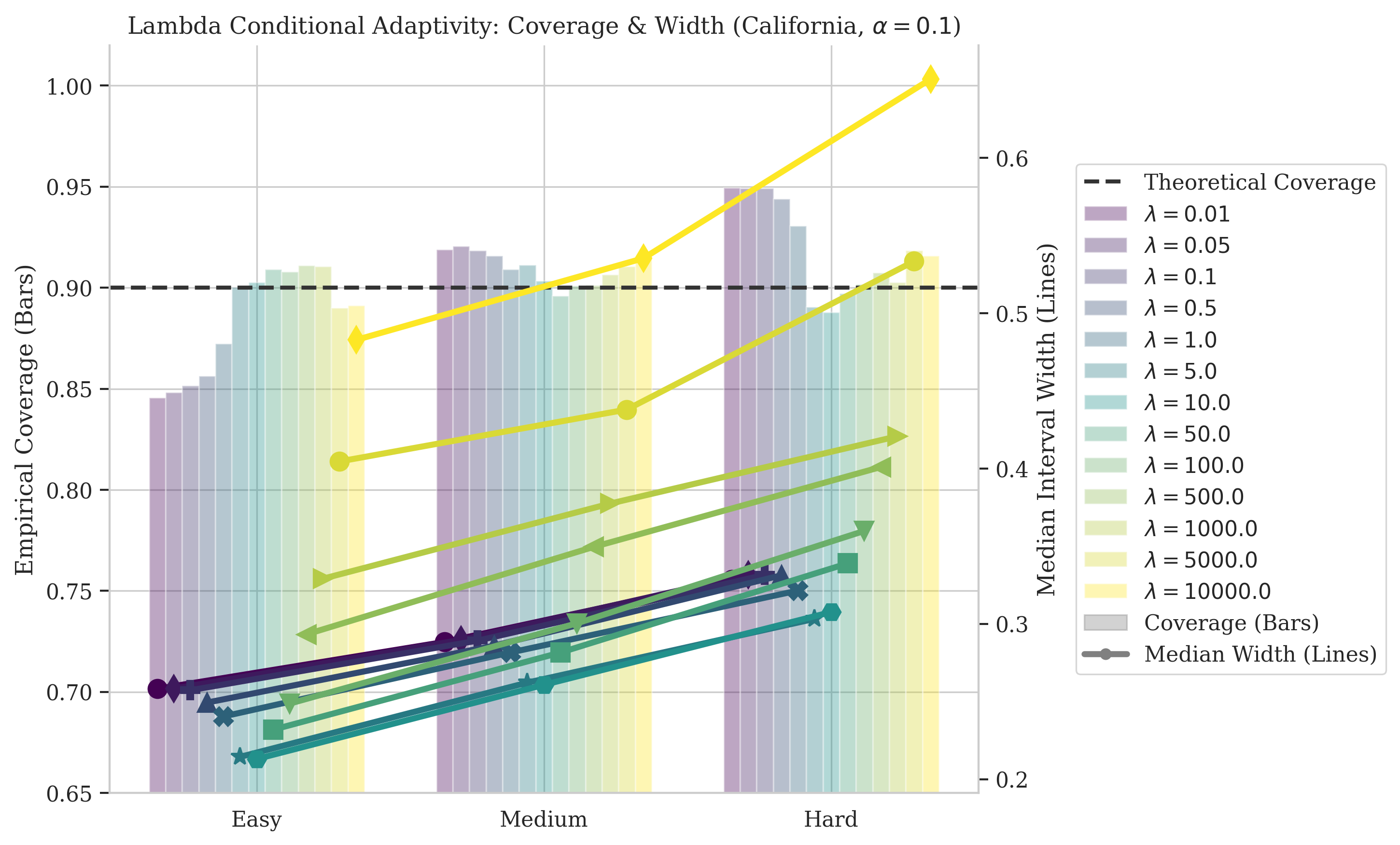}
    \caption{Conditional coverage (bars) and efficiency (lines) by internal difficulty.}
    \label{fig:box_plot_lambda_California}
  \end{subfigure}  
    \hfill
  \begin{subfigure}[t]{0.32\textwidth}
    \includegraphics[width=\textwidth]{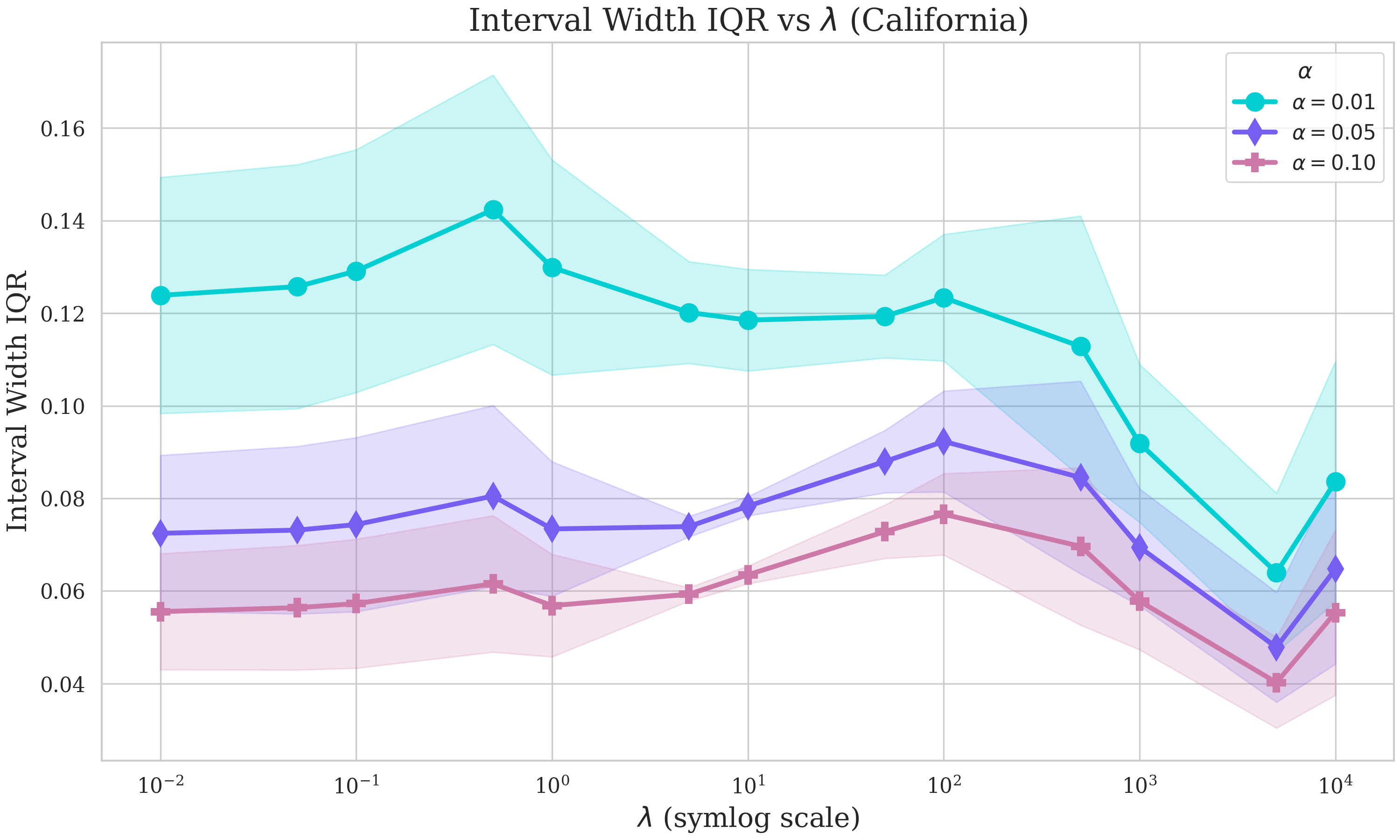}
    \caption{Interval width IQR.}
    \label{fig:iqr_lambda_California}
  \end{subfigure}
    \caption{Sensitivity analysis of the regularization parameter $\lambda$ on the California dataset.}
  \label{fig:lambda_figures_California}
\end{figure*}

\begin{figure*}[!ht]
  \centering
  \begin{subfigure}[t]{0.32\textwidth}
    \includegraphics[width=\textwidth]{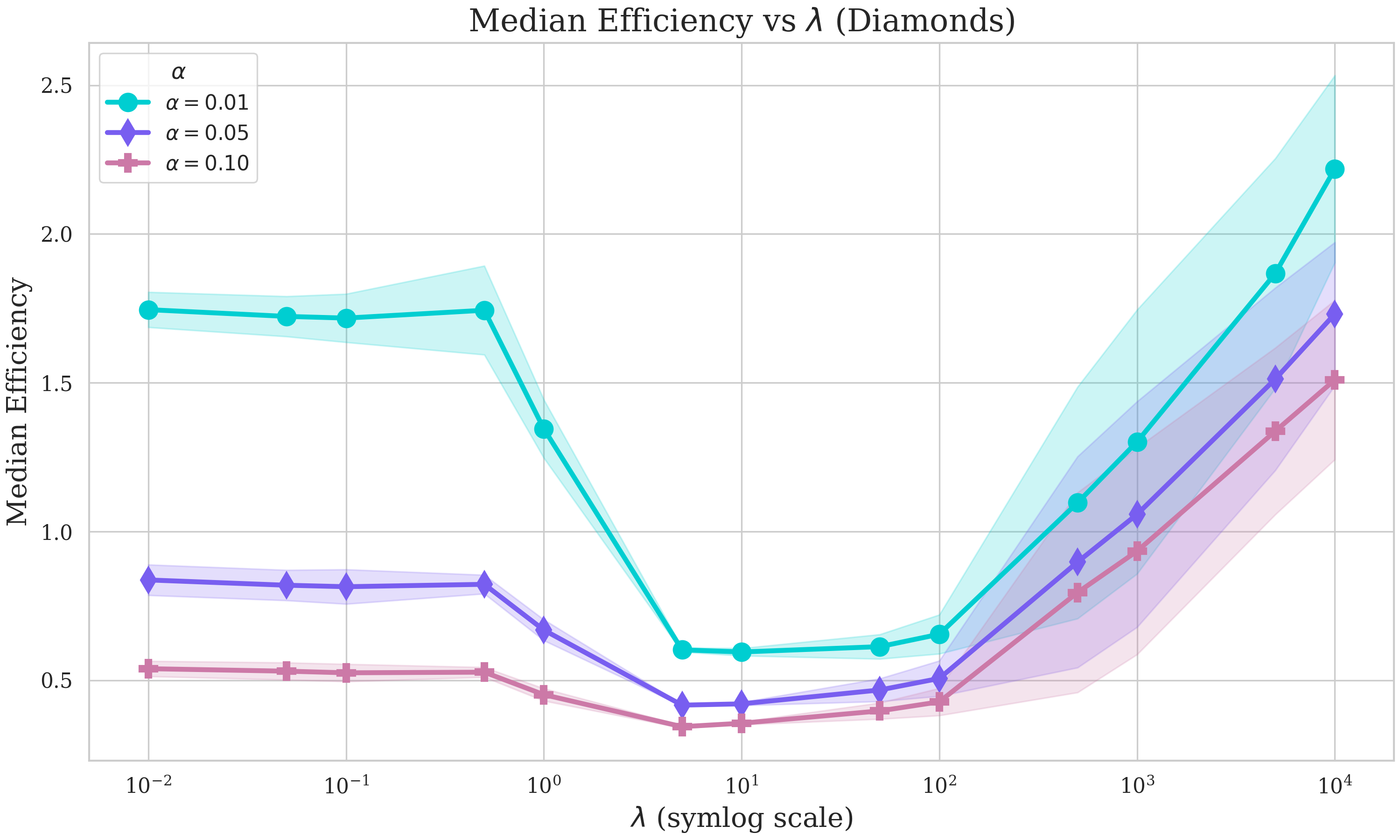}
    \caption{Median efficiency.}
    \label{fig:efficiency_lambda_Diamonds}
  \end{subfigure}
  \hfill
  \begin{subfigure}[t]{0.32\textwidth}
    \includegraphics[width=\textwidth]{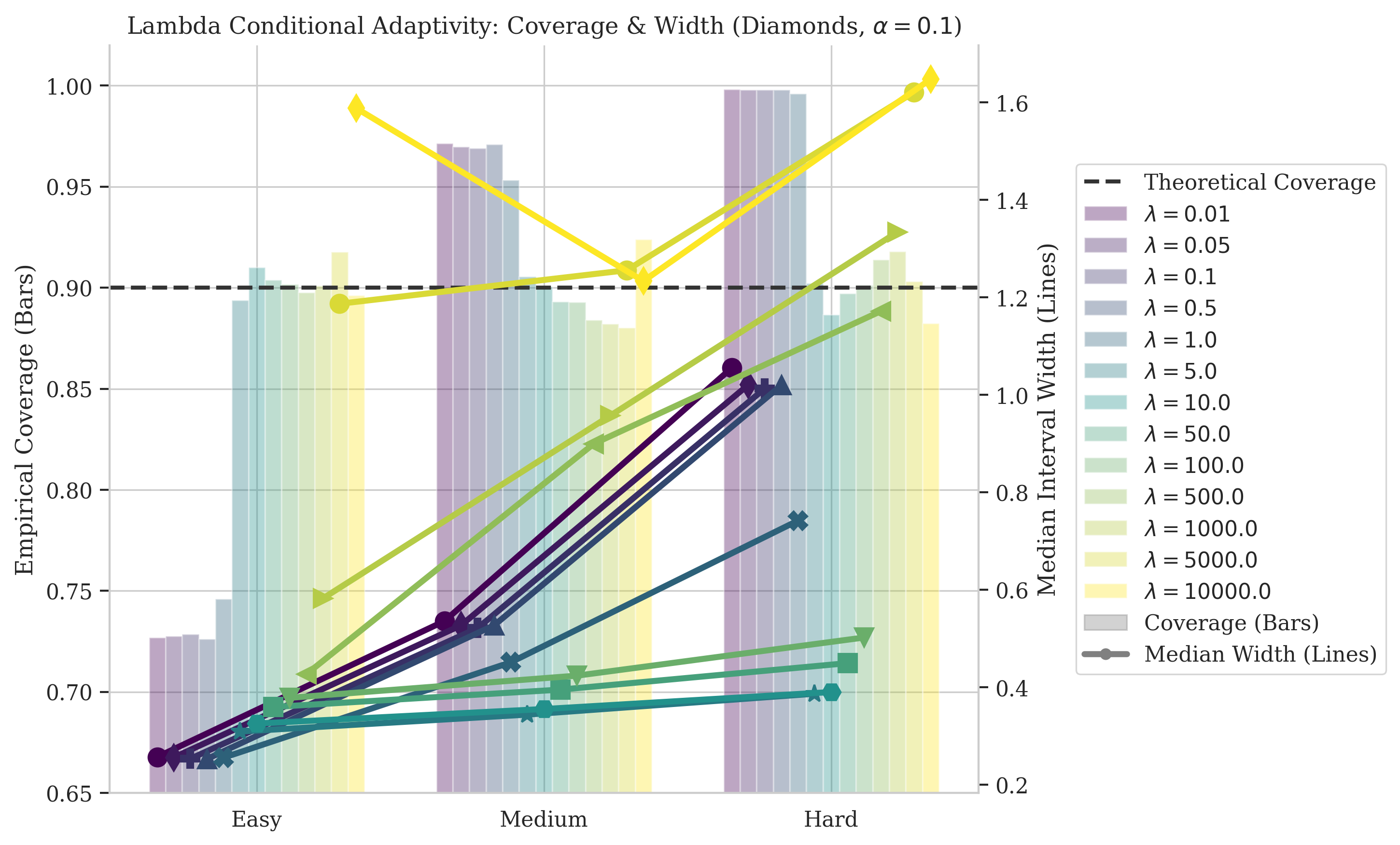}
    \caption{Conditional coverage (bars) and efficiency (lines) by internal difficulty.}
    \label{fig:box_plot_lambda_Diamonds}
  \end{subfigure}  
    \hfill
  \begin{subfigure}[t]{0.32\textwidth}
    \includegraphics[width=\textwidth]{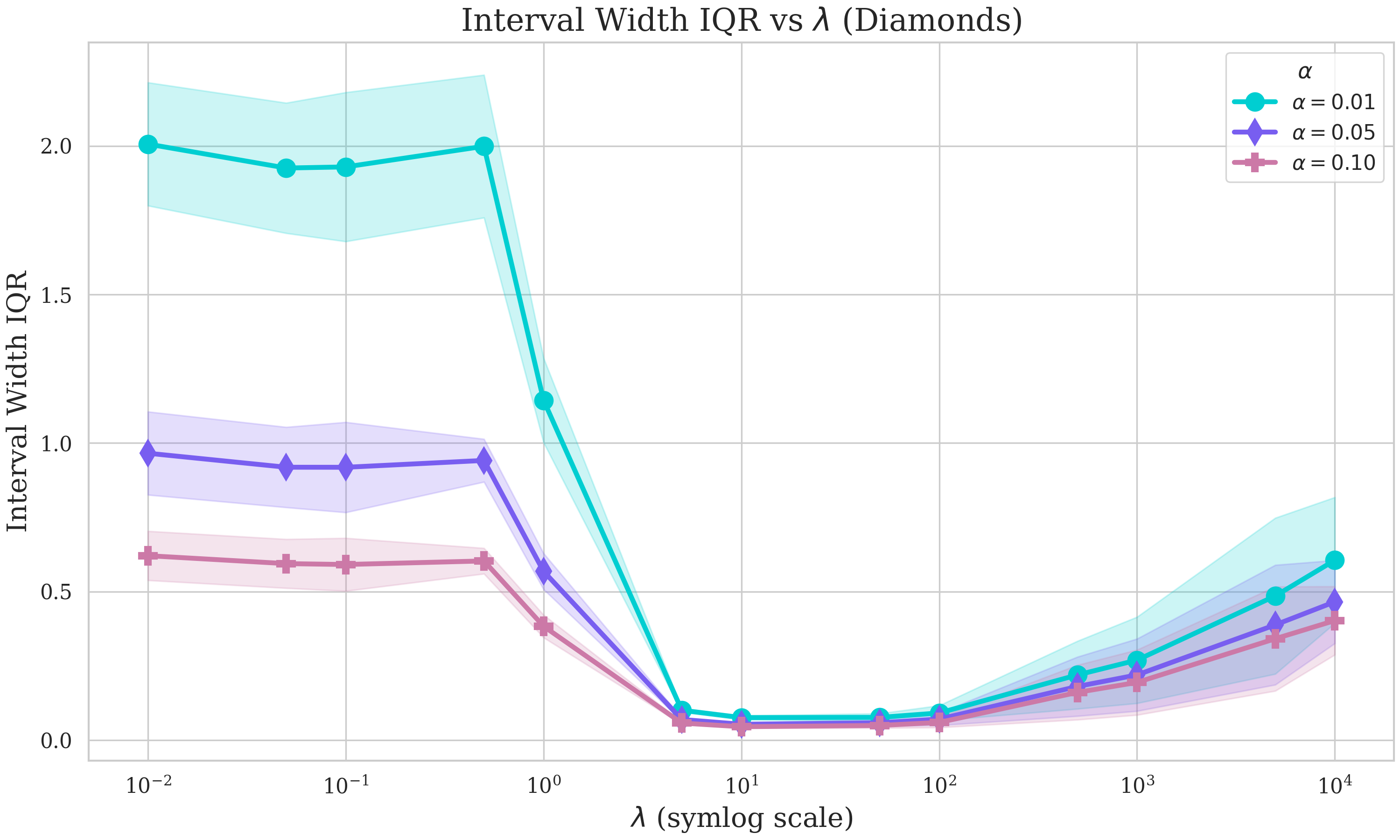}
    \caption{Interval width IQR.}
    \label{fig:iqr_lambda_Diamonds}
  \end{subfigure}
    \caption{Sensitivity analysis of the regularization parameter $\lambda$ on the Diamonds dataset.}
  \label{fig:lambda_figures_Diamonds}
\end{figure*}

\begin{figure*}[!ht]
  \centering
  \begin{subfigure}[t]{0.32\textwidth}
    \includegraphics[width=\textwidth]{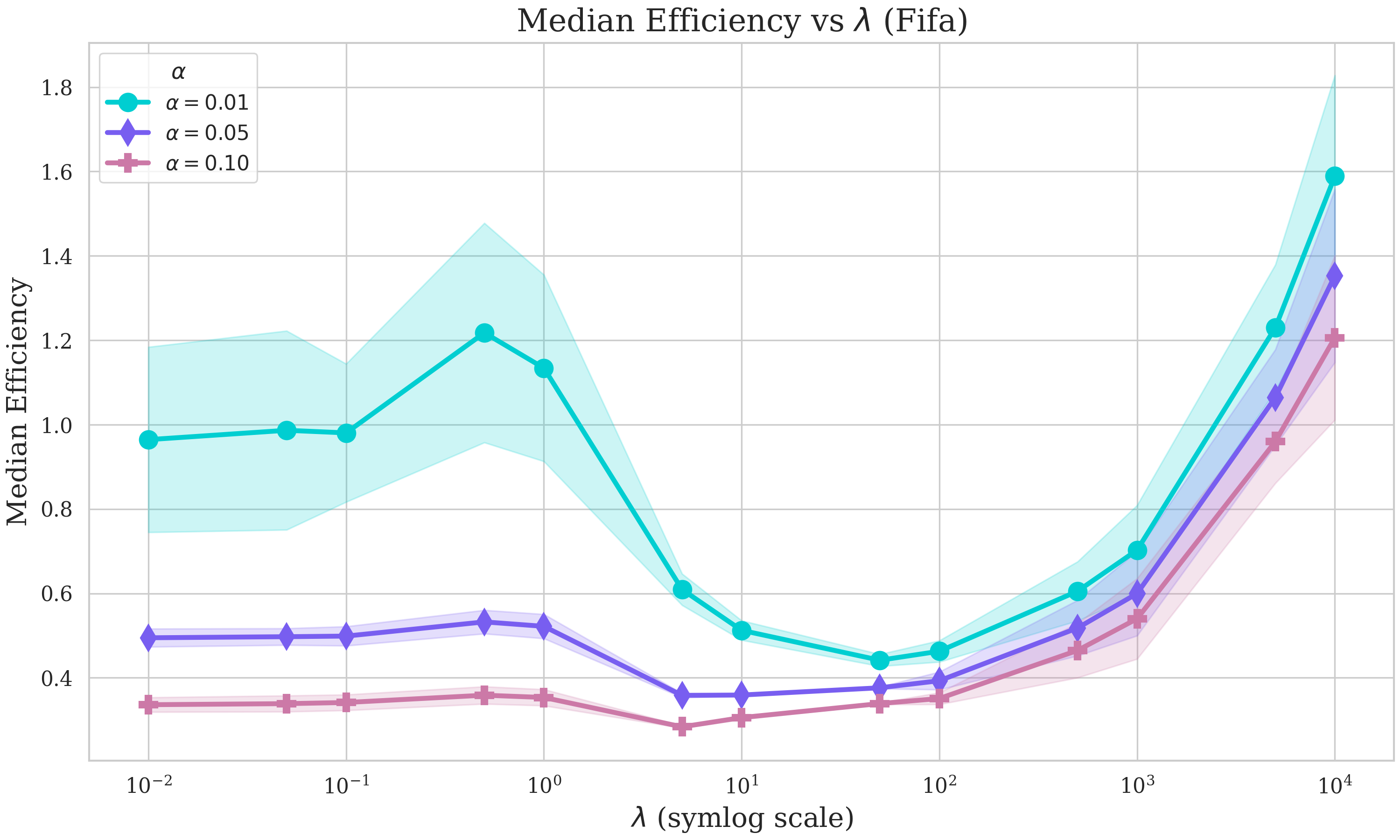}
    \caption{Median efficiency.}
    \label{fig:efficiency_lambda_Fifa}
  \end{subfigure}
  \hfill
  \begin{subfigure}[t]{0.32\textwidth}
    \includegraphics[width=\textwidth]{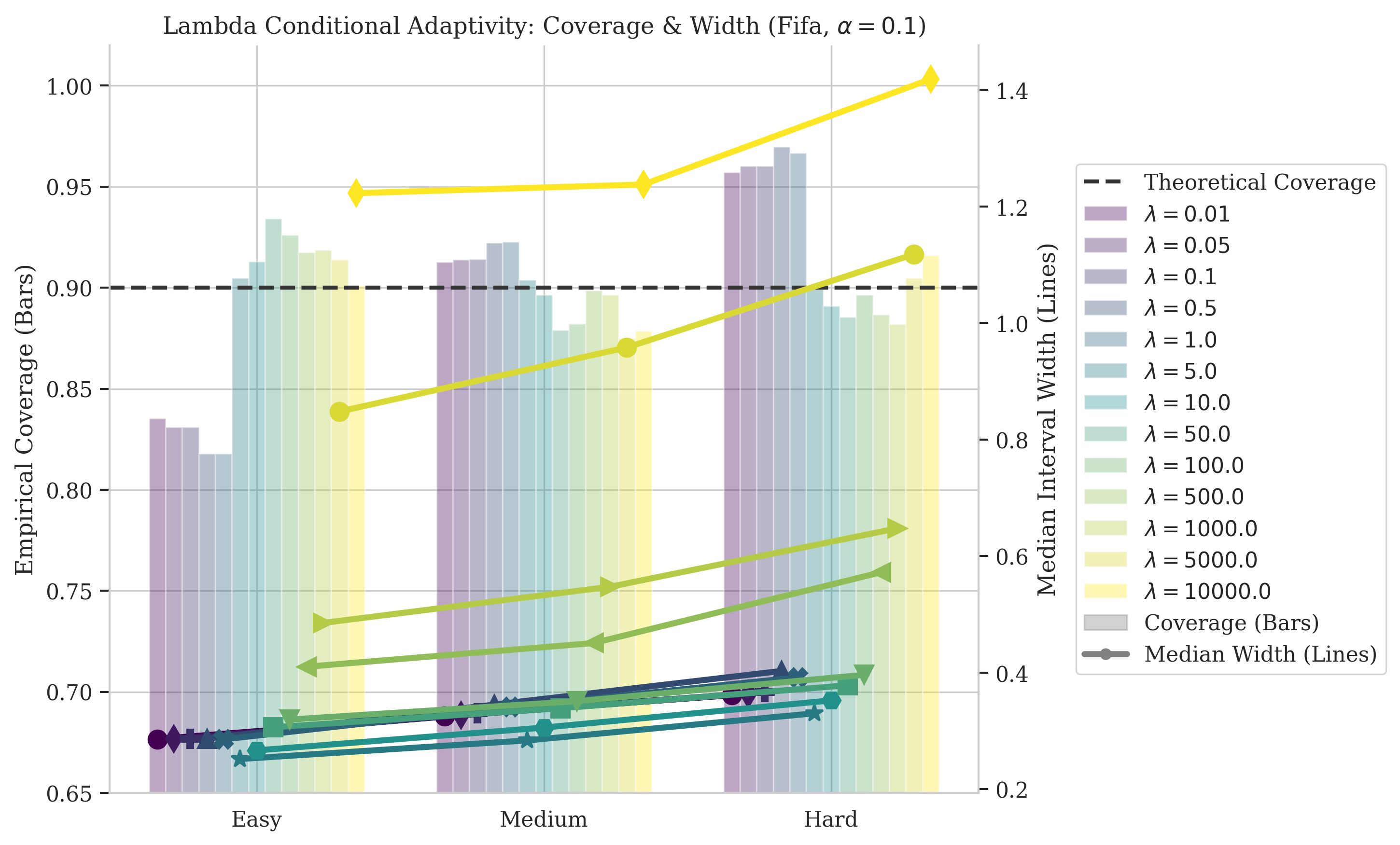}
    \caption{Conditional coverage (bars) and efficiency (lines) by internal difficulty.}
    \label{fig:box_plot_lambda_Fifa}
  \end{subfigure}  
    \hfill
  \begin{subfigure}[t]{0.32\textwidth}
    \includegraphics[width=\textwidth]{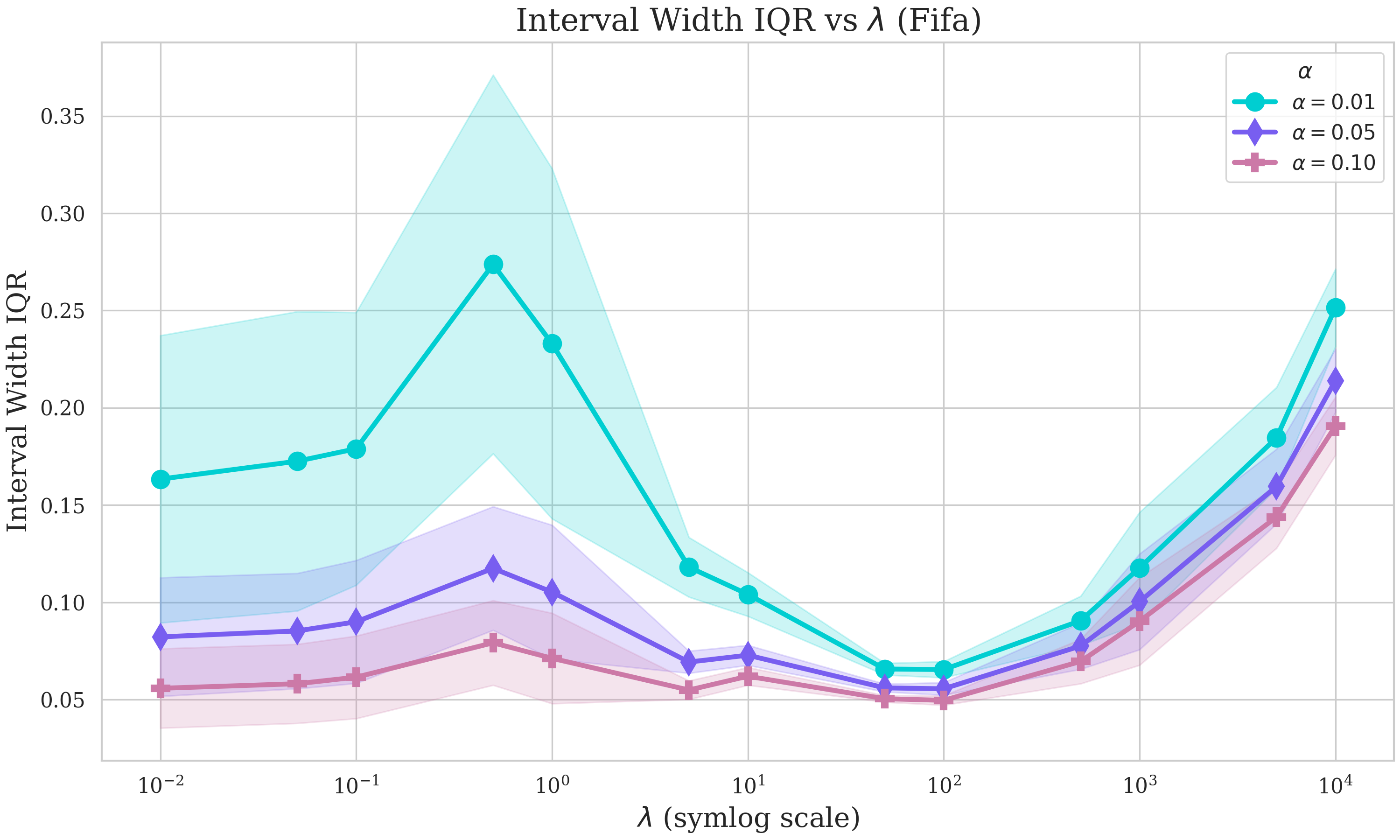}
    \caption{Interval width IQR.}
    \label{fig:iqr_lambda_Fifa}
  \end{subfigure}
    \caption{Sensitivity analysis of the regularization parameter $\lambda$ on the Fifa dataset.}
  \label{fig:lambda_figures_Fifa}
\end{figure*}

\begin{figure*}[!ht]
  \centering
  \begin{subfigure}[t]{0.32\textwidth}
    \includegraphics[width=\textwidth]{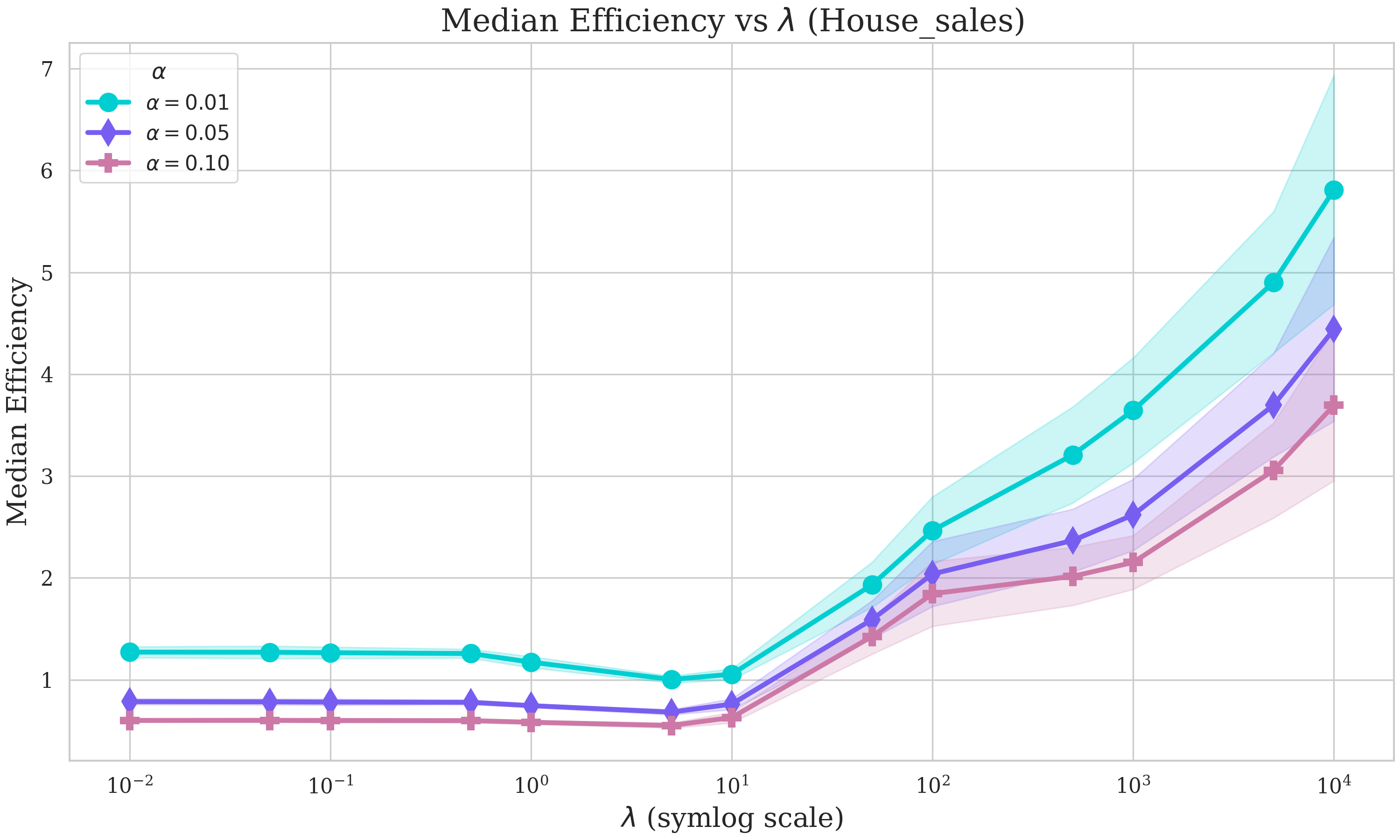}
    \caption{Median efficiency.}
    \label{fig:efficiency_lambda_House_sales}
  \end{subfigure}
  \hfill
  \begin{subfigure}[t]{0.32\textwidth}
    \includegraphics[width=\textwidth]{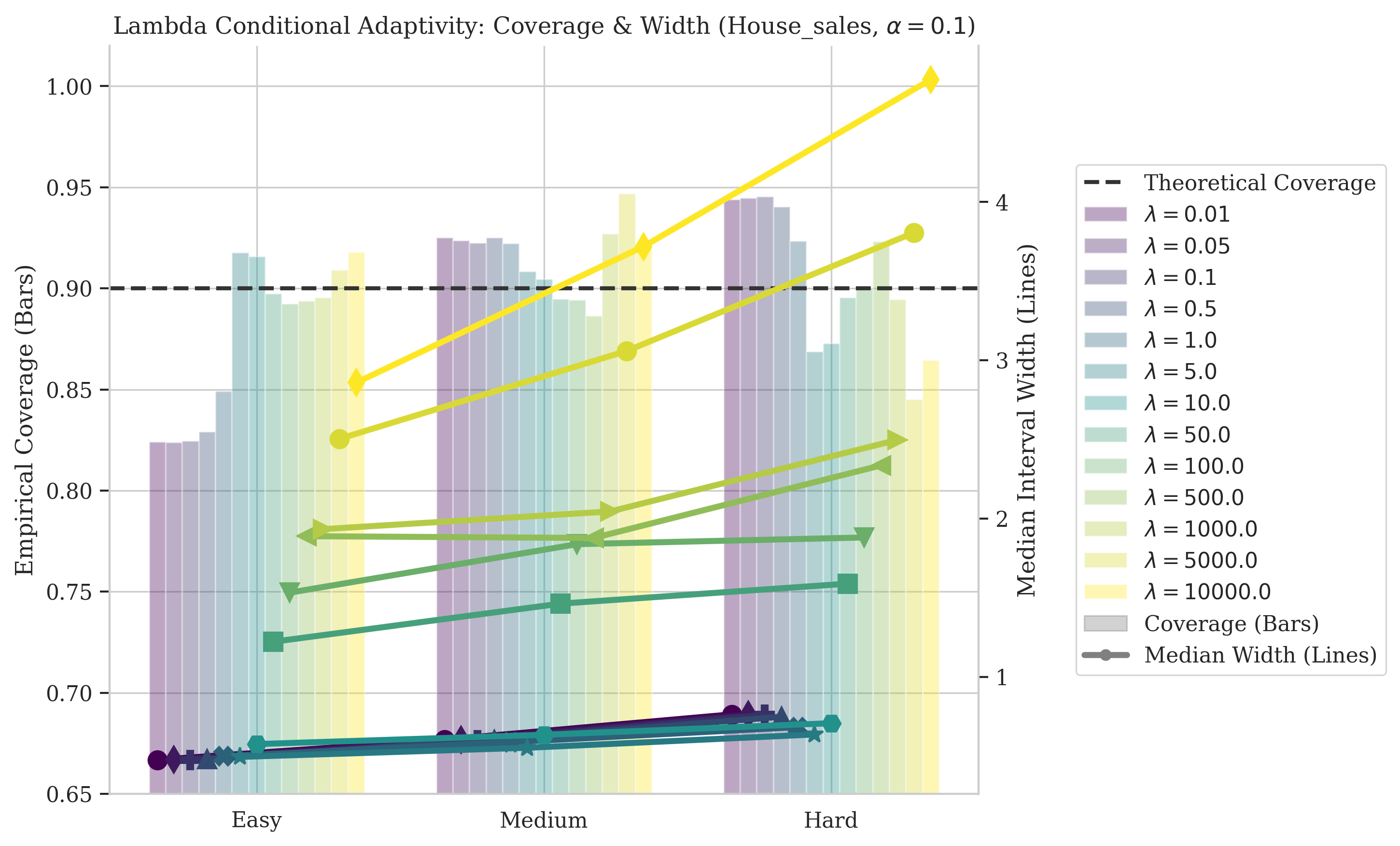}
    \caption{Conditional coverage (bars) and efficiency (lines) by internal difficulty.}
    \label{fig:box_plot_lambda_House_sales}
  \end{subfigure}  
    \hfill
  \begin{subfigure}[t]{0.32\textwidth}
    \includegraphics[width=\textwidth]{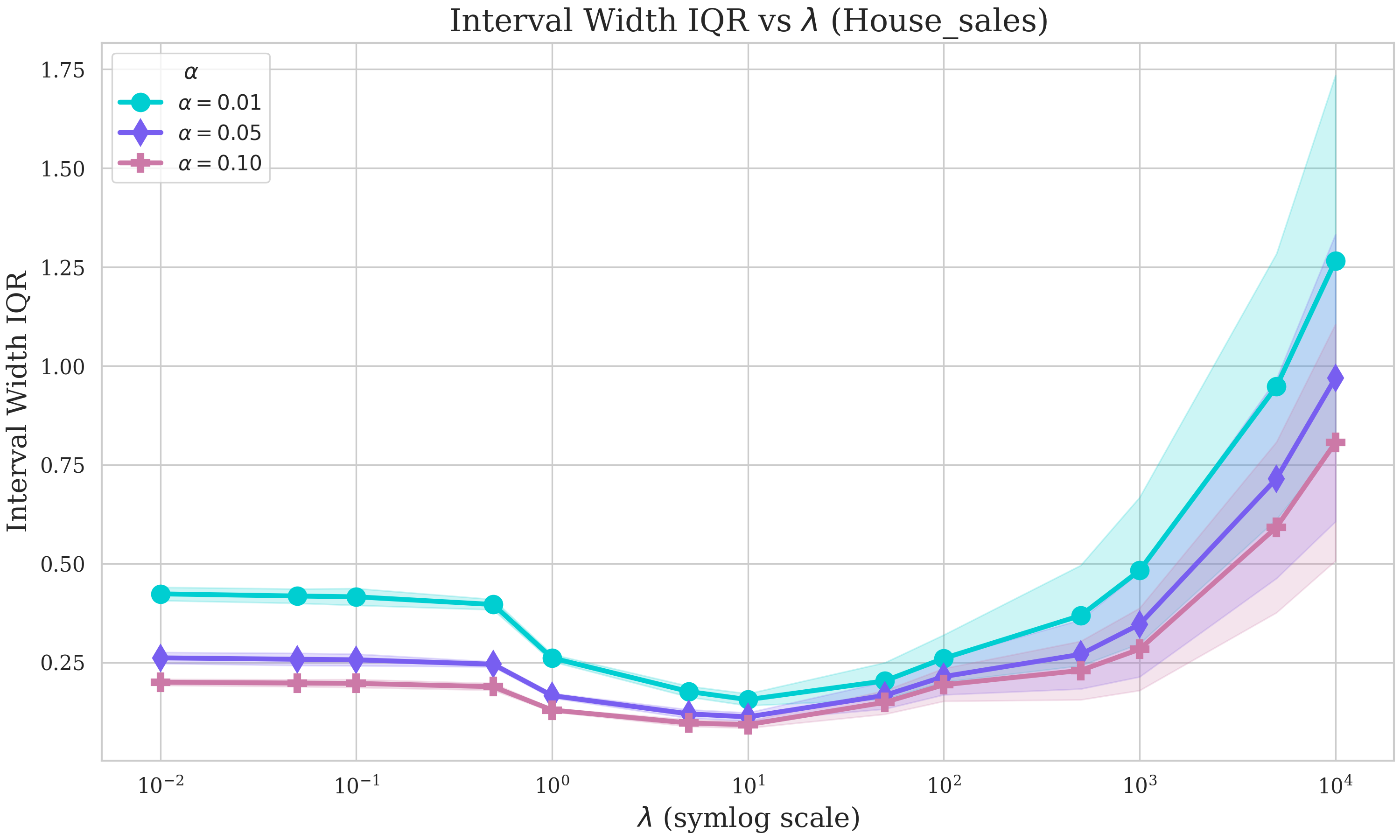}
    \caption{Interval width IQR.}
    \label{fig:iqr_lambda_House_sales}
  \end{subfigure}
    \caption{Sensitivity analysis of the regularization parameter $\lambda$ on the House Sales dataset.}
  \label{fig:lambda_figures_House_sales}
\end{figure*}

\begin{figure*}[!ht]
  \centering
  \begin{subfigure}[t]{0.32\textwidth}
    \includegraphics[width=\textwidth]{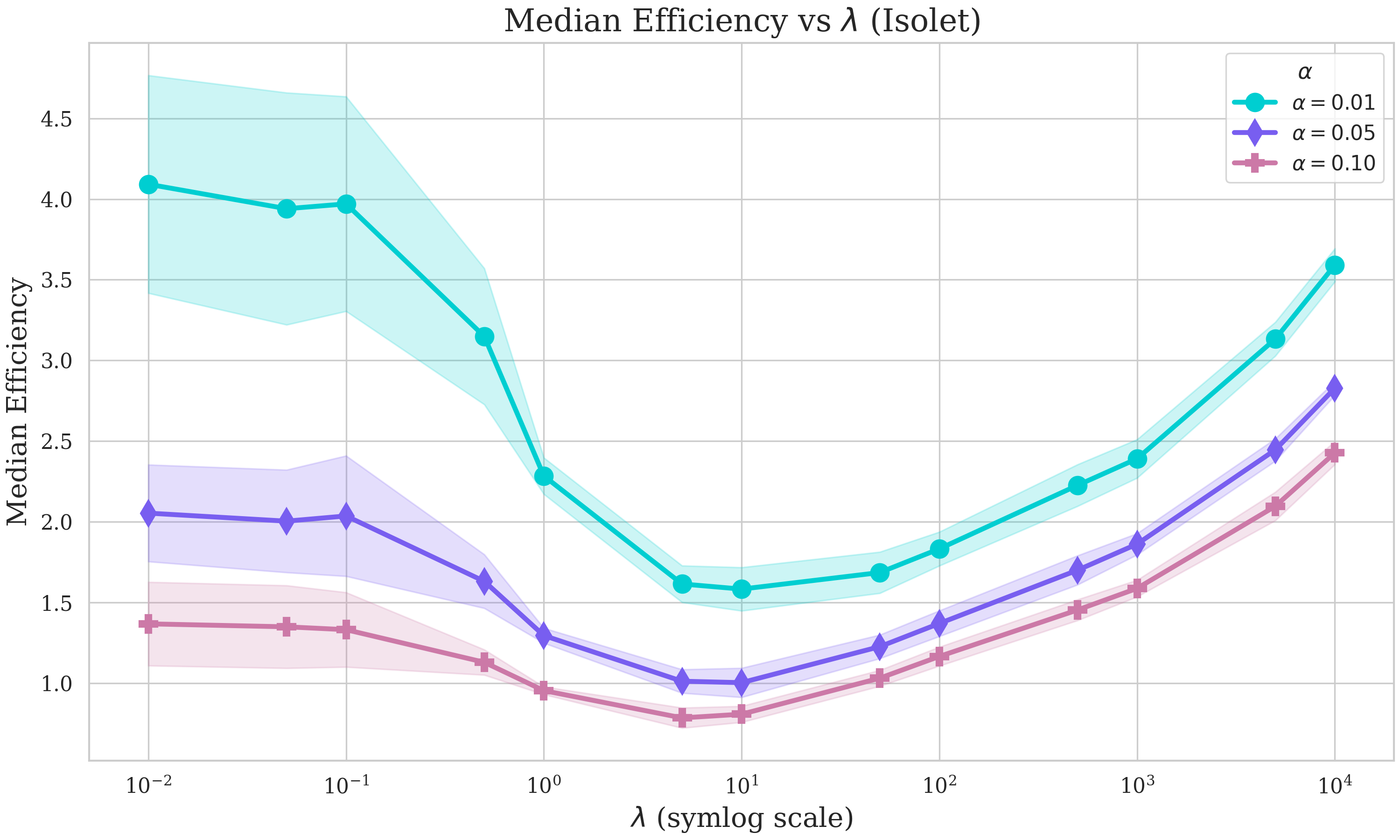}
    \caption{Median efficiency.}
    \label{fig:efficiency_lambda_Isolet}
  \end{subfigure}
  \hfill
  \begin{subfigure}[t]{0.32\textwidth}
    \includegraphics[width=\textwidth]{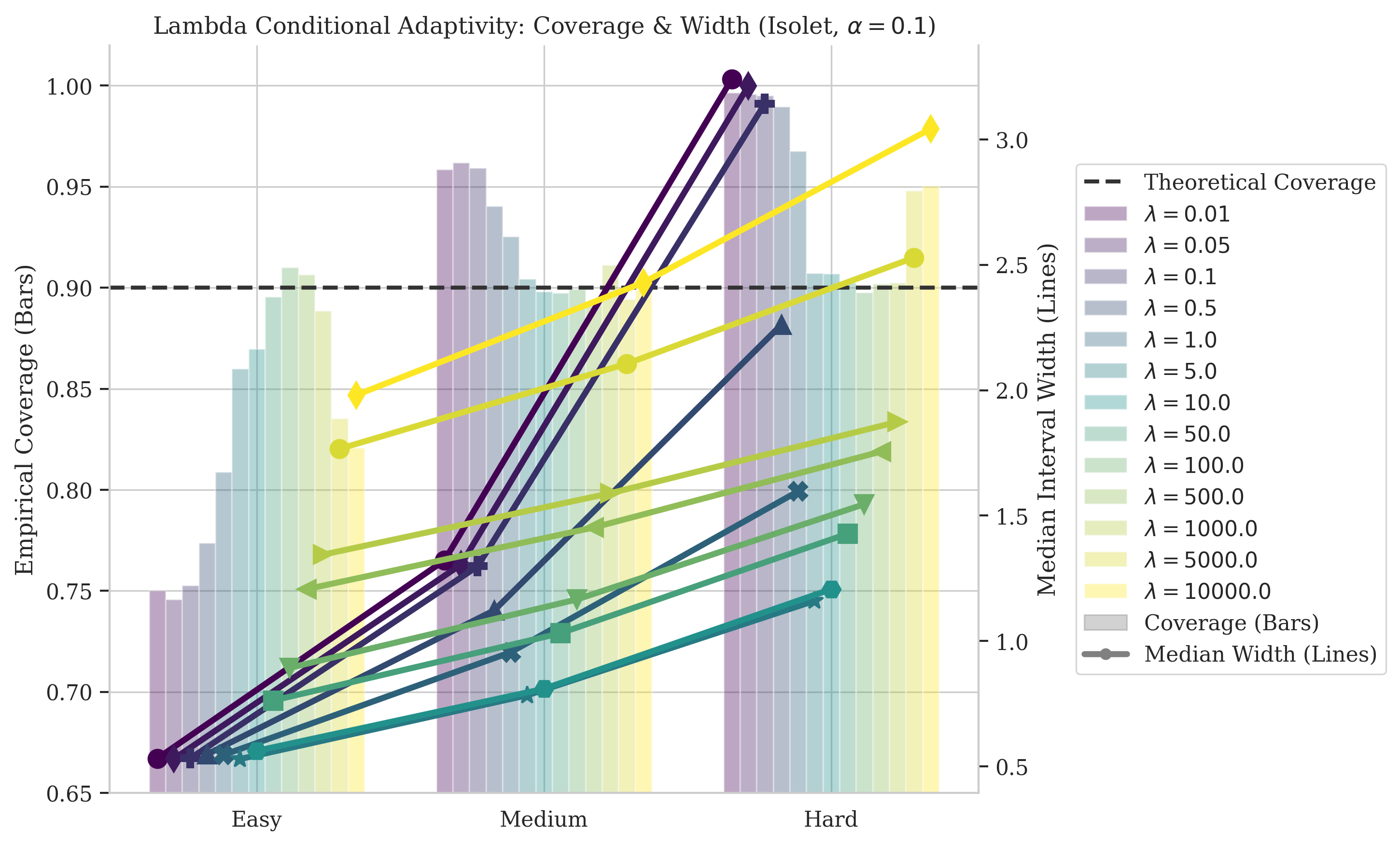}
    \caption{Conditional coverage (bars) and efficiency (lines) by internal difficulty.}
    \label{fig:box_plot_lambda_Isolet}
  \end{subfigure}  
    \hfill
  \begin{subfigure}[t]{0.32\textwidth}
    \includegraphics[width=\textwidth]{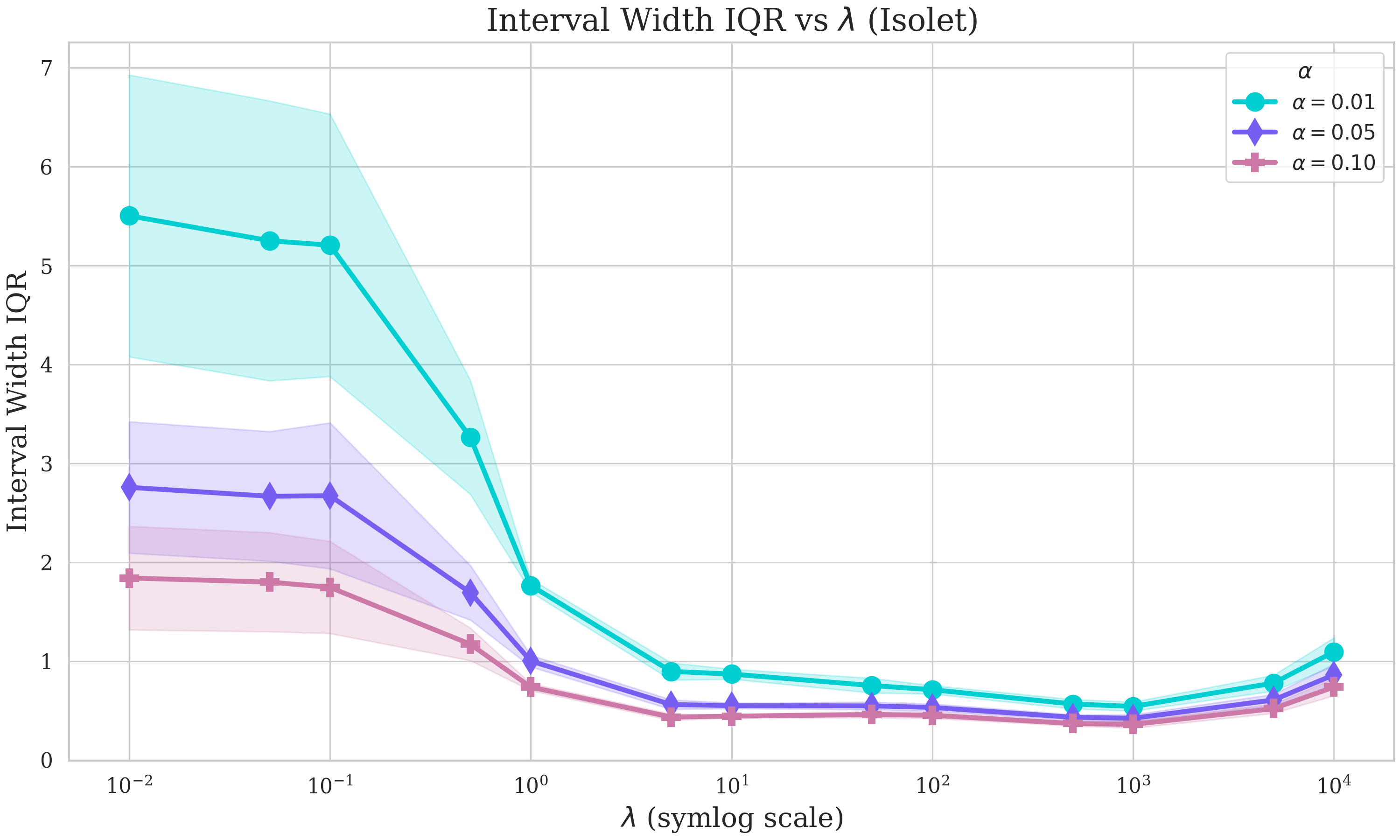}
    \caption{Interval width IQR.}
    \label{fig:iqr_lambda_Isolet}
  \end{subfigure}
    \caption{Sensitivity analysis of the regularization parameter $\lambda$ on the Isolet dataset.}
  \label{fig:lambda_figures_Isolet}
\end{figure*}

\begin{figure*}[!ht]
  \centering
  \begin{subfigure}[t]{0.32\textwidth}
    \includegraphics[width=\textwidth]{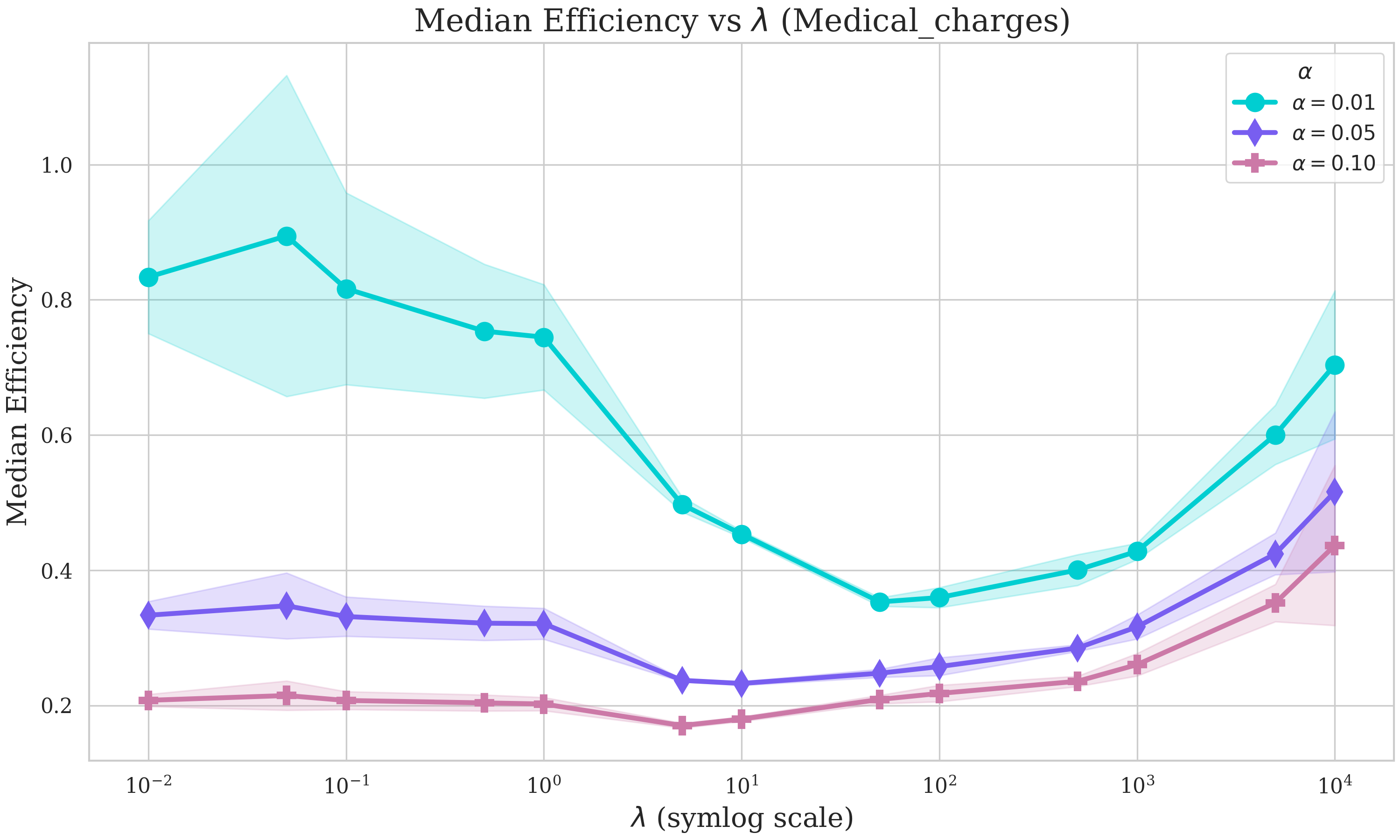}
    \caption{Median efficiency.}
    \label{fig:efficiency_lambda_Medical_charges}
  \end{subfigure}
  \hfill
  \begin{subfigure}[t]{0.32\textwidth}
    \includegraphics[width=\textwidth]{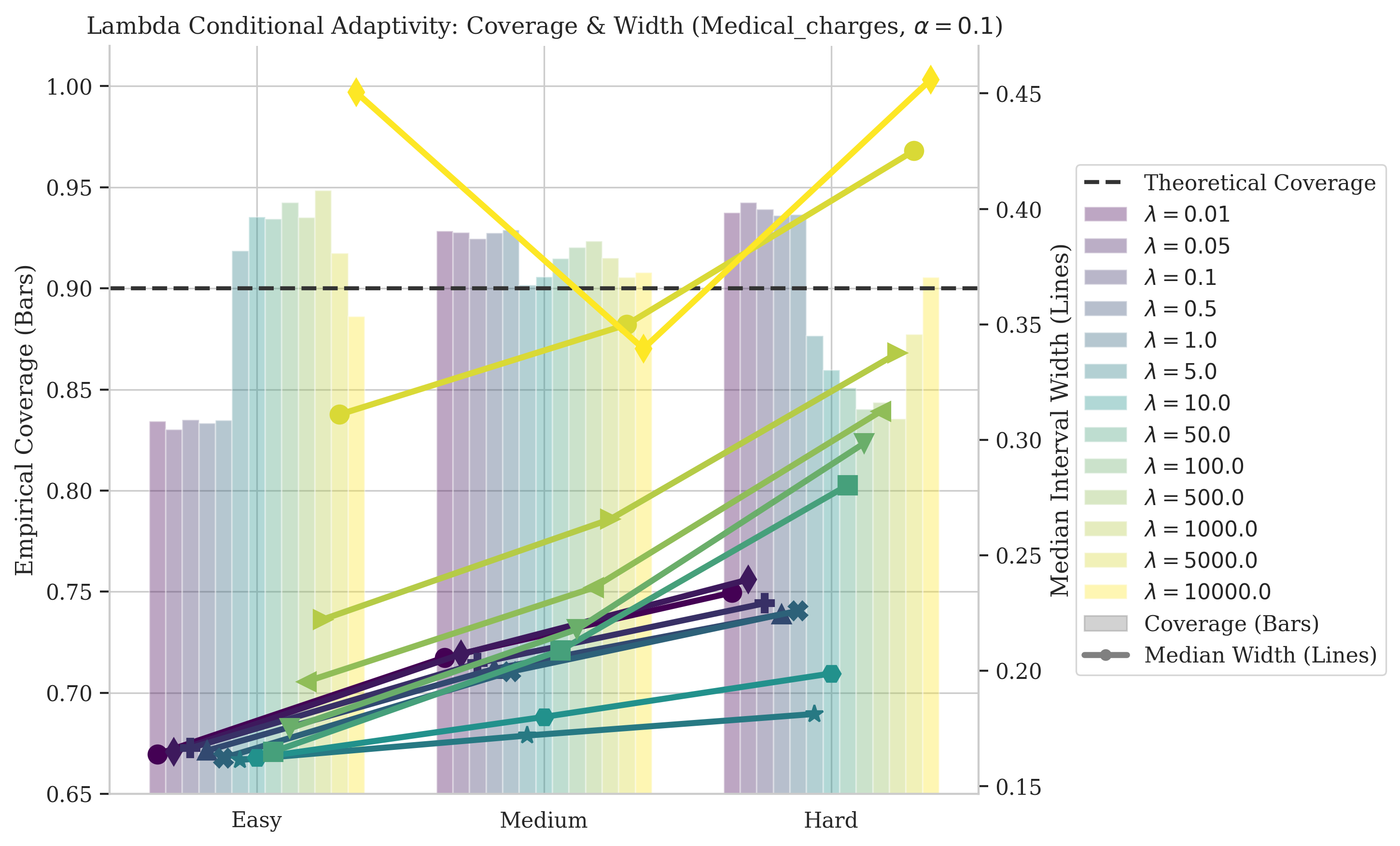}
    \caption{Conditional coverage (bars) and efficiency (lines) by internal difficulty.}
    \label{fig:box_plot_lambda_Medical_charges}
  \end{subfigure}
  \hfill
  \begin{subfigure}[t]{0.32\textwidth}
    \includegraphics[width=\textwidth]{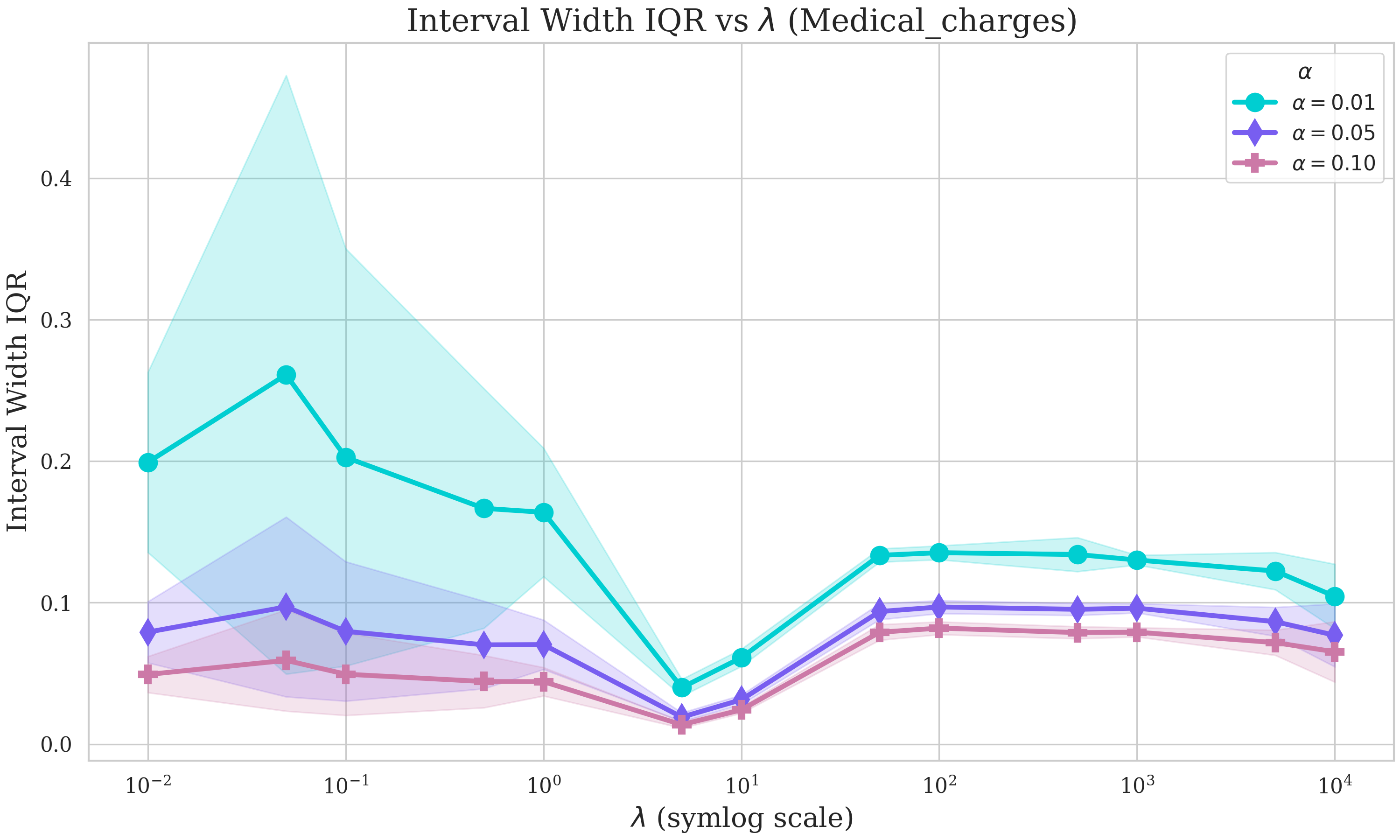}
    \caption{Interval width IQR.}
    \label{fig:iqr_lambda_Medical_charges}
  \end{subfigure}
    \caption{Sensitivity analysis of the regularization parameter $\lambda$ on the Medical Charges dataset.}
  \label{fig:lambda_figures_Medical_charges}
\end{figure*}

\begin{figure*}[!ht]
  \centering
  \begin{subfigure}[t]{0.32\textwidth}
    \includegraphics[width=\textwidth]{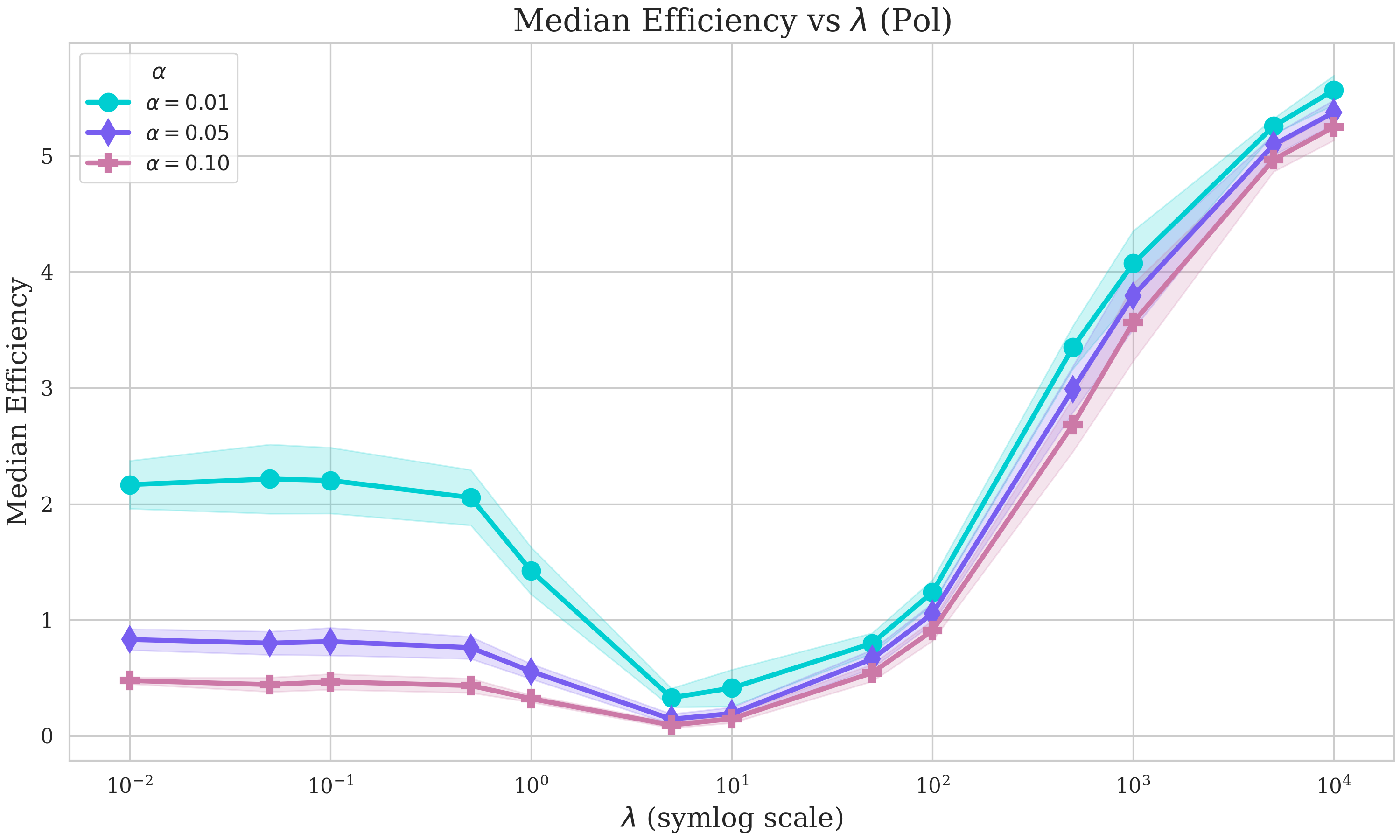}
    \caption{Median efficiency.}
    \label{fig:efficiency_lambda_Pol}
  \end{subfigure}
  \hfill
  \begin{subfigure}[t]{0.32\textwidth}
    \includegraphics[width=\textwidth]{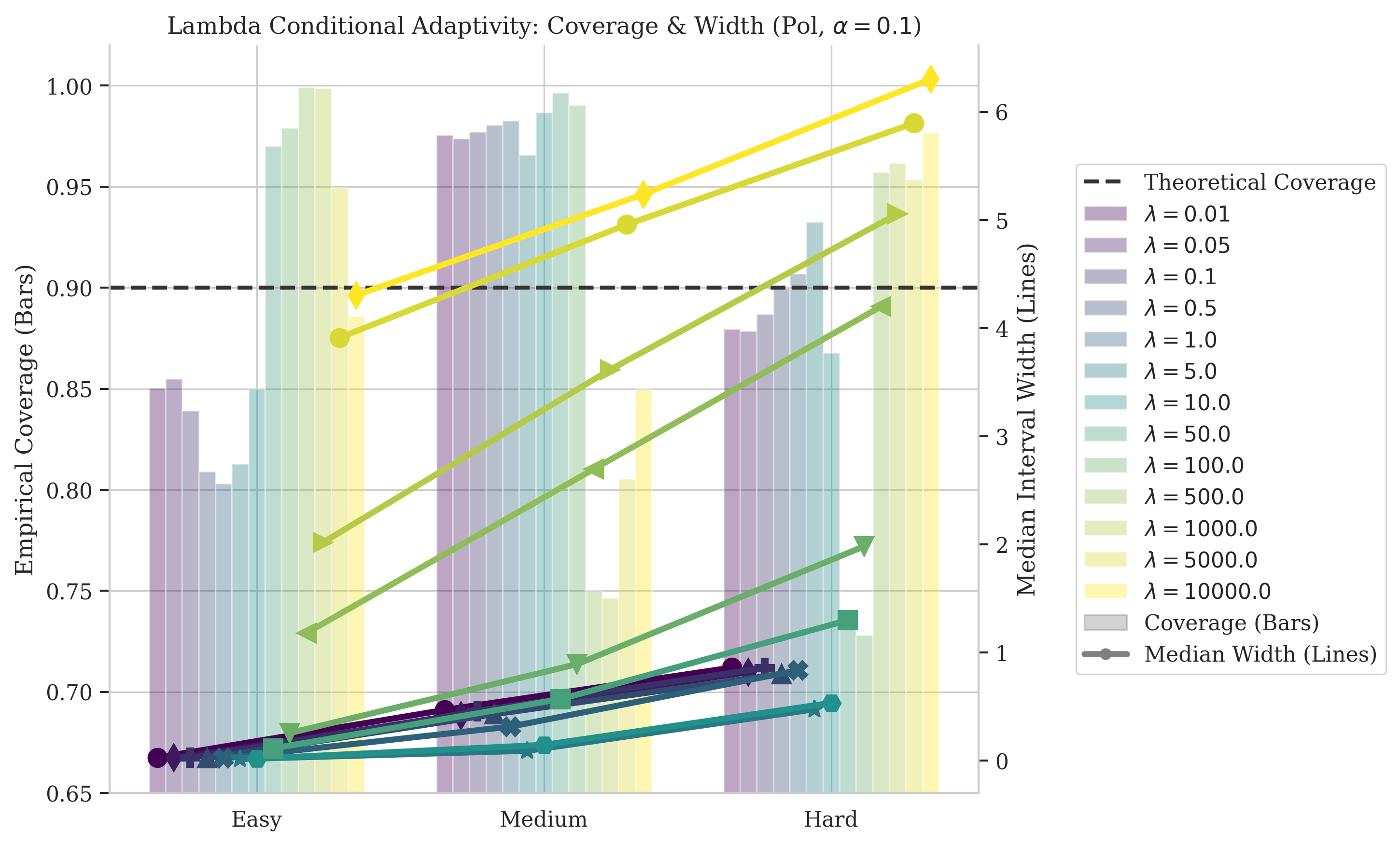}
    \caption{Conditional coverage (bars) and efficiency (lines) by internal difficulty.}
    \label{fig:box_plot_lambda_Pol}
  \end{subfigure}  
    \hfill
  \begin{subfigure}[t]{0.32\textwidth}
    \includegraphics[width=\textwidth]{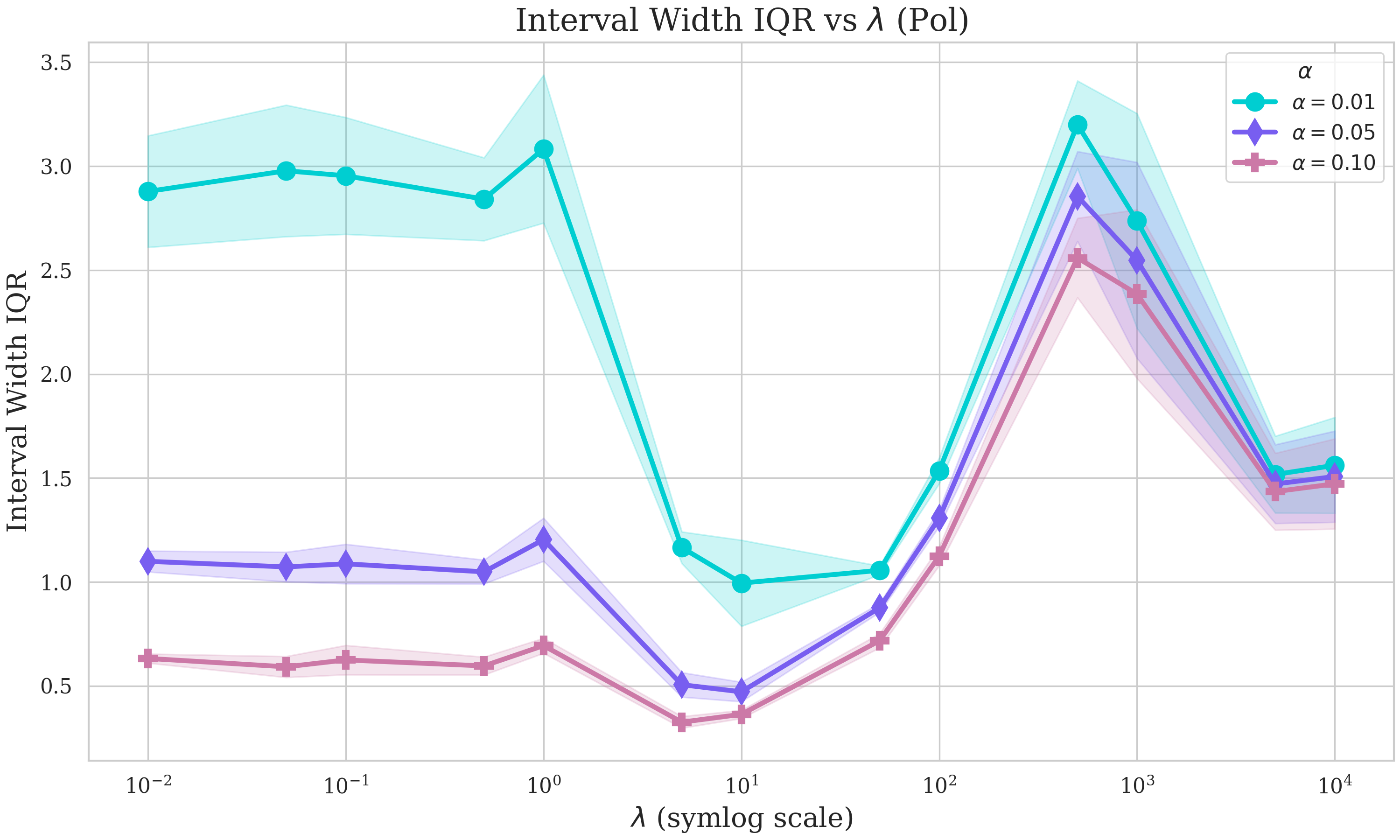}
    \caption{Interval width IQR.}
    \label{fig:iqr_lambda_Pol}
  \end{subfigure}
    \caption{Sensitivity analysis of the regularization parameter $\lambda$ on the Pol dataset.}
  \label{fig:lambda_figures_Pol}
\end{figure*}

\begin{figure*}[!ht]
  \centering
  \begin{subfigure}[t]{0.32\textwidth}
    \includegraphics[width=\textwidth]{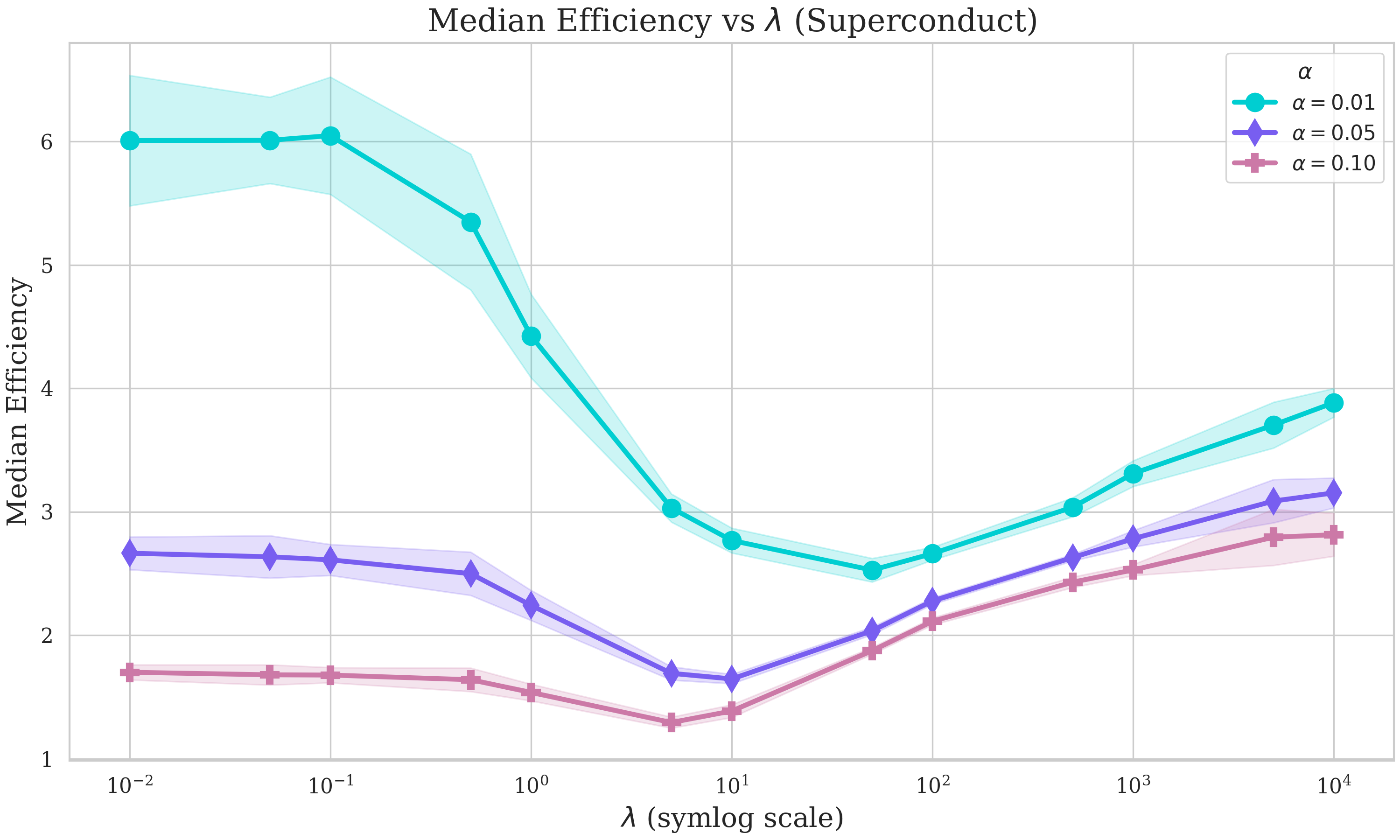}
    \caption{Median efficiency.}
    \label{fig:efficiency_lambda_Superconduct}
  \end{subfigure}
  \hfill
  \begin{subfigure}[t]{0.32\textwidth}
    \includegraphics[width=\textwidth]{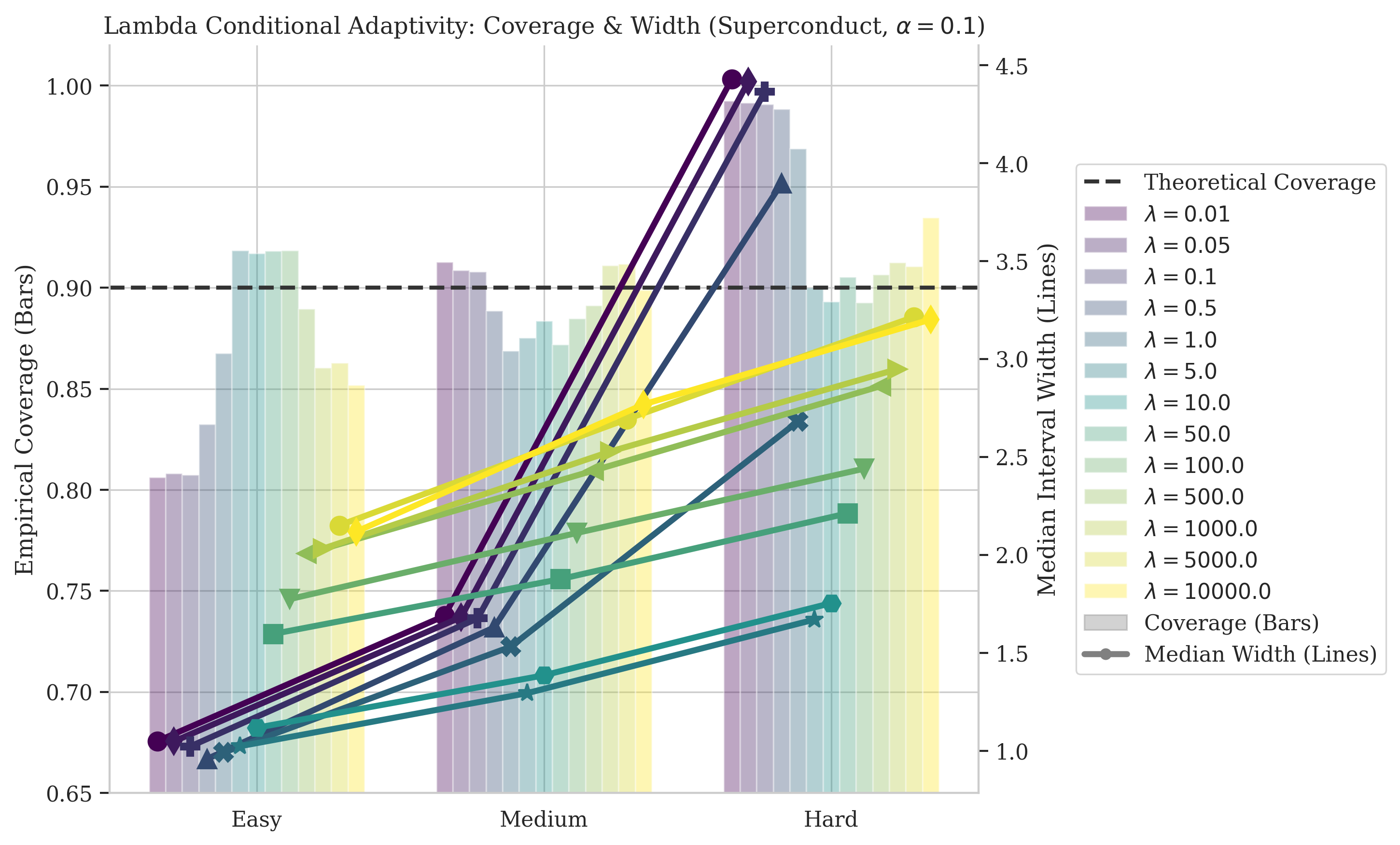}
    \caption{Conditional coverage (bars) and efficiency (lines) by internal difficulty.}
    \label{fig:box_plot_lambda_Superconduct}
  \end{subfigure}  
    \hfill
  \begin{subfigure}[t]{0.32\textwidth}
    \includegraphics[width=\textwidth]{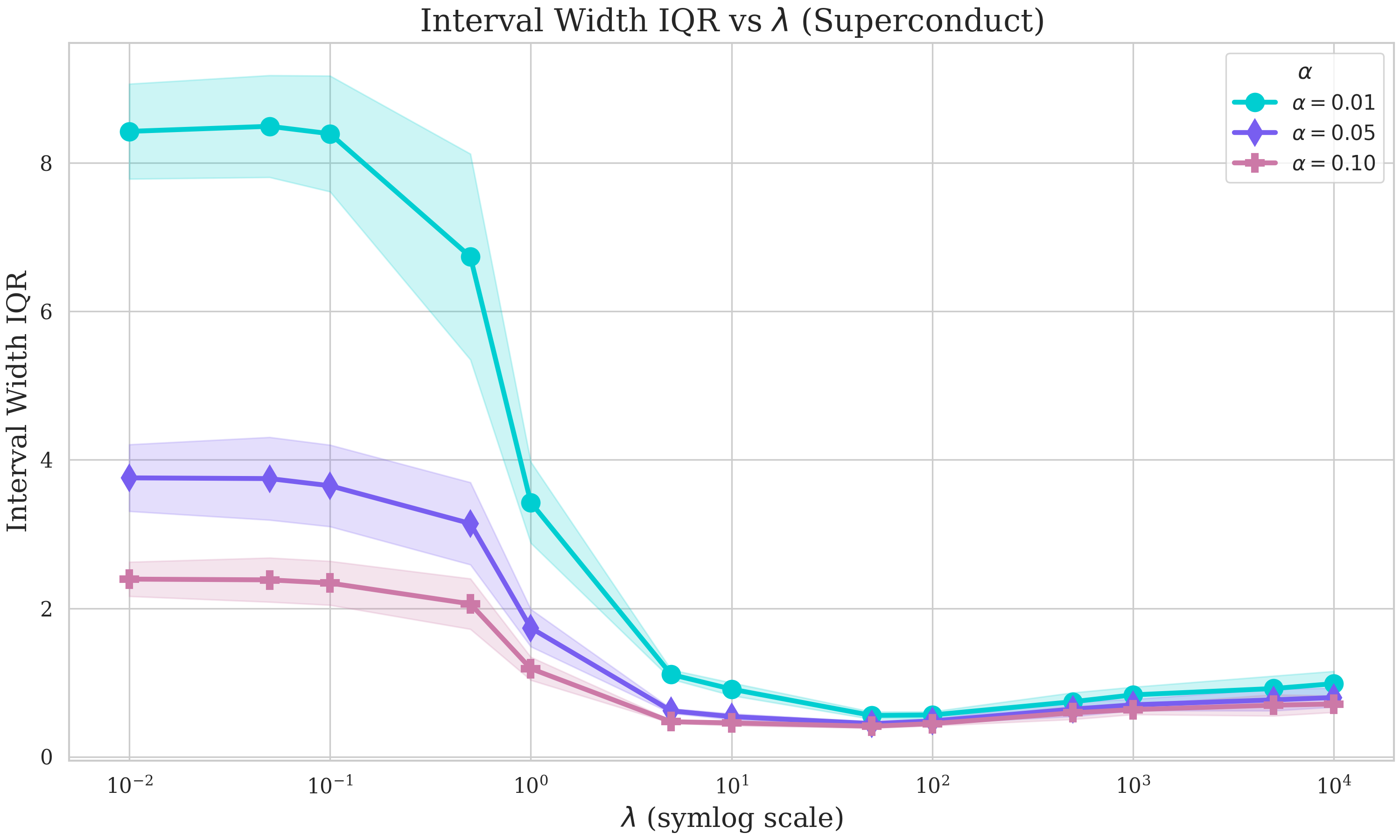}
    \caption{Interval width IQR.}
    \label{fig:iqr_lambda_Superconduct}
  \end{subfigure}
    \caption{Sensitivity analysis of the regularization parameter $\lambda$ on the Superconduct dataset.}
  \label{fig:lambda_figures_Superconduct}
\end{figure*}

\begin{figure*}[!ht]
  \centering
  \begin{subfigure}[t]{0.32\textwidth}
    \includegraphics[width=\textwidth]{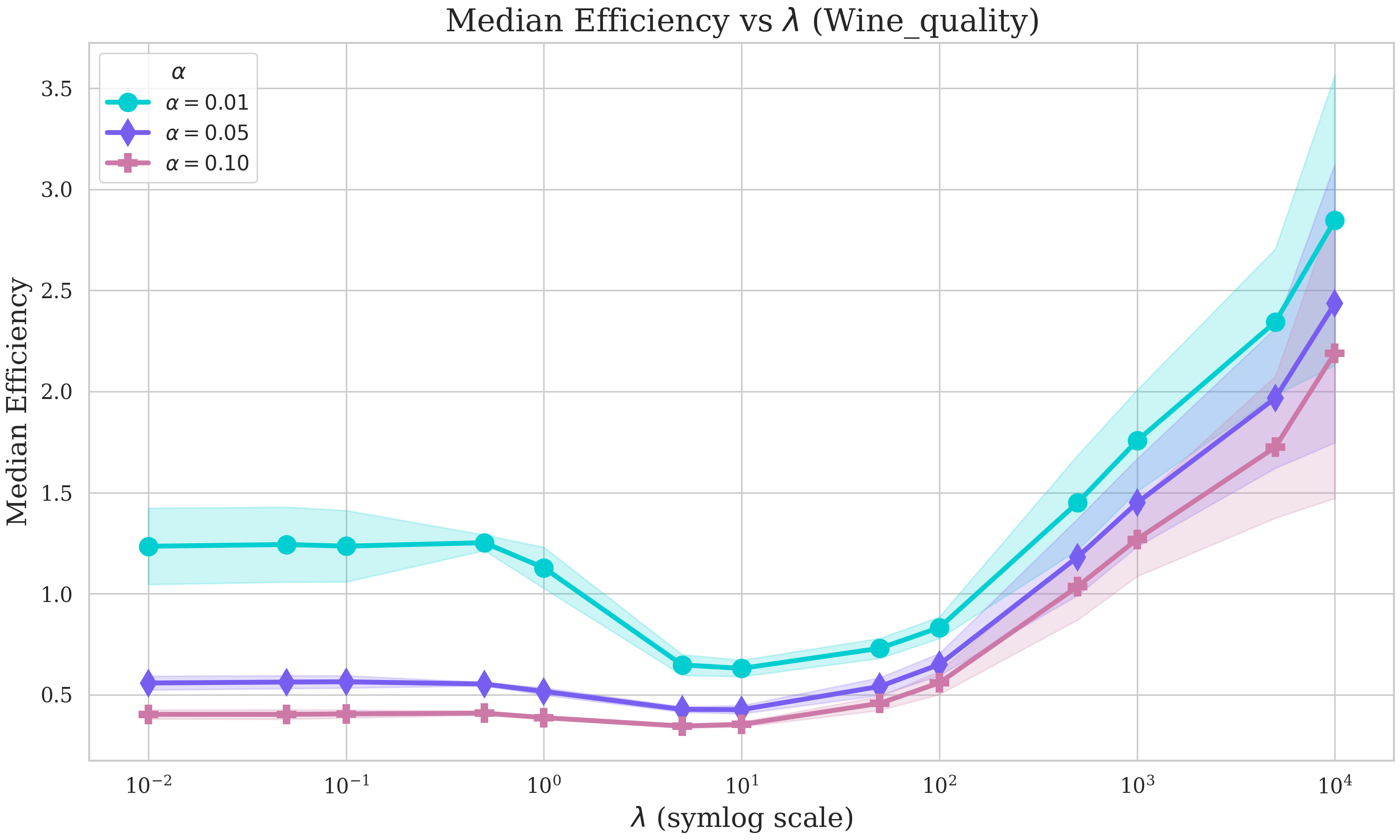}
    \caption{Median efficiency.}
    \label{fig:efficiency_lambda_Wine_quality}
  \end{subfigure}
  \hfill
  \begin{subfigure}[t]{0.32\textwidth}
    \includegraphics[width=\textwidth]{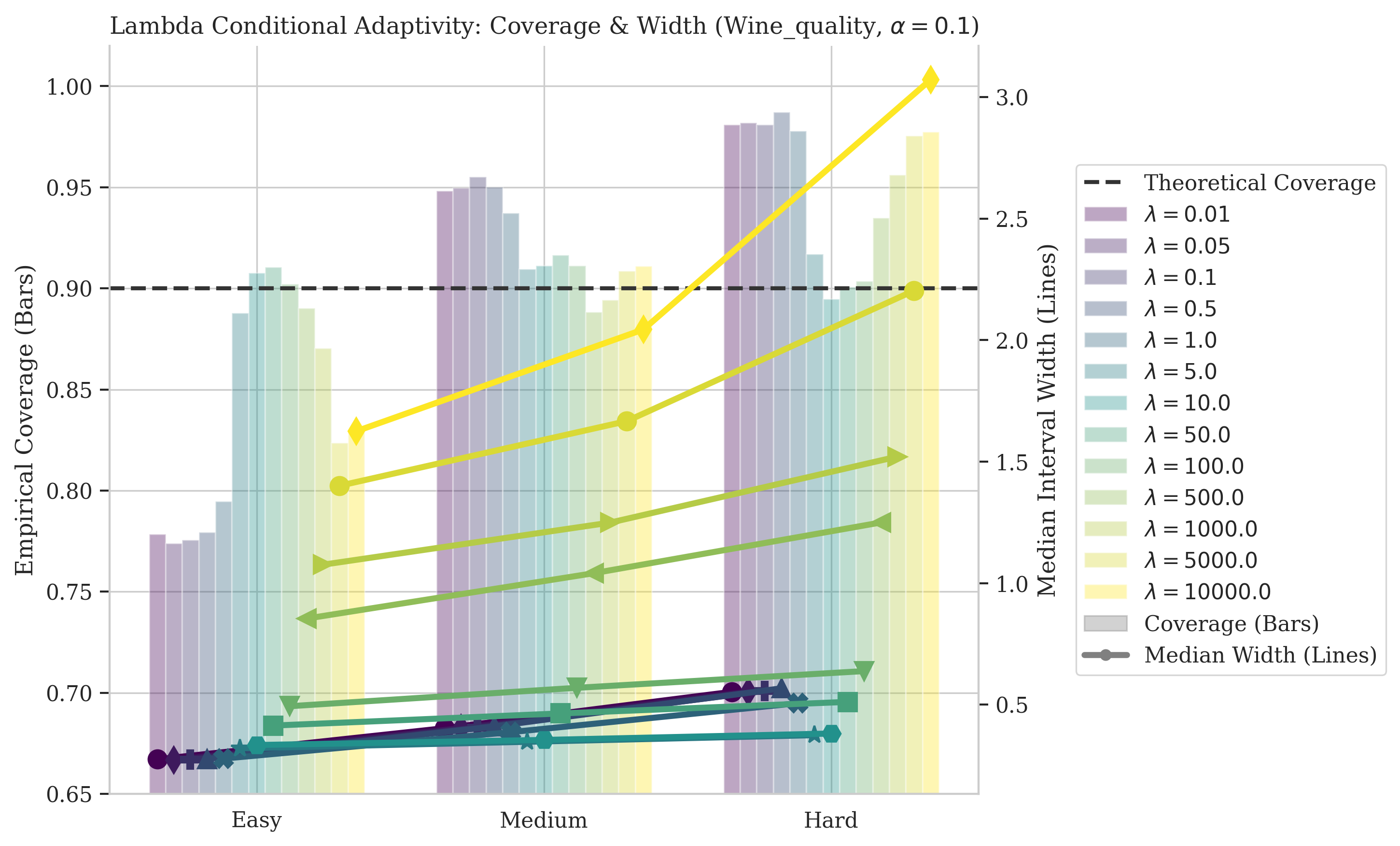}
    \caption{Conditional coverage (bars) and efficiency (lines) by internal difficulty.}
    \label{fig:box_plot_lambda_Wine_quality}
  \end{subfigure}  
    \hfill
  \begin{subfigure}[t]{0.32\textwidth}
    \includegraphics[width=\textwidth]{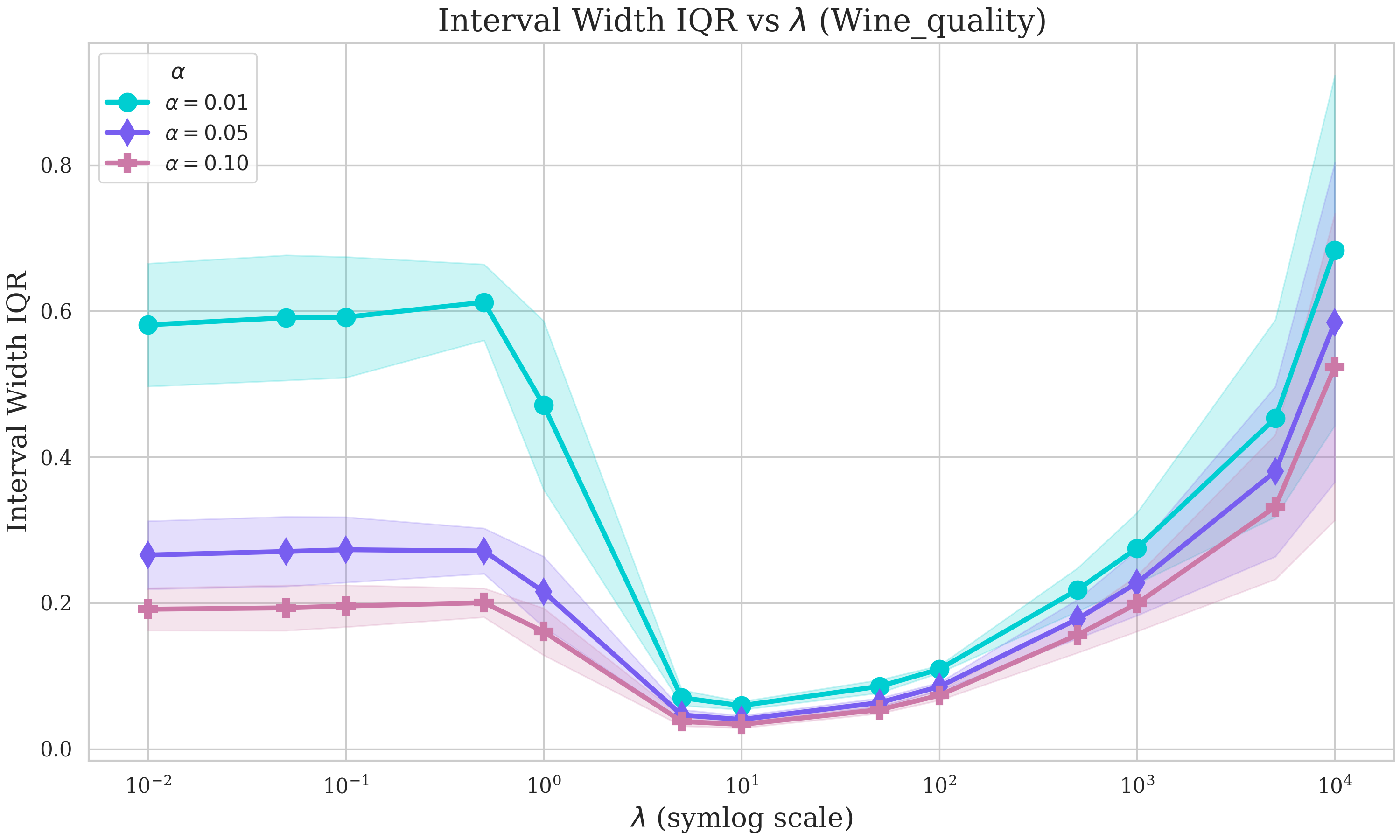}
    \caption{Interval width IQR.}
    \label{fig:iqr_lambda_Wine_quality}
  \end{subfigure}
    \caption{Sensitivity analysis of the regularization parameter $\lambda$ on the Wine Quality dataset.}
  \label{fig:lambda_figures_Wine_quality}
\end{figure*}

\begin{figure*}[!ht]
  \centering
  \begin{subfigure}[t]{0.32\textwidth}
    \includegraphics[width=\textwidth]{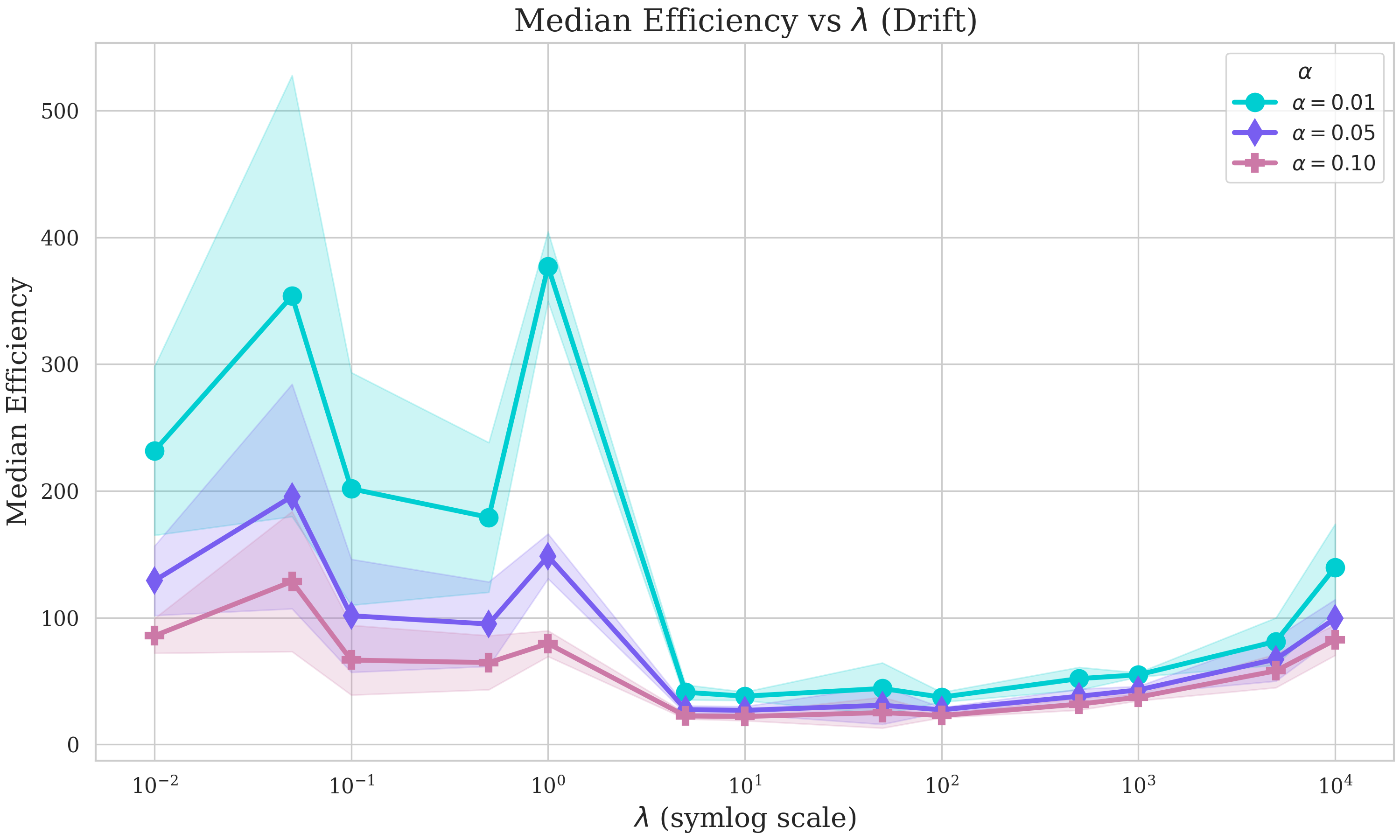}
    \caption{Median efficiency.}
    \label{fig:efficiency_lambda_Drift}
  \end{subfigure}
  \hfill
  \begin{subfigure}[t]{0.32\textwidth}
    \includegraphics[width=\textwidth]{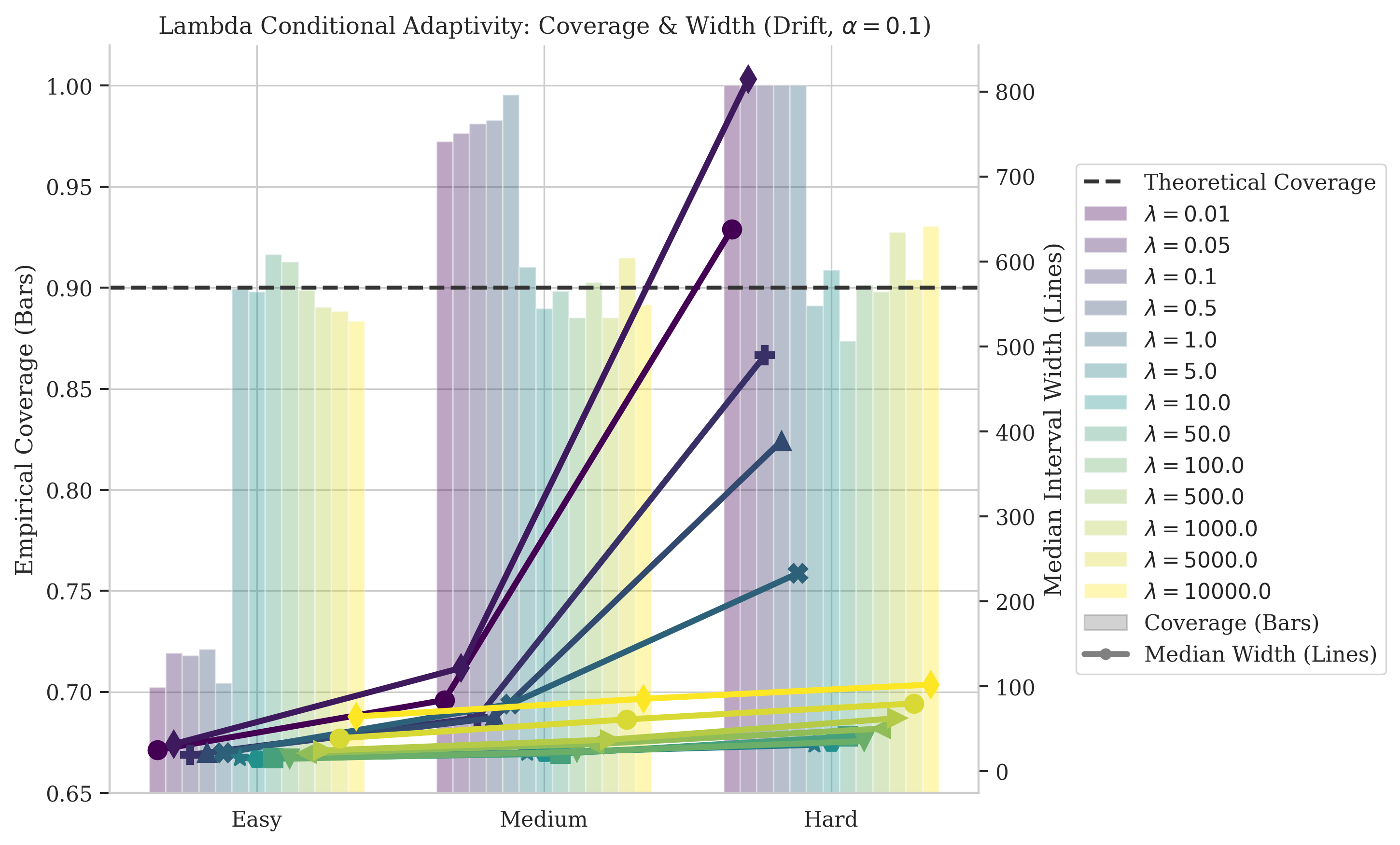}
    \caption{Conditional coverage (bars) and efficiency (lines) by internal difficulty.}
    \label{fig:box_plot_lambda_Drift}
  \end{subfigure}
    \hfill
  \begin{subfigure}[t]{0.32\textwidth}
    \includegraphics[width=\textwidth]{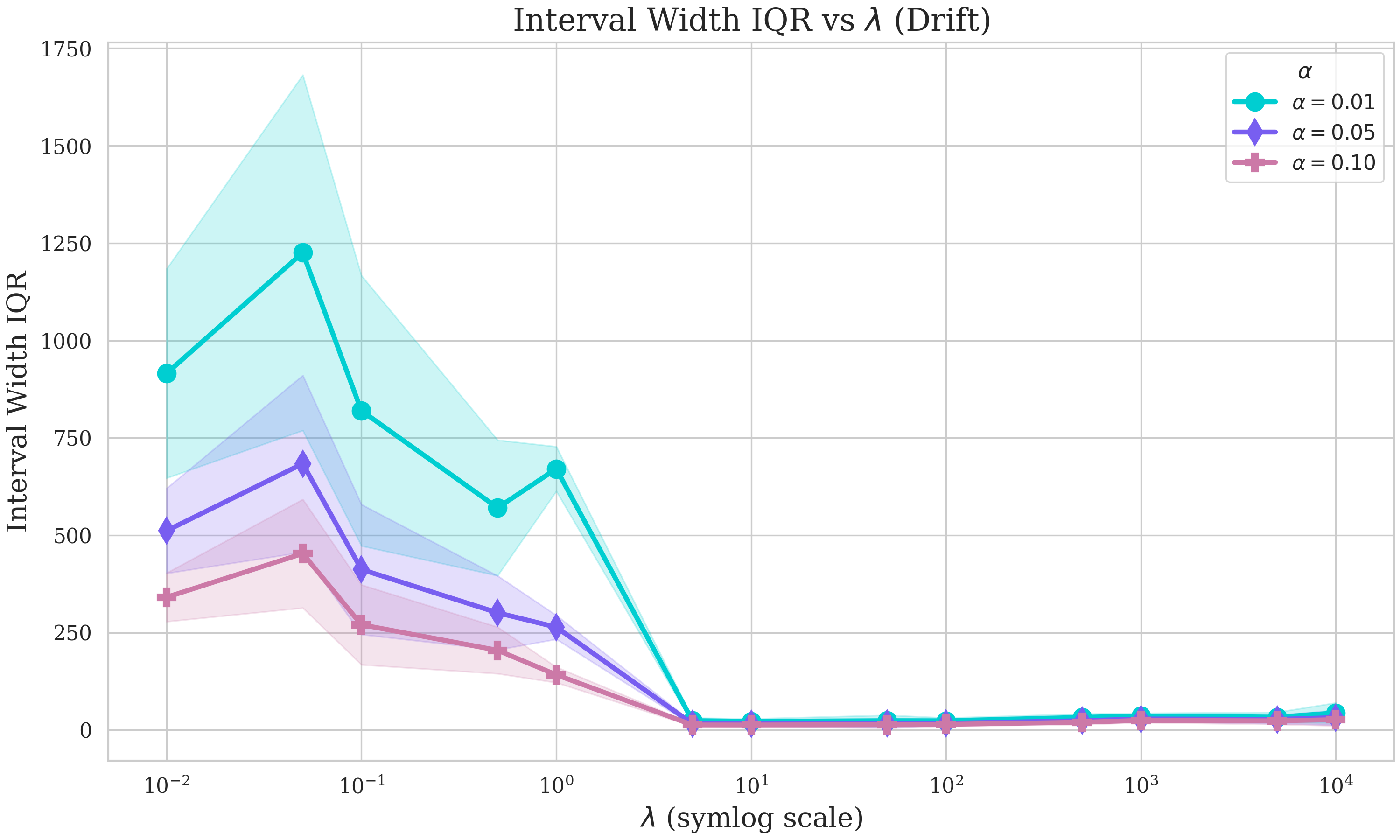}
    \caption{Interval width IQR.}
    \label{fig:iqr_lambda_Drift}
  \end{subfigure}
    \caption{Sensitivity analysis of the regularization parameter $\lambda$ on the Drift dataset (image regression).}
  \label{fig:lambda_figures_Drift}
\end{figure*}

Tables~\ref{tab:lambda_results_cov_mae_all},~\ref{tab:lambda_results_eff_all_1} and~\ref{tab:lambda_results_eff_all_2} and Figures~\ref{fig:lambda_figures_Bike_sharing}, \ref{fig:lambda_figures_California}, \ref{fig:lambda_figures_Diamonds}, \ref{fig:lambda_figures_Fifa}, \ref{fig:lambda_figures_House_sales}, \ref{fig:lambda_figures_Isolet}, \ref{fig:lambda_figures_Medical_charges}, \ref{fig:lambda_figures_Pol}, \ref{fig:lambda_figures_Superconduct}, \ref{fig:lambda_figures_Wine_quality} and \ref{fig:lambda_figures_Drift} illustrate how SPACR balances coverage and efficiency across different values of $\lambda$ on the remaining datasets.

\begin{table*}[ht]
\caption{Impact of $\lambda$ on MAE and Marginal Coverage across different confidence levels ($1-\alpha$). Best MAE results per dataset are highlighted in bold. The row corresponding to $\lambda = 5$ is shaded in gray.}
\centering\scriptsize\resizebox{\textwidth}{!}{
\begin{tabular}{lccccccc@{\hspace{2em}}lcccccc}
\toprule
\textbf{Dataset} & \textbf{Lambda} & \textbf{MAE $\downarrow$} & \multicolumn{3}{c}{\textbf{Coverage (\%)}} & \textbf{Dataset} & \textbf{Lambda} & \textbf{MAE $\downarrow$} & \multicolumn{3}{c}{\textbf{Coverage (\%)}} \\
\cmidrule(lr){4-6} \cmidrule(lr){10-12}
& \textbf{($\lambda$)} & & $\alpha = 0.10$ & $\alpha = 0.05$ & $\alpha = 0.01$ & & \textbf{($\lambda$)} & & $\alpha = 0.10$ & $\alpha = 0.05$ & $\alpha = 0.01$ \\ \midrule
\multirow{13}{*}{\begin{tabular}{@{}l@{}}\textbf{Bike} \\ \textbf{Sharing}\end{tabular}}
& 0.01 & 0.31 $\pm$ 0.01 & 90.10 $\pm$ 0.25 & 95.51 $\pm$ 0.61 & 99.04 $\pm$ 0.16 & \multirow{13}{*}{\begin{tabular}{@{}l@{}}\textbf{Brazilian} \\ \textbf{Houses}\end{tabular}} & 0.01 & 0.13 $\pm$ 0.02 & 90.12 $\pm$ 1.25 & 95.12 $\pm$ 0.54 & 99.10 $\pm$ 0.23 \\
& 0.05 & 0.31 $\pm$ 0.01 & 90.25 $\pm$ 0.18 & 95.42 $\pm$ 0.41 & 99.02 $\pm$ 0.18 & & 0.05 & \textbf{0.12 $\pm$ 0.02} & 90.01 $\pm$ 0.84 & 95.21 $\pm$ 0.54 & 99.16 $\pm$ 0.22 \\
& 0.1 & 0.31 $\pm$ 0.01 & 90.32 $\pm$ 0.21 & 95.48 $\pm$ 0.50 & 99.06 $\pm$ 0.18 & & 0.1 & 0.13 $\pm$ 0.02 & 90.11 $\pm$ 0.83 & 95.22 $\pm$ 0.41 & 99.15 $\pm$ 0.23 \\
& 0.5 & 0.31 $\pm$ 0.01 & 89.99 $\pm$ 0.16 & 95.54 $\pm$ 0.42 & 99.15 $\pm$ 0.22 & & 0.5 & 0.14 $\pm$ 0.02 & 90.20 $\pm$ 0.83 & 95.07 $\pm$ 0.42 & 99.14 $\pm$ 0.32 \\
& 1 & \textbf{0.30 $\pm$ 0.01} & 89.95 $\pm$ 0.71 & 95.11 $\pm$ 0.44 & 99.22 $\pm$ 0.13 & & 1 & 0.14 $\pm$ 0.02 & 90.12 $\pm$ 0.16 & 94.86 $\pm$ 0.37 & 99.06 $\pm$ 0.32 \\
& \cellcolor{gray!20} 5 & \cellcolor{gray!20} 0.32 $\pm$ 0.02 & \cellcolor{gray!20} 90.06 $\pm$ 0.52 & \cellcolor{gray!20} 95.09 $\pm$ 0.41 & \cellcolor{gray!20} 99.02 $\pm$ 0.22 & & \cellcolor{gray!20} 5 & \cellcolor{gray!20} 0.34 $\pm$ 0.03 & \cellcolor{gray!20} 89.86 $\pm$ 0.69 & \cellcolor{gray!20} 95.00 $\pm$ 0.88 & \cellcolor{gray!20} 99.21 $\pm$ 0.22 \\
& 10 & 0.36 $\pm$ 0.02 & 90.12 $\pm$ 0.80 & 95.07 $\pm$ 0.52 & 98.99 $\pm$ 0.21 & & 10 & 0.41 $\pm$ 0.03 & 90.14 $\pm$ 0.52 & 95.29 $\pm$ 0.74 & 99.22 $\pm$ 0.18 \\
& 50 & 0.51 $\pm$ 0.02 & 90.29 $\pm$ 0.49 & 95.06 $\pm$ 0.59 & 99.02 $\pm$ 0.25 & & 50 & 0.43 $\pm$ 0.04 & 90.13 $\pm$ 0.89 & 94.98 $\pm$ 0.50 & 99.01 $\pm$ 0.44 \\
& 100 & 0.65 $\pm$ 0.03 & 90.25 $\pm$ 0.72 & 94.90 $\pm$ 0.25 & 98.98 $\pm$ 0.42 & & 100 & 0.48 $\pm$ 0.10 & 90.07 $\pm$ 0.56 & 95.18 $\pm$ 0.39 & 98.96 $\pm$ 0.40 \\
& 500 & 0.90 $\pm$ 0.04 & 89.99 $\pm$ 0.40 & 94.97 $\pm$ 0.50 & 99.00 $\pm$ 0.24 & & 500 & 1.02 $\pm$ 0.81 & 90.33 $\pm$ 0.99 & 95.06 $\pm$ 0.90 & 99.13 $\pm$ 0.34 \\
& 1000 & 0.91 $\pm$ 0.03 & 89.94 $\pm$ 0.53 & 94.97 $\pm$ 0.34 & 99.03 $\pm$ 0.37 & & 1000 & 1.41 $\pm$ 1.15 & 90.43 $\pm$ 0.80 & 95.27 $\pm$ 0.54 & 99.12 $\pm$ 0.31 \\
& 5000 & 0.92 $\pm$ 0.04 & 89.59 $\pm$ 0.77 & 95.05 $\pm$ 0.51 & 99.11 $\pm$ 0.26 & & 5000 & 2.92 $\pm$ 1.74 & 90.48 $\pm$ 0.81 & 95.18 $\pm$ 0.67 & 99.20 $\pm$ 0.20 \\
& 10000 & 0.98 $\pm$ 0.07 & 89.76 $\pm$ 0.46 & 94.80 $\pm$ 0.51 & 99.10 $\pm$ 0.16 & & 10000 & 3.37 $\pm$ 1.54 & 90.49 $\pm$ 0.97 & 95.13 $\pm$ 0.75 & 99.15 $\pm$ 0.04 \\
\midrule
\multirow{13}{*}{\textbf{California}}
& 0.01 & 0.07 $\pm$ 0.00 & 90.46 $\pm$ 1.09 & 95.38 $\pm$ 0.75 & 99.09 $\pm$ 0.18 & \multirow{13}{*}{\begin{tabular}{@{}l@{}}\textbf{CPU} \\ \textbf{Act}\end{tabular}} & 0.01 & \textbf{0.04 $\pm$ 0.01} & 89.59 $\pm$ 1.36 & 94.74 $\pm$ 1.16 & 98.97 $\pm$ 0.30 \\
& 0.05 & 0.07 $\pm$ 0.00 & 90.60 $\pm$ 0.92 & 95.30 $\pm$ 0.77 & 99.09 $\pm$ 0.20 & & 0.05 & \textbf{0.04 $\pm$ 0.01} & 89.49 $\pm$ 1.07 & 94.80 $\pm$ 1.30 & 99.02 $\pm$ 0.26 \\
& 0.1 & 0.07 $\pm$ 0.00 & 90.63 $\pm$ 0.84 & 95.29 $\pm$ 0.75 & 99.10 $\pm$ 0.15 & & 0.1 & \textbf{0.04 $\pm$ 0.01} & 89.46 $\pm$ 1.23 & 94.70 $\pm$ 1.26 & 99.02 $\pm$ 0.22 \\
& 0.5 & \textbf{0.06 $\pm$ 0.00} & 90.53 $\pm$ 0.89 & 95.43 $\pm$ 0.69 & 99.13 $\pm$ 0.05 & & 0.5 & \textbf{0.04 $\pm$ 0.01} & 89.05 $\pm$ 0.78 & 94.77 $\pm$ 0.84 & 98.94 $\pm$ 0.11 \\
& 1 & \textbf{0.06 $\pm$ 0.00} & 90.40 $\pm$ 0.96 & 95.20 $\pm$ 0.48 & 99.07 $\pm$ 0.05 & & 1 & \textbf{0.04 $\pm$ 0.01} & 88.82 $\pm$ 1.03 & 94.94 $\pm$ 0.85 & 99.05 $\pm$ 0.16 \\
& \cellcolor{gray!20} 5 & \cellcolor{gray!20} \textbf{0.06 $\pm$ 0.00} & \cellcolor{gray!20} 90.06 $\pm$ 0.94 & \cellcolor{gray!20} 95.04 $\pm$ 0.57 & \cellcolor{gray!20} 99.02 $\pm$ 0.15 & & \cellcolor{gray!20} 5 & \cellcolor{gray!20} 0.08 $\pm$ 0.02 & \cellcolor{gray!20} 89.96 $\pm$ 0.52 & \cellcolor{gray!20} 94.78 $\pm$ 0.71 & \cellcolor{gray!20} 99.13 $\pm$ 0.15 \\
& 10 & \textbf{0.06 $\pm$ 0.00} & 89.80 $\pm$ 1.13 & 95.00 $\pm$ 0.66 & 98.95 $\pm$ 0.15 & & 10 & 0.11 $\pm$ 0.03 & 90.01 $\pm$ 0.59 & 95.14 $\pm$ 0.50 & 99.08 $\pm$ 0.26 \\
& 50 & 0.07 $\pm$ 0.00 & 90.07 $\pm$ 0.70 & 94.86 $\pm$ 0.53 & 99.02 $\pm$ 0.15 & & 50 & 0.29 $\pm$ 0.06 & 90.01 $\pm$ 0.62 & 94.81 $\pm$ 0.48 & 99.15 $\pm$ 0.54 \\
& 100 & 0.08 $\pm$ 0.00 & 90.36 $\pm$ 0.42 & 95.11 $\pm$ 0.50 & 99.06 $\pm$ 0.15 & & 100 & 0.39 $\pm$ 0.08 & 90.26 $\pm$ 1.20 & 95.08 $\pm$ 0.63 & 99.15 $\pm$ 0.40 \\
& 500 & 0.09 $\pm$ 0.00 & 90.65 $\pm$ 0.51 & 95.49 $\pm$ 0.52 & 99.08 $\pm$ 0.17 & & 500 & 0.66 $\pm$ 0.11 & 90.27 $\pm$ 0.85 & 95.23 $\pm$ 0.54 & 99.21 $\pm$ 0.34 \\
& 1000 & 0.09 $\pm$ 0.00 & 90.65 $\pm$ 0.31 & 95.38 $\pm$ 0.51 & 99.09 $\pm$ 0.06 & & 1000 & 0.80 $\pm$ 0.13 & 90.13 $\pm$ 0.63 & 95.24 $\pm$ 0.53 & 99.23 $\pm$ 0.29 \\
& 5000 & 0.11 $\pm$ 0.02 & 90.63 $\pm$ 0.46 & 95.52 $\pm$ 0.33 & 99.10 $\pm$ 0.09 & & 5000 & 1.54 $\pm$ 0.53 & 90.19 $\pm$ 0.83 & 94.91 $\pm$ 0.23 & 99.10 $\pm$ 0.22 \\
& 10000 & 0.14 $\pm$ 0.02 & 90.74 $\pm$ 0.46 & 95.45 $\pm$ 0.25 & 99.14 $\pm$ 0.18 & & 10000 & 1.86 $\pm$ 0.54 & 90.23 $\pm$ 0.71 & 95.23 $\pm$ 0.52 & 99.06 $\pm$ 0.31 \\
\midrule
\multirow{13}{*}{\textbf{Diamonds}}
& 0.01 & \textbf{0.09 $\pm$ 0.00} & 89.88 $\pm$ 0.41 & 94.98 $\pm$ 0.29 & 99.07 $\pm$ 0.09 & \multirow{13}{*}{\textbf{Fifa}} & 0.01 & \textbf{0.07 $\pm$ 0.00} & 90.17 $\pm$ 0.57 & 95.29 $\pm$ 0.25 & 99.19 $\pm$ 0.07 \\
& 0.05 & \textbf{0.09 $\pm$ 0.00} & 89.84 $\pm$ 0.43 & 94.93 $\pm$ 0.35 & 99.09 $\pm$ 0.08 & & 0.05 & \textbf{0.07 $\pm$ 0.00} & 90.17 $\pm$ 0.60 & 95.31 $\pm$ 0.26 & 99.16 $\pm$ 0.09 \\
& 0.1 & \textbf{0.09 $\pm$ 0.00} & 89.86 $\pm$ 0.39 & 95.00 $\pm$ 0.29 & 99.06 $\pm$ 0.09 & & 0.1 & \textbf{0.07 $\pm$ 0.00} & 90.18 $\pm$ 0.64 & 95.30 $\pm$ 0.26 & 99.15 $\pm$ 0.09 \\
& 0.5 & \textbf{0.09 $\pm$ 0.00} & 89.84 $\pm$ 0.34 & 95.03 $\pm$ 0.33 & 99.03 $\pm$ 0.12 & & 0.5 & \textbf{0.07 $\pm$ 0.00} & 90.33 $\pm$ 0.56 & 95.24 $\pm$ 0.26 & 99.14 $\pm$ 0.07 \\
& 1 & \textbf{0.09 $\pm$ 0.00} & 89.85 $\pm$ 0.30 & 95.10 $\pm$ 0.31 & 99.03 $\pm$ 0.09 & & 1 & \textbf{0.07 $\pm$ 0.00} & 90.25 $\pm$ 0.61 & 95.17 $\pm$ 0.36 & 99.14 $\pm$ 0.16 \\
& \cellcolor{gray!20} 5 & \cellcolor{gray!20} \textbf{0.09 $\pm$ 0.00} & \cellcolor{gray!20} 90.05 $\pm$ 0.36 & \cellcolor{gray!20} 95.02 $\pm$ 0.11 & \cellcolor{gray!20} 99.00 $\pm$ 0.08 & & \cellcolor{gray!20} 5 & \cellcolor{gray!20} \textbf{0.07 $\pm$ 0.00} & \cellcolor{gray!20} 90.27 $\pm$ 0.51 & \cellcolor{gray!20} 94.90 $\pm$ 0.57 & \cellcolor{gray!20} 98.93 $\pm$ 0.16 \\
& 10 & \textbf{0.09 $\pm$ 0.00} & 89.90 $\pm$ 0.51 & 95.04 $\pm$ 0.19 & 98.98 $\pm$ 0.13 & & 10 & 0.08 $\pm$ 0.00 & 90.01 $\pm$ 0.81 & 95.05 $\pm$ 0.25 & 99.00 $\pm$ 0.21 \\
& 50 & 0.10 $\pm$ 0.01 & 89.81 $\pm$ 0.27 & 94.76 $\pm$ 0.19 & 98.97 $\pm$ 0.10 & & 50 & 0.09 $\pm$ 0.00 & 89.95 $\pm$ 0.59 & 94.94 $\pm$ 0.54 & 99.01 $\pm$ 0.28 \\
& 100 & 0.11 $\pm$ 0.01 & 89.86 $\pm$ 0.35 & 94.76 $\pm$ 0.16 & 98.97 $\pm$ 0.10 & & 100 & 0.10 $\pm$ 0.01 & 90.15 $\pm$ 0.55 & 95.05 $\pm$ 0.23 & 98.92 $\pm$ 0.19 \\
& 500 & 0.25 $\pm$ 0.13 & 89.85 $\pm$ 0.18 & 94.99 $\pm$ 0.18 & 99.02 $\pm$ 0.13 & & 500 & 0.13 $\pm$ 0.02 & 90.09 $\pm$ 0.35 & 94.87 $\pm$ 0.25 & 98.88 $\pm$ 0.18 \\
& 1000 & 0.29 $\pm$ 0.12 & 90.03 $\pm$ 0.27 & 95.01 $\pm$ 0.13 & 99.05 $\pm$ 0.12 & & 1000 & 0.16 $\pm$ 0.03 & 89.90 $\pm$ 0.64 & 95.07 $\pm$ 0.39 & 99.06 $\pm$ 0.27 \\
& 5000 & 0.40 $\pm$ 0.08 & 90.04 $\pm$ 0.18 & 95.07 $\pm$ 0.19 & 99.01 $\pm$ 0.09 & & 5000 & 0.29 $\pm$ 0.04 & 89.79 $\pm$ 0.78 & 95.11 $\pm$ 0.48 & 99.07 $\pm$ 0.37 \\
& 10000 & 0.48 $\pm$ 0.15 & 90.08 $\pm$ 0.41 & 95.08 $\pm$ 0.36 & 99.00 $\pm$ 0.07 & & 10000 & 0.35 $\pm$ 0.07 & 89.85 $\pm$ 0.81 & 94.91 $\pm$ 0.72 & 99.04 $\pm$ 0.47 \\
\midrule
\multirow{13}{*}{\begin{tabular}{@{}l@{}}\textbf{House} \\ \textbf{Sales}\end{tabular}}
& 0.01 & \textbf{0.13 $\pm$ 0.00} & 89.77 $\pm$ 0.77 & 94.79 $\pm$ 0.41 & 98.91 $\pm$ 0.25 & \multirow{13}{*}{\textbf{Isolet}} & 0.01 & 0.21 $\pm$ 0.01 & 90.17 $\pm$ 1.13 & 95.27 $\pm$ 0.62 & 99.12 $\pm$ 0.22 \\
& 0.05 & \textbf{0.13 $\pm$ 0.00} & 89.74 $\pm$ 0.62 & 94.74 $\pm$ 0.43 & 98.90 $\pm$ 0.30 & & 0.05 & 0.21 $\pm$ 0.01 & 90.12 $\pm$ 1.40 & 95.17 $\pm$ 0.46 & 99.06 $\pm$ 0.21 \\
& 0.1 & \textbf{0.13 $\pm$ 0.00} & 89.74 $\pm$ 0.53 & 94.71 $\pm$ 0.39 & 98.90 $\pm$ 0.25 & & 0.1 & \textbf{0.20 $\pm$ 0.01} & 90.23 $\pm$ 1.12 & 95.36 $\pm$ 0.65 & 99.14 $\pm$ 0.19 \\
& 0.5 & \textbf{0.13 $\pm$ 0.00} & 89.82 $\pm$ 0.71 & 94.83 $\pm$ 0.47 & 98.94 $\pm$ 0.18 & & 0.5 & \textbf{0.20 $\pm$ 0.01} & 90.13 $\pm$ 0.68 & 95.04 $\pm$ 0.81 & 99.19 $\pm$ 0.17 \\
& 1 & \textbf{0.13 $\pm$ 0.00} & 89.84 $\pm$ 0.84 & 94.74 $\pm$ 0.33 & 98.86 $\pm$ 0.19 & & 1 & \textbf{0.20 $\pm$ 0.01} & 90.06 $\pm$ 1.02 & 94.85 $\pm$ 0.87 & 99.05 $\pm$ 0.22 \\
& \cellcolor{gray!20} 5.0 & \cellcolor{gray!20} 0.14 $\pm$ 0.01 & \cellcolor{gray!20} 89.83 $\pm$ 1.04 & \cellcolor{gray!20} 94.74 $\pm$ 0.41 & \cellcolor{gray!20} 98.87 $\pm$ 0.31 & & \cellcolor{gray!20} 5 & \cellcolor{gray!20} \textbf{0.20 $\pm$ 0.01} & \cellcolor{gray!20} 89.05 $\pm$ 1.26 & \cellcolor{gray!20} 94.54 $\pm$ 0.70 & \cellcolor{gray!20} 99.13 $\pm$ 0.28 \\
& 10 & 0.17 $\pm$ 0.01 & 89.78 $\pm$ 0.79 & 94.94 $\pm$ 0.33 & 98.86 $\pm$ 0.15 & & 10 & 0.22 $\pm$ 0.01 & 89.15 $\pm$ 1.10 & 93.96 $\pm$ 1.34 & 99.22 $\pm$ 0.26 \\
& 50 & 0.48 $\pm$ 0.07 & 89.59 $\pm$ 0.46 & 94.99 $\pm$ 0.37 & 99.07 $\pm$ 0.14 & & 50 & 0.27 $\pm$ 0.01 & 89.83 $\pm$ 1.15 & 94.91 $\pm$ 0.80 & 99.04 $\pm$ 0.15 \\
& 100 & 0.62 $\pm$ 0.15 & 89.56 $\pm$ 0.89 & 94.80 $\pm$ 0.48 & 99.10 $\pm$ 0.15 & & 100 & 0.31 $\pm$ 0.01 & 90.23 $\pm$ 0.67 & 95.06 $\pm$ 0.79 & 99.24 $\pm$ 0.08 \\
& 500 & 0.56 $\pm$ 0.08 & 90.11 $\pm$ 0.34 & 95.13 $\pm$ 0.41 & 99.04 $\pm$ 0.22 & & 500 & 0.39 $\pm$ 0.02 & 90.03 $\pm$ 0.88 & 94.68 $\pm$ 0.68 & 99.18 $\pm$ 0.23 \\
& 1000 & 0.56 $\pm$ 0.07 & 90.57 $\pm$ 0.45 & 95.34 $\pm$ 0.21 & 99.07 $\pm$ 0.14 & & 1000 & 0.42 $\pm$ 0.01 & 90.06 $\pm$ 0.74 & 95.21 $\pm$ 0.50 & 99.14 $\pm$ 0.10 \\
& 5000 & 0.82 $\pm$ 0.11 & 90.03 $\pm$ 0.62 & 95.32 $\pm$ 0.30 & 99.05 $\pm$ 0.20 & & 5000 & 0.57 $\pm$ 0.03 & 89.26 $\pm$ 0.20 & 94.94 $\pm$ 0.58 & 98.94 $\pm$ 0.33 \\
& 10000 & 1.00 $\pm$ 0.18 & 90.09 $\pm$ 0.88 & 94.99 $\pm$ 0.45 & 98.95 $\pm$ 0.18 & & 10000 & 0.68 $\pm$ 0.05 & 89.00 $\pm$ 0.89 & 95.04 $\pm$ 0.59 & 98.82 $\pm$ 0.30 \\
\midrule
\multirow{13}{*}{\begin{tabular}{@{}l@{}}\textbf{Medical} \\ \textbf{Charges}\end{tabular}}
& 0.01 & \textbf{0.04 $\pm$ 0.00} & 90.01 $\pm$ 0.14 & 94.97 $\pm$ 0.28 & 98.96 $\pm$ 0.09 & \multirow{13}{*}{\textbf{Pol}} & 0.01 & 0.11 $\pm$ 0.01 & 90.17 $\pm$ 0.68 & 95.37 $\pm$ 0.71 & 99.05 $\pm$ 0.26 \\
& 0.05 & \textbf{0.04 $\pm$ 0.00} & 90.02 $\pm$ 0.20 & 94.96 $\pm$ 0.27 & 98.96 $\pm$ 0.09 & & 0.05 & 0.11 $\pm$ 0.01 & 90.24 $\pm$ 0.76 & 95.31 $\pm$ 0.70 & 99.03 $\pm$ 0.26 \\
& 0.1 & \textbf{0.04 $\pm$ 0.00} & 89.97 $\pm$ 0.23 & 94.91 $\pm$ 0.28 & 98.95 $\pm$ 0.07 & & 0.1 & 0.11 $\pm$ 0.01 & 90.11 $\pm$ 0.65 & 95.27 $\pm$ 0.36 & 99.05 $\pm$ 0.21 \\
& 0.5 & \textbf{0.04 $\pm$ 0.00} & 89.90 $\pm$ 0.15 & 94.95 $\pm$ 0.24 & 98.97 $\pm$ 0.09 & & 0.5 & 0.10 $\pm$ 0.01 & 89.65 $\pm$ 0.71 & 94.83 $\pm$ 0.80 & 99.09 $\pm$ 0.18 \\
& 1 & \textbf{0.04 $\pm$ 0.00} & 90.01 $\pm$ 0.14 & 94.99 $\pm$ 0.19 & 98.98 $\pm$ 0.06 & & 1 & \textbf{0.09 $\pm$ 0.01} & 89.75 $\pm$ 1.03 & 94.91 $\pm$ 0.63 & 99.03 $\pm$ 0.29 \\
& \cellcolor{gray!20} 5 & \cellcolor{gray!20} \textbf{0.04 $\pm$ 0.00} & \cellcolor{gray!20} 89.90 $\pm$ 0.09 & \cellcolor{gray!20} 95.00 $\pm$ 0.20 & \cellcolor{gray!20} 99.00 $\pm$ 0.07 & & \cellcolor{gray!20} 5 & \cellcolor{gray!20} \textbf{0.09 $\pm$ 0.01} & \cellcolor{gray!20} 90.37 $\pm$ 0.43 & \cellcolor{gray!20} 95.67 $\pm$ 0.37 & \cellcolor{gray!20} 99.14 $\pm$ 0.17 \\
& 10 & 0.05 $\pm$ 0.00 & 90.03 $\pm$ 0.14 & 94.98 $\pm$ 0.13 & 99.03 $\pm$ 0.05 & & 10 & 0.11 $\pm$ 0.01 & 90.15 $\pm$ 0.52 & 95.35 $\pm$ 0.42 & 99.19 $\pm$ 0.35 \\
& 50 & 0.06 $\pm$ 0.00 & 90.00 $\pm$ 0.14 & 95.05 $\pm$ 0.12 & 99.07 $\pm$ 0.04 & & 50 & 0.22 $\pm$ 0.03 & 89.92 $\pm$ 0.36 & 94.83 $\pm$ 0.62 & 99.16 $\pm$ 0.18 \\
& 100 & 0.06 $\pm$ 0.00 & 90.11 $\pm$ 0.11 & 95.04 $\pm$ 0.08 & 99.02 $\pm$ 0.08 & & 100 & 0.32 $\pm$ 0.02 & 89.91 $\pm$ 0.82 & 95.13 $\pm$ 0.43 & 99.03 $\pm$ 0.23 \\
& 500 & 0.07 $\pm$ 0.00 & 90.08 $\pm$ 0.27 & 95.04 $\pm$ 0.08 & 99.00 $\pm$ 0.04 & & 500 & 0.73 $\pm$ 0.08 & 90.21 $\pm$ 0.87 & 95.13 $\pm$ 0.60 & 99.10 $\pm$ 0.21 \\
& 1000 & 0.07 $\pm$ 0.01 & 89.97 $\pm$ 0.18 & 94.94 $\pm$ 0.13 & 99.01 $\pm$ 0.04 & & 1000 & 1.05 $\pm$ 0.13 & 90.23 $\pm$ 0.72 & 95.27 $\pm$ 0.76 & 99.19 $\pm$ 0.15 \\
& 5000 & 0.10 $\pm$ 0.01 & 90.02 $\pm$ 0.37 & 95.11 $\pm$ 0.20 & 99.00 $\pm$ 0.10 & & 5000 & 1.70 $\pm$ 0.05 & 90.27 $\pm$ 0.61 & 95.10 $\pm$ 0.52 & 98.81 $\pm$ 0.20 \\
& 10000 & 0.12 $\pm$ 0.04 & 89.98 $\pm$ 0.22 & 95.10 $\pm$ 0.22 & 99.01 $\pm$ 0.05 & & 10000 & 1.85 $\pm$ 0.03 & 90.41 $\pm$ 0.60 & 95.05 $\pm$ 0.22 & 98.93 $\pm$ 0.16 \\
\midrule
\multirow{13}{*}{\textbf{Superconduct}}
& 0.01 & 0.36 $\pm$ 0.01 & 90.38 $\pm$ 0.53 & 95.16 $\pm$ 0.58 & 99.10 $\pm$ 0.25 & \multirow{13}{*}{\begin{tabular}{@{}l@{}}\textbf{Wine} \\ \textbf{Quality}\end{tabular}} & 0.01 & \textbf{0.08 $\pm$ 0.00} & 90.26 $\pm$ 1.30 & 95.38 $\pm$ 0.28 & 98.95 $\pm$ 0.46 \\
& 0.05 & 0.36 $\pm$ 0.01 & 90.28 $\pm$ 0.56 & 95.12 $\pm$ 0.65 & 99.13 $\pm$ 0.24 & & 0.05 & \textbf{0.08 $\pm$ 0.00} & 90.18 $\pm$ 1.48 & 95.51 $\pm$ 0.42 & 98.97 $\pm$ 0.47 \\
& 0.1 & 0.36 $\pm$ 0.01 & 90.20 $\pm$ 0.46 & 95.05 $\pm$ 0.48 & 99.17 $\pm$ 0.23 & & 0.1 & \textbf{0.08 $\pm$ 0.00} & 90.40 $\pm$ 1.17 & 95.45 $\pm$ 0.38 & 98.95 $\pm$ 0.50 \\
& 0.5 & \textbf{0.35 $\pm$ 0.01} & 90.31 $\pm$ 0.65 & 95.04 $\pm$ 0.38 & 99.06 $\pm$ 0.27 & & 0.5 & \textbf{0.08 $\pm$ 0.00} & 90.55 $\pm$ 1.00 & 95.14 $\pm$ 0.36 & 98.98 $\pm$ 0.31 \\
& 1 & \textbf{0.35 $\pm$ 0.01} & 90.17 $\pm$ 0.39 & 94.80 $\pm$ 0.46 & 98.98 $\pm$ 0.12 & & 1 & \textbf{0.08 $\pm$ 0.00} & 90.32 $\pm$ 0.87 & 94.83 $\pm$ 0.36 & 99.05 $\pm$ 0.26 \\
& \cellcolor{gray!20} 5 & \cellcolor{gray!20} 0.38 $\pm$ 0.01 & \cellcolor{gray!20} 89.79 $\pm$ 0.36 & \cellcolor{gray!20} 95.02 $\pm$ 0.27 & \cellcolor{gray!20} 98.83 $\pm$ 0.16 & & \cellcolor{gray!20} 5 & \cellcolor{gray!20} \textbf{0.08 $\pm$ 0.00} & \cellcolor{gray!20} 90.48 $\pm$ 0.85 & \cellcolor{gray!20} 95.31 $\pm$ 0.80 & \cellcolor{gray!20} 99.02 $\pm$ 0.33 \\
& 10 & 0.42 $\pm$ 0.01 & 89.79 $\pm$ 0.55 & 94.96 $\pm$ 0.23 & 98.85 $\pm$ 0.20 & & 10 & \textbf{0.08 $\pm$ 0.00} & 90.46 $\pm$ 0.69 & 95.31 $\pm$ 0.87 & 99.03 $\pm$ 0.36 \\
& 50 & 0.56 $\pm$ 0.01 & 89.84 $\pm$ 0.31 & 94.95 $\pm$ 0.53 & 98.99 $\pm$ 0.20 & & 50 & 0.11 $\pm$ 0.01 & 90.92 $\pm$ 0.59 & 95.52 $\pm$ 0.66 & 99.08 $\pm$ 0.19 \\
& 100 & 0.63 $\pm$ 0.01 & 89.86 $\pm$ 0.28 & 94.92 $\pm$ 0.52 & 98.97 $\pm$ 0.24 & & 100 & 0.14 $\pm$ 0.02 & 90.57 $\pm$ 0.62 & 95.38 $\pm$ 0.62 & 99.23 $\pm$ 0.31 \\
& 500 & 0.72 $\pm$ 0.03 & 89.57 $\pm$ 0.57 & 94.68 $\pm$ 0.75 & 98.94 $\pm$ 0.22 & & 500 & 0.27 $\pm$ 0.05 & 90.45 $\pm$ 0.81 & 95.31 $\pm$ 0.88 & 99.12 $\pm$ 0.57 \\
& 1000 & 0.74 $\pm$ 0.01 & 89.47 $\pm$ 0.18 & 94.62 $\pm$ 0.29 & 99.10 $\pm$ 0.17 & & 1000 & 0.33 $\pm$ 0.06 & 90.69 $\pm$ 1.27 & 95.77 $\pm$ 1.02 & 99.11 $\pm$ 0.57 \\
& 5000 & 0.78 $\pm$ 0.06 & 89.50 $\pm$ 0.71 & 94.80 $\pm$ 0.35 & 99.11 $\pm$ 0.16 & & 5000 & 0.47 $\pm$ 0.14 & 90.26 $\pm$ 1.64 & 95.60 $\pm$ 1.22 & 99.25 $\pm$ 0.30 \\
& 10000 & 0.78 $\pm$ 0.06 & 89.51 $\pm$ 0.50 & 94.86 $\pm$ 0.42 & 99.06 $\pm$ 0.29 & & 10000 & 0.65 $\pm$ 0.32 & 90.57 $\pm$ 1.60 & 95.54 $\pm$ 1.22 & 99.28 $\pm$ 0.30 \\
\midrule
\multirow{13}{*}{\textbf{Drift}}
& 0.01 & 7.61 $\pm$ 1.29 & 89.16 $\pm$ 0.64 & 94.87 $\pm$ 0.29 & 98.98 $\pm$ 0.13 & \multirow{13}{*}{\begin{tabular}{@{}l@{}}\textbf{UTK} \\ \textbf{Face}\end{tabular}} & 0.01 & 7.80 $\pm$ 2.48 & 90.02 $\pm$ 0.65 & 95.06 $\pm$ 0.42 & 99.01 $\pm$ 0.19 \\
& 0.05 & 9.30 $\pm$ 2.69 & 89.86 $\pm$ 0.80 & 95.09 $\pm$ 0.54 & 99.00 $\pm$ 0.20 & & 0.05 & 8.11 $\pm$ 2.55 & 90.19 $\pm$ 0.72 & 94.85 $\pm$ 0.55 & 99.02 $\pm$ 0.27 \\
& 0.1 & 6.10 $\pm$ 1.37 & 89.97 $\pm$ 0.69 & 95.07 $\pm$ 0.38 & 99.18 $\pm$ 0.22 & & 0.1 & 5.98 $\pm$ 0.38 & 90.13 $\pm$ 0.48 & 94.78 $\pm$ 0.54 & 98.86 $\pm$ 0.19 \\
& 0.5 & 6.10 $\pm$ 0.69 & 90.13 $\pm$ 0.35 & 94.94 $\pm$ 0.44 & 99.12 $\pm$ 0.21 & & 0.5 & 6.01 $\pm$ 0.20 & 89.79 $\pm$ 0.46 & 94.71 $\pm$ 0.39 & 98.88 $\pm$ 0.22 \\
& 1 & \textbf{5.72 $\pm$ 0.22} & 90.00 $\pm$ 0.70 & 95.04 $\pm$ 0.42 & 98.92 $\pm$ 0.23 & & 1 & 5.89 $\pm$ 0.40 & 89.68 $\pm$ 0.76 & 94.52 $\pm$ 0.56 & 98.81 $\pm$ 0.22 \\
& \cellcolor{gray!20} 5 & \cellcolor{gray!20} 6.00 $\pm$ 0.36 & \cellcolor{gray!20} 90.03 $\pm$ 1.29 & \cellcolor{gray!20} 95.17 $\pm$ 0.71 & \cellcolor{gray!20} 99.27 $\pm$ 0.28 & & \cellcolor{gray!20} 5 & \cellcolor{gray!20} \textbf{5.43 $\pm$ 0.18} & \cellcolor{gray!20} 90.13 $\pm$ 0.46 & \cellcolor{gray!20} 95.02 $\pm$ 0.30 & \cellcolor{gray!20} 98.95 $\pm$ 0.14 \\
& 10 & 5.90 $\pm$ 1.05 & 89.89 $\pm$ 0.67 & 95.27 $\pm$ 0.52 & 98.92 $\pm$ 0.35 & & 10 & 5.53 $\pm$ 0.23 & 90.02 $\pm$ 0.20 & 95.09 $\pm$ 0.31 & 99.01 $\pm$ 0.25 \\
& 50 & 7.11 $\pm$ 3.57 & 89.62 $\pm$ 0.59 & 94.87 $\pm$ 0.59 & 99.02 $\pm$ 0.40 & & 50 & 5.65 $\pm$ 0.30 & 90.66 $\pm$ 0.85 & 94.97 $\pm$ 0.45 & 99.03 $\pm$ 0.06 \\
& 100 & 6.32 $\pm$ 0.89 & 89.99 $\pm$ 1.25 & 94.98 $\pm$ 0.85 & 98.84 $\pm$ 0.22 & & 100 & 5.96 $\pm$ 0.14 & 89.81 $\pm$ 0.38 & 95.07 $\pm$ 0.18 & 98.92 $\pm$ 0.17 \\
& 500 & 8.71 $\pm$ 1.62 & 89.98 $\pm$ 0.70 & 95.11 $\pm$ 1.16 & 99.05 $\pm$ 0.45 & & 500 & 7.72 $\pm$ 1.36 & 90.30 $\pm$ 0.34 & 95.39 $\pm$ 0.56 & 99.08 $\pm$ 0.11 \\
& 1000 & 11.36 $\pm$ 1.08 & 90.10 $\pm$ 0.96 & 95.27 $\pm$ 0.51 & 99.14 $\pm$ 0.24 & & 1000 & 8.27 $\pm$ 1.04 & 89.73 $\pm$ 0.64 & 94.85 $\pm$ 0.38 & 98.95 $\pm$ 0.21 \\
& 5000 & 15.78 $\pm$ 2.81 & 90.24 $\pm$ 1.13 & 94.87 $\pm$ 0.55 & 99.16 $\pm$ 0.31 & & 5000 & 14.01 $\pm$ 2.92 & 90.08 $\pm$ 0.35 & 94.98 $\pm$ 0.15 & 98.99 $\pm$ 0.22 \\
& 10000 & 20.23 $\pm$ 1.65 & 90.19 $\pm$ 0.82 & 95.29 $\pm$ 0.57 & 99.20 $\pm$ 0.05 & & 10000 & 19.20 $\pm$ 1.61 & 90.00 $\pm$ 0.34 & 95.03 $\pm$ 0.45 & 98.91 $\pm$ 0.24 \\
\bottomrule
\end{tabular}}
\label{tab:lambda_results_cov_mae_all}\end{table*}

\begin{table*}[t]
\caption{Impact of $\lambda$ on Efficiency across different confidence levels ($1-\alpha$). Best results per dataset are highlighted in bold. The row corresponding to $\lambda = 5$ is shaded in gray. (Part 1).}
\centering\scriptsize\resizebox{\textwidth}{!}{
\begin{tabular}{lcccccccccc}
\toprule
\textbf{Dataset} & \textbf{Lambda} & \multicolumn{3}{c}{$\alpha = 0.10$} & \multicolumn{3}{c}{$\alpha = 0.05$} & \multicolumn{3}{c}{$\alpha = 0.01$} \\
\cmidrule(lr){3-5} \cmidrule(lr){6-8} \cmidrule(lr){9-11}
& \textbf{($\lambda$)} & Med. Eff. $\downarrow$ & Mean Eff. $\downarrow$ & IQR $\downarrow$ & Med. Eff. $\downarrow$ & Mean Eff. $\downarrow$ & IQR $\downarrow$ & Med. Eff. $\downarrow$ & Mean Eff. $\downarrow$ & IQR $\downarrow$ \\ \midrule
\multirow{13}{*}{\begin{tabular}{@{}l@{}}\textbf{Bike} \\ \textbf{Sharing}\end{tabular}}
& 0.01 & 1.57 $\pm$ 0.15 & 2.46 $\pm$ 0.12 & 1.98 $\pm$ 0.16 & 2.21 $\pm$ 0.30 & 3.45 $\pm$ 0.21 & 2.77 $\pm$ 0.26 & 3.88 $\pm$ 1.15 & 5.97 $\pm$ 1.17 & 4.80 $\pm$ 0.96 \\
& 0.05 & 1.57 $\pm$ 0.13 & 2.52 $\pm$ 0.16 & 2.05 $\pm$ 0.20 & 2.20 $\pm$ 0.26 & 3.51 $\pm$ 0.19 & 2.86 $\pm$ 0.27 & 3.83 $\pm$ 1.14 & 6.01 $\pm$ 1.03 & 4.89 $\pm$ 0.82 \\
& 0.1 & 1.55 $\pm$ 0.12 & 2.46 $\pm$ 0.16 & 1.98 $\pm$ 0.19 & 2.18 $\pm$ 0.23 & 3.44 $\pm$ 0.19 & 2.77 $\pm$ 0.25 & 3.78 $\pm$ 1.07 & 5.87 $\pm$ 0.93 & 4.72 $\pm$ 0.71 \\
& 0.5 & 1.52 $\pm$ 0.09 & 2.57 $\pm$ 0.49 & 1.99 $\pm$ 0.48 & 2.13 $\pm$ 0.15 & 3.59 $\pm$ 0.67 & 2.79 $\pm$ 0.67 & 3.71 $\pm$ 0.79 & 6.13 $\pm$ 0.98 & 4.72 $\pm$ 0.81 \\
& 1 & \textbf{1.31 $\pm$ 0.05} & 1.83 $\pm$ 0.31 & 1.20 $\pm$ 0.32 & 1.73 $\pm$ 0.09 & 2.42 $\pm$ 0.45 & 1.59 $\pm$ 0.45 & 2.46 $\pm$ 0.14 & 3.44 $\pm$ 0.67 & 2.26 $\pm$ 0.66 \\
& \cellcolor{gray!20} 5 & \cellcolor{gray!20} 1.29 $\pm$ 0.09 & \cellcolor{gray!20} \textbf{1.35 $\pm$ 0.08} & \cellcolor{gray!20} \textbf{0.49 $\pm$ 0.06} & \cellcolor{gray!20} \textbf{1.54 $\pm$ 0.11} & \cellcolor{gray!20} \textbf{1.60 $\pm$ 0.10} & \cellcolor{gray!20} \textbf{0.58 $\pm$ 0.07} & \cellcolor{gray!20} \textbf{2.07 $\pm$ 0.09} & \cellcolor{gray!20} \textbf{2.16 $\pm$ 0.08} & \cellcolor{gray!20} 0.79 $\pm$ 0.12 \\
& 10 & 1.41 $\pm$ 0.08 & 1.48 $\pm$ 0.08 & 0.54 $\pm$ 0.07 & 1.64 $\pm$ 0.10 & 1.72 $\pm$ 0.10 & 0.63 $\pm$ 0.08 & 2.11 $\pm$ 0.07 & 2.21 $\pm$ 0.06 & 0.81 $\pm$ 0.09 \\
& 50 & 1.90 $\pm$ 0.03 & 2.04 $\pm$ 0.06 & 0.84 $\pm$ 0.19 & 2.14 $\pm$ 0.03 & 2.30 $\pm$ 0.07 & 0.95 $\pm$ 0.21 & 2.55 $\pm$ 0.05 & 2.74 $\pm$ 0.08 & 1.13 $\pm$ 0.25 \\
& 100 & 2.46 $\pm$ 0.11 & 2.60 $\pm$ 0.09 & 1.22 $\pm$ 0.10 & 2.72 $\pm$ 0.11 & 2.87 $\pm$ 0.09 & 1.35 $\pm$ 0.11 & 3.15 $\pm$ 0.13 & 3.32 $\pm$ 0.12 & 1.57 $\pm$ 0.14 \\
& 500 & 3.34 $\pm$ 0.10 & 3.37 $\pm$ 0.11 & 0.94 $\pm$ 0.05 & 3.72 $\pm$ 0.09 & 3.76 $\pm$ 0.11 & 1.05 $\pm$ 0.05 & 4.25 $\pm$ 0.10 & 4.30 $\pm$ 0.12 & 1.20 $\pm$ 0.06 \\
& 1000 & 3.38 $\pm$ 0.05 & 3.42 $\pm$ 0.06 & 0.78 $\pm$ 0.07 & 3.80 $\pm$ 0.05 & 3.84 $\pm$ 0.05 & 0.88 $\pm$ 0.08 & 4.46 $\pm$ 0.06 & 4.51 $\pm$ 0.06 & 1.04 $\pm$ 0.10 \\
& 5000 & 3.61 $\pm$ 0.07 & 3.63 $\pm$ 0.07 & 0.53 $\pm$ 0.04 & 4.19 $\pm$ 0.07 & 4.22 $\pm$ 0.06 & 0.61 $\pm$ 0.05 & 5.12 $\pm$ 0.14 & 5.16 $\pm$ 0.13 & \textbf{0.75 $\pm$ 0.06} \\
& 10000 & 3.92 $\pm$ 0.11 & 3.93 $\pm$ 0.11 & 0.51 $\pm$ 0.07 & 4.54 $\pm$ 0.16 & 4.55 $\pm$ 0.15 & 0.59 $\pm$ 0.07 & 5.64 $\pm$ 0.27 & 5.66 $\pm$ 0.26 & 0.73 $\pm$ 0.09 \\
\midrule
\multirow{13}{*}{\begin{tabular}{@{}l@{}}\textbf{Brazilian} \\ \textbf{Houses}\end{tabular}}
& 0.01 & 0.73 $\pm$ 0.17 & 0.78 $\pm$ 0.19 & 0.59 $\pm$ 0.17 & 0.99 $\pm$ 0.27 & 1.06 $\pm$ 0.30 & 0.81 $\pm$ 0.27 & 1.69 $\pm$ 0.42 & 1.81 $\pm$ 0.47 & 1.39 $\pm$ 0.46 \\
& 0.05 & 0.69 $\pm$ 0.17 & 0.74 $\pm$ 0.18 & 0.56 $\pm$ 0.16 & 0.94 $\pm$ 0.26 & 1.01 $\pm$ 0.29 & 0.77 $\pm$ 0.25 & 1.65 $\pm$ 0.43 & 1.76 $\pm$ 0.47 & 1.35 $\pm$ 0.45 \\
& 0.1 & 0.72 $\pm$ 0.16 & 0.77 $\pm$ 0.18 & 0.58 $\pm$ 0.17 & 0.98 $\pm$ 0.27 & 1.05 $\pm$ 0.30 & 0.80 $\pm$ 0.27 & 1.69 $\pm$ 0.44 & 1.81 $\pm$ 0.49 & 1.39 $\pm$ 0.48 \\
& 0.5 & 0.79 $\pm$ 0.14 & 0.84 $\pm$ 0.16 & 0.61 $\pm$ 0.16 & 1.03 $\pm$ 0.21 & 1.11 $\pm$ 0.23 & 0.81 $\pm$ 0.23 & 1.68 $\pm$ 0.37 & 1.80 $\pm$ 0.42 & 1.32 $\pm$ 0.44 \\
& 1 & \textbf{0.65 $\pm$ 0.11} & \textbf{0.67 $\pm$ 0.11} & 0.33 $\pm$ 0.12 & \textbf{0.80 $\pm$ 0.14} & \textbf{0.82 $\pm$ 0.14} & 0.41 $\pm$ 0.15 & \textbf{1.23 $\pm$ 0.23} & \textbf{1.26 $\pm$ 0.24} & 0.64 $\pm$ 0.26 \\
& \cellcolor{gray!20} 5 & \cellcolor{gray!20} 1.07 $\pm$ 0.12 & \cellcolor{gray!20} 818.78 $\pm$ 1828.32 & \cellcolor{gray!20} \textbf{0.08 $\pm$ 0.01} & \cellcolor{gray!20} 1.22 $\pm$ 0.15 & \cellcolor{gray!20} 918.38 $\pm$ 2050.66 & \cellcolor{gray!20} \textbf{0.09 $\pm$ 0.01} & \cellcolor{gray!20} 1.59 $\pm$ 0.18 & \cellcolor{gray!20} 1203.90 $\pm$ 2688.22 & \cellcolor{gray!20} \textbf{0.12 $\pm$ 0.02} \\
& 10 & 1.30 $\pm$ 0.13 & $(4.90 \pm 11.00) \times 10^7$ & 0.12 $\pm$ 0.04 & 1.47 $\pm$ 0.16 & $(5.60 \pm 12.50) \times 10^7$ & 0.14 $\pm$ 0.05 & 1.87 $\pm$ 0.22 & $(6.80 \pm 15.20) \times 10^7$ & 0.18 $\pm$ 0.06 \\
& 50 & 1.49 $\pm$ 0.21 & $(2.15 \pm 4.80) \times 10^8$ & 0.21 $\pm$ 0.07 & 1.73 $\pm$ 0.25 & $(2.53 \pm 5.66) \times 10^8$ & 0.24 $\pm$ 0.08 & 2.31 $\pm$ 0.46 & $(3.20 \pm 7.15) \times 10^8$ & 0.33 $\pm$ 0.12 \\
& 100 & 1.72 $\pm$ 0.38 & $(1.10 \pm 2.46) \times 10^9$ & 0.24 $\pm$ 0.11 & 2.03 $\pm$ 0.45 & $(1.31 \pm 2.92) \times 10^9$ & 0.29 $\pm$ 0.14 & 2.74 $\pm$ 0.70 & $(1.71 \pm 3.82) \times 10^9$ & 0.39 $\pm$ 0.20 \\
& 500 & 3.41 $\pm$ 2.41 & $(1.06 \pm 2.38) \times 10^9$ & 0.65 $\pm$ 0.65 & 3.96 $\pm$ 2.64 & $(1.28 \pm 2.86) \times 10^9$ & 0.75 $\pm$ 0.73 & 5.25 $\pm$ 3.34 & $(1.63 \pm 3.63) \times 10^9$ & 0.99 $\pm$ 0.93 \\
& 1000 & 4.67 $\pm$ 3.28 & $(5.67 \pm 12.70) \times 10^9$ & 1.02 $\pm$ 0.87 & 5.36 $\pm$ 3.64 & $(6.57 \pm 14.70) \times 10^9$ & 1.17 $\pm$ 0.97 & 6.97 $\pm$ 4.76 & $(8.09 \pm 18.10) \times 10^9$ & 1.51 $\pm$ 1.27 \\
& 5000 & 9.04 $\pm$ 4.18 & $(5.20 \pm 11.60) \times 10^{18}$ & 2.19 $\pm$ 1.18 & 10.34 $\pm$ 4.36 & $(6.84 \pm 15.30) \times 10^{18}$ & 2.52 $\pm$ 1.30 & 13.57 $\pm$ 4.89 & $(1.13 \pm 2.53) \times 10^{19}$ & 3.34 $\pm$ 1.67 \\
& 10000 & 10.46 $\pm$ 3.53 & $(1.56 \pm 3.49) \times 10^{21}$ & 2.71 $\pm$ 1.87 & 11.95 $\pm$ 3.84 & $(2.01 \pm 4.50) \times 10^{21}$ & 3.16 $\pm$ 2.25 & 15.43 $\pm$ 4.74 & $(3.26 \pm 7.28) \times 10^{21}$ & 4.24 $\pm$ 3.15 \\
\midrule
\multirow{13}{*}{\textbf{California}}
& 0.01 & 0.30 $\pm$ 0.01 & 0.30 $\pm$ 0.01 & 0.06 $\pm$ 0.01 & 0.39 $\pm$ 0.02 & 0.39 $\pm$ 0.02 & 0.07 $\pm$ 0.02 & 0.66 $\pm$ 0.03 & 0.67 $\pm$ 0.02 & 0.12 $\pm$ 0.03 \\
& 0.05 & 0.30 $\pm$ 0.01 & 0.30 $\pm$ 0.01 & 0.06 $\pm$ 0.01 & 0.38 $\pm$ 0.02 & 0.39 $\pm$ 0.02 & 0.07 $\pm$ 0.02 & 0.66 $\pm$ 0.03 & 0.67 $\pm$ 0.03 & 0.13 $\pm$ 0.03 \\
& 0.1 & 0.30 $\pm$ 0.01 & 0.30 $\pm$ 0.01 & 0.06 $\pm$ 0.01 & 0.38 $\pm$ 0.02 & 0.39 $\pm$ 0.02 & 0.07 $\pm$ 0.02 & 0.67 $\pm$ 0.03 & 0.68 $\pm$ 0.02 & 0.13 $\pm$ 0.03 \\
& 0.5 & 0.29 $\pm$ 0.01 & 0.29 $\pm$ 0.01 & 0.06 $\pm$ 0.01 & 0.38 $\pm$ 0.02 & 0.38 $\pm$ 0.02 & 0.08 $\pm$ 0.02 & 0.68 $\pm$ 0.05 & 0.68 $\pm$ 0.04 & 0.14 $\pm$ 0.03 \\
& 1 & 0.28 $\pm$ 0.00 & 0.28 $\pm$ 0.01 & 0.06 $\pm$ 0.01 & 0.36 $\pm$ 0.01 & 0.36 $\pm$ 0.01 & 0.07 $\pm$ 0.01 & 0.65 $\pm$ 0.04 & 0.64 $\pm$ 0.03 & 0.13 $\pm$ 0.02 \\
& \cellcolor{gray!20} 5 & \cellcolor{gray!20} \textbf{0.26 $\pm$ 0.01} & \cellcolor{gray!20} \textbf{0.26 $\pm$ 0.01} & \cellcolor{gray!20} 0.06 $\pm$ 0.00 & \cellcolor{gray!20} 0.33 $\pm$ 0.01 & \cellcolor{gray!20} \textbf{0.32 $\pm$ 0.01} & \cellcolor{gray!20} 0.07 $\pm$ 0.00 & \cellcolor{gray!20} 0.53 $\pm$ 0.03 & \cellcolor{gray!20} 0.52 $\pm$ 0.03 & \cellcolor{gray!20} 0.12 $\pm$ 0.01 \\
& 10 & 0.26 $\pm$ 0.01 & 0.26 $\pm$ 0.01 & 0.06 $\pm$ 0.00 & \textbf{0.32 $\pm$ 0.01} & \textbf{0.32 $\pm$ 0.01} & 0.08 $\pm$ 0.00 & 0.48 $\pm$ 0.02 & 0.49 $\pm$ 0.03 & 0.12 $\pm$ 0.01 \\
& 50 & 0.28 $\pm$ 0.01 & 0.29 $\pm$ 0.01 & 0.07 $\pm$ 0.01 & 0.34 $\pm$ 0.01 & 0.35 $\pm$ 0.01 & 0.09 $\pm$ 0.01 & \textbf{0.46 $\pm$ 0.01} & \textbf{0.47 $\pm$ 0.01} & 0.12 $\pm$ 0.01 \\
& 100 & 0.30 $\pm$ 0.01 & 0.31 $\pm$ 0.01 & 0.08 $\pm$ 0.01 & 0.36 $\pm$ 0.01 & 0.37 $\pm$ 0.01 & 0.09 $\pm$ 0.01 & 0.48 $\pm$ 0.02 & 0.49 $\pm$ 0.02 & 0.12 $\pm$ 0.01 \\
& 500 & 0.34 $\pm$ 0.02 & 0.36 $\pm$ 0.02 & 0.07 $\pm$ 0.02 & 0.42 $\pm$ 0.03 & 0.44 $\pm$ 0.02 & 0.08 $\pm$ 0.02 & 0.56 $\pm$ 0.04 & 0.58 $\pm$ 0.03 & 0.11 $\pm$ 0.03 \\
& 1000 & 0.37 $\pm$ 0.03 & 0.40 $\pm$ 0.04 & 0.06 $\pm$ 0.01 & 0.45 $\pm$ 0.03 & 0.48 $\pm$ 0.05 & 0.07 $\pm$ 0.01 & 0.59 $\pm$ 0.04 & 0.63 $\pm$ 0.06 & 0.09 $\pm$ 0.02 \\
& 5000 & 0.45 $\pm$ 0.06 & 0.48 $\pm$ 0.05 & \textbf{0.04 $\pm$ 0.01} & 0.54 $\pm$ 0.06 & 0.57 $\pm$ 0.05 & \textbf{0.05 $\pm$ 0.01} & 0.72 $\pm$ 0.08 & 0.76 $\pm$ 0.06 & \textbf{0.06 $\pm$ 0.02} \\
& 10000 & 0.55 $\pm$ 0.08 & 0.82 $\pm$ 0.53 & 0.06 $\pm$ 0.02 & 0.65 $\pm$ 0.08 & 0.96 $\pm$ 0.61 & 0.06 $\pm$ 0.02 & 0.84 $\pm$ 0.09 & 1.24 $\pm$ 0.79 & 0.08 $\pm$ 0.03 \\
\midrule
\multirow{13}{*}{\begin{tabular}{@{}l@{}}\textbf{CPU} \\ \textbf{Act}\end{tabular}}
& 0.01 & \textbf{0.15 $\pm$ 0.01} & 0.19 $\pm$ 0.02 & 0.04 $\pm$ 0.01 & \textbf{0.21 $\pm$ 0.02} & 0.26 $\pm$ 0.03 & 0.06 $\pm$ 0.01 & 1.04 $\pm$ 0.24 & 1.29 $\pm$ 0.25 & 0.30 $\pm$ 0.05 \\
& 0.05 & \textbf{0.15 $\pm$ 0.01} & 0.19 $\pm$ 0.03 & 0.04 $\pm$ 0.01 & \textbf{0.21 $\pm$ 0.02} & 0.27 $\pm$ 0.03 & 0.06 $\pm$ 0.01 & 1.06 $\pm$ 0.29 & 1.31 $\pm$ 0.33 & 0.31 $\pm$ 0.05 \\
& 0.1 & \textbf{0.15 $\pm$ 0.01} & 0.19 $\pm$ 0.03 & 0.04 $\pm$ 0.01 & 0.22 $\pm$ 0.02 & 0.27 $\pm$ 0.03 & 0.06 $\pm$ 0.01 & 1.06 $\pm$ 0.20 & 1.32 $\pm$ 0.22 & 0.31 $\pm$ 0.04 \\
& 0.5 & \textbf{0.15 $\pm$ 0.01} & 0.18 $\pm$ 0.03 & 0.04 $\pm$ 0.01 & \textbf{0.21 $\pm$ 0.02} & 0.26 $\pm$ 0.03 & 0.06 $\pm$ 0.01 & 0.98 $\pm$ 0.19 & 1.18 $\pm$ 0.20 & 0.26 $\pm$ 0.05 \\
& 1 & \textbf{0.15 $\pm$ 0.02} & \textbf{0.17 $\pm$ 0.03} & \textbf{0.03 $\pm$ 0.02} & \textbf{0.21 $\pm$ 0.02} & \textbf{0.24 $\pm$ 0.03} & \textbf{0.04 $\pm$ 0.02} & 0.82 $\pm$ 0.19 & 0.90 $\pm$ 0.17 & 0.16 $\pm$ 0.03 \\
& \cellcolor{gray!20} 5 & \cellcolor{gray!20} 0.24 $\pm$ 0.05 & \cellcolor{gray!20} 0.28 $\pm$ 0.06 & \cellcolor{gray!20} 0.06 $\pm$ 0.02 & \cellcolor{gray!20} 0.27 $\pm$ 0.05 & \cellcolor{gray!20} 0.32 $\pm$ 0.06 & \cellcolor{gray!20} 0.07 $\pm$ 0.02 & \cellcolor{gray!20} \textbf{0.37 $\pm$ 0.05} & \cellcolor{gray!20} \textbf{0.44 $\pm$ 0.06} & \cellcolor{gray!20} \textbf{0.10 $\pm$ 0.02} \\
& 10 & 0.33 $\pm$ 0.07 & 0.40 $\pm$ 0.09 & 0.10 $\pm$ 0.03 & 0.37 $\pm$ 0.07 & 0.45 $\pm$ 0.09 & 0.11 $\pm$ 0.03 & 0.47 $\pm$ 0.06 & 0.58 $\pm$ 0.08 & 0.14 $\pm$ 0.03 \\
& 50 & 0.79 $\pm$ 0.15 & 0.95 $\pm$ 0.18 & 0.29 $\pm$ 0.07 & 0.91 $\pm$ 0.17 & 1.08 $\pm$ 0.20 & 0.33 $\pm$ 0.08 & 1.21 $\pm$ 0.18 & 1.44 $\pm$ 0.21 & 0.44 $\pm$ 0.08 \\
& 100 & 1.14 $\pm$ 0.20 & 1.33 $\pm$ 0.23 & 0.43 $\pm$ 0.10 & 1.33 $\pm$ 0.22 & 1.55 $\pm$ 0.25 & 0.50 $\pm$ 0.11 & 1.77 $\pm$ 0.24 & 2.06 $\pm$ 0.28 & 0.66 $\pm$ 0.13 \\
& 500 & 2.19 $\pm$ 0.33 & 2.57 $\pm$ 0.48 & 0.69 $\pm$ 0.19 & 2.57 $\pm$ 0.33 & 3.01 $\pm$ 0.51 & 0.81 $\pm$ 0.20 & 3.29 $\pm$ 0.29 & 3.86 $\pm$ 0.46 & 1.04 $\pm$ 0.22 \\
& 1000 & 2.71 $\pm$ 0.37 & 3.21 $\pm$ 0.55 & 0.72 $\pm$ 0.25 & 3.20 $\pm$ 0.34 & 3.79 $\pm$ 0.58 & 0.85 $\pm$ 0.27 & 4.12 $\pm$ 0.32 & 4.86 $\pm$ 0.56 & 1.08 $\pm$ 0.29 \\
& 5000 & 4.84 $\pm$ 1.33 & 5.81 $\pm$ 1.60 & 1.27 $\pm$ 0.39 & 5.50 $\pm$ 1.35 & 6.60 $\pm$ 1.69 & 1.45 $\pm$ 0.46 & 6.79 $\pm$ 1.39 & 8.14 $\pm$ 1.72 & 1.79 $\pm$ 0.50 \\
& 10000 & 5.89 $\pm$ 1.61 & 7.46 $\pm$ 2.45 & 1.87 $\pm$ 0.74 & 6.60 $\pm$ 1.64 & 8.36 $\pm$ 2.59 & 2.10 $\pm$ 0.81 & 8.10 $\pm$ 1.77 & 10.23 $\pm$ 2.88 & 2.56 $\pm$ 0.93 \\
\midrule
\multirow{13}{*}{\textbf{Diamonds}}
& 0.01 & 0.54 $\pm$ 0.03 & 0.63 $\pm$ 0.05 & 0.62 $\pm$ 0.08 & 0.84 $\pm$ 0.05 & 0.98 $\pm$ 0.09 & 0.97 $\pm$ 0.14 & 1.75 $\pm$ 0.06 & 2.03 $\pm$ 0.10 & 2.01 $\pm$ 0.21 \\
& 0.05 & 0.53 $\pm$ 0.03 & 0.61 $\pm$ 0.05 & 0.59 $\pm$ 0.08 & 0.82 $\pm$ 0.05 & 0.95 $\pm$ 0.08 & 0.92 $\pm$ 0.14 & 1.72 $\pm$ 0.07 & 1.99 $\pm$ 0.11 & 1.93 $\pm$ 0.22 \\
& 0.1 & 0.53 $\pm$ 0.03 & 0.61 $\pm$ 0.05 & 0.59 $\pm$ 0.09 & 0.81 $\pm$ 0.06 & 0.94 $\pm$ 0.09 & 0.92 $\pm$ 0.15 & 1.72 $\pm$ 0.08 & 1.98 $\pm$ 0.12 & 1.93 $\pm$ 0.25 \\
& 0.5 & 0.53 $\pm$ 0.02 & 0.61 $\pm$ 0.03 & 0.60 $\pm$ 0.04 & 0.82 $\pm$ 0.03 & 0.95 $\pm$ 0.05 & 0.94 $\pm$ 0.07 & 1.74 $\pm$ 0.15 & 2.02 $\pm$ 0.20 & 2.00 $\pm$ 0.24 \\
& 1 & 0.45 $\pm$ 0.02 & 0.49 $\pm$ 0.02 & 0.38 $\pm$ 0.04 & 0.67 $\pm$ 0.04 & 0.72 $\pm$ 0.04 & 0.57 $\pm$ 0.06 & 1.35 $\pm$ 0.10 & 1.45 $\pm$ 0.11 & 1.14 $\pm$ 0.14 \\
& \cellcolor{gray!20} 5 & \cellcolor{gray!20} \textbf{0.34 $\pm$ 0.00} & \cellcolor{gray!20} \textbf{0.35 $\pm$ 0.00} & \cellcolor{gray!20} 0.06 $\pm$ 0.00 & \cellcolor{gray!20} \textbf{0.42 $\pm$ 0.00} & \cellcolor{gray!20} \textbf{0.42 $\pm$ 0.00} & \cellcolor{gray!20} 0.07 $\pm$ 0.01 & \cellcolor{gray!20} \textbf{0.60 $\pm$ 0.01} & \cellcolor{gray!20} \textbf{0.61 $\pm$ 0.01} & \cellcolor{gray!20} 0.10 $\pm$ 0.01 \\
& 10 & 0.36 $\pm$ 0.00 & 0.38 $\pm$ 0.03 & \textbf{0.05 $\pm$ 0.00} & \textbf{0.42 $\pm$ 0.01} & 0.44 $\pm$ 0.04 & \textbf{0.05 $\pm$ 0.00} & \textbf{0.60 $\pm$ 0.01} & 0.63 $\pm$ 0.05 & \textbf{0.08 $\pm$ 0.01} \\
& 50 & 0.40 $\pm$ 0.03 & 0.43 $\pm$ 0.04 & \textbf{0.05 $\pm$ 0.01} & 0.47 $\pm$ 0.04 & 0.50 $\pm$ 0.06 & 0.06 $\pm$ 0.01 & 0.61 $\pm$ 0.04 & 0.66 $\pm$ 0.06 & \textbf{0.08 $\pm$ 0.01} \\
& 100 & 0.43 $\pm$ 0.05 & 0.48 $\pm$ 0.09 & 0.06 $\pm$ 0.02 & 0.51 $\pm$ 0.06 & 0.57 $\pm$ 0.11 & 0.07 $\pm$ 0.02 & 0.65 $\pm$ 0.07 & 0.74 $\pm$ 0.13 & 0.09 $\pm$ 0.03 \\
& 500 & 0.79 $\pm$ 0.34 & 0.96 $\pm$ 0.47 & 0.16 $\pm$ 0.09 & 0.90 $\pm$ 0.35 & 1.09 $\pm$ 0.50 & 0.18 $\pm$ 0.10 & 1.10 $\pm$ 0.39 & 1.32 $\pm$ 0.57 & 0.22 $\pm$ 0.11 \\
& 1000 & 0.93 $\pm$ 0.35 & 1.07 $\pm$ 0.41 & 0.19 $\pm$ 0.11 & 1.06 $\pm$ 0.38 & 1.21 $\pm$ 0.45 & 0.22 $\pm$ 0.12 & 1.30 $\pm$ 0.44 & 1.49 $\pm$ 0.53 & 0.27 $\pm$ 0.15 \\
& 5000 & 1.34 $\pm$ 0.28 & 1.61 $\pm$ 0.38 & 0.34 $\pm$ 0.18 & 1.51 $\pm$ 0.31 & 1.83 $\pm$ 0.48 & 0.39 $\pm$ 0.20 & 1.87 $\pm$ 0.39 & 2.29 $\pm$ 0.76 & 0.49 $\pm$ 0.26 \\
& 10000 & 1.51 $\pm$ 0.27 & 1.90 $\pm$ 0.52 & 0.40 $\pm$ 0.11 & 1.73 $\pm$ 0.24 & 2.21 $\pm$ 0.74 & 0.47 $\pm$ 0.14 & 2.22 $\pm$ 0.31 & 2.90 $\pm$ 1.29 & 0.61 $\pm$ 0.21 \\
\midrule
\multirow{13}{*}{\textbf{Fifa}}
& 0.01 & 0.34 $\pm$ 0.02 & 0.32 $\pm$ 0.02 & 0.06 $\pm$ 0.02 & 0.49 $\pm$ 0.02 & 0.48 $\pm$ 0.02 & 0.08 $\pm$ 0.03 & 0.96 $\pm$ 0.22 & 0.93 $\pm$ 0.21 & 0.16 $\pm$ 0.07 \\
& 0.05 & 0.34 $\pm$ 0.02 & 0.32 $\pm$ 0.02 & 0.06 $\pm$ 0.02 & 0.50 $\pm$ 0.02 & 0.48 $\pm$ 0.02 & 0.09 $\pm$ 0.03 & 0.99 $\pm$ 0.24 & 0.95 $\pm$ 0.22 & 0.17 $\pm$ 0.08 \\
& 0.1 & 0.34 $\pm$ 0.02 & 0.33 $\pm$ 0.02 & 0.06 $\pm$ 0.02 & 0.50 $\pm$ 0.02 & 0.48 $\pm$ 0.02 & 0.09 $\pm$ 0.03 & 0.98 $\pm$ 0.16 & 0.94 $\pm$ 0.15 & 0.18 $\pm$ 0.07 \\
& 0.5 & 0.36 $\pm$ 0.02 & 0.34 $\pm$ 0.02 & 0.08 $\pm$ 0.02 & 0.53 $\pm$ 0.03 & 0.51 $\pm$ 0.02 & 0.12 $\pm$ 0.03 & 1.22 $\pm$ 0.26 & 1.15 $\pm$ 0.24 & 0.27 $\pm$ 0.10 \\
& 1 & 0.35 $\pm$ 0.02 & 0.33 $\pm$ 0.01 & 0.07 $\pm$ 0.02 & 0.52 $\pm$ 0.03 & 0.49 $\pm$ 0.02 & 0.11 $\pm$ 0.03 & 1.13 $\pm$ 0.22 & 1.07 $\pm$ 0.20 & 0.23 $\pm$ 0.09 \\
& \cellcolor{gray!20} 5 & \cellcolor{gray!20} \textbf{0.28 $\pm$ 0.00} & \cellcolor{gray!20} \textbf{0.29 $\pm$ 0.00} & \cellcolor{gray!20} \textbf{0.05 $\pm$ 0.00} & \cellcolor{gray!20} \textbf{0.36 $\pm$ 0.01} & \cellcolor{gray!20} \textbf{0.36 $\pm$ 0.00} & \cellcolor{gray!20} 0.07 $\pm$ 0.01 & \cellcolor{gray!20} 0.61 $\pm$ 0.04 & \cellcolor{gray!20} 0.62 $\pm$ 0.04 & \cellcolor{gray!20} 0.12 $\pm$ 0.02 \\
& 10 & 0.31 $\pm$ 0.00 & 0.31 $\pm$ 0.00 & 0.06 $\pm$ 0.00 & \textbf{0.36 $\pm$ 0.00} & \textbf{0.36 $\pm$ 0.00} & 0.07 $\pm$ 0.01 & 0.51 $\pm$ 0.02 & 0.52 $\pm$ 0.02 & 0.10 $\pm$ 0.01 \\
& 50 & 0.34 $\pm$ 0.00 & 0.35 $\pm$ 0.00 & \textbf{0.05 $\pm$ 0.00} & 0.38 $\pm$ 0.00 & 0.38 $\pm$ 0.00 & \textbf{0.06 $\pm$ 0.00} & \textbf{0.44 $\pm$ 0.01} & \textbf{0.45 $\pm$ 0.01} & \textbf{0.07 $\pm$ 0.00} \\
& 100 & 0.35 $\pm$ 0.01 & 0.36 $\pm$ 0.02 & \textbf{0.05 $\pm$ 0.00} & 0.39 $\pm$ 0.02 & 0.41 $\pm$ 0.02 & \textbf{0.06 $\pm$ 0.00} & 0.46 $\pm$ 0.03 & 0.48 $\pm$ 0.03 & \textbf{0.07 $\pm$ 0.00} \\
& 500 & 0.46 $\pm$ 0.06 & 0.50 $\pm$ 0.07 & 0.07 $\pm$ 0.01 & 0.52 $\pm$ 0.07 & 0.55 $\pm$ 0.07 & 0.08 $\pm$ 0.01 & 0.61 $\pm$ 0.07 & 0.65 $\pm$ 0.07 & 0.09 $\pm$ 0.01 \\
& 1000 & 0.54 $\pm$ 0.10 & 0.58 $\pm$ 0.11 & 0.09 $\pm$ 0.02 & 0.60 $\pm$ 0.10 & 0.64 $\pm$ 0.11 & 0.10 $\pm$ 0.02 & 0.70 $\pm$ 0.11 & 0.75 $\pm$ 0.12 & 0.12 $\pm$ 0.03 \\
& 5000 & 0.96 $\pm$ 0.10 & 1.02 $\pm$ 0.11 & 0.14 $\pm$ 0.02 & 1.06 $\pm$ 0.11 & 1.13 $\pm$ 0.12 & 0.16 $\pm$ 0.02 & 1.23 $\pm$ 0.15 & 1.31 $\pm$ 0.16 & 0.18 $\pm$ 0.03 \\
& 10000 & 1.21 $\pm$ 0.19 & 1.30 $\pm$ 0.21 & 0.19 $\pm$ 0.01 & 1.35 $\pm$ 0.21 & 1.45 $\pm$ 0.22 & 0.21 $\pm$ 0.02 & 1.59 $\pm$ 0.24 & 1.71 $\pm$ 0.25 & 0.25 $\pm$ 0.02 \\
\midrule
\multirow{13}{*}{\begin{tabular}{@{}l@{}}\textbf{House} \\ \textbf{Sales}\end{tabular}}
& 0.01 & 0.60 $\pm$ 0.02 & 0.61 $\pm$ 0.02 & 0.20 $\pm$ 0.01 & 0.79 $\pm$ 0.03 & 0.80 $\pm$ 0.03 & 0.26 $\pm$ 0.01 & 1.27 $\pm$ 0.05 & 1.30 $\pm$ 0.05 & 0.42 $\pm$ 0.02 \\
& 0.05 & 0.60 $\pm$ 0.02 & 0.61 $\pm$ 0.01 & 0.20 $\pm$ 0.01 & 0.78 $\pm$ 0.03 & 0.80 $\pm$ 0.03 & 0.26 $\pm$ 0.02 & 1.27 $\pm$ 0.06 & 1.30 $\pm$ 0.06 & 0.42 $\pm$ 0.02 \\
& 0.1 & 0.60 $\pm$ 0.02 & 0.61 $\pm$ 0.02 & 0.20 $\pm$ 0.01 & 0.78 $\pm$ 0.03 & 0.80 $\pm$ 0.03 & 0.26 $\pm$ 0.01 & 1.27 $\pm$ 0.06 & 1.29 $\pm$ 0.05 & 0.42 $\pm$ 0.02 \\
& 0.5 & 0.60 $\pm$ 0.02 & 0.61 $\pm$ 0.02 & 0.19 $\pm$ 0.01 & 0.78 $\pm$ 0.02 & 0.79 $\pm$ 0.02 & 0.25 $\pm$ 0.01 & 1.26 $\pm$ 0.04 & 1.28 $\pm$ 0.04 & 0.40 $\pm$ 0.01 \\
& 1 & 0.58 $\pm$ 0.02 & 0.59 $\pm$ 0.02 & 0.13 $\pm$ 0.00 & 0.75 $\pm$ 0.02 & 0.75 $\pm$ 0.02 & 0.17 $\pm$ 0.01 & 1.17 $\pm$ 0.06 & 1.18 $\pm$ 0.05 & 0.26 $\pm$ 0.01 \\
& \cellcolor{gray!20} 5 & \cellcolor{gray!20} \textbf{0.55 $\pm$ 0.02} & \cellcolor{gray!20} \textbf{0.57 $\pm$ 0.02} & \cellcolor{gray!20} 0.10 $\pm$ 0.01 & \cellcolor{gray!20} \textbf{0.68 $\pm$ 0.03} & \cellcolor{gray!20} \textbf{0.70 $\pm$ 0.02} & \cellcolor{gray!20} 0.12 $\pm$ 0.01 & \cellcolor{gray!20} \textbf{1.00 $\pm$ 0.03} & \cellcolor{gray!20} \textbf{1.03 $\pm$ 0.04} & \cellcolor{gray!20} 0.18 $\pm$ 0.01 \\
& 10 & 0.63 $\pm$ 0.05 & 0.65 $\pm$ 0.05 & \textbf{0.09 $\pm$ 0.01} & 0.76 $\pm$ 0.05 & 0.79 $\pm$ 0.05 & \textbf{0.11 $\pm$ 0.01} & 1.05 $\pm$ 0.05 & 1.09 $\pm$ 0.06 & \textbf{0.16 $\pm$ 0.02} \\
& 50 & 1.43 $\pm$ 0.17 & 1.48 $\pm$ 0.18 & 0.15 $\pm$ 0.03 & 1.59 $\pm$ 0.18 & 1.65 $\pm$ 0.19 & 0.17 $\pm$ 0.03 & 1.93 $\pm$ 0.22 & 2.01 $\pm$ 0.24 & 0.20 $\pm$ 0.05 \\
& 100 & 1.84 $\pm$ 0.32 & 1.92 $\pm$ 0.33 & 0.19 $\pm$ 0.04 & 2.04 $\pm$ 0.32 & 2.12 $\pm$ 0.33 & 0.21 $\pm$ 0.05 & 2.47 $\pm$ 0.33 & 2.56 $\pm$ 0.35 & 0.26 $\pm$ 0.06 \\
& 500 & 2.02 $\pm$ 0.29 & 2.09 $\pm$ 0.31 & 0.23 $\pm$ 0.07 & 2.37 $\pm$ 0.31 & 2.46 $\pm$ 0.34 & 0.27 $\pm$ 0.09 & 3.21 $\pm$ 0.47 & 3.33 $\pm$ 0.51 & 0.37 $\pm$ 0.13 \\
& 1000 & 2.15 $\pm$ 0.26 & 2.25 $\pm$ 0.29 & 0.28 $\pm$ 0.10 & 2.62 $\pm$ 0.35 & 2.73 $\pm$ 0.39 & 0.35 $\pm$ 0.13 & 3.64 $\pm$ 0.52 & 3.80 $\pm$ 0.57 & 0.48 $\pm$ 0.19 \\
& 5000 & 3.05 $\pm$ 0.47 & 3.26 $\pm$ 0.48 & 0.59 $\pm$ 0.22 & 3.70 $\pm$ 0.51 & 3.94 $\pm$ 0.52 & 0.72 $\pm$ 0.25 & 4.90 $\pm$ 0.69 & 5.23 $\pm$ 0.72 & 0.95 $\pm$ 0.34 \\
& 10000 & 3.70 $\pm$ 0.74 & 3.98 $\pm$ 0.76 & 0.81 $\pm$ 0.30 & 4.44 $\pm$ 0.90 & 4.78 $\pm$ 0.92 & 0.97 $\pm$ 0.36 & 5.81 $\pm$ 1.12 & 6.26 $\pm$ 1.16 & 1.27 $\pm$ 0.47 \\
\midrule
\multirow{13}{*}{\textbf{Isolet}}
& 0.01 & 1.37 $\pm$ 0.26 & 2.03 $\pm$ 0.47 & 1.84 $\pm$ 0.52 & 2.05 $\pm$ 0.30 & 3.04 $\pm$ 0.59 & 2.76 $\pm$ 0.67 & 4.09 $\pm$ 0.67 & 6.07 $\pm$ 1.28 & 5.50 $\pm$ 1.42 \\
& 0.05 & 1.35 $\pm$ 0.26 & 1.99 $\pm$ 0.43 & 1.80 $\pm$ 0.50 & 2.00 $\pm$ 0.32 & 2.95 $\pm$ 0.56 & 2.67 $\pm$ 0.66 & 3.94 $\pm$ 0.72 & 5.81 $\pm$ 1.23 & 5.25 $\pm$ 1.42 \\
& 0.1 & 1.33 $\pm$ 0.23 & 1.95 $\pm$ 0.37 & 1.75 $\pm$ 0.47 & 2.04 $\pm$ 0.37 & 2.98 $\pm$ 0.61 & 2.67 $\pm$ 0.74 & 3.97 $\pm$ 0.67 & 5.81 $\pm$ 1.11 & 5.21 $\pm$ 1.33 \\
& 0.5 & 1.13 $\pm$ 0.08 & 1.45 $\pm$ 0.13 & 1.17 $\pm$ 0.16 & 1.63 $\pm$ 0.17 & 2.09 $\pm$ 0.26 & 1.69 $\pm$ 0.28 & 3.15 $\pm$ 0.42 & 4.03 $\pm$ 0.61 & 3.26 $\pm$ 0.58 \\
& 1 & 0.95 $\pm$ 0.02 & 1.08 $\pm$ 0.04 & 0.74 $\pm$ 0.04 & 1.29 $\pm$ 0.05 & 1.47 $\pm$ 0.06 & 1.00 $\pm$ 0.06 & 2.28 $\pm$ 0.11 & 2.60 $\pm$ 0.12 & 1.77 $\pm$ 0.07 \\
& \cellcolor{gray!20} 5 & \cellcolor{gray!20} \textbf{0.78 $\pm$ 0.06} & \cellcolor{gray!20} \textbf{0.85 $\pm$ 0.07} & \cellcolor{gray!20} 0.43 $\pm$ 0.03 & \cellcolor{gray!20} 1.01 $\pm$ 0.07 & \cellcolor{gray!20} \textbf{1.09 $\pm$ 0.07} & \cellcolor{gray!20} 0.56 $\pm$ 0.05 & \cellcolor{gray!20} 1.61 $\pm$ 0.11 & \cellcolor{gray!20} 1.75 $\pm$ 0.11 & \cellcolor{gray!20} 0.90 $\pm$ 0.09 \\
& 10 & 0.81 $\pm$ 0.05 & 0.89 $\pm$ 0.05 & 0.44 $\pm$ 0.02 & \textbf{1.00 $\pm$ 0.09} & 1.10 $\pm$ 0.10 & 0.55 $\pm$ 0.03 & \textbf{1.58 $\pm$ 0.13} & \textbf{1.74 $\pm$ 0.14} & 0.87 $\pm$ 0.05 \\
& 50 & 1.03 $\pm$ 0.05 & 1.09 $\pm$ 0.04 & 0.46 $\pm$ 0.03 & 1.23 $\pm$ 0.07 & 1.30 $\pm$ 0.07 & 0.55 $\pm$ 0.04 & 1.69 $\pm$ 0.13 & 1.79 $\pm$ 0.13 & 0.75 $\pm$ 0.07 \\
& 100 & 1.17 $\pm$ 0.06 & 1.22 $\pm$ 0.05 & 0.45 $\pm$ 0.03 & 1.37 $\pm$ 0.08 & 1.43 $\pm$ 0.07 & 0.53 $\pm$ 0.04 & 1.83 $\pm$ 0.10 & 1.92 $\pm$ 0.10 & 0.71 $\pm$ 0.04 \\
& 500 & 1.45 $\pm$ 0.06 & 1.48 $\pm$ 0.06 & 0.37 $\pm$ 0.02 & 1.70 $\pm$ 0.09 & 1.74 $\pm$ 0.09 & 0.43 $\pm$ 0.03 & 2.23 $\pm$ 0.13 & 2.27 $\pm$ 0.13 & 0.57 $\pm$ 0.05 \\
& 1000 & 1.59 $\pm$ 0.05 & 1.61 $\pm$ 0.05 & \textbf{0.36 $\pm$ 0.04} & 1.86 $\pm$ 0.07 & 1.89 $\pm$ 0.06 & \textbf{0.42 $\pm$ 0.04} & 2.39 $\pm$ 0.12 & 2.43 $\pm$ 0.12 & \textbf{0.54 $\pm$ 0.05} \\
& 5000 & 2.10 $\pm$ 0.09 & 2.15 $\pm$ 0.10 & 0.52 $\pm$ 0.05 & 2.45 $\pm$ 0.07 & 2.51 $\pm$ 0.08 & 0.61 $\pm$ 0.06 & 3.13 $\pm$ 0.11 & 3.22 $\pm$ 0.10 & 0.78 $\pm$ 0.08 \\
& 10000 & 2.43 $\pm$ 0.07 & 2.52 $\pm$ 0.07 & 0.74 $\pm$ 0.09 & 2.83 $\pm$ 0.04 & 2.94 $\pm$ 0.05 & 0.86 $\pm$ 0.11 & 3.59 $\pm$ 0.10 & 3.73 $\pm$ 0.13 & 1.09 $\pm$ 0.14 \\
\bottomrule\end{tabular}}
\label{tab:lambda_results_eff_all_1}\end{table*}

\clearpage

\begin{table*}[t]
\caption{Impact of $\lambda$ on Efficiency across different confidence levels ($1-\alpha$). Best results per dataset are highlighted in bold. The row corresponding to $\lambda = 5$ is shaded in gray. (Part 2).}
\centering\scriptsize\resizebox{\textwidth}{!}{
\begin{tabular}{lcccccccccc}
\toprule
\textbf{Dataset} & \textbf{Lambda} & \multicolumn{3}{c}{$\alpha = 0.10$} & \multicolumn{3}{c}{$\alpha = 0.05$} & \multicolumn{3}{c}{$\alpha = 0.01$} \\
\cmidrule(lr){3-5} \cmidrule(lr){6-8} \cmidrule(lr){9-11}
& \textbf{($\lambda$)} & Med. Eff. $\downarrow$ & Mean Eff. $\downarrow$ & IQR $\downarrow$ & Med. Eff. $\downarrow$ & Mean Eff. $\downarrow$ & IQR $\downarrow$ & Med. Eff. $\downarrow$ & Mean Eff. $\downarrow$ & IQR $\downarrow$ \\ 
\midrule
\multirow{13}{*}{\begin{tabular}{@{}l@{}}\textbf{Medical} \\ \textbf{Charges}\end{tabular}}
& 0.01 & 0.21 $\pm$ 0.01 & 0.20 $\pm$ 0.01 & 0.05 $\pm$ 0.01 & 0.33 $\pm$ 0.02 & 0.32 $\pm$ 0.02 & 0.08 $\pm$ 0.02 & 0.83 $\pm$ 0.08 & 0.80 $\pm$ 0.08 & 0.20 $\pm$ 0.06 \\
& 0.05 & 0.21 $\pm$ 0.02 & 0.21 $\pm$ 0.02 & 0.06 $\pm$ 0.04 & 0.35 $\pm$ 0.05 & 0.33 $\pm$ 0.05 & 0.10 $\pm$ 0.06 & 0.89 $\pm$ 0.24 & 0.86 $\pm$ 0.23 & 0.26 $\pm$ 0.21 \\
& 0.1 & 0.21 $\pm$ 0.01 & 0.20 $\pm$ 0.01 & 0.05 $\pm$ 0.03 & 0.33 $\pm$ 0.03 & 0.32 $\pm$ 0.03 & 0.08 $\pm$ 0.05 & 0.82 $\pm$ 0.14 & 0.78 $\pm$ 0.13 & 0.20 $\pm$ 0.15 \\
& 0.5 & 0.20 $\pm$ 0.01 & 0.20 $\pm$ 0.01 & 0.04 $\pm$ 0.02 & 0.32 $\pm$ 0.03 & 0.31 $\pm$ 0.02 & 0.07 $\pm$ 0.03 & 0.75 $\pm$ 0.10 & 0.72 $\pm$ 0.09 & 0.17 $\pm$ 0.08 \\
& 1 & 0.20 $\pm$ 0.01 & 0.19 $\pm$ 0.01 & 0.04 $\pm$ 0.01 & 0.32 $\pm$ 0.02 & 0.31 $\pm$ 0.02 & 0.07 $\pm$ 0.02 & 0.74 $\pm$ 0.08 & 0.71 $\pm$ 0.07 & 0.16 $\pm$ 0.05 \\
& \cellcolor{gray!20} 5 & \cellcolor{gray!20} \textbf{0.17 $\pm$ 0.00} & \cellcolor{gray!20} \textbf{0.17 $\pm$ 0.00} & \cellcolor{gray!20} \textbf{0.01 $\pm$ 0.00} & \cellcolor{gray!20} 0.24 $\pm$ 0.00 & \cellcolor{gray!20} \textbf{0.24 $\pm$ 0.00} & \cellcolor{gray!20} \textbf{0.02 $\pm$ 0.00} & \cellcolor{gray!20} 0.50 $\pm$ 0.01 & \cellcolor{gray!20} 0.50 $\pm$ 0.02 & \cellcolor{gray!20} \textbf{0.04 $\pm$ 0.01} \\
& 10 & 0.18 $\pm$ 0.00 & 0.18 $\pm$ 0.00 & 0.02 $\pm$ 0.00 & \textbf{0.23 $\pm$ 0.00} & \textbf{0.24 $\pm$ 0.00} & 0.03 $\pm$ 0.00 & 0.45 $\pm$ 0.01 & 0.46 $\pm$ 0.01 & 0.06 $\pm$ 0.01 \\
& 50 & 0.21 $\pm$ 0.01 & 0.23 $\pm$ 0.01 & 0.08 $\pm$ 0.01 & 0.25 $\pm$ 0.01 & 0.27 $\pm$ 0.01 & 0.09 $\pm$ 0.01 & \textbf{0.35 $\pm$ 0.01} & \textbf{0.38 $\pm$ 0.00} & 0.13 $\pm$ 0.00 \\
& 100 & 0.22 $\pm$ 0.01 & 0.24 $\pm$ 0.01 & 0.08 $\pm$ 0.00 & 0.26 $\pm$ 0.01 & 0.28 $\pm$ 0.01 & 0.10 $\pm$ 0.00 & 0.36 $\pm$ 0.01 & 0.40 $\pm$ 0.01 & 0.14 $\pm$ 0.00 \\
& 500 & 0.24 $\pm$ 0.01 & 0.26 $\pm$ 0.01 & 0.08 $\pm$ 0.00 & 0.29 $\pm$ 0.00 & 0.31 $\pm$ 0.01 & 0.10 $\pm$ 0.00 & 0.40 $\pm$ 0.02 & 0.44 $\pm$ 0.03 & 0.13 $\pm$ 0.01 \\
& 1000 & 0.26 $\pm$ 0.02 & 0.28 $\pm$ 0.02 & 0.08 $\pm$ 0.00 & 0.32 $\pm$ 0.02 & 0.34 $\pm$ 0.02 & 0.10 $\pm$ 0.00 & 0.43 $\pm$ 0.01 & 0.46 $\pm$ 0.01 & 0.13 $\pm$ 0.00 \\
& 5000 & 0.35 $\pm$ 0.03 & 0.37 $\pm$ 0.03 & 0.07 $\pm$ 0.01 & 0.42 $\pm$ 0.03 & 0.45 $\pm$ 0.03 & 0.09 $\pm$ 0.01 & 0.60 $\pm$ 0.04 & 0.63 $\pm$ 0.04 & 0.12 $\pm$ 0.01 \\
& 10000 & 0.44 $\pm$ 0.12 & 0.45 $\pm$ 0.12 & 0.07 $\pm$ 0.02 & 0.52 $\pm$ 0.12 & 0.53 $\pm$ 0.12 & 0.08 $\pm$ 0.02 & 0.70 $\pm$ 0.11 & 0.73 $\pm$ 0.11 & 0.10 $\pm$ 0.02 \\
\midrule
\multirow{13}{*}{\textbf{Pol}}
& 0.01 & 0.48 $\pm$ 0.03 & 0.49 $\pm$ 0.04 & 0.63 $\pm$ 0.02 & 0.83 $\pm$ 0.09 & 0.86 $\pm$ 0.04 & 1.10 $\pm$ 0.05 & 2.16 $\pm$ 0.21 & 2.25 $\pm$ 0.26 & 2.88 $\pm$ 0.27 \\
& 0.05 & 0.44 $\pm$ 0.06 & 0.46 $\pm$ 0.03 & 0.59 $\pm$ 0.05 & 0.80 $\pm$ 0.10 & 0.83 $\pm$ 0.04 & 1.07 $\pm$ 0.07 & 2.21 $\pm$ 0.30 & 2.31 $\pm$ 0.25 & 2.98 $\pm$ 0.32 \\
& 0.1 & 0.47 $\pm$ 0.07 & 0.49 $\pm$ 0.06 & 0.63 $\pm$ 0.07 & 0.81 $\pm$ 0.12 & 0.85 $\pm$ 0.07 & 1.09 $\pm$ 0.09 & 2.20 $\pm$ 0.28 & 2.30 $\pm$ 0.23 & 2.95 $\pm$ 0.28 \\
& 0.5 & 0.43 $\pm$ 0.06 & 0.45 $\pm$ 0.04 & 0.60 $\pm$ 0.04 & 0.76 $\pm$ 0.10 & 0.78 $\pm$ 0.05 & 1.05 $\pm$ 0.06 & 2.05 $\pm$ 0.24 & 2.12 $\pm$ 0.16 & 2.84 $\pm$ 0.20 \\
& 1 & 0.32 $\pm$ 0.03 & 0.44 $\pm$ 0.03 & 0.69 $\pm$ 0.04 & 0.56 $\pm$ 0.07 & 0.77 $\pm$ 0.08 & 1.20 $\pm$ 0.10 & 1.42 $\pm$ 0.20 & 1.97 $\pm$ 0.27 & 3.08 $\pm$ 0.36 \\
& \cellcolor{gray!20} 5 & \cellcolor{gray!20} \textbf{0.09 $\pm$ 0.03} & \cellcolor{gray!20} \textbf{0.34 $\pm$ 0.05} & \cellcolor{gray!20} \textbf{0.33 $\pm$ 0.03} & \cellcolor{gray!20} \textbf{0.14 $\pm$ 0.04} & \cellcolor{gray!20} 0.54 $\pm$ 0.10 & \cellcolor{gray!20} 0.51 $\pm$ 0.06 & \cellcolor{gray!20} \textbf{0.33 $\pm$ 0.08} & \cellcolor{gray!20} 1.23 $\pm$ 0.15 & \cellcolor{gray!20} 1.17 $\pm$ 0.08 \\
& 10 & 0.15 $\pm$ 0.04 & 0.37 $\pm$ 0.03 & 0.36 $\pm$ 0.02 & 0.19 $\pm$ 0.05 & \textbf{0.48 $\pm$ 0.06} & \textbf{0.47 $\pm$ 0.05} & 0.41 $\pm$ 0.16 & 1.02 $\pm$ 0.23 & 0.99 $\pm$ 0.21 \\
& 50 & 0.54 $\pm$ 0.07 & 0.74 $\pm$ 0.07 & 0.72 $\pm$ 0.03 & 0.66 $\pm$ 0.08 & 0.91 $\pm$ 0.06 & 0.88 $\pm$ 0.02 & 0.80 $\pm$ 0.09 & \textbf{1.09 $\pm$ 0.07} & \textbf{1.06 $\pm$ 0.02} \\
& 100 & 0.91 $\pm$ 0.09 & 1.08 $\pm$ 0.08 & 1.12 $\pm$ 0.05 & 1.05 $\pm$ 0.09 & 1.25 $\pm$ 0.08 & 1.31 $\pm$ 0.04 & 1.24 $\pm$ 0.11 & 1.47 $\pm$ 0.11 & 1.54 $\pm$ 0.06 \\
& 500 & 2.68 $\pm$ 0.23 & 2.69 $\pm$ 0.31 & 2.56 $\pm$ 0.19 & 2.99 $\pm$ 0.20 & 2.99 $\pm$ 0.29 & 2.86 $\pm$ 0.22 & 3.35 $\pm$ 0.18 & 3.35 $\pm$ 0.27 & 3.20 $\pm$ 0.21 \\
& 1000 & 3.56 $\pm$ 0.33 & 3.58 $\pm$ 0.32 & 2.39 $\pm$ 0.41 & 3.79 $\pm$ 0.29 & 3.81 $\pm$ 0.27 & 2.55 $\pm$ 0.47 & 4.07 $\pm$ 0.28 & 4.09 $\pm$ 0.26 & 2.74 $\pm$ 0.52 \\
& 5000 & 4.97 $\pm$ 0.10 & 4.91 $\pm$ 0.08 & 1.43 $\pm$ 0.19 & 5.09 $\pm$ 0.09 & 5.03 $\pm$ 0.07 & 1.47 $\pm$ 0.19 & 5.25 $\pm$ 0.07 & 5.19 $\pm$ 0.04 & 1.52 $\pm$ 0.18 \\
& 10000 & 5.25 $\pm$ 0.11 & 5.28 $\pm$ 0.10 & 1.47 $\pm$ 0.22 & 5.37 $\pm$ 0.11 & 5.41 $\pm$ 0.09 & 1.51 $\pm$ 0.22 & 5.57 $\pm$ 0.13 & 5.60 $\pm$ 0.10 & 1.56 $\pm$ 0.23 \\
\midrule
\multirow{13}{*}{\textbf{Superconduct}}
& 0.01 & 1.70 $\pm$ 0.06 & 2.78 $\pm$ 0.16 & 2.39 $\pm$ 0.23 & 2.66 $\pm$ 0.13 & 4.36 $\pm$ 0.33 & 3.76 $\pm$ 0.45 & 6.01 $\pm$ 0.53 & 9.81 $\pm$ 0.85 & 8.42 $\pm$ 0.64 \\
& 0.05 & 1.68 $\pm$ 0.08 & 2.75 $\pm$ 0.18 & 2.38 $\pm$ 0.30 & 2.64 $\pm$ 0.17 & 4.32 $\pm$ 0.38 & 3.75 $\pm$ 0.56 & 6.01 $\pm$ 0.35 & 9.85 $\pm$ 0.63 & 8.49 $\pm$ 0.69 \\
& 0.1 & 1.68 $\pm$ 0.06 & 2.71 $\pm$ 0.16 & 2.34 $\pm$ 0.30 & 2.61 $\pm$ 0.13 & 4.22 $\pm$ 0.34 & 3.65 $\pm$ 0.55 & 6.05 $\pm$ 0.48 & 9.74 $\pm$ 0.70 & 8.39 $\pm$ 0.78 \\
& 0.5 & 1.64 $\pm$ 0.10 & 2.44 $\pm$ 0.21 & 2.06 $\pm$ 0.34 & 2.50 $\pm$ 0.18 & 3.73 $\pm$ 0.36 & 3.14 $\pm$ 0.55 & 5.35 $\pm$ 0.55 & 7.98 $\pm$ 1.06 & 6.74 $\pm$ 1.39 \\
& 1 & 1.54 $\pm$ 0.07 & 1.81 $\pm$ 0.07 & 1.19 $\pm$ 0.16 & 2.24 $\pm$ 0.12 & 2.65 $\pm$ 0.14 & 1.74 $\pm$ 0.25 & 4.42 $\pm$ 0.34 & 5.22 $\pm$ 0.41 & 3.43 $\pm$ 0.55 \\
& \cellcolor{gray!20} 5 & \cellcolor{gray!20} \textbf{1.29 $\pm$ 0.04} & \cellcolor{gray!20} \textbf{1.36 $\pm$ 0.03} & \cellcolor{gray!20} 0.47 $\pm$ 0.02 & \cellcolor{gray!20} 1.69 $\pm$ 0.05 & \cellcolor{gray!20} 1.78 $\pm$ 0.05 & \cellcolor{gray!20} 0.62 $\pm$ 0.03 & \cellcolor{gray!20} 3.03 $\pm$ 0.11 & \cellcolor{gray!20} 3.19 $\pm$ 0.14 & \cellcolor{gray!20} 1.11 $\pm$ 0.06 \\
& 10 & 1.39 $\pm$ 0.05 & 1.45 $\pm$ 0.05 & 0.46 $\pm$ 0.04 & \textbf{1.65 $\pm$ 0.04} & \textbf{1.72 $\pm$ 0.03} & 0.54 $\pm$ 0.04 & 2.77 $\pm$ 0.10 & 2.90 $\pm$ 0.13 & 0.91 $\pm$ 0.09 \\
& 50 & 1.87 $\pm$ 0.03 & 1.92 $\pm$ 0.03 & \textbf{0.41 $\pm$ 0.02} & 2.04 $\pm$ 0.03 & 2.09 $\pm$ 0.03 & \textbf{0.45 $\pm$ 0.03} & \textbf{2.53 $\pm$ 0.10} & \textbf{2.59 $\pm$ 0.09} & \textbf{0.56 $\pm$ 0.05} \\
& 100 & 2.11 $\pm$ 0.03 & 2.13 $\pm$ 0.02 & 0.45 $\pm$ 0.03 & 2.28 $\pm$ 0.03 & 2.30 $\pm$ 0.01 & 0.48 $\pm$ 0.03 & 2.66 $\pm$ 0.05 & 2.69 $\pm$ 0.06 & \textbf{0.56 $\pm$ 0.05} \\
& 500 & 2.43 $\pm$ 0.04 & 2.45 $\pm$ 0.07 & 0.59 $\pm$ 0.09 & 2.63 $\pm$ 0.03 & 2.66 $\pm$ 0.05 & 0.64 $\pm$ 0.10 & 3.04 $\pm$ 0.08 & 3.07 $\pm$ 0.07 & 0.74 $\pm$ 0.12 \\
& 1000 & 2.53 $\pm$ 0.04 & 2.52 $\pm$ 0.04 & 0.64 $\pm$ 0.07 & 2.78 $\pm$ 0.07 & 2.77 $\pm$ 0.04 & 0.70 $\pm$ 0.08 & 3.31 $\pm$ 0.10 & 3.30 $\pm$ 0.08 & 0.83 $\pm$ 0.11 \\
& 5000 & 2.79 $\pm$ 0.23 & 2.72 $\pm$ 0.20 & 0.70 $\pm$ 0.14 & 3.09 $\pm$ 0.18 & 3.00 $\pm$ 0.12 & 0.77 $\pm$ 0.15 & 3.70 $\pm$ 0.19 & 3.60 $\pm$ 0.11 & 0.92 $\pm$ 0.17 \\
& 10000 & 2.81 $\pm$ 0.17 & 2.72 $\pm$ 0.16 & 0.71 $\pm$ 0.11 & 3.15 $\pm$ 0.12 & 3.04 $\pm$ 0.09 & 0.80 $\pm$ 0.13 & 3.88 $\pm$ 0.12 & 3.75 $\pm$ 0.07 & 0.99 $\pm$ 0.17 \\
\midrule
\multirow{13}{*}{\begin{tabular}{@{}l@{}}\textbf{Wine} \\ \textbf{Quality}\end{tabular}}
& 0.01 & 0.40 $\pm$ 0.02 & 0.41 $\pm$ 0.03 & 0.19 $\pm$ 0.03 & 0.56 $\pm$ 0.03 & 0.57 $\pm$ 0.04 & 0.27 $\pm$ 0.05 & 1.23 $\pm$ 0.19 & 1.26 $\pm$ 0.19 & 0.58 $\pm$ 0.08 \\
& 0.05 & 0.40 $\pm$ 0.02 & 0.41 $\pm$ 0.03 & 0.19 $\pm$ 0.03 & 0.56 $\pm$ 0.03 & 0.58 $\pm$ 0.04 & 0.27 $\pm$ 0.05 & 1.24 $\pm$ 0.19 & 1.27 $\pm$ 0.19 & 0.59 $\pm$ 0.09 \\
& 0.1 & 0.41 $\pm$ 0.02 & 0.42 $\pm$ 0.02 & 0.20 $\pm$ 0.03 & 0.56 $\pm$ 0.03 & 0.58 $\pm$ 0.04 & 0.27 $\pm$ 0.04 & 1.24 $\pm$ 0.18 & 1.26 $\pm$ 0.18 & 0.59 $\pm$ 0.08 \\
& 0.5 & 0.41 $\pm$ 0.01 & 0.42 $\pm$ 0.01 & 0.20 $\pm$ 0.02 & 0.55 $\pm$ 0.01 & 0.57 $\pm$ 0.01 & 0.27 $\pm$ 0.03 & 1.25 $\pm$ 0.04 & 1.29 $\pm$ 0.04 & 0.61 $\pm$ 0.05 \\
& 1 & 0.39 $\pm$ 0.01 & 0.39 $\pm$ 0.01 & 0.16 $\pm$ 0.03 & 0.52 $\pm$ 0.02 & 0.52 $\pm$ 0.02 & 0.22 $\pm$ 0.05 & 1.13 $\pm$ 0.10 & 1.14 $\pm$ 0.11 & 0.47 $\pm$ 0.12 \\
& \cellcolor{gray!20} 5 & \cellcolor{gray!20} \textbf{0.35 $\pm$ 0.01} & \cellcolor{gray!20} \textbf{0.35 $\pm$ 0.01} & \cellcolor{gray!20} 0.04 $\pm$ 0.01 & \cellcolor{gray!20} \textbf{0.43 $\pm$ 0.01} & \cellcolor{gray!20} \textbf{0.43 $\pm$ 0.01} & \cellcolor{gray!20} 0.05 $\pm$ 0.01 & \cellcolor{gray!20} 0.65 $\pm$ 0.05 & \cellcolor{gray!20} 0.65 $\pm$ 0.05 & \cellcolor{gray!20} 0.07 $\pm$ 0.01 \\
& 10 & \textbf{0.35 $\pm$ 0.01} & 0.36 $\pm$ 0.01 & \textbf{0.03 $\pm$ 0.00} & \textbf{0.43 $\pm$ 0.02} & \textbf{0.43 $\pm$ 0.02} & \textbf{0.04 $\pm$ 0.01} & \textbf{0.63 $\pm$ 0.04} & \textbf{0.63 $\pm$ 0.04} & \textbf{0.06 $\pm$ 0.01} \\
& 50 & 0.46 $\pm$ 0.03 & 0.47 $\pm$ 0.03 & 0.05 $\pm$ 0.01 & 0.54 $\pm$ 0.04 & 0.55 $\pm$ 0.04 & 0.06 $\pm$ 0.01 & 0.73 $\pm$ 0.05 & 0.74 $\pm$ 0.05 & 0.09 $\pm$ 0.01 \\
& 100 & 0.56 $\pm$ 0.06 & 0.57 $\pm$ 0.06 & 0.07 $\pm$ 0.01 & 0.65 $\pm$ 0.05 & 0.66 $\pm$ 0.05 & 0.09 $\pm$ 0.01 & 0.83 $\pm$ 0.05 & 0.85 $\pm$ 0.05 & 0.11 $\pm$ 0.01 \\
& 500 & 1.04 $\pm$ 0.16 & 1.06 $\pm$ 0.17 & 0.16 $\pm$ 0.02 & 1.18 $\pm$ 0.19 & 1.21 $\pm$ 0.20 & 0.18 $\pm$ 0.03 & 1.45 $\pm$ 0.23 & 1.49 $\pm$ 0.25 & 0.22 $\pm$ 0.03 \\
& 1000 & 1.27 $\pm$ 0.18 & 1.30 $\pm$ 0.19 & 0.20 $\pm$ 0.04 & 1.45 $\pm$ 0.22 & 1.49 $\pm$ 0.22 & 0.23 $\pm$ 0.04 & 1.76 $\pm$ 0.25 & 1.80 $\pm$ 0.26 & 0.27 $\pm$ 0.05 \\
& 5000 & 1.72 $\pm$ 0.35 & 1.80 $\pm$ 0.36 & 0.33 $\pm$ 0.10 & 1.97 $\pm$ 0.35 & 2.05 $\pm$ 0.36 & 0.38 $\pm$ 0.12 & 2.34 $\pm$ 0.36 & 2.44 $\pm$ 0.38 & 0.45 $\pm$ 0.14 \\
& 10000 & 2.19 $\pm$ 0.72 & 2.34 $\pm$ 0.80 & 0.52 $\pm$ 0.21 & 2.44 $\pm$ 0.69 & 2.60 $\pm$ 0.78 & 0.58 $\pm$ 0.22 & 2.85 $\pm$ 0.72 & 3.04 $\pm$ 0.81 & 0.68 $\pm$ 0.24 \\
\midrule
\multirow{13}{*}{\textbf{Drift}}
& 0.01 & 85.77 $\pm$ 13.72 & 485.24 $\pm$ 139.09 & 340.83 $\pm$ 62.37 & 129.24 $\pm$ 27.41 & 711.75 $\pm$ 127.62 & 511.50 $\pm$ 109.12 & 231.64 $\pm$ 66.53 & 1245.08 $\pm$ 159.81 & 915.42 $\pm$ 268.53 \\
& 0.05 & 128.65 $\pm$ 55.28 & 546.47 $\pm$ 130.17 & 452.92 $\pm$ 139.28 & 195.72 $\pm$ 88.58 & 815.22 $\pm$ 163.47 & 683.63 $\pm$ 226.99 & 353.86 $\pm$ 174.01 & 1443.59 $\pm$ 267.66 & 1225.39 $\pm$ 456.33 \\
& 0.1 & 66.60 $\pm$ 27.49 & 384.38 $\pm$ 87.21 & 270.27 $\pm$ 102.54 & 101.60 $\pm$ 44.58 & 584.75 $\pm$ 147.23 & 412.20 $\pm$ 166.88 & 201.74 $\pm$ 91.76 & 1160.53 $\pm$ 315.22 & 819.73 $\pm$ 346.97 \\
& 0.5 & 64.62 $\pm$ 21.31 & 393.58 $\pm$ 79.36 & 204.61 $\pm$ 59.75 & 95.03 $\pm$ 33.34 & 575.55 $\pm$ 108.95 & 300.99 $\pm$ 94.93 & 179.22 $\pm$ 59.02 & 1093.16 $\pm$ 213.72 & 570.41 $\pm$ 173.99 \\
& 1 & 79.64 $\pm$ 10.16 & 133.81 $\pm$ 17.41 & 141.85 $\pm$ 19.99 & 148.52 $\pm$ 17.69 & 248.95 $\pm$ 24.40 & 263.94 $\pm$ 30.69 & 377.03 $\pm$ 27.50 & 634.12 $\pm$ 61.13 & 670.45 $\pm$ 57.30 \\
& \cellcolor{gray!20} 5 & \cellcolor{gray!20} 22.51 $\pm$ 2.28 & \cellcolor{gray!20} \textbf{24.72 $\pm$ 2.46} & \cellcolor{gray!20} \textbf{13.04 $\pm$ 3.21} & \cellcolor{gray!20} 27.59 $\pm$ 2.98 & \cellcolor{gray!20} \textbf{30.31 $\pm$ 3.34} & \cellcolor{gray!20} \textbf{16.02 $\pm$ 4.19} & \cellcolor{gray!20} 41.19 $\pm$ 5.90 & \cellcolor{gray!20} 45.28 $\pm$ 6.74 & \cellcolor{gray!20} 24.02 $\pm$ 7.20 \\
& 10 & 22.08 $\pm$ 3.21 & 24.55 $\pm$ 4.37 & 12.95 $\pm$ 4.35 & 26.96 $\pm$ 3.11 & 29.95 $\pm$ 4.40 & 15.79 $\pm$ 4.99 & 38.15 $\pm$ 3.36 & \textbf{42.35 $\pm$ 4.88} & \textbf{22.30 $\pm$ 6.78} \\
& 50 & 25.20 $\pm$ 12.18 & 28.04 $\pm$ 13.44 & 13.69 $\pm$ 8.50 & 31.00 $\pm$ 15.03 & 34.50 $\pm$ 16.58 & 16.85 $\pm$ 10.49 & 44.21 $\pm$ 20.26 & 49.21 $\pm$ 22.37 & 23.92 $\pm$ 14.24 \\
& 100 & \textbf{22.86 $\pm$ 1.62} & 26.39 $\pm$ 3.63 & 14.40 $\pm$ 3.94 & \textbf{27.40 $\pm$ 1.81} & 31.64 $\pm$ 4.26 & 17.28 $\pm$ 4.80 & \textbf{37.37 $\pm$ 3.88} & 43.22 $\pm$ 7.48 & 23.71 $\pm$ 7.60 \\
& 500 & 31.85 $\pm$ 4.66 & 36.66 $\pm$ 6.64 & 19.99 $\pm$ 5.62 & 38.08 $\pm$ 5.74 & 43.83 $\pm$ 8.07 & 23.90 $\pm$ 6.74 & 51.99 $\pm$ 8.98 & 59.75 $\pm$ 11.59 & 32.47 $\pm$ 9.00 \\
& 1000 & 37.36 $\pm$ 2.79 & 47.63 $\pm$ 5.55 & 24.70 $\pm$ 5.41 & 43.06 $\pm$ 2.86 & 54.87 $\pm$ 5.73 & 28.43 $\pm$ 5.91 & 54.99 $\pm$ 1.66 & 70.12 $\pm$ 6.53 & 36.29 $\pm$ 7.02 \\
& 5000 & 58.13 $\pm$ 13.22 & 65.14 $\pm$ 16.25 & 23.56 $\pm$ 8.94 & 67.31 $\pm$ 17.24 & 75.53 $\pm$ 21.54 & 27.31 $\pm$ 11.00 & 81.31 $\pm$ 18.65 & 91.19 $\pm$ 23.45 & 33.07 $\pm$ 12.80 \\
& 10000 & 82.49 $\pm$ 12.06 & 87.21 $\pm$ 10.27 & 26.64 $\pm$ 15.71 & 99.55 $\pm$ 14.85 & 105.23 $\pm$ 12.58 & 31.97 $\pm$ 18.65 & 139.84 $\pm$ 34.11 & 146.92 $\pm$ 28.50 & 43.91 $\pm$ 26.11 \\
\midrule
\multirow{13}{*}{\begin{tabular}{@{}l@{}}\textbf{UTK} \\ \textbf{Face}\end{tabular}}
& 0.01 & 138.69 $\pm$ 42.87 & 378.64 $\pm$ 138.88 & 235.76 $\pm$ 98.26 & 208.32 $\pm$ 55.31 & 568.11 $\pm$ 176.38 & 351.79 $\pm$ 133.48 & 344.87 $\pm$ 72.34 & 934.46 $\pm$ 209.20 & 580.14 $\pm$ 192.72 \\
& 0.05 & 113.36 $\pm$ 32.09 & 440.49 $\pm$ 284.55 & 204.10 $\pm$ 102.21 & 159.39 $\pm$ 30.99 & 601.83 $\pm$ 308.36 & 282.22 $\pm$ 108.78 & 265.22 $\pm$ 46.35 & 961.65 $\pm$ 317.08 & 459.76 $\pm$ 120.96 \\
& 0.1 & 123.09 $\pm$ 25.53 & 295.86 $\pm$ 21.04 & 195.18 $\pm$ 28.97 & 189.65 $\pm$ 41.42 & 454.28 $\pm$ 19.14 & 300.17 $\pm$ 44.72 & 341.30 $\pm$ 86.09 & 811.58 $\pm$ 45.20 & 538.98 $\pm$ 99.85 \\
& 0.5 & 119.50 $\pm$ 26.40 & 297.73 $\pm$ 50.03 & 213.17 $\pm$ 54.07 & 183.79 $\pm$ 40.42 & 456.44 $\pm$ 64.41 & 326.68 $\pm$ 79.70 & 327.50 $\pm$ 75.78 & 807.36 $\pm$ 74.45 & 579.72 $\pm$ 139.87 \\
& 1 & 97.68 $\pm$ 13.83 & 111.02 $\pm$ 23.27 & 106.05 $\pm$ 31.99 & 171.63 $\pm$ 23.31 & 194.56 $\pm$ 36.36 & 184.91 $\pm$ 49.37 & 363.66 $\pm$ 48.17 & 412.46 $\pm$ 79.39 & 392.12 $\pm$ 108.11 \\
& \cellcolor{gray!20} 5 & \cellcolor{gray!20} \textbf{21.84 $\pm$ 0.84} & \cellcolor{gray!20} \textbf{22.54 $\pm$ 0.43} & \cellcolor{gray!20} 12.18 $\pm$ 1.61 & \cellcolor{gray!20} 27.49 $\pm$ 0.98 & \cellcolor{gray!20} 28.39 $\pm$ 0.91 & \cellcolor{gray!20} 15.36 $\pm$ 2.26 & \cellcolor{gray!20} 42.51 $\pm$ 2.06 & \cellcolor{gray!20} 43.93 $\pm$ 2.76 & \cellcolor{gray!20} 23.82 $\pm$ 4.13 \\
& 10 & 21.96 $\pm$ 1.16 & 22.22 $\pm$ 0.94 & \textbf{12.34 $\pm$ 1.18} & 27.79 $\pm$ 1.66 & 28.12 $\pm$ 1.12 & \textbf{15.61 $\pm$ 1.39} & 42.41 $\pm$ 1.56 & 42.92 $\pm$ 0.45 & 23.80 $\pm$ 1.50 \\
& 50 & 22.20 $\pm$ 2.17 & 22.71 $\pm$ 0.77 & 11.97 $\pm$ 2.00 & 27.17 $\pm$ 2.55 & 27.80 $\pm$ 1.04 & 14.67 $\pm$ 2.55 & 40.26 $\pm$ 2.33 & 41.29 $\pm$ 1.57 & 21.88 $\pm$ 4.48 \\
& 100 & 22.38 $\pm$ 0.66 & 23.40 $\pm$ 0.55 & 12.65 $\pm$ 1.90 & \textbf{26.87 $\pm$ 0.79} & \textbf{28.10 $\pm$ 0.56} & 15.18 $\pm$ 2.18 & \textbf{37.44 $\pm$ 0.73} & \textbf{39.16 $\pm$ 1.28} & 21.17 $\pm$ 3.23 \\
& 500 & 29.87 $\pm$ 3.94 & 31.26 $\pm$ 5.54 & 15.93 $\pm$ 2.13 & 35.25 $\pm$ 4.07 & 36.86 $\pm$ 5.81 & 18.82 $\pm$ 2.40 & 47.47 $\pm$ 3.85 & 49.44 $\pm$ 4.38 & 25.38 $\pm$ 2.95 \\
& 1000 & 31.78 $\pm$ 2.96 & 32.90 $\pm$ 2.64 & 14.13 $\pm$ 3.13 & 37.36 $\pm$ 2.49 & 38.69 $\pm$ 2.15 & 16.69 $\pm$ 4.00 & 49.66 $\pm$ 3.48 & 51.42 $\pm$ 3.14 & 22.27 $\pm$ 5.78 \\
& 5000 & 56.83 $\pm$ 9.02 & 60.51 $\pm$ 10.58 & 22.25 $\pm$ 13.65 & 63.41 $\pm$ 12.67 & 67.55 $\pm$ 14.68 & 25.16 $\pm$ 16.82 & 73.31 $\pm$ 16.82 & 78.11 $\pm$ 19.30 & 29.33 $\pm$ 20.77 \\
& 10000 & 76.74 $\pm$ 4.66 & 79.09 $\pm$ 4.97 & 12.56 $\pm$ 4.47 & 84.29 $\pm$ 4.72 & 86.87 $\pm$ 5.04 & 13.83 $\pm$ 5.02 & 92.08 $\pm$ 4.38 & 94.87 $\pm$ 4.09 & \textbf{15.22 $\pm$ 5.88} \\
\bottomrule\end{tabular}}
\label{tab:lambda_results_eff_all_2}\end{table*}

\end{document}